\documentclass[11pt]{book}
\usepackage{csquotes}
\usepackage{amsmath}
\usepackage{amsthm}
\usepackage{amsfonts}
\usepackage{amssymb}
\usepackage{url}
\usepackage{hyperref}
\usepackage{algorithm,algorithmic}
\usepackage{xspace}
\usepackage{color}
\usepackage{mathtools}
\usepackage{enumerate}
\usepackage{subfigure}
\usepackage[T1]{fontenc}
\usepackage{makeidx}
\usepackage[totoc]{idxlayout}

\usepackage{tikz}
\usetikzlibrary{arrows}
\usetikzlibrary{shapes}
\tikzset{vertex/.style = {shape=circle,draw,minimum size=1.5em}}
\tikzset{edge/.style = {->,> = latex'}}
\usetikzlibrary{arrows.meta, positioning, decorations.pathreplacing}

\newtheorem{theorem}{Theorem}[chapter]
\newtheorem{definition}[theorem]{Definition}

\newtheorem{corollary}[theorem]{Corollary}
\newtheorem{lemma}[theorem]{Lemma}
\newtheorem{fact}[theorem]{Fact}

\newcommand{\citep}{\cite}
\newcommand{\citet}{\cite}

\newif\ifarxiv
\arxivtrue

\usepackage{newfloat}
\DeclareFloatingEnvironment[
	fileext=lob,
	listname={List of Boxes},
	name={MP},
	placement=htp,
]{optbox}

\DeclareMathOperator*{\argmin}{\arg\min}

\newcommand{\defeq}{\equaldef}

\def\bpi{{\boldsymbol{\pi}}}

\def\F{{\mathcal F}}

\def\reals{{\mathbb R}}
\def\R{{\mathcal R}}

\def\norm#1{\mathopen\| #1 \mathclose\|}

\newcommand{\proj}{\mathop{\Pi}}

\newcommand{\rank}{\mathop{\mbox{\rm rank}}}

\newcommand{\myspan}{\mathop{\mbox{\rm span}}}

\newcommand{\ignore}[1]{}
\newcommand{\equaldef}{\stackrel{\text{\tiny def}}{=}}
\newcommand{\equaltri}{\equaldef}

\def\trace{{\bf Tr}}

\def\reals{{\mathbb R}}
\newcommand{\E}{\mathop{\mbox{\bf E}}}
\newcommand\ball{\mathbb{B}}

\def\bzero{\mathbf{0}}

\def\mA{{\mathcal A}}

\def\bold0{\mathbf{0}}

\newcommand\mycases[4] {{
\left\{
\begin{array}{ll}
    {#1}, & {#2} \\\\
    {#3}, & {#4}
\end{array}
\right. }}

\def\bB{\mathbf{B}}
\def\bK{\mathbf{K}}
\def\bM{\mathbf{M}}
\def\bC{\mathbf{C}}

\renewcommand{\a}{\mathbf{a}}
\def\bx{\mathbf{x}}

\def\w{\mathbf{w}}
\def\by{\mathbf{y}}

\def\bp{\mathbf{p}}

\def\br{\mathbf{r}}

\def\bv{\mathbf{v}}
\def\xnat{\mathbf{x}^{\mathbf{nat}}}
\def\ynat{\mathbf{y}^{\mathbf{nat}}}

\def\bw{\mathbf{w}}
\def\ba{\mathbf{a}}
\def\bA{\mathbf{A}}

\def\bB{\mathbf{B}}
\def\bC{\mathbf{C}}

\def\bP{\mathbf{P}}

\def\bone{\mathbf{1}}

\def\trace{{\bf Tr}}

\newcommand{\eps}{\varepsilon}

\def\bone{\mathbf{1}}
\newcommand{\sphere}{\ensuremath{\mathbb {S}}}

\newcommand{\K}{\ensuremath{\mathcal K}}

\def\mA{{\mathcal A}}
\newcommand{\x}{\ensuremath{\mathbf x}}
\newcommand{\s}{\ensuremath{\mathbf s}}

\newcommand{\y}{\ensuremath{\mathbf y}}
\newcommand{\z}{\ensuremath{\mathbf z}}

\newcommand{\xv}[1][t]{\ensuremath{\mathbf x_{#1}}}

\newcommand{\yv}[1][t]{\ensuremath{\mathbf y_{#1}}}

\newcommand{\uv}{\ensuremath{\mathbf u}}
\ifdefined\vv
	\renewcommand{\vv}{\ensuremath{\mathbf v}}
\else
	\newcommand{\vv}{\ensuremath{\mathbf v}}
\fi

\def\regret{\ensuremath{\mathrm{{regret}}}}

\def\bx{\mathbf{x}}
\def\by{\mathbf{y}}

\def\bp{\mathbf{p}}

\def\br{\mathbf{r}}
\def\bv{\mathbf{v}}
\def\ba{\mathbf{a}}

\def\bP{\mathbf{P}}

\def\eps{\varepsilon}
\def\epsilon{\varepsilon}

\def\R{\ensuremath{\mathcal R}}

\usepackage{chngcntr}

\newcounter{exercise}[chapter]
\counterwithin{exercise}{chapter}
\ifdefined\exercise
\renewenvironment{exercise}{
\refstepcounter{exercise}
\textbf{\arabic{chapter}.\arabic{exercise}.}
}{
	\medskip
}
\else
\newenvironment{exercise}{
	\refstepcounter{exercise}
	\textbf{\arabic{chapter}.\arabic{exercise}.}
}{
	\medskip
}
\fi

\newcounter{subexercise}[exercise]
\newenvironment{subexercise}{
	\refstepcounter{subexercise}
	\textbf{\Alph{subexercise}.}
}{
}

\title{
    \tikz[remember picture,overlay] \node[opacity=1.0,inner sep=0pt] at (current page.center){\includegraphics[width=\paperwidth,height=\paperheight]{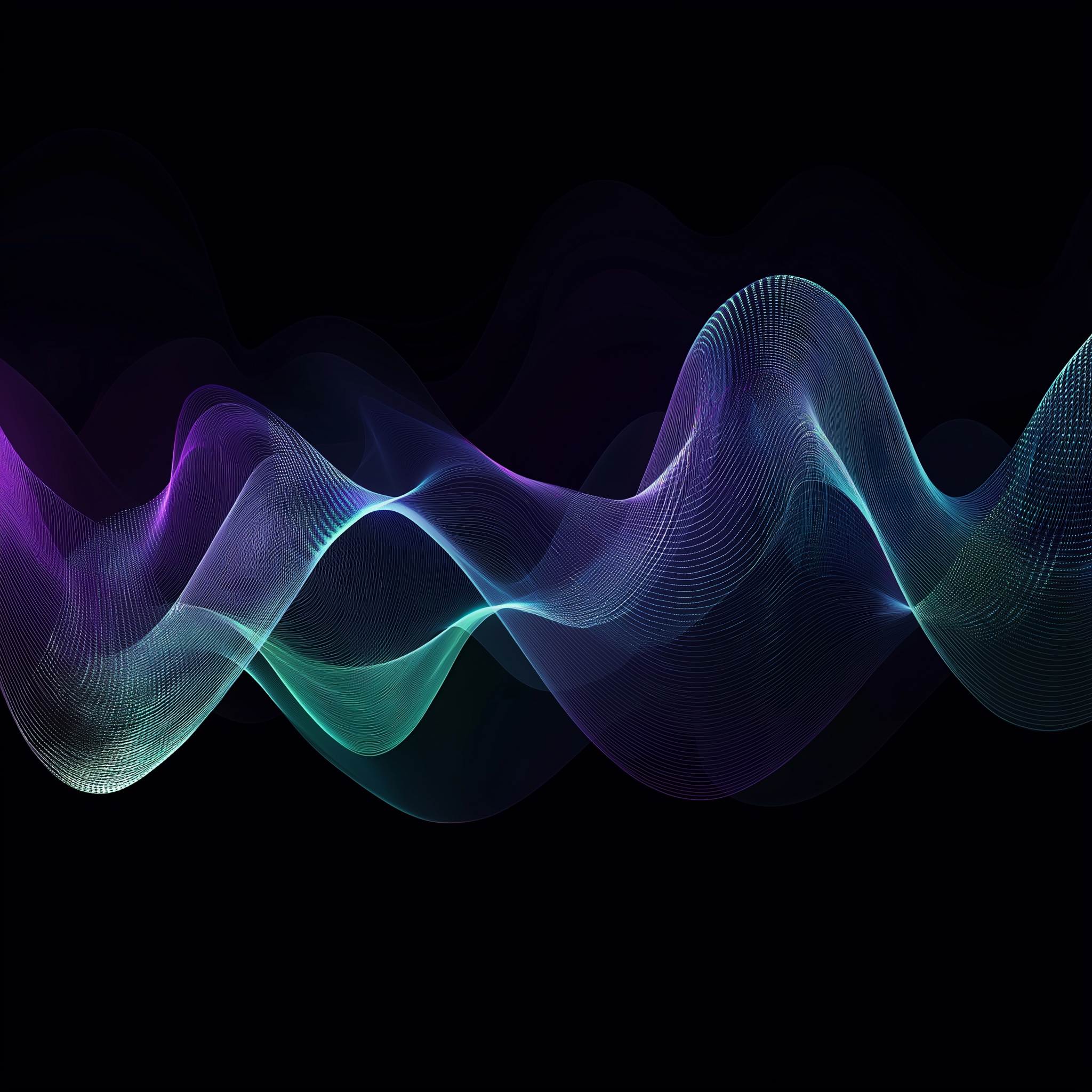}};
    \tikz[remember picture, overlay, opacity=0.2] \fill [white] (0.1,1.1) rectangle (11.2,-3.1);
    \tikz[remember picture, overlay, opacity=0.8] \fill [black] (0,1) rectangle (10.9,-3);
    {\fontfamily{lmss}\selectfont \Huge\color{white}
    Introduction to  Online Control \\ \vspace{0.1em}
    {\normalsize \em version 1.2 $\beta$}}
}
\date{}
\author{\fontfamily{lmss}\selectfont\Large\color{white} Elad Hazan \and \fontfamily{lmss}\selectfont\Large\color{white} Karan Singh}
\makeindex

\begin{document}
\frontmatter
\maketitle

\newpage
\thispagestyle{empty}
\tikz[remember picture,overlay] \node[opacity=1.0,inner sep=0pt] at (current page.center){\includegraphics[width=\paperwidth,height=\paperheight]{images/book-cover.png}};
\tikz[remember picture, overlay, opacity=0.55] \fill [black] (-0.2,-2.0) rectangle (11.5,-10.5);
\tikz[remember picture, overlay, opacity=0.55] \fill [black] (-0.3,-11.3) rectangle (11.4,-17.8);
\vspace*{\fill}
\begin{center}
    \vspace{1cm}
    
    \parbox{0.9\textwidth}{\fontfamily{lmss}\color{white}
    
        \textit{We are in a golden age for control and decision making. A proliferation of new applications including self-driving vehicles, humanoid robots, and artificially intelligent drones open a new set of challenges for control theory to address.  Hazan and Singh have written the definitive book on the New Control Theory-- non-stochastic control. The phrase ``a paradigm shift'' has become cliche from overuse, but here it is truly well deserved; the authors have revisited the foundations by focusing on building controllers that perform nearly as well as if they knew future disturbances in advance, rather than relying on probabilistic or worst-case models. The non-stochastic control approach has extended one of the most profound ideas in mathematics of the 20th century, online (no-regret) learning, to master sequential decision making with continuous actions. This leads to high performance in benign environments and resilience in adversarial ones. The book, authored by pioneers in the field, presents both foundational concepts and the latest research, making it an invaluable resource.}\\[1ex]
        \hspace*{0pt}\hfill --- \textbf{Drew Bagnell}        
        \vspace{1cm}
        
        \textit{As someone who has worked extensively on learning theory and online learning, and later applied these ideas in domains such as autonomous driving and humanoid robotics, I find this book both timely and inspiring. It introduces a regret-minimization framework for control that draws on the elegance and power of online learning. Traditional control theory often models noise either as stochastic—sometimes unrealistically optimistic—or adversarial—often overly conservative. This book charts a new path by asking a deeper question: while we cannot predict noise, can we perform nearly as well as if we could? The answer, developed here, is a novel and exciting paradigm that bridges learning theory and control, and I believe it will have a lasting impact on both research and practice.} \\ [1ex]
        \hspace*{0pt}\hfill --- \textbf{Shai Shalev-Shwartz}
        }

\end{center}
\vspace*{\fill}
\newpage

\chapter*{}
\begin{center}
To my family: Dana, Hadar, Yoav, Oded, and Deluca, \\ 
---EH \\
\end{center}
\hspace*{\fill}
\begin{center}
To my parents-- A \& V. \\ 
---KS \\
\end{center}

\ifarxiv
\hspace*{\fill}
\vfill
\textcopyright\ [2025] Elad Hazan and Karan Singh
\fi

\chapter*{Preface}
\ifarxiv
\addcontentsline{toc}{chapter}{Preface}
     \markboth{\sffamily\slshape Preface}
       {\sffamily\slshape Preface}
\fi

This text presents an introduction to an emerging paradigm in control of dynamical systems and differentiable reinforcement learning called {\em online nonstochastic control}. The new approach applies techniques from online convex optimization and convex relaxations to obtain new methods with provable guarantees for classical settings in optimal and robust control. 

The primary distinction between online nonstochastic control and other frameworks is the objective. In optimal control, robust control, and other control methodologies that assume stochastic noise,  the goal is to perform comparably to an offline optimal strategy. In online nonstochastic control, both the cost functions and the perturbations from the assumed dynamical model are chosen by an adversary. Thus, the optimal policy is not defined a priori. 
Rather, the goal is to attain low regret against the best policy in hindsight from a benchmark class of policies. 

This objective suggests the use of the decision-making framework of online convex optimization as an algorithmic methodology. The resulting methods are based on iterative mathematical optimization algorithms and are accompanied by finite-time regret and computational complexity guarantees. 
\chapter*{Acknowledgments}
\ifarxiv
     \addcontentsline{toc}{chapter}{Acknowledgments}
     \markboth{\sffamily\slshape Acknowledgments}
       {\sffamily\slshape Acknowledgments}
\fi

This book was a journey that started in Princeton at the Google DeepMind Lab in the summer of 2018. At that time Sham Kakade visited the group, which consisted of the authors together with Naman Agarwal, Brian Bullins, Xinyi Chen, Cyril Zhang, and Yi Zhang, henceforth {\em the gang}. The {\em gang} worked on various aspects of optimization, reinforcement learning, and dynamical systems. The first paper, in which online nonstochastic control was basically invented, started in room 401 in the computer science building at Princeton University. We remember and cherish the exact moment along with the mathematical writings in white chalk on that blackboard. 

Ever since then, the gem of an idea turned out into a complete theory that is now mainstream in machine learning and control, with numerous applications and refinements. Notable events that contributed to this theory include the following. Max Simchowitz, who was a graduate student then, visited Princeton during 2019-2020, where he collaborated with the group, and the theory was extended to partial observed systems. Paula Gradu, who was then an undergraduate, and Edgar Minasyan, while a graduate student, contributed to the study of time-varying and nonlinear dynamical systems. Anirudha Majumdar joined for several important projects, notably for iterative linear control. Udaya Ghai and Arushi Gupta studied extending the GPC to model-free RL. 

Most recently, this theory has been used to design new architectures for sequence prediction. The theory of spectral filtering was extended by the aforementioned members of the group, and newer folks joined, including Daniel Suo and Annie Marsden. 


This book was further refined through teaching. Elad Hazan gratefully acknowledges the numerous contributions and insights of the students of the course {\em computational control theory} delivered at Princeton during COVID over Zoom during
2020-2021. The text was further enhanced by students of {\em theoretical machine learning} taught
at Princeton University in the fall of 2024-2025.

The following colleagues and students were particularly helpful, making insightful comments, finding various typos and suggesting improvements: Gon Buzaglo, Anand Brahmbhatt, Sofiia Druchyna,  Jonathan Pillow, and many more.

Throughout the writing of this book, we engaged in extensive discussions with ChatGPT, an AI assistant. Although it does not claim authorship, its ability to refine mathematical reasoning and clarify complex topics has been a remarkable collaboration in human-AI interaction. In fact, it wrote this acknowledgment all by itself!  

\chapter*{List of Symbols}
\ifarxiv
\addcontentsline{toc}{chapter}{List of Symbols}
\fi

\subsection*{General}

\begingroup
\renewcommand{\arraystretch}{1.5}
\setlength{\tabcolsep}{10pt} 
\begin{tabular}{ll}
$\equaldef$ & definition \\
$\argmin\{ \}$ &  the argument minimizing the expression in braces \\
$[n]$ & the set of integers $\{1,2,\ldots,n\}$ \\
$\bone_{A}$ & the indicator function, equals one if $A$ is true, else zero
\end{tabular}  
\endgroup

\subsection*{Linear Algebra, Geometry and Calculus}

\begingroup
\renewcommand{\arraystretch}{1.5}
\setlength{\tabcolsep}{10pt} 

\begin{tabular}{ll}
$\reals^d$ & $d$ dimensional Euclidean space \\
$\Delta_d$ & $d$ dimensional simplex, $\{ \sum_i \x_i=1, \x_i \geq 0\}$    \\
$\sphere$ & $d$ dimensional sphere, $\{ \|\x\| =1\}$  \\
$\ball$ & $d$ dimensional ball, $\{ \|\x\| \leq 1 \}$  \\
$\reals$ & real numbers \\
$\mathbb{C}$ & complex numbers \\
$|A|$ & determinant of matrix $A$  \\
$\trace(A)$ & trace of matrix $A$ \\
$ A \succeq 0$ & $A$ is positive semi-definite, $\forall v \ ,  \ v^\top A v \geq 0$ \\
$ A \succeq B$ & $A-B$ is positive semi-definite 
\end{tabular}  
\endgroup

\subsection*{Control}

\begingroup
\renewcommand{\arraystretch}{1.5}
\setlength{\tabcolsep}{10pt} 
\begin{tabular}{ll}
 $f$  & dynamics function \\
 $\x_t \in \reals^{d_\x}$  & state at time $t$  \\
 $\uv_t \in \reals^{d_\uv} $  & control at time $t$ \\
 $\w_t \in \reals^{d_\w} $  & perturbation at time $t$ \\
 $\y_t \in \reals^{d_\y} $  & observation at time $t$ \\
 $A_t,B_t,C_t$ & system matrices for linear dynamical system \\
 $\gamma $ & BIBO stabilizability  $\|\w_t\| \leq 1 \longrightarrow \|\x_t\| \leq \gamma$ \\
 $h$ & Number of parameters in a policy class \\
 $\kappa$ & norm bound on policy parameters \\
 $K_t$ & stabilizing linear controller at time $t$ \\
 $K_r(A,B)$ & Kalman controllability matrix of order $r$ \\
 $ \hat{K}(A,C)$ & Kalman observability matrix 
\end{tabular}
\endgroup

\subsection*{Reinforcement learning}
\begingroup
\renewcommand{\arraystretch}{1.5}
\setlength{\tabcolsep}{10pt} 
\begin{tabular}{ll}
 $S,A,\gamma, R,P$  & Markov Decision Process with states $S$, actions $A$, \\
 & discount factor $\gamma$, rewards $R$ and transition probability matrix $P$ \\
 $R_t$  & reward at time $t$ \\
$\bP_{ss'}^{a}$  & probability of transition from $s$ to $s'$ given action $a$ \\
$\vv : S \to \reals$  & value function \\
$Q : S \times A \to \reals$  & $Q$ function \\
$\pi : S \to  A $  & policy 
\end{tabular}
\endgroup

\subsection*{Optimization}

\begingroup
\renewcommand{\arraystretch}{1.5}
\setlength{\tabcolsep}{10pt} 
\begin{tabular}{ll}
$\x$ & vectors in the decision set \\
 $\K$ & decision set \\
 $\nabla^k f $ & the $k$'th differential of  $f$; note $\nabla^k f \in \reals^{d^k}$\\
 $\nabla^{-2} f $ & the inverse Hessian of  $f$ \\
 $\nabla f $ & the gradient of $f$ \\
 $\nabla_t$ & the gradient of $f$ at point $\x_t$ \\
 $\x^\star$ & the global or local optima of objective $f$ \\
 $h_t$ & objective value distance to optimality, $h_t = f(\x_t) - f(\x^\star)$ \\
 $d_t$ & Euclidean distance to optimality $d_t =  \|\x_t - \x^\star\| $ \\
 $G$ & upper bound on norm of subgradients \\
 $D$ & upper bound on Euclidean diameter \\
 $D_p,G_p$ & upper bound on the $p$-norm of the subgradients/diameter 
\end{tabular}
\endgroup

\tableofcontents
\mainmatter

\part{Background in Control and Reinforcement Learning}

\chapter{Introduction}

\section{What is This Book About? }

Control theory is the engineering science of manipulating physical systems to achieve the desired functionality. Its history is centuries old and spans rich areas from the mathematical sciences. A brief list includes differential equations, operator theory and linear algebra, functional analysis, harmonic analysis, and more.  
However, it mostly does not concern modern algorithmic theory or computational complexity from computer science. 

The advance of the deep learning revolution has brought to light the power and robustness of algorithms based on gradient methods.  The following quote is attributed to deep learning pioneer Yann Lecun during a Barbados workshop on deep learning theory circa February 2020. 
\begin{displayquote}
``Control = reinforcement learning with gradients.'' 
\end{displayquote}

This book concerns this exact observation: {\it How can we exploit the differentiable structure in natural physical environments to create efficient and robust gradient-based algorithms for control? } 

The answer we give is {\bf not} to apply gradient descent to any existing algorithm. Rather, we rethink the paradigm of control from the start. What can be achieved in a given dynamical system, whether known or unknown, fully observed or not, linear or nonlinear, and without assumptions on the noise sequence? 

This is a deep and difficult question that we fall short of answering. However, we do propose a different way of answering this question: we define the control problem from the point of view of an online player in a repeated game. This gives rise to a simple but powerful framework called nonstochastic control theory. The framework is a natural fit to applying techniques from online convex optimization, resulting in new, efficient, gradient-based control methods. 

\section{The Origins of Control}

The industrial revolution was in many ways shaped by the steam engine that came to prominence in the eighteenth century. One of its most critical components was introduced by the Scottish inventor James Watt: the {\it centrifugal governor}, which is amongst the earliest examples of a control mechanism. A major source of efficiency losses of the earlier designs was the difficulty in regulating the engine. Too much or too little steam could easily cause engine failure. The centrifugal governor was created to solve this problem. 

\begin{figure}[h]
\centering 
\includegraphics[scale=0.1]{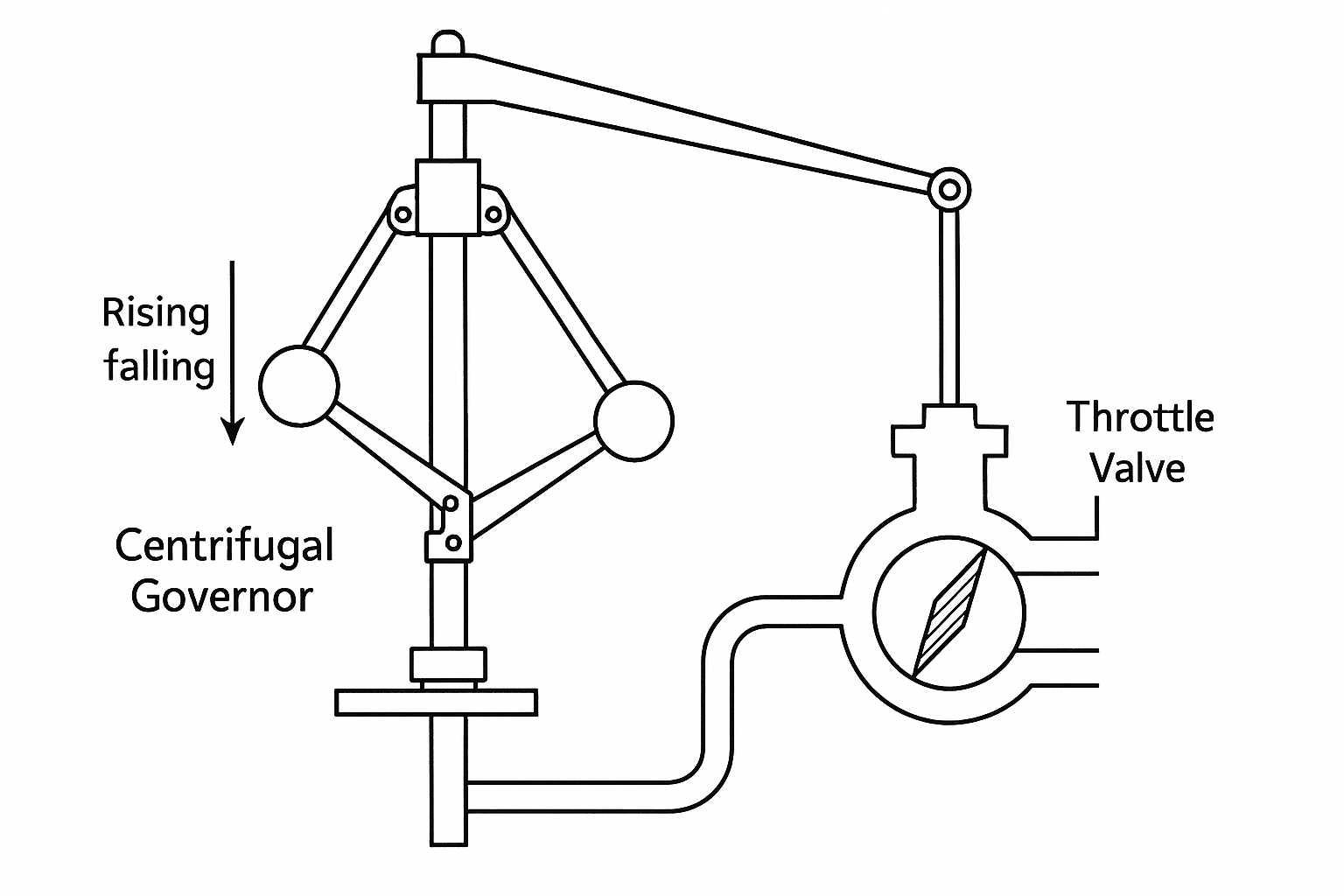}
\caption{A centrifugal governor.}
\end{figure}

The centrifugal governor consists of a shaft connected to two metal balls that are allowed to rotate. The increased shaft speed causes the balls to spin faster and the resulting centrifugal force causes the balls to lift. The balls are in turn attached to a pin that controls the amount of steam entering the engine, so as the balls spin faster, the pin lifts and limits the amount of fuel that can enter the system. However, if the balls spin slower, the pin is lowered, allowing more steam to enter the system. This provides a natural, direct, and proportional control to the system.

The impact of the centrifugal governor on the steam engine, and in turn on the industrial revolution, illustrates the importance of the field of control in engineering and the sciences. Since the eighteenth century, our world has changed quite a bit, and the importance of control has only grown. 

The challenges of controlling systems have moved from the steam engine to the design of aircraft, robotics, and autonomous vehicles. The importance of designing efficient and robust methods remains as crucial as in the industrial revolution. 

\section{Formalization and Examples of a Control Problem}

With the physical example of the centrifugal governor in mind, we give a mathematical definition of a control problem.  
The first step is to formally define a dynamical system in the following way. 

\begin{definition}[Dynamical system] \label{def:dynamical-system-general}
A dynamical system evolving from an initial state $\x_1$ is given by the following equations:
$$ \x_{t+1} = f_t(\x_t, \uv_t, \w_t), \quad \y_t = g_t(\x_t). $$
Here  $\x_t \in \reals^{d_x} $ is the state of the system, $\y_t \in \reals^{d_y} $ is the observation,  $\uv_t \in \reals^{d_u}$ is the control input, and $\w_t \in \reals^{d_w}$ is the perturbation. 

The function $g_t$ is a mapping from the state to the observed state, and the function $f_t$ is the transition function, given the current state $\x_t$, the input control $\uv_t$, and the perturbation $\w_t$. The subscript indicates the relevant quantities at time step $t$. 
\end{definition}

Given a dynamical system, there are several possible objectives that we can formulate. We give a generic formulation below that captures many common control formulations, notably the optimal control problem and the robust control problem. These will be addressed in much more detail later in this book. 
In this problem formulation, the goal is to minimize the long-term horizon cost given by a sequence of cost functions. 

\begin{definition}[A generic control problem] \label{def:generic-control}
Given a dynamical system with starting state $(\x_1,\y_1)$,  iteratively construct controls $u_t$ to minimize a sequence of loss functions $c_t(\y_t, \uv_t)$, $c_t \colon \reals^{d_y + d_u} \to \reals$. This is given by the following mathematical program:
\begin{align*}
& \min_{\uv_{1:T} }  \sum_t c_t(\y_t, \uv_t) \\
 s.t. \ \  & \x_{t+1} = f_t(\x_t, \uv_t, \w_t), \quad \y_t = g_t(\x_t) .
\end{align*}
\end{definition}

At this point, many aspects of the generic problem have not been specified. These include:
\begin{itemize}
    \item Are the transition functions and observation functions known or unknown to the controller?
    \item How accurately is the transition function known?
    \item What is the mechanism that generates the perturbations $\w_t$?
    \item Are the cost functions known or unknown ahead of time?
\end{itemize}

Different theories of control differ by the answers they give to the above questions, and these in turn give rise to different methods. 

Instead of postulating a hypothesis about how our world should behave, we proceed to give an intuitive example of a control problem. We can then reason about the different questions and motivate the nonstochastic control setting.

\subsection{Example: Control of a Medical Ventilator}
A recent relevant example of control is in the case of medical ventilators. The ventilator connects the trachea in the lungs to a central pressure chamber, which helps maintain positive-end expiratory pressure (PEEP) to prevent the lungs from collapsing and simulate healthy breathing in patients with respiratory problems. 

The ventilator must take into account the structure of the lung to determine the optimal pressure to induce. Such structural factors include \textit{compliance}, or the change in lung volume per unit pressure, and \textit{resistance}, or the change in pressure per unit flow. 

A simplistic formalization of the dynamical system of the ventilator and lung can be derived from the physics of a connected two-balloon system. The dynamics equations can be written as
\begin{align*} &p_t = C_0 + C_1 v_t^{-1/3}+ C_2 v_t^{5/3}, \\ &v_{t+1} = v_t + \Delta_t u_t, \end{align*}
where $p_t$ is the observed pressure, $v_t$ is the volume of the lung, $u_t$ is the input flow (control) and $\Delta_t$ is the unit of time resolution of the device. $C_0$, $C_1$, and $C_2$ are constants that depend on the particular system/lung-ventilator pair. 

The goal of ventilator control is to regulate the pressure according to some predetermined pattern. As a dynamical system, we can write $v_{t+1} = f(v_t, u_t, w_t)$, $p_t = g(v_t)$, and we can define the cost function to be $c_t(p_t, u_t) = |p_t - p_t^*|^2,$ where $p_t^*$ is the desired pressure at time $t$. The goal, then, is to choose a sequence of control signals $\{ u_t\}_{t=1}^T$ to  minimize the total cost over $T$ iterations. 

\begin{figure}[h]
\centering 
\includegraphics[scale=0.6]{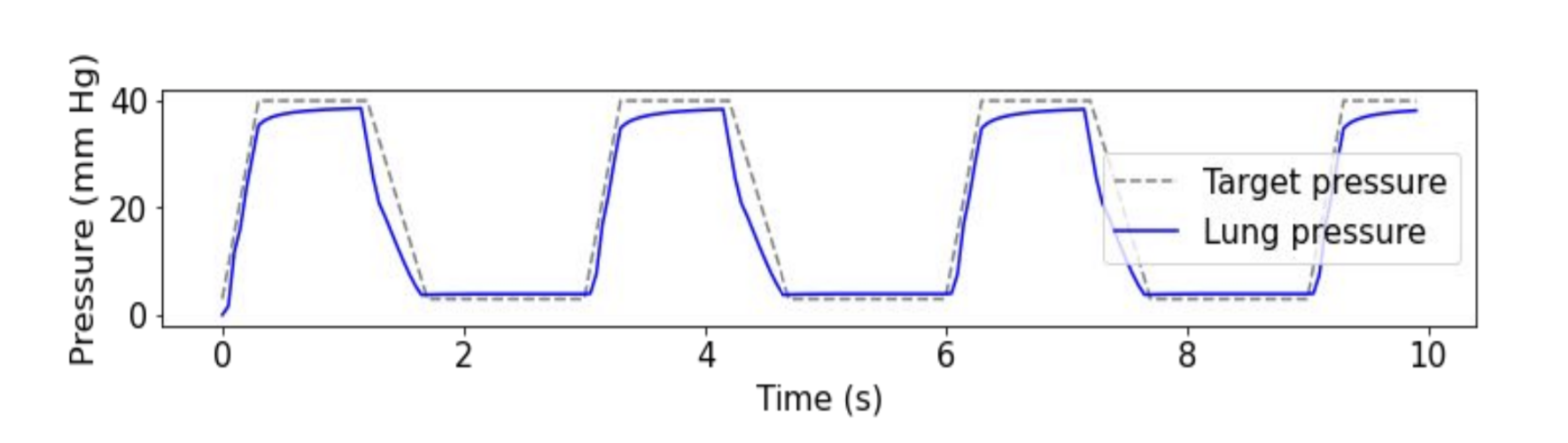}
\caption{Performance of the PID controller on a mechanical ventilator, from \citet{suo2021machine}.}
\label{fig:vent_pressure}
\end{figure}

In figure \ref{fig:vent_pressure} we see an illustration of a simple controller, which we will soon describe, called the PID controller, as it performs on a ventilator and lung simulator. 

\section{Simple Control Algorithms}

The previous section detailed the importance of control problems, specifically through the example of a critical care medical device. In this section, we detail methods for controlling a medical ventilator, or indeed any physical device that can be modeled as a dynamical system. 

The following methods are simplistic, but still widely used. For example, the PID algorithm is still used in most medical ventilators today. In later sections, we describe a modern optimization-based framework that yields more sophisticated algorithms that are capable of better performance.

\subsection{The Bang-Bang Controller}
Bang-bang control is one of the simplest control strategies. Consider a scalar dynamical system, where it is desirable to keep the state $x_t$ within a prespecified dynamic range $[x_{\min}, x_{\max}]$ (which can vary over time) and where the scalar control input $u_t$ must at all times be within the interval $[u_{\min}, u_{\max}]$, which is assumed to contain zero for simplicity. A bang-bang controller chooses the control input at time $t$ as
$$ u_t = \bone_{\{ x_t<x_{\min} \}} \times u_{\max} + \bone_{\{ x_t>x_{\max} \}} \times u_{min} . $$
A bang-bang controller saturates its inputs to extreme values in an attempt to confine the state of the system to a desirable range. For example, in the context of mechanical ventilation, one could choose $[p_t^*-\Delta p, p_t^* + \Delta p]$ as a reasonable region around the prescribed pressure $p^*_t$ as a desirable range for $p_t$. In such a case, a bang-bang controller would allow oxygen to enter at the highest possible rate whenever the pressure drops below $p^*_t-\Delta p$.

This discussion highlights two unsatisfactory aspects of the bang-bang control scheme: one, it ignores the cost associated with acting with the extreme values of the control input; two, it produces oscillatory and rapidly changing behavior in the control state.

\subsection{The PID Controller}

A more sophisticated control algorithm is the PID control method. 
Consider a control method that chooses the control to be a linear function of the observed states, that is: 
$$ \uv_t = \sum_{i=0}^k {A}_i \x_{t-i} ,$$ 
for some coefficients $A_i$, where we consider only the $k+1$ most recent states.
One of the most useful special cases of this linear family is the PID controller. 

A special subfamily of linear controllers is that of linear controllers that optimize on a predefined basis of historical observations. More precisely, 

\begin{enumerate}
	\item Proportional Control (e.g. Centrifugal Governor): $\uv_t = \alpha_0 \x_t.$
	\item Integral Control: $\uv_t = \beta \sum_{i < t} \x_{t-i}.$
	\item Derivative Control: $\uv_t = \gamma (\x_t - \x_{t-1}).$
\end{enumerate}

The PID controller, named after the nature of the three components that make up it, specifies three coefficients $\alpha,\beta,\gamma$ to generate a control signal. In that sense, it is extremely sparse, which from a learning-theoretic perspective is a good indication of generalization. 

In fact, the PID controller is the method of choice for numerous engineering applications, including control of medical ventilators.

\section{Classical Theory: Optimal and Robust Control}

Classical theories of control differ within the generic control problem formulation \ref{def:generic-control} primarily in one significant aspect: the perturbation model. A rough divide between the two main theories -- {\it optimal control} and {\it robust control} -- is as follows. 

Optimal control postulates a probabilistic model for the perturbations, which are typically i.i.d. draws from a Gaussian distribution. However, other types of distribution can also be considered. As such, the control problem can be rephrased in a more precise stochastic mathematical program,
\begin{align*}
& \min_{\uv_{1:T} } \E_{\w_{1:T}} \left[  \sum_t c_t(\y_t, \uv_t) \right] \\
 s.t. \ \  & \x_{t+1} = f_t(\x_t, \uv_t, \w_t), \quad \y_t = g(\x_t)  .
\end{align*}

In contrast, robust control theory plans for the worst-case noise from a given set of constraints, denoted $\K$.  
Thus, the robust control problem can be rephrased as the following mathematical program,
\begin{align*}
& \min_{\uv_{1:T} } \max_{\w_{1:T} \in \K} \left[  \sum_t c_t(\y_t, \uv_t) \right] \\
 s.t. \ \  & \x_{t+1} = f_t(\x_t, \uv_t, \w_t), \quad \y_t = g(\x_t)  .
\end{align*}

The two formulations have been debated for decades, with pros and cons for each side. 
Optimal control focuses on the average case and often yields simpler formulations for optimization, resulting in more efficient methods. Robust control, in contrast, allows for more difficult noise models and some amount of model misspecification. However, it is too pessimistic for many natural problems.

Nonstochastic control theory takes the best of both worlds, simultaneously granting robustness to adversarial noise and allowing the use of optimistic behavior when appropriate while providing computationally efficient methods.

\section{The Need for a New Theory}

Consider the problem of flying a drone from the source to the destination subject to unknown weather conditions. 
The aerodynamics of flight can be modeled sufficiently well by time-varying linear dynamical systems, and existing techniques are perfectly capable of doing a great job for indoor flight. However, the wind conditions, rain, and other uncertainties are a different story. Certainly, the wind is not an i.i.d. Gaussian random variable! Optimal control theory, which assumes this zero-mean noise, is therefore overly optimistic and inaccurate for our problem. 

The designer might resort to robust control theory to account for all possible wind conditions, but this is overly pessimistic. What if we encounter a bright sunny day after all? Planning for the worst case would mean slow and fuel-inefficient flight. 

We would like an adaptive control theory that allows us to attain the best of both worlds: an efficient and fast flight when the weather permits, and a careful drone maneuvering when this is required by the conditions. Can we design a control theory that will allow us to take into account the specific instance perturbations and misspecifications, and give us finite-time provable guarantees?  This is the subject of online nonstochastic control!

\subsection{Online Nonstochastic Control Theory}

This book is concerned with nonstochastic noise. When dealing with non-stochastic, that is, arbitrary or even adversarial perturbations, the optimal policy, that is, a decision making rule mapping observations to control inputs,  is not clear a priori. Rather, an optimal policy for the observed perturbations can be determined {\it only in hindsight}.  This is a significant difference from optimal and robust control theory, where the mapping from states to actions can be completely predetermined based upon the noise model. 

Since optimality is defined only in hindsight, the goal is not predetermined. We thus shift to a different performance metric borrowed from the world of learning in repeated games, namely regret. Specifically, we consider regret with respect to a reference class of policies. In online control, the controller iteratively chooses a control $\uv_t$. The controller then observes the next state of the system $\x_{t+1}$ and suffers a loss of $c_t(\x_t,\uv_t)$ according to an adversarially chosen loss function. For simplicity,  we consider the case of full state observation where $\y_t = \x_t$. 
Let $\Pi = \{ \pi \colon \x \to \uv \} $ be a class of policies. The regret of the controller with respect to $\Pi$ is defined as follows. 
\begin{definition} \label{defn:regret}
Given a dynamical system, the regret of an online control algorithm $\mathcal{A}$ over $T$ iterations with respect to a class of policies $\Pi$ is given by:
\begin{align*}
\regret_T(\mA,\Pi) = \max_{\mathbf{w}_{1:T}: \norm{\mathbf{w}_t} \le 1} &\left(\sum_{t=1}^{T} c_t (\mathbf{x}_t, \mathbf{u}_t) - \min_{\pi \in \Pi} \sum_{t=1}^{T} c_t (\x_t^\pi,  \uv_t^\pi) \right) ,
\end{align*}
where $\uv_t = \mA(\x_t)$ are the controls generated by $\mA$, and $\x_t^\pi,\uv_t^\pi$ are the counterfactual state sequence and controls under the policy $\pi$, that is,
\begin{align*}
    \uv_t^\pi & = \pi(\x_t^\pi) \\ 
    \x_{t+1}^\pi & = f_t(\x_t^\pi, \uv_t^\pi, \mathbf{w}_t) .
\end{align*}
\end{definition}

Henceforth, if $T,\Pi$ and $\mA$ are clear from the context, we drop these symbols when defining $\regret$.

At this point, the reader may wonder why we should compare to a reference class, as opposed to all possible policies. There are two main answers to this question: 
\begin{enumerate}
\item 
First, the best policy in hindsight out of all possible policies may be very complicated to describe and to reason about. This argument was made by the mathematician Tyrell Rockafellar; see the bibliographic section for more details. 
\item
Secondly, it can be shown that it is impossible, in general, to obtain sublinear regret versus the best policy in hindsight. Rather, a different performance metric called {\it competitive ratio} can be analyzed, as we discuss later in the book. 
\end{enumerate}

The non-stochastic control problem can now be stated as finding an efficient algorithm that minimizes the worst-case regret vs. meaningful policy classes that we examine henceforth. 
It is important to mention that {\it the algorithm does not have to belong to the comparator set of policies!} Indeed, in the most powerful results, we will see that the algorithm will learn and operate over a policy class that is strictly larger than the comparator class.

\subsection{A New  Family of Algorithms } \label{sec:intro-gpc}

The policy regret minimization viewpoint in control leads to a new approach to algorithm design. Instead of a priori computation of the optimal controls, online convex optimization suggests modifying the controller according to the costs and dynamics, in the flavor of adaptive control, in order to achieve provable bounds on the policy regret. 

As a representative example, consider the {\it Gradient Perturbation Controller} described in Algorithm \ref{algo:GPC-generic}. The algorithm maintains matrices that are used to create the control as a linear function of past perturbations, according to line \ref{algline:control-gpc}. These parameters change over time according to a gradient-based update rule, where the gradient is calculated as the derivative of a counterfactual cost function with respect to the parametrization, according to line \ref{algline:grad-gpc}.  

\begin{algorithm}[h]
\begin{algorithmic}[1] 
\STATE {Input:}  $h$, $\eta$, initial parameters ${M}_{1:h}^1$, dynamics $f$.
\FOR{$t$ = $1 \ldots T$}
        \STATE \label{algline:control-gpc} $\mbox{Use control } \mathbf{u}_t = \sum_{i=1}^{h} {{M}}_i^t \mathbf{w}_{t-i} $
        \STATE  Observe state $\mathbf{x}_{t+1}$, compute perturbation $\mathbf{w}_t = \mathbf{x}_{t+1} - f(\x_t,\uv_t) $
        \STATE  \label{algline:grad-gpc} Construct loss $\ell_t({M}_{1:h}) = c_t(\mathbf{x}_t({M}_{1:h}), \mathbf{u}_t({M}_{1:h}))$\\
       Update ${M}_{1:h}^{t+1} \leftarrow  {M}_{1:h}^t -\eta \nabla \ell_t({M}_{1:h}^t)$
\ENDFOR
 \caption{Gradient Perturbation Controller(GPC) - simplified version}
 \label{algo:GPC-generic}
\end{algorithmic}
\end{algorithm}

The version given here assumes that the time-invariant dynamics $f$ are known. The GPC computes a control that is a linear function of past perturbations. It modifies the linear parameters according to the loss function via an iterative gradient method. In Chapter \ref{chap:gpc}, we prove that this algorithm attains sublinear regret bounds for a large class of time-varying and time-invariant dynamical systems. In later chapters, we explore extensions to nonlinear dynamical systems and unknown systems, which lie at the heart of nonstochastic control theory.

\ifarxiv
\newpage
\fi
\section{Bibliographic Remarks}
Control theory is a mature and active discipline with both deep mathematical foundations and a rich history of applications. In terms of application, self-regulating feedback mechanisms that limit the flow of water, much like the modern float valve, were reportedly used in ancient Greece over two thousand years ago. The first mathematical analysis of such systems appeared in a paper by James Clerk Maxwell on the topic of governors \citep{wellmax}, shortly after Maxwell had published his treatise on the unification of electricity and magnetism. For an enlightening historical account, see \citet{fernandez2003control}. 

For a historical account on the development of the PID controller, see \citet{248006}. The theoretical and practical properties of this controller are described in detail in \citet{aastrom1995pid}.

There are many excellent textbooks on control theory, which we point to in later chapters. The book of \citet{bertsekas2007dynamic} offers extensive coverage of topics from the point of view of dynamic programming.
The texts \citet{stengel1994optimal, zhou1996robust}, in contrast, delve deeply into results on linear control, the most well-understood branch of control theory. 
Rockafellar \citet{rockafellar1987linear} proposed the use of convex cost functions to model the cost in a linear dynamical system to incorporate constraints on state and control and suggested computational difficulties with this approach. 
The text \citet{slotine1991applied} gives a thorough introduction to adaptive control theory.

Online nonstochastic control theory and the Gradient Perturbation Control algorithm were proposed in \citet{agarwal2019online}.  A flurry of novel methods and regret bounds for online nonstochastic control have appeared in recent years, many of which are surveyed in \cite{hazan2021tutorial}. Notable among them are logarithmic regret algorithms \citep{agarwal2019logarithmic,foster2020logarithmic,simchowitz2020making}, nonstochastic control for unknown systems \citep{hazan2019nonstochastic}, black-box online nonstochastic control \citep{chen2021black}, and nonstochastic control with partial observability \citep{simchowitz2020improper}.  

The theory is heavily based on recent developments in decision making and online learning, notably the framework of online convex optimization \citep{hazan2016introduction}, and especially its extensions to learning with memory \citep{anava2013online}.  

\chapter{Dynamical Systems} \label{chap:dynamical_systems}

In this chapter, we return to the general problem of control from Definition \ref{def:generic-control}: choosing a sequence of controls to minimize a sequence of cost functions. Before moving on to algorithms for control, it is useful to ask what kinds of goals are natural and what kinds of guarantees are even possible for a dynamical system.

To that end, we consider several examples of dynamical systems and introduce a few basic solution concepts, including equilibrium, stability, stabilizability, and controllability. These examples also illustrate the kinds of observation models and control objectives that arise in real applications.

\section{Examples of Dynamical Systems}

\subsection{Medical Ventilator} \label{subsec:ventilator}
In the task of mechanical ventilation, the control algorithm dictates the amount of oxygen inflow $u_t$ (in $\mathrm{m^3/s}$) into the lungs of a patient. This inflow approximately determines the volume of the lungs $v_t$ via the simple relation
\[
v_{t+1} = v_t + \Delta u_t,
\]
where $\Delta$ is the time elapsed between successive actions of the discrete-time control algorithm. Being difficult to directly measure or estimate, the quantity $v_t$ is a hidden or latent variable from the controller's perspective. Instead, the controller observes the air pressure $p_t$. A simplified physical model relating pressure and volume is
\[
p_t = C_0 + C_1 v_t^{-1/3} + C_2 v_t^{5/3}.
\]
This relation can be derived from stress--strain considerations by modeling the left and right lungs, connected by the trachea, as in a two-balloon experiment.

\begin{figure}[ht]
	\centering
	\includegraphics[scale=0.5]{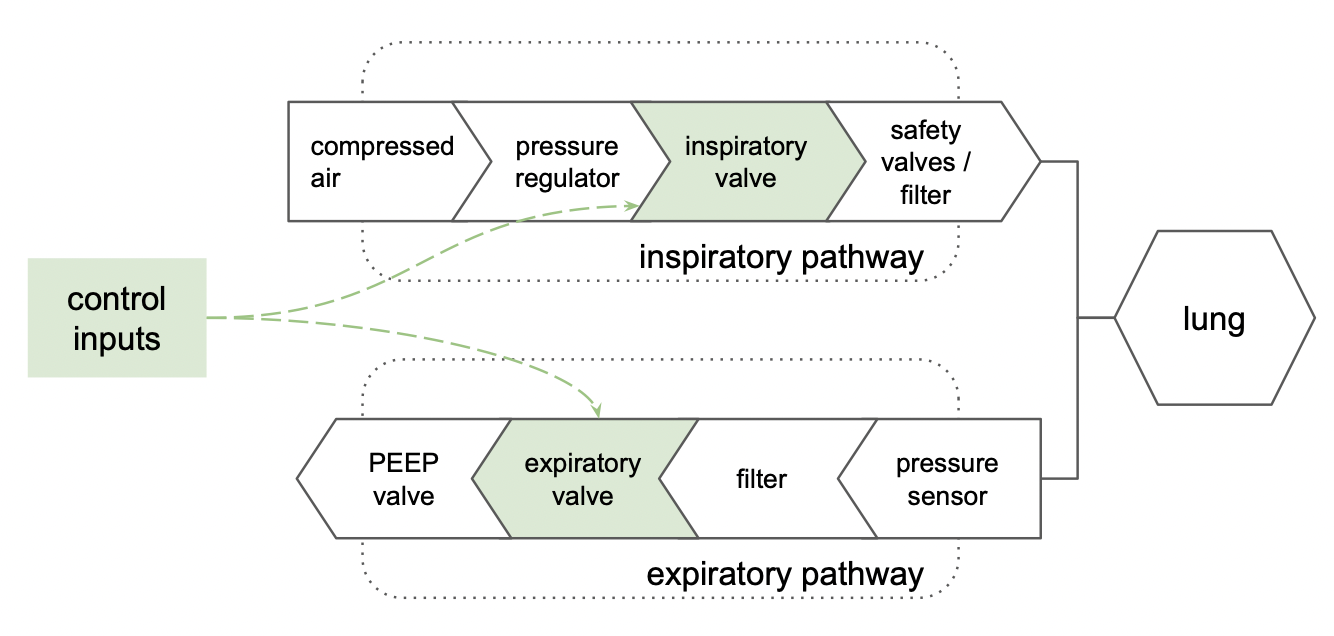}
	\caption{A schematic of the respiratory circuit from \citet{suo2021machine}. \label{fig:vent_scheme}}
\end{figure}

The objective of the control algorithm is to follow an ideal pressure trajectory $p_t^\star$ prescribed by the physician as closely as possible, for example via the stage cost
\[
c_t(p_t, u_t) = |p_t-p_t^\star|^2 + |u_t|^2,
\]
while penalizing large airflow rates so as to limit patient discomfort. See Figure~\ref{fig:vent_scheme} for a schematic representation of airflow when a lung is connected to a mechanical ventilator.

\subsection{Double Integrator} \label{section:double-integrator}

The double integrator is a common example of a {\em linear} dynamical system. In its simplest form, it models a one-dimensional point object moving along the real line. The state $\x_t\in\reals^2$ is two-dimensional, with the first coordinate indicating position and the second denoting velocity. The control algorithm chooses a scalar actuation $u_t\in\reals$ that directly affects the velocity coordinate at each time step; after an appropriate normalization, $u_t$ may be interpreted as a force or acceleration input applied over one sampling interval. The dynamics equation can be written as
\begin{align*}
	\x_{t+1} = \begin{bmatrix}
		1 & \Delta \\ 0 & 1
	\end{bmatrix}\x_t + \begin{bmatrix}
		0 \\ 1
	\end{bmatrix} u_t + \w_t .
\end{align*}
The position is linearly related to the velocity through the sampling period $\Delta$. Optionally, the state is also affected by external perturbations $\w_t$ outside the influence of the controller.

The goal here is to ensure that the point object arrives and then comes to rest at some coordinate $x^\star\in \mathbb{R}$ as quickly as possible, while also not expending too much energy. A natural stage cost is
\[
c(\x_t, u_t) = \left\|\x_t - \begin{bmatrix}
	x^\star \\ 0
\end{bmatrix}\right\|^2 + u_t^2.
\]

\begin{figure}
\begin{center}
\begin{tikzpicture}[thick,>=Stealth,scale=0.75]

\draw[thick] (-1,0) -- (7,0);

\node[draw, rectangle, minimum width=1.5cm, minimum height=1cm, fill=blue!10] (block) at (3,0.75) {$m$};

\draw[<->, thick] (0,-0.6) -- (3,-0.6);
\node[fill=white] at (1.7,-0.6) {$x_1(t)$ (position)};

\draw[->, very thick, blue] (3,1.5) -- (4.5,1.5);
\node[blue,above] at (3.75,1.5) {$x_2(t)$ (velocity)};

\draw[->, very thick, red] (4,0.75) -- (5.5,0.75);
\node[red,above] at (4.75,0.75) {$u(t)$};
\end{tikzpicture}
\end{center}
\caption{Double integrator illustration showing state coordinate $x_1(t)$ (position), state coordinate $x_2(t)$ (velocity), and control input $u(t)$.}
\end{figure}

\subsection{Pendulum Swing-Up} \label{subsec:inv_pend}
The swing-up pendulum is one of the simplest examples of rotational motion. It consists of a point mass $m$ attached to the end of a massless rod of length $l$, with the other end fixed at a pivot. The system is subject to gravity with acceleration $g$, and the controller applies a torque $u_t\in\mathbb{R}$ at each time step. If $\theta_t$ denotes the angle of the rod measured from the downward vertical, then a discretization of the continuous-time dynamics gives
\begin{align*}
    \theta_{t+1} &= \theta_t + \Delta \dot{\theta}_t \ \ , \ \  \  \dot{\theta}_{t+1} = \dot{\theta}_t + \Delta\left(\frac{u_t}{ml^2} - \frac{g}{l}\sin\theta_t\right).
\end{align*}
Thus the state is $\x_t = [\theta_t,\dot{\theta}_t]^\top$, and the dynamics are nonlinear because of the term $\sin\theta_t$.

\begin{figure}
\begin{center}
\begin{tikzpicture}[thick,>=Stealth,scale=0.7]
\def\rodlength{4}
\def\massradius{0.2}
\def\angle{45}

\filldraw[fill=black] (0,0) circle (0.05);
\node[above left] at (0,0) {Pivot};

\draw[thick] (0,0) -- (\angle:\rodlength);

\draw[fill=blue!20] (\angle:\rodlength) circle (\massradius);
\node[right] at (\angle:\rodlength) {$m$};

\draw[dashed] (0,0) -- (0,-\rodlength-0.5);

\draw[->] (0,-1) arc[start angle=-90,end angle=\angle-90,radius=1.5];
\node[fill=white] at ({(\angle/2)-90}:1.5) {Angle $\theta_t$};

\draw[->,red,thick] (\angle:\rodlength) -- +(0,-1) node[midway,right] {$m \mathbf{g}$};

\draw[->,thick,green!70!black] (0.7,0) arc[start angle=0,end angle=\angle-20,radius=1.3];
\node[green!70!black] at ({(\angle/2)-10}:1.1) {$u_t$};

\node[fill=white] at ({\angle}:\rodlength/2) {$l$};
\end{tikzpicture}
\end{center}
\caption{Pendulum swing-up illustration with marked angle $\theta_t$, gravitational force $mg$, control torque $u_t$, mass $m$, and rod length $l$.}
\end{figure}

The objective here is to bring the rod to rest in the upright position $\theta=\pi$, using as little torque as possible. A simple stage cost is
\[
c(\x_t, u_t) = \left\|\x_t - \begin{bmatrix}
	\pi \\ 0
\end{bmatrix}\right\|^2 + u_t^2.
\]
When interpreting this objective globally, the angular error should be understood modulo $2\pi$.

\subsection{Autoregressive Language Models}

A modern example of a dynamical system is an autoregressive language model. Let $\mathcal{V}$ be a finite vocabulary, and let $y_t \in \mathcal{V}$ denote the token observed at time $t$. The model maintains a hidden state $h_t$ that summarizes the preceding context. For a recurrent neural network, one may take $h_t \in \mathbb{R}^d$ to be a fixed-dimensional latent vector. For a transformer, it is more natural to regard $h_t$ as the model's internal cache or prefix representation associated with the sequence $y_{1:t}$; this state is typically very high-dimensional and is not directly observed.

A simple state-space description is
\[
h_{t+1} = f_\theta(h_t, e(y_t)),
\]
where $e(y_t)$ is the embedding of token $y_t$ and $f_\theta$ is a learned nonlinear update map. From the hidden state, the model produces a probability distribution over the next token,
\[
p_{t+1} = g_\theta(h_{t+1}) \in \Delta(\mathcal{V}),
\]
where $\Delta(\mathcal{V})$ denotes the probability simplex over the vocabulary.

The objective in next-token prediction is to assign high probability to the realized next token $y_{t+1}$. A natural stage cost is therefore the log loss
\[
c_t(h_t,y_{t+1}) = -\log p_{t+1}(y_{t+1}).
\]
Equivalently, one may view training as minimizing the sum of these stage costs over time.

This example illustrates several features that are common in modern machine learning systems. The hidden state is extremely large and latent, the observations are discrete symbols rather than physical measurements, and the dynamics are highly nonlinear. At generation time, the model recursively predicts a distribution over the next token, chooses or samples a token from that distribution, appends it to the context, and repeats the process.

\subsection{Epidemiological Models}
The COVID-19 pandemic renewed public interest in policy interventions for controlling the spread of infectious diseases. A basic compartmental model is the SIR model. Let $S_t$, $I_t$, and $R_t$ denote the fractions of the population that are susceptible, infected, and removed (recovered or immune), respectively, so that $S_t+I_t+R_t=1$. Let $u_t\ge 0$ denote the vaccination effort at time $t$, and let $\alpha\in[0,1]$ denote vaccine efficacy. A discrete-time controlled SIR model is
\begin{align*}
	S_{t+1} &= S_t - \beta S_t I_t - \alpha u_t,\\
	I_{t+1} &= I_t + \beta S_t I_t - \gamma I_t, \\
	R_{t+1} &= R_t + \gamma I_t + \alpha u_t.
\end{align*}
Here, $\beta$ measures transmissibility, $\gamma$ is the recovery rate, and $\alpha u_t$ is the fraction of susceptible individuals effectively moved into the removed/immune compartment by vaccination. Under these dynamics, the identity $S_t+I_t+R_t=1$ is preserved whenever it holds initially.

A natural control objective is to keep infections small while also limiting intervention cost, for example through the stage cost
\[
c_t(S_t,I_t,R_t,u_t) = I_t^2 + \lambda u_t^2,
\]
for some tradeoff parameter $\lambda>0$, subject to $u_t\ge 0$ and $\alpha u_t \le S_t$.

\begin{figure}[ht]
	\centering
	\includegraphics[scale=0.3]{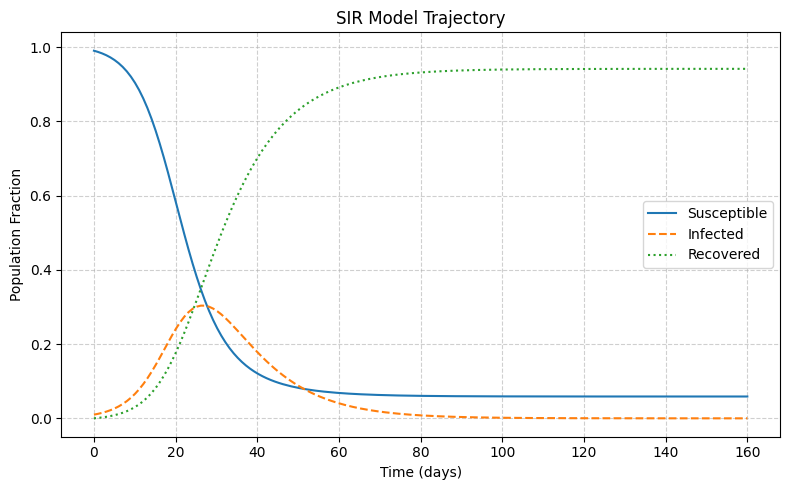}
	\caption{Typical trajectory of a continuous-time SIR model without vaccination, with parameters $\beta = 0.3$ and $\gamma = 0.1$.}
\end{figure}

These examples span hidden-state and fully observed systems, linear and nonlinear dynamics, and a variety of natural control objectives. We next abstract away from the application details and formalize several general solution concepts.

\section{Solution Concepts for Dynamical Systems}

Before considering how to control a dynamical system, we should decide what success means. In this section, we discuss several solution concepts, including equilibrium, stability, stabilizability, and controllability.

Given a dynamical system $\x_{t+1} = f(\x_t,\uv_t,\w_t)$, its deterministic counterpart
\[
\x_{t+1} = f(\x_t,\uv_t) = f(\x_t,\uv_t,\mathbf{0})
\]
models the evolution of the system in the absence of perturbations. The autonomous system
\[
\x_{t+1} = f(\x_t) = f(\x_t,\mathbf{0},\mathbf{0})
\]
describes the state transitions in the absence of both control inputs and perturbations. In both cases, we use the same symbol $f$ when the omitted arguments are understood to be zero.

\subsection{Equilibrium of Dynamical Systems}

An example of an equilibrium is a ball sitting at rest at the bottom of a pit. If the ball is moved slightly, it returns to the bottom of the pit. Thus this equilibrium is both stable and attractive.

If a ball is at rest on the top of a mountain, then pushing it in any direction causes it to roll away. This is an equilibrium that is unstable. We now formalize these notions.

\begin{definition}
A point $\x_e$ is an \textbf{equilibrium point} of an autonomous dynamical system $\x_{t+1}=f(\x_t)$ if and only if $f(\x_e)=\x_e$.

The equilibrium $\x_e$ is \textbf{stable} if for every $\varepsilon>0$ there exists $\delta>0$ such that
\[
\|\x_0-\x_e\|<\delta
\quad\Longrightarrow\quad
\|\x_t-\x_e\|<\varepsilon
\qquad \text{for all } t\ge 0.
\]

The equilibrium $\x_e$ is \textbf{attractive} if there exists $\delta>0$ such that
\[
\|\x_0-\x_e\|<\delta
\quad\Longrightarrow\quad
\lim_{t\to\infty}\x_t=\x_e.
\]

The equilibrium $\x_e$ is \textbf{asymptotically stable} if it is both stable and attractive. If the attraction property holds for every initial state in the state space, we say that the equilibrium is \textbf{globally attractive}; if it is also stable, we say that it is \textbf{globally asymptotically stable}.
\end{definition}

Another example is given by a linear transformation in Euclidean space. Consider the system
\[
\x_{t+1} = A \x_t .
\]
Clearly $\bzero$ is an equilibrium point, since $A\bzero=\bzero$ for every matrix $A$. If every eigenvalue of $A$ has modulus strictly less than one, then $A^t\x_0 \to \bzero$ for every initial state $\x_0$. Thus, $\bzero$ is a globally asymptotically stable equilibrium.

In general, determining or even describing equilibrium sets and long-run behavior can be difficult, as the examples later in this chapter illustrate.

\subsection{Stabilizability and Controllability}

Two fundamental properties of dynamical systems are stabilizability and controllability.

\begin{definition}
A dynamical system is \textbf{asymptotically stabilizable} if there exists a control policy $\pi$ such that, for every initial state, the deterministic closed-loop trajectory satisfies
\[
\lim_{t\to\infty}\x_t^\pi = \bzero.
\]
\end{definition}

\begin{definition}
A dynamical system is \textbf{controllable} if, for every initial state $\x_0$ and every target state $\x^\star$, there exist a finite horizon $T$ and controls $\uv_0,\ldots,\uv_{T-1}$ such that the deterministic counterpart of the system satisfies $\x_T=\x^\star$.
\end{definition}

Controllability implies finite-time reachability of the zero state. If the zero state is an equilibrium that can be maintained once reached, then controllability also implies asymptotic stabilizability.

Some texts reserve the word \emph{stabilizable} for a weaker algebraic property of linear systems. Here we use the more direct trajectory-based notion above, since it is convenient for our purposes.

In the next subsection, we introduce a more quantitative robust-boundedness notion that is useful in nonstochastic control.

\subsection{Quantitative Definitions of Stabilizability}

The notion below is stronger in one sense and weaker in another: it allows adversarial bounded disturbances, but it asks only that the state remain uniformly bounded rather than converge to zero. For linear systems, related ideas include robust invariance and input-to-state stability.

\begin{definition} \label{def:gamma-stabilizing}
A control policy $\pi$ for a dynamical system $f$ is \textbf{$\gamma$-stabilizing} if the following holds. For every initial state satisfying $\|\x_0^\pi\|\le 1$ and every disturbance sequence $\w_0,\ldots,\w_{T-1}$ such that $\|\w_t\|\le 1$ for all $t\in\{0,\ldots,T-1\}$, let $\x_t^\pi$ be the trajectory generated by policy $\pi$:
\[
\x_{t+1}^\pi = f(\x_t^\pi,\pi(\x_t^\pi),\w_t).
\]
Then
\[
\|\x_t^\pi\| \le \gamma
\qquad \text{for all } t=0,1,\ldots,T.
\]

A dynamical system is said to be \textbf{$\gamma$-stabilizable} if and only if it admits a $\gamma$-stabilizing control policy.
\end{definition}

\section{Complexity of Dynamical Systems and Hardness of Stabilizability}

Before formally proving the computational hardness of stabilizability, we begin with two examples that illustrate how complicated invariant sets and attractors can become even for simple dynamical rules.

\begin{figure}[h]
    \centering
    \begin{minipage}{0.45\textwidth}
        \centering
        \includegraphics[width=0.9\textwidth]{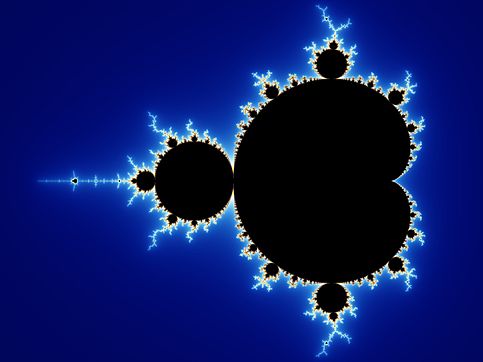}
        \caption{Mandelbrot set.}\label{fig:hard1}
    \end{minipage}\hfill
    \begin{minipage}{0.45\textwidth}
        \centering
        \includegraphics[width=0.68\textwidth]{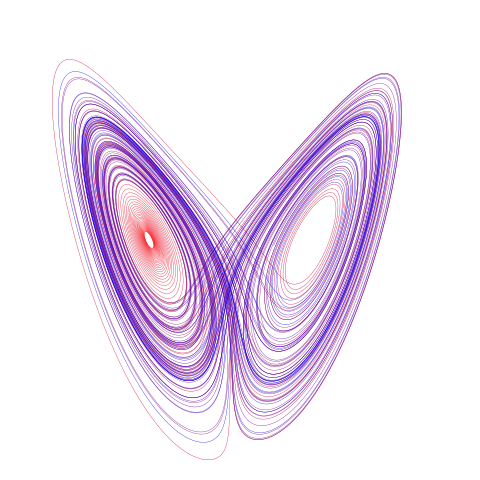}
        \caption{Lorenz attractor. Visualization by D Schwen, licensed under CC-BY-2.5.}\label{fig:hard2}
    \end{minipage}
\end{figure}

Consider the complex quadratic map
\[
z_{t+1} = f_c(z_t) = z_t^2+c.
\]
Its \emph{filled Julia set} is
\[
K_c = \left\{ z \in \mathbb{C} \ \colon\ \sup_{t\ge 0}|f_c^t(z)| < \infty \right\},
\]
where $f_c^t(z)$ denotes $t$ successive applications of the function $f_c$. The \emph{Julia set} is the boundary $J_c=\partial K_c$. Depending on the parameter $c$, these sets can have extremely intricate geometry, and there are strong negative results on the computability of Julia sets for certain values of $c$; see, for example, \citet{braverman2008computability}. A related parameter-space object is the Mandelbrot set, shown in Figure~\ref{fig:hard1}.

Another compelling visual illustration of complexity in dynamical systems is the Lorenz system, governed by the continuous-time equations
\begin{align*}
    \dot{x} &= \sigma(y-x),\\
    \dot{y} &= x(\rho-z)-y,\\
    \dot{z} &= xy-\beta z.
\end{align*}
For classical parameter values, trajectories are attracted to a complicated invariant set known as the Lorenz attractor, depicted in Figure~\ref{fig:hard2}. For background on this attractor, see \citet{Strogatz2014,tucker1999lorenz,Lorenz1963}. These examples show that even describing the long-run behavior of a seemingly simple dynamical system can be a formidable task.

\subsection{Computational Hardness of Stabilizability}

Consider the following decision problem: given a dynamical system, can we determine whether it is asymptotically stabilizable, that is, whether there exists a control policy that drives the system to the zero state from every initial condition? We now show that this decision problem is computationally hard.

\begin{theorem} \label{thm:hardness_of_stabilizability}
For the family of discrete-time polynomial dynamical systems with integer coefficients, deciding asymptotic stabilizability is NP-hard.
\end{theorem}

\begin{proof}
We reduce from \textsc{3SAT}. Let
\[
F(x_{1:n}) = c_1 \land c_2 \land \cdots \land c_m
\]
be a Boolean formula in conjunctive normal form with three literals per clause. Introduce real variables $u_1,\ldots,u_n$. For each literal $\ell$, define
\[
r(\ell)=
\begin{cases}
1-u_i, & \text{if } \ell=x_i,\\
u_i, & \text{if } \ell=\neg x_i.
\end{cases}
\]
For a clause $c=\ell_1 \lor \ell_2 \lor \ell_3$, let
\[
g_c(\uv) = r(\ell_1)\,r(\ell_2)\,r(\ell_3).
\]
On Boolean assignments $\uv\in\{0,1\}^n$, the quantity $g_c(\uv)$ equals $1$ exactly when the clause $c$ is false and $0$ otherwise.

Now define the polynomial
\[
p_F(\uv)
=
\sum_{i=1}^n \bigl(u_i(1-u_i)\bigr)^2
+
\sum_{j=1}^m g_{c_j}(\uv)^2.
\]
Then $p_F(\uv)=0$ if and only if $\uv\in\{0,1\}^n$ encodes a satisfying assignment of $F$. Indeed, the first sum vanishes if and only if each $u_i\in\{0,1\}$, and the second sum vanishes if and only if no clause is false. Hence deciding whether a degree-six polynomial with integer coefficients has a real zero is NP-hard.

Consider now the scalar dynamical system
\[
x_{t+1} = p_F(\uv_t).
\]
If $F$ is satisfiable, choose a satisfying assignment $\uv^\star$ and apply the constant control $\uv_t\equiv \uv^\star$. Then $p_F(\uv^\star)=0$, so $x_t=0$ for all $t\ge 1$, regardless of the initial state. Therefore the system is asymptotically stabilizable.

Conversely, suppose $F$ is unsatisfiable. Then $p_F(\uv)>0$ for all $\uv\in\reals^n$. Since $p_F$ is continuous and
\[
\sum_{i=1}^n \bigl(u_i(1-u_i)\bigr)^2 \to \infty
\qquad \text{as } \|\uv\|\to\infty,
\]
the function $p_F$ attains a strictly positive global minimum $m>0$. Hence, for every control sequence,
\[
x_{t+1} = p_F(\uv_t) \ge m
\qquad \text{for all } t\ge 0.
\]
So the state cannot converge to $0$. Therefore the system is asymptotically stabilizable if and only if $F$ is satisfiable, proving NP-hardness.
\end{proof}

The proof already shows hardness for an especially simple family of polynomial control systems. Exercise \ref{exercise:hardness_of_stabilizability} asks the reader to strengthen this result and show that deciding asymptotic stabilizability for polynomial systems of degree four is also NP-hard.

\ifarxiv
\newpage
\fi
\section{Bibliographic Remarks}
The text \citet{tedrake} offers several examples and characterizations of dynamical systems useful for robotic manipulation. A more mathematically involved introduction appears in \citet{Strogatz2014}.

\citet{adam2020special} discusses further aspects of simulations used in the early days of the COVID-19 pandemic. See \citet{suo2021machine} for examples of control algorithms used for noninvasive mechanical ventilation.

The Lorenz equations were introduced to model the phenomenon of atmospheric convection \citep{Lorenz1963}. For detailed information on the mathematical properties of the Lorenz attractor, see \citet{Strogatz2014,tucker1999lorenz}. For discussion of computability questions for Julia sets and related sets, see \citet{braverman2008computability}.

Nonlinear dynamical systems admit many stability notions beyond those introduced here, including Lyapunov stability, local and global asymptotic stability, robust invariance, and input-to-state stability. Lyapunov's direct method shows that suitable potential functions can certify stability without requiring explicit trajectory formulas; see \citet{lyapunov1992general,khalil2015nonlinear} for further discussion.

An excellent survey on the computational complexity of control is \citet{blondel2000survey}. For a more recent and stronger result on stabilizability, see \citet{ahmadi2012difficulty}. The NP-hardness theorem in this chapter appeared in \citet{aaac}. The computational complexity of mathematical programming---both constrained and unconstrained---is commented upon in \citet{hazan2019lecture}.

\ifarxiv
\newpage
\fi
\begin{exercises}

\begin{exercise} \label{exercise:hardness_of_stabilizability}
Strengthen Theorem \ref{thm:hardness_of_stabilizability} to show that deciding whether a dynamical system given by a polynomial of degree four is asymptotically stabilizable is NP-hard. Hint: Reduce from the NP-hard problem of MAX-2-SAT instead of 3-SAT.
\end{exercise}

\begin{exercise}
Let $A \in \R^{d \times d}$ be a diagonalizable matrix. Consider the following dynamical system:
\[
\x_{t + 1} =
\begin{cases}
\dfrac{A \x_t}{\|\x_t\|}, & \x_t \neq \bzero,\\[1ex]
\bzero, & \x_t = \bzero.
\end{cases}
\]
What are the equilibrium points $\x \in \R^d$ of the system? Which of them are stable, attractive, or asymptotically stable? Are any of them globally attractive? Your answer may depend on $A$.
\end{exercise}

\begin{exercise}
Let $\alpha, \beta, \gamma > 0$. Does the SIR model, as described in this chapter, define an asymptotically stabilizable dynamical system for a natural fixed choice of target state, such as a disease-free equilibrium? What about controllability? We assume that only starting states with the property $S_0 + I_0 + R_0 = 1$ are allowed and that $S_t, I_t, R_t, u_t \geq 0$ and $\alpha u_t \le S_t$ hold at all times.
\end{exercise}

\begin{exercise}
Consider the double integrator system:
\[
\mathbf{x}_{t+1} = A\mathbf{x}_t + B u_t + \mathbf{w}_t,
\]
where
\[
A = \begin{bmatrix}1 & \Delta \\ 0 & 1\end{bmatrix}, \quad B = \begin{bmatrix}0 \\ 1\end{bmatrix}, \quad \mathbf{x}_t = \begin{bmatrix}x_t \\ v_t\end{bmatrix} \in \mathbb{R}^2, \quad u_t \in \mathbb{R}, \quad \Delta >0.
\]
Assume throughout this exercise that $\mathbf{w}_t \equiv \mathbf{0}$.

The goal is to ensure that the system reaches the desired state \([x^*,0]^\top\) as \(t \to \infty\) from any initial condition.
\begin{enumerate}
    \item Prove that \((A,B)\) is controllable. That is, show that for any initial state and any target state, there exists a finite sequence of inputs that transitions the system from the initial state to the target state.

    \item Design a stabilizing state-feedback law
\[
u_t = -K(\mathbf{x}_t - [x^* \; 0]^\top),
\]
with \(K=[k_1 \; k_2]\), which places all eigenvalues of \(A-BK\) strictly within the unit circle. Find conditions on \(\Delta, k_1, k_2\) that ensure that all eigenvalues of \(A-BK\) are real and lie strictly between \(-1\) and \(1\), thus guaranteeing \(\mathbf{x}_t \to [x^*,0]^\top\).
\end{enumerate}
\end{exercise}

\end{exercises}
\chapter{Markov Decision Processes} \label{chap:RL}
\chaptermark{Markov Decision Processes} 

The previous chapters described and motivated the problem of control in dynamical systems. We have seen that, in full generality, even simple questions about dynamical systems such as stabilizability and controllability can be computationally hard. In this chapter, we study a more structured model for sequential decision making for which computationally efficient methods are available.

This chapter focuses on \emph{finite-state, finite-action, discounted, fully observed} Markov decision processes. In this setting, the state space and transition dynamics are explicitly represented, and this extra structure makes efficient planning algorithms possible. In addition, throughout this chapter we assume that the transition probabilities and rewards are known. When these quantities are unknown and must be inferred from interaction with the environment, one arrives at reinforcement learning in the narrower sense.

After introducing the model of Markov decision processes, we discuss dynamic programming and the accompanying characterization of optimal policies via the Bellman equation. We then present two algorithmic consequences: a linear-programming formulation and the value-iteration algorithm.

\section{Reinforcement Learning} 

Reinforcement learning is a subfield of machine learning that studies learning through interaction with a reactive environment. This is in contrast to learning from labeled examples, as in supervised learning, or inference in a probabilistic model, as in Bayesian statistics. In reinforcement learning, we have an agent and an environment. The agent observes its current state in the environment and takes an action; the environment then transitions to a new state and emits a reward.

\begin{figure}[!htb]
        \center{\includegraphics[scale = 0.4]
        {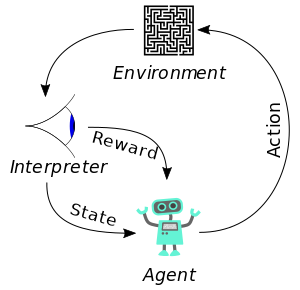}}
\caption{The agent--environment interaction loop in reinforcement learning.}
\end{figure}

Typically, the goal is to learn behavior that maximizes the total reward accumulated over time. This viewpoint has led to some notable successes: machines have learned how to play Atari games and how to defeat the strongest human players at Go.

A crucial modeling assumption is the following.
\begin{optbox}
\fbox{\parbox{0.97\textwidth}{
{\bf The Markovian assumption:} Conditional on the current state and current action, the distribution of the next state and immediate reward is independent of the past.
}}
\end{optbox}

The reader can imagine situations where such an assumption is reasonable. For example, in the inverted pendulum setting from Section~\ref{subsec:inv_pend}, the assumption is natural: once the current angle and angular velocity are known, the future evolution under a chosen torque does not depend on how that state was reached.

Other problems call this assumption into question. In the medical ventilation problem, for instance, future treatment may depend on the medical history of the patient and not only on the present physiological variables. One possible remedy is to expand the state description so that it includes enough information to make the process Markovian. In the extreme, the state can contain the entire history, in which case the Markovian assumption becomes true by construction. Of course, such a representation may be computationally inefficient.

In this chapter we restrict attention to the fully observed setting, in which the state itself is available to the agent. When only partial observations are available, the appropriate formal model is a partially observed Markov decision process (POMDP), which is beyond our present scope.

\subsection{A Classical Example: Bubbo's World} 
Before diving into the formalism, we start with a classical example from the text of \ifarxiv Russel and Norvig \fi \citet{norvig2002modern}. In this example, a robot starts in the bottom-left square, and its goal is to reach the terminal state with reward $+1$. The square marked $-1$ is also terminal, but undesirable. The block `X' represents an obstacle through which the robot cannot travel.

\begin{figure}[h]
\centering
\begin{tikzpicture}
\draw[step=1cm,color=gray] (0,0) grid (4,3);
\node at (0.5, 0.5) {{Start}};
\node at (0.5,1.5) {};
\node at (0.5,2.5) {};
\node at (1.5,2.5) {};
\node at (2.5,2.5) {};
\node at (1.5,0.5) {};
\node at (2.5,0.5) {};
\node at (3.5,0.5) {};
\node at (2.5,1.5) {};
\node at (1.5, 1.5) {X};
\node at (3.5, 2.5) {$+1$};
\node at (3.5, 1.5) {$-1$};
\end{tikzpicture}
\caption{A visual representation of Bubbo's world.}
\end{figure}

The robot can choose to move left, right, up, or down. However, the controls are noisy. With probability $0.8$, the robot moves in the intended direction; with the remaining probability $0.2$, it moves sideways, with probability $0.1$ to each side.

Notice that the Markovian assumption clearly holds in this example. Once the robot's current square and chosen action are specified, the distribution of the next square and the immediate reward no longer depend on the earlier history.

A solution to this problem is a mapping from states to actions. Such a mapping is called a \emph{policy}. The best policy depends not only on the transition structure but also on how rewards are assigned. To connect the example to the discounted infinite-horizon formalism used later in the chapter, it is convenient to model terminal states as absorbing states: entering the upper terminal state yields reward $+1$, entering the lower terminal state yields reward $-1$, and thereafter the process remains in that terminal state with reward $0$.

For example, if the reward for each non-terminal step is mildly negative, say $-0.04$, then Bubbo prefers to steer toward the $+1$ terminal while avoiding the risk of falling into the $-1$ terminal, as in Figure~\ref{fig:bubbo-3}.

\begin{figure}[h]
\centering
\begin{tikzpicture}
\draw[step=1cm,color=gray] (0,0) grid (4,3);
\node at (0.5, 0.5) {$\uparrow$};
\node at (0.5,1.5) {$\uparrow$};
\node at (0.5,2.5) {$\rightarrow$};
\node at (1.5,2.5) {$\rightarrow$};
\node at (2.5,2.5) {$\rightarrow$};
\node at (1.5,0.5) {$\leftarrow$};
\node at (2.5,0.5) {$\leftarrow$};
\node at (3.5,0.5) {$\leftarrow$};
\node at (2.5,1.5) {$\uparrow$};
\node at (1.5, 1.5) {X};
\node at (3.5, 2.5) {$+1$};
\node at (3.5, 1.5) {$-1$};
\end{tikzpicture}
\caption{An optimal policy when the reward for each non-terminal step is $-0.04$. \label{fig:bubbo-3}}
\end{figure}

Conversely, if the step cost is very negative, then Bubbo prefers to terminate the episode as quickly as possible, even at the risk of ending in the $-1$ terminal. This leads to a different optimal policy, as in Figure~\ref{fig:bubbo-negative}.

\begin{figure}
\centering
\begin{tikzpicture}
\draw[step=1cm,color=gray] (0,0) grid (4,3);
\node at (0.5, 0.5) {$\rightarrow$};
\node at (0.5,1.5) {$\uparrow$};
\node at (0.5,2.5) {$\rightarrow$};
\node at (1.5,2.5) {$\rightarrow$};
\node at (2.5,2.5) {$\rightarrow$};
\node at (1.5,0.5) {$\rightarrow$};
\node at (2.5,0.5) {$\rightarrow$};
\node at (3.5,0.5) {$\uparrow$};
\node at (2.5,1.5) {$\rightarrow$};
\node at (1.5, 1.5) {X};
\node at (3.5, 2.5) {$+1$};
\node at (3.5, 1.5) {$-1$};
\end{tikzpicture}
\caption{If the reward of each non-terminal step is changed to $-2$, the optimal policy changes: the robot is now more concerned with ending the game quickly than with reaching the $+1$ terminal. \label{fig:bubbo-negative}}
\end{figure}
\newpage

We next describe a formal model that captures this and many other reinforcement-learning problems.

\section{Markov Decision Processes}

We now describe the mathematical model used in this chapter: finite discounted Markov decision processes. To ease the exposition, we begin with Markov chains, proceed to Markov reward processes, and then give the full Markov decision process definition.

\subsection{Markov Chains}

A Markov chain is a stochastic process that evolves without exogenous actions. It consists of a finite set of states together with a transition matrix that specifies the probability of moving from one state to another.

Formally, let $S$ be a finite set of states. A Markov chain on $S$ is specified by a row-stochastic matrix $P \in [0,1]^{S \times S}$, where the entry $P_{ij}$ is the probability of moving from state $i$ to state $j$ in one step. Figure~\ref{fig:MDP-illustration} describes a Markov chain corresponding to a simple weather process with transition matrix
\[
P = \begin{pmatrix}
    0.9&0.1&0&0\\
    0.5&0&0.4&0.1\\
    0&0.5&0&0.5\\
    0&0&0&1\\
\end{pmatrix} .
\]

\begin{figure}[h] 
\centering
\begin{tikzpicture}[node distance = 5cm, auto]
	\node[vertex] (1) at  (0,2) {Rain};
	\node[vertex] (2) at  (5,0) {Sunny day};
	\node[vertex] (3) at  (8,-2) {Cloudy};
	\node[vertex] (4) at  (8,2) {Snow};
    \draw[edge] (1) to[bend left] node{0.1}(2);
    \draw[edge] (1) to[loop above] node{0.9} (1);
	\draw[edge] (2) to[bend left=40] node{0.5} (1);
	\draw[edge] (2) to[bend left=40] node{0.1} (4);
	\draw[edge] (2) to[bend left=40] node{0.4} (3);
	\draw[edge] (3) to[bend left=40] node{0.5} (2);
	\draw[edge] (3) to[bend right=40] node{0.5} (4);
	\draw[edge] (4) to[loop above] node{1} (4);
\end{tikzpicture}
\caption{A graphical description of a Markov chain. Each state is the weather on a given day, and the transition probabilities describe the distribution of the next day's weather. \label{fig:MDP-illustration}}
\end{figure}

Denote the distribution over states at time $t$ by $\bp_t$. Then
\[
\bp_{t+1} = \bp_t P = \bp_0 P^{t+1},
\]
where $\bp_0$ is the initial state distribution.

A central object in the study of Markov chains is a stationary distribution.

\begin{definition}
A distribution $\bpi$ over the state space $S$ is a \textbf{stationary distribution} of a Markov chain with transition matrix $P$ if and only if
\[
\bpi = \bpi P.
\]
\end{definition}

For finite Markov chains, a stationary distribution always exists. It need not be unique, however, and convergence from an arbitrary initial distribution need not occur without further assumptions.

Two standard structural properties are the following.
\begin{enumerate}
    \item  {\bf Irreducibility.}
    A Markov chain is \emph{irreducible} if and only if
    \[
    \forall i,j \in S \; \exists t \ge 0 \text{ such that } (P^t)_{ij} > 0.
    \]
    In words, every state is reachable from every other state.
    
    \item {\bf Aperiodicity.}
    A Markov chain is \emph{aperiodic} if and only if every state has period one. The period of a state $s_i$ is defined by
\[
 d(s_i) = \gcd \left\{t \in \mathbb{N}_+ \ \middle| \ (P^t)_{ii} > 0 \right\} .
\]
\end{enumerate}

The following finite-state ergodic theorem summarizes the role of these properties.

\begin{theorem}[Ergodic theorem for finite Markov chains]
Let $P$ be a finite Markov chain.
\begin{enumerate}
    \item $P$ has at least one stationary distribution.
    \item If $P$ is irreducible, then the stationary distribution is unique.
    \item If $P$ is irreducible and aperiodic, then for every initial distribution $\bp_0$,
    \[
    \lim_{t\to\infty} \bp_0 P^t = \bpi,
    \]
    where $\bpi$ is the unique stationary distribution.
\end{enumerate}
\end{theorem}

Its proof is a beautiful application of spectral graph theory. Exercise~\ref{exer:ergodic} guides the reader through one standard proof of the convergence statement.

\begin{figure}
\centering
\begin{tikzpicture}
\node[vertex] (A) at  (0,-5) {A};
\node[vertex] (B) at  (3,-5) {B};

\draw[edge] (A) to[bend left] node[above]{1}(B);
\draw[edge] (B) to[bend left] node[above]{1} (A);
\end{tikzpicture}
\caption{This two-state Markov chain is irreducible but periodic with period $2$. It has a unique stationary distribution, namely $\bpi = (\frac{1}{2},\frac{1}{2})$, but the distribution of the chain does not converge to $\bpi$ from every initial state because the chain alternates deterministically between the two states.}
\end{figure}

\begin{figure}
\centering
\begin{tikzpicture}
\node[vertex] (1) at  (0,-5) {1};
\node[vertex] (2) at  (6,-5) {2};

\draw[edge] (1) to[bend left=50] node[above]{0.5}(2);
\draw[edge] (2) to[bend left=50] node[above]{0.5} (1);
\draw[edge] (1) to[loop above] node[above]{0.5} (1);
\draw[edge] (2) to[loop above] node[above]{0.5} (2);

\end{tikzpicture}
\caption{Adding a self-loop to every node makes the period equal to one. This chain is irreducible and aperiodic, so its stationary distribution exists, is unique, and equals $\bpi = (\frac{1}{2}, \frac{1}{2})$.}
\end{figure}

An important special case is the reversible Markov chain.
\begin{definition}
A Markov chain $P$ is said to be \textbf{reversible with respect to a distribution} $\bpi$ if and only if, for every two states $i,j$,
\[
\bpi_i P_{ij} = \bpi_j P_{ji} .
\]
A Markov chain is said to be \textbf{reversible} if there exists some distribution with respect to which it is reversible.
\end{definition}

Exercise~\ref{exer:reversibleMC} asks the reader to prove that any distribution satisfying the reversibility equations is automatically stationary.

\subsection{Markov Reward Processes}
Markov reward processes add rewards to Markov chains. At each time step, the process transitions between states and emits a reward. For simplicity, we assume rewards lie in the range $[0,1]$. This normalization loses no essential generality and keeps the formulas uncluttered.

\begin{definition}
A \textbf{Markov reward process} is a tuple $(S,P,R,\gamma)$, where:
\begin{itemize}
\item $S$ is a finite set of states.
\item $P \in [0,1]^{S \times S}$ is a row-stochastic transition matrix.
\item $R \in [0,1]^{S \times S}$ is a reward function, where $R(s,s')$ is the reward obtained when moving from state $s$ to state $s'$.
\item $\gamma \in [0,1)$ is the discount factor.
\end{itemize}
\end{definition}

We henceforth use the following notation.
\begin{enumerate}
    \item $S_t \in S$ denotes the random state of the process at time $t$.
    
    \item $R_t$ denotes the random reward obtained when transitioning from $S_t$ to $S_{t+1}$.

    \item $\hat{R} \in \reals^S$ denotes the vector of expected one-step rewards:
    \[
    \hat{R}(s) = \E[ R_t \mid S_t = s].
    \]
    
    \item $G_t$ denotes the discounted return from time $t$ onward:
    \[
    G_t = \sum_{i=0}^{\infty} \gamma^{i} R_{t+i} .
    \]
\end{enumerate}

Since $\gamma<1$ and rewards are bounded, the random variable $G_t$ is well defined.

A natural question is: what quantity should summarize the future behavior of a Markov reward process? For an ordinary Markov chain, one often studies long-run state frequencies via stationary distributions. Once rewards and discounting are introduced, however, the central quantity becomes the expected discounted return from a given starting state. This dependence on the starting state is unavoidable: for example, when $\gamma=0$, the return reduces to the immediate reward, which can vary from state to state.

This motivates the central object in the theory of Markov reward processes: the value function. Because the process is time-homogeneous, the conditional expectation $\E[G_t \mid S_t = s]$ depends only on the state $s$, and not on the particular time index $t$.

\begin{definition}
The \textbf{value function} of a Markov reward process maps states to expected discounted return:
\[
\bv(s) = \mathbb{E}[G_t \mid S_t = s] .
\]
\end{definition}

The value function measures how desirable each state is in terms of expected future reward. It satisfies the Bellman equation for a Markov reward process:
\[
\bv(s) = \hat{R}(s) + \gamma \sum_{s' \in S} P_{s,s'} \bv(s') .
\]
To verify this identity, write
\begin{align*}
\bv(s) &= \mathbb{E}[G_t \mid S_t = s] \\
&= \mathbb{E}\left[\sum_{i=0}^{\infty} \gamma^{i} R_{t+i} \; \middle| \; S_t = s\right] \\
&= \mathbb{E}\left[R_t + \gamma \sum_{i=0}^{\infty} \gamma^i R_{t+1+i} \; \middle| \; S_t = s\right] \\
&= \mathbb{E}[R_t + \gamma G_{t+1} \mid S_t = s] \\
&= \mathbb{E}[R_t + \gamma \bv(S_{t+1}) \mid S_t = s] \\
&= \hat{R}(s) + \gamma \sum_{s' \in S} P_{s,s'} \bv(s').
\end{align*}

In matrix form, the Bellman equation is
\[
\bv = \hat{R} + \gamma P \bv.
\]
Equivalently, $\bv$ is the unique solution to the linear system
\[
(I - \gamma P)\bv = \hat{R}.
\]
Since $\gamma < 1$ and $P$ is a stochastic matrix, the matrix $I - \gamma P$ is invertible, and this system has a unique solution.

Thus, for a known Markov reward process, the value function is the solution to a linear system and can be computed in polynomial time in the number of states.

\subsection{Markov Decision Processes}

The final layer of abstraction is to allow actions, or controls in the language of control theory. This gives the full Markov decision process model.

\begin{definition}
A \textbf{Markov decision process} is a tuple $(S,P,R,A,\gamma)$ such that:
\begin{enumerate}
\item $S$ is a finite set of states.
\item $A$ is a finite set of possible actions.
\item $P$ is a transition kernel, written as
\[
P^a_{ss'} = \Pr[ S_{t+1} = s' \mid S_t = s, A_t = a].
\]
For each action $a \in A$, the matrix $P^a$ is row-stochastic.
\item $R$ is a reward function on state--action--next-state triples, where $R_{ss'a}$ is the reward for moving from $s$ to $s'$ using action $a$. We also write
\[
R_{sa} = \sum_{s' \in S} P^a_{ss'} R_{ss'a}
\]
for the corresponding expected immediate reward. As above, we assume rewards lie in $[0,1]$.
\item $\gamma \in [0,1)$ is the discount factor.
\end{enumerate}
\end{definition}

The central solution concept is a policy.

\begin{definition}
A \textbf{policy} is a rule for choosing actions as the process evolves. A policy may in principle depend on the entire history. A {stationary policy} is a policy that depends only on the current state, and is therefore a mapping
\[
\pi : S \to \Delta(A),
\qquad
\pi(a\mid s) = \Pr[A_t = a \mid S_t = s].
\]
If each $\pi(\cdot\mid s)$ is a point mass, the policy is called {deterministic}.
\end{definition}

In this chapter we consider infinite-horizon discounted return. Starting from a state $s$, the value of a policy $\pi$ is
\[
\bv_\pi(s) = \E_{\pi,P} \left[ \sum_{t=0}^\infty \gamma^t R_t \; \middle| \; S_0 = s \right].
\]
The optimal value function is defined pointwise by
\[
\bv^*(s) = \sup_{\pi} \bv_\pi(s).
\]

A fundamental theorem of discounted MDPs is the following.
\begin{theorem} \label{thm:fundamental-rl}
For every finite discounted Markov decision process, there exists a deterministic stationary optimal policy $\pi^*$. Moreover, every optimal policy has the same value function $\bv^*$.
\end{theorem}

The existence of an optimal stationary policy is a consequence of the Markovian structure of the problem and the Bellman optimality equation, which we now discuss.

\section{The Bellman Equation}

A fundamental property of Markov decision processes is that they satisfy a recursive optimality principle. This principle is expressed by the Bellman equation and is the basis for dynamic programming.

For a fixed stationary policy $\pi$, the Bellman equation takes the form
\[
\bv_\pi(s) = \sum_{a \in A} \pi(a\mid s) \left[ R_{sa}  +  \gamma \sum_{s' \in S} P_{ss'}^a \bv_\pi(s')  \right].
\]
In matrix--vector notation this can be written as
\[
\bv_\pi = \mathbf{R^\pi} + \gamma P^\pi \bv_\pi ,
\]
where
\begin{enumerate}
\item $R^\pi$ is the expected reward vector induced by $\pi$, defined by
\[
R^\pi(s) = \E_{a \sim \pi(\cdot\mid s)} [R_{sa}],
\]
\item $P^\pi$ is the transition matrix induced by $\pi$, defined by
\[
P^\pi_{ss'} =  \sum_{a \in A } P^{a}_{ss'} \, \pi(a\mid s).
\]
\end{enumerate}

Since $\gamma<1$, the matrix $I-\gamma P^\pi$ is invertible, and therefore the Bellman equation for a fixed policy has a unique solution.

The Bellman equation becomes especially powerful when applied to the optimal value function.
\begin{optbox}
\fbox{\parbox{0.97\textwidth}{
\[
\bv^*(s) = \max_{a \in A} \left\{ R_{sa} + \gamma \sum_{s' \in S} P^a_{ss'} \bv^*(s') \right\}
\]
}}
\end{optbox}

This is the \textbf{Bellman optimality equation}. It says that the optimal value of a state is obtained by choosing the action with the largest one-step reward plus discounted continuation value.

The proof is a one-step decomposition. Fix a state $s$. For any policy, the value obtained from $s$ is bounded above by the best possible choice of first action plus the optimal continuation value thereafter, so
\[
\bv^*(s) \le \max_{a \in A} \left\{ R_{sa} + \gamma \sum_{s' \in S} P^a_{ss'} \bv^*(s') \right\} .
\]
Conversely, choose an action $a^*$ achieving the maximum on the right-hand side. For each successor state $s'$, choose a policy that is $\varepsilon$-optimal from $s'$. By first taking action $a^*$ and then following the chosen continuation policy, one achieves a value at least
\[
R_{sa^*} + \gamma \sum_{s' \in S} P^{a^*}_{ss'} \bv^*(s') ,
\]
and the reverse inequality follows.

The Bellman optimality equation immediately implies the existence of a deterministic optimal policy. Indeed, define a policy $\pi^*$ by choosing at each state any maximizing action,
\[
\pi^*(s) \in \arg\max_{a \in A} \left\{ R_{sa} + \gamma \sum_{s' \in S} P^a_{ss'} \bv^*(s') \right\} .
\]
Then $\bv^*$ satisfies
\[
\bv^*(s) = R_{s\pi^*(s)} + \gamma \sum_{s' \in S} P^{\pi^*(s)}_{ss'} \bv^*(s')
\qquad \text{for all } s \in S,
\]
which is exactly the Bellman equation for the fixed policy $\pi^*$. By uniqueness of the solution of the Bellman equation for a fixed policy, we obtain $\bv_{\pi^*}=\bv^*$, proving that $\pi^*$ is optimal. This proves Theorem~\ref{thm:fundamental-rl}.

\subsection{Linear Programming Formulation}

The Bellman optimality equation characterizes $v^*$ as a fixed point. An equivalent characterization views $v^*$ as the smallest function satisfying the Bellman inequalities.

Consider the following linear program:
\begin{align*}
\min_{\bv \in \mathbb{R}^{|S|}} \quad & \sum_{s \in S} \bv(s) \\
\text{s.t.} \quad 
& \bv(s) \ge r(s,a) + \gamma \sum_{s' \in S} P(s' \mid s,a)\, \bv(s') \quad \forall s \in S, \ a \in A.
\end{align*}

This program has $|S|$ variables and $|S|\,|A|$ constraints, so it can be solved in polynomial time in the numbers of states and actions. The next theorem shows that its unique optimum is precisely the optimal value function.

\begin{theorem}
The unique optimal solution of the above linear program is the optimal value function $\bv^*$.
\end{theorem}

\begin{proof}
First, $\bv^*$ is feasible, since it satisfies the Bellman optimality equation and therefore all Bellman inequalities.

Now let $\bv$ be any feasible solution. Then for all $s$,
\[
\bv(s) \ge \max_a \left\{ r(s,a) + \gamma \sum_{s'} P(s'\mid s,a)\, \bv(s') \right\}
= (T \bv)(s),
\]
where $T$ is the Bellman optimality operator. Thus $\bv \ge T \bv$. Iterating,
\[
\bv \ge T \bv \ge T^2 \bv \ge \cdots.
\]
Since $T^k \bv \to \bv^*$ as $k \to \infty$, it follows that $\bv \ge \bv^*$ componentwise.

Therefore $\bv^*$ is the smallest feasible vector, and hence minimizes any strictly increasing objective such as $\sum_s \bv(s)$. Uniqueness follows.
\end{proof}

\section{The Value Iteration Algorithm} \label{sec:vi}

The Bellman optimality equation also leads to one of the most fundamental algorithms for discounted MDPs: value iteration. Starting from an initial estimate of the value function, the algorithm repeatedly applies the Bellman optimality update.

\begin{algorithm}[h!]
\caption{Value Iteration} \label{alg:value-iteration}
\begin{algorithmic}
\STATE \textbf{Input:} MDP
\STATE Set $\bv_0(s) = 0$ for all states $s$
\FOR{$t = 0$ to $T-1$}
\STATE Update for all states,
\[
\bv_{t+1}(s) = \max_{a \in A} \left\{ R_{sa} + \gamma \sum_{s' \in S} P_{ss'}^a \, \bv_t(s') \right\}
\]
\ENDFOR
\STATE Define a greedy policy with respect to $\bv_T$ by
\[
\pi_T(s) \in \arg\max_{a \in A} \left\{ R_{sa} +  \gamma \sum_{s' \in S} P_{ss'}^a \, \bv_T(s') \right\}
\]
\RETURN $\bv_T$ and $\pi_T$
\end{algorithmic}
\end{algorithm}

The next theorem gives the basic convergence guarantee.
\begin{theorem}
Let $(\bv_t)_{t\ge 0}$ be the sequence produced by Value Iteration with initialization $\bv_0=\mathbf{0}$. Then:
\begin{enumerate}
    \item $\mathbf{0} \le \bv_t \le \bv_{t+1} \le \bv^*$ entry-wise for every $t \ge 0$.
    \item If we define the residual vector $\br_t = \bv^* - \bv_t$, then
    \[
    \|\br_{t+1}\|_\infty \le \gamma \|\br_t\|_\infty
    \qquad \text{for every } t \ge 0.
    \]
    Consequently,
    \[
    \|\bv^* - \bv_t\|_\infty \le \frac{\gamma^t}{1-\gamma} .
    \]
\end{enumerate}
In particular, $\bv_t \to \bv^*$ geometrically fast.
\end{theorem}

\begin{proof}
Monotonicity and the upper bound are left as an exercise. 
We now prove the geometric decay of the residual.

For every state we have by choice of the algorithm that
$$\bv_{t+1}(s) =  \max_a \left\{ R_{sa} + \gamma \sum_{s'} P^a_{ss'} \bv_t(s') \right\}  \geq  R_{sa^*} + \gamma \sum_{s'} P^{a^*}_{ss'} \bv_t(s'),   $$
where $a^*=\bpi^*(s)$ is the action of an optimal policy at state $s$.
On the other hand, from Bellman optimality equation, we have that
$$\bv^*(s) =   R_{sa^*} + \gamma \sum_{s'} P^{a^*}_{ss'} \bv^*(s').   $$
Subtracting the two equations we get for all states $s$ that
\begin{align*}
\br_{t+1}(s) & =  \bv^*(s) - \bv_{t+1} (s) \\
& \leq \gamma \sum_{s'} P^{a^*}_{ss'} (\bv^*(s') - \bv_t(s') ) \\  
& = \gamma \sum_{s'} P^{a^*}_{ss'} \br_t(s') .
\end{align*}
Or in vector form, $\br_{t+1} \leq \gamma P^{\ba^*} \br_t $. Furthermore, $\br_{t} \geq 0 $ is entry-wise nonnegative for all $t$. (See Exercise~\ref{exer:vinc}.) Therefore, it holds that
$$\|\br_{t+1}\|_\infty \leq \gamma \|P^{\ba^*} \br_t\|_\infty \leq \gamma \|\br_t\|_\infty,$$
where we use the fact that each row of $P^{\ba^*}$ sums to one. Applying this repeatedly, we have
$$ \| \br_{t+1} \|_\infty \leq \gamma^t  \|\br_1\|_\infty .$$
Finally, because rewards lie in $[0,1]$, the total discounted reward is at most $1+\gamma+\gamma^2+\cdots = 1/(1-\gamma)$, so
\[
\|\br_0\|_\infty = \|\bv^*\|_\infty \le \frac{1}{1-\gamma}.
\]
Combining the two inequalities yields
\[
\|\bv^* - \bv_t\|_\infty = \|\br_t\|_\infty \le \frac{\gamma^t}{1-\gamma}.
\]
\end{proof}

\ifarxiv
\newpage
\fi
\section{Bibliographic Remarks}
The section on reinforcement learning in the text \citet{norvig2002modern} introduces basic concepts and places them in the broader context of artificial intelligence. The classical textbook on reinforcement learning is \citet{Sutton1998}, which develops the subject from both the algorithmic and conceptual perspectives.

Another excellent reference is the set of lecture notes \citet{SilverLecNotes}. See \citet{agarwal2019lectures} for an introduction to the theoretical aspects of reinforcement learning, including sample-complexity guarantees and reductions to supervised learning. The reader may also refer to \citet{bertsekas2019reinforcement} for a broad survey of computational approaches to learning in unknown Markov decision processes.

Dynamic programming is at the heart of Markov decision processes. This viewpoint is emphasized in \citet{bertsekas2007dynamic}, which discusses the Bellman equations and related algorithmic ideas at length. It is also a useful source for the finite-horizon and average-reward settings, which we have only mentioned briefly here.

Although tabular MDPs do not compactly represent stochastic linear dynamical systems, function-approximation schemes \citep{jin2020provably, du2021bilinear, jiang2017contextual, kakade2020information} can successfully capture problems such as stochastic LQR.

\ifarxiv
\newpage
\fi
\begin{exercises}

\begin{exercise}\label{exer:ergodic}
In this exercise, we give a guided proof of the convergence part of the ergodic theorem. Consider a finite irreducible and aperiodic Markov chain with transition matrix $P$. \\
\begin{subexercise}
Show that the all-ones vector is a right eigenvector of $P$ with eigenvalue $1$. Conclude that $1$ is also an eigenvalue of $P^\top$.
\end{subexercise}\\
\begin{subexercise}
Show that all eigenvalues of $P$ have magnitude at most one.
\end{subexercise}

The irreducibility and aperiodicity of the Markov chain imply that the eigenvalue $1$ of $P^\top$ is simple and that all other eigenvalues have magnitude strictly smaller than one. Assume this fact, and let $\vv_1$ denote the corresponding left eigenvector normalized to sum to one. \\
\begin{subexercise}
Consider an arbitrary initial distribution $\bp_0$, and the distribution after $t$ iterations of the Markov chain, $\bp_t = \bp_0 P^t$. Prove that
\[
\lim_{t\to \infty} \bp_t = \vv_1.
\]
\end{subexercise}
\end{exercise}

\begin{exercise}
Prove the geometric convergence of the value-iteration algorithm for rewards that can be arbitrary bounded real numbers, as opposed to lying in the range $[0,1]$.
\end{exercise}

\begin{exercise}
Let $\bv^*$ be the optimal value function of a discounted Markov decision process, and define a greedy policy by
\[
\pi(s) \in \arg\max_{a \in A} \left\{ R_{sa} + \gamma \sum_{s' \in S} P^a_{ss'} \bv^*(s') \right\}.
\]
Prove that $\pi$ is optimal.
\end{exercise}

\begin{exercise}\label{exer:reversibleMC}
Let Markov chain $P$ be reversible with respect to a distribution $\bpi$, i.e.
\[
\forall i,j \ , \ \bpi_i P_{ij} = \bpi_j P_{ji} .
\]
Prove that $\bpi$ is stationary for $P$.
\end{exercise}

\begin{exercise}\label{exer:vinc}
Suppose Value Iteration is initialized at an arbitrary vector $\bv_0$ satisfying
\[
\bv_0(s) \le \max_{a \in A} \left\{ R_{sa} + \gamma \sum_{s' \in S} P^a_{ss'} \bv_0(s') \right\}
\qquad \text{for all } s \in S.
\]
Prove that the successive iterates $\bv_t$ increase monotonically, that is, $\bv_{t+1} \geq \bv_t$ for all $t$, and that $\bv_t \leq \bv^\star$.
\end{exercise}

\end{exercises}

\include{4.LDS-intro}

\chapter{Linear Dynamical Systems}
\label{chap:lds}

In Chapter \ref{chap:dynamical_systems}, we observed that even simple questions about general dynamical systems are intractable. This motivates restricting attention to a special class of dynamical systems for which we can design efficient algorithms with provable guarantees.

Our treatment in this chapter is restricted to linear dynamical systems (LDS): dynamical systems that evolve linearly according to the equation
\[
    \x_{t + 1} = {A}_t \x_t + {B}_t \uv_t + \w_t.
\]

If the matrices ${A}_t={A}$ and ${B}_t={B}$ are fixed, we say that the dynamical system is linear time-invariant (LTI). We have already seen examples of LTI systems in Chapter \ref{chap:dynamical_systems}.

\begin{figure}
	\centering
	\includegraphics[scale=0.15]{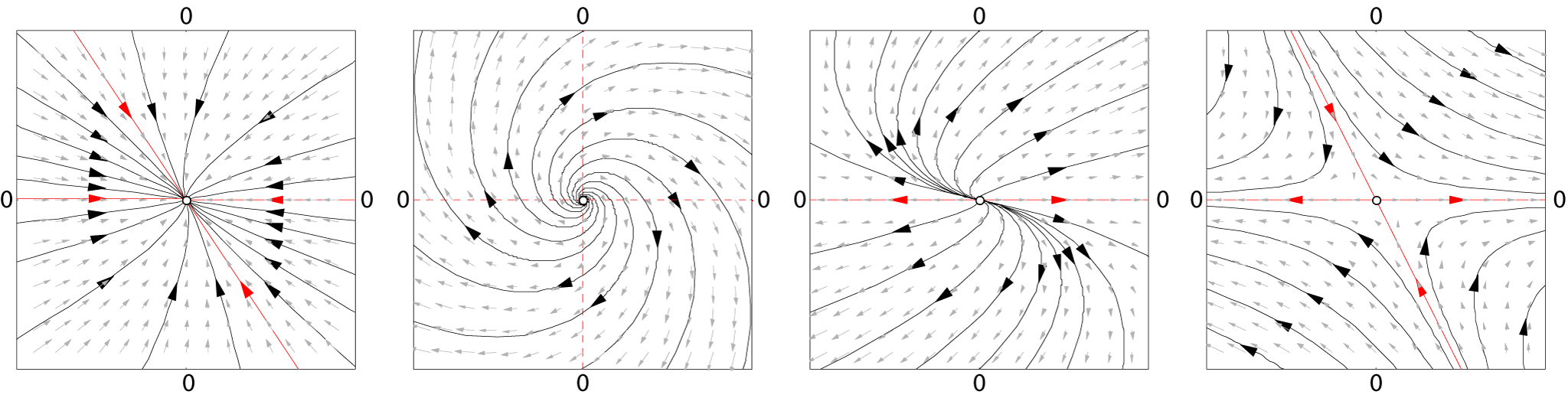}
	\caption{Four different deterministic linear time-invariant dynamical systems represented via their induced vector fields and a few chosen trajectories. Created by XaosBits using the Mathematica packages CurvesGraphics.nb by Gianluca Gorni and DrawGraphics.nb by David and Alice Park, licensed under CC-BY-2.5.}
\end{figure}

A more general class of linear dynamical systems arises when the matrices ${A}_t$ and ${B}_t$ change over time. These are called linear time-varying (LTV) systems. The class of LTV systems can express more general dynamics via linearization, as we show next.

\section{Approximating General Dynamics via LTV Systems}

Consider deterministic dynamics of the form
\[
    \x_{t+1} = f(\x_t,\uv_t),
\]
where $f$ is twice continuously differentiable. Fix a nominal state-control trajectory $(\bar{\x}_t,\bar{\uv}_t)$ satisfying
\[
    \bar{\x}_{t+1} = f(\bar{\x}_t,\bar{\uv}_t),
\]
and define the deviation variables
\[
    \delta \x_t = \x_t-\bar{\x}_t,
    \qquad
    \delta \uv_t = \uv_t-\bar{\uv}_t.
\]

A first-order Taylor expansion around $(\bar{\x}_t,\bar{\uv}_t)$ gives
\[
    f(\x_t,\uv_t)
    =
    f(\bar{\x}_t,\bar{\uv}_t)
    + J_{\x}f(\bar{\x}_t,\bar{\uv}_t)(\x_t-\bar{\x}_t)
    + J_{\uv}f(\bar{\x}_t,\bar{\uv}_t)(\uv_t-\bar{\uv}_t)
    + r_t,
\]
where $J_{\x}f(\bar{\x}_t,\bar{\uv}_t)$ and $J_{\uv}f(\bar{\x}_t,\bar{\uv}_t)$ are the Jacobians of $f$ with respect to $\x$ and $\uv$, evaluated at $(\bar{\x}_t,\bar{\uv}_t)$, and where the remainder satisfies
\[
    \|r_t\| = O\!\left(\|\delta \x_t\|^2 + \|\delta \uv_t\|^2\right)
\]
locally around the nominal trajectory. Writing
\[
    A_t \defeq J_{\x}f(\bar{\x}_t,\bar{\uv}_t),
    \qquad
    B_t \defeq J_{\uv}f(\bar{\x}_t,\bar{\uv}_t),
\]
and subtracting the nominal dynamics yields the deviation dynamics
\[
    \delta \x_{t+1} = A_t \delta \x_t + B_t \delta \uv_t + r_t.
\]

Thus, near a nominal trajectory, a smooth nonlinear dynamical system is well approximated by a time-varying linear system in the deviation variables. If the original nonlinear system also has additive perturbations, then those perturbations simply enter additively in the linearized deviation dynamics as well.

This derivation explains why in practice the class of time-varying linear dynamical systems is very expressive, at least locally for smooth dynamical systems. In chapter \ref{chap:koopman} we see another reduction to LDS from nonlinear dynamics.

As we shall see next, stabilizability and controllability can be checked efficiently for linear systems, in contrast to general dynamics.

\section{Stabilizability of Linear Systems}
\label{subsec:stabilizability}

Verifying stability of a system, and more so stabilizability, was shown to be intractable in our discussion of Chapter \ref{chap:dynamical_systems}. For linear dynamical systems, however, the picture is quite different: stability and stabilizability can be efficiently verified. Furthermore, the theoretical analysis of this fact is elegant.

For simplicity, in the rest of this section we consider time-invariant linear dynamical systems of the form
\[
    \x_{t + 1} = {A} \x_t + {B} \uv_t + \w_t,
\]
although much of our discussion can be generalized to LTV systems.

Recall that a system is stabilizable if, in the absence of perturbations, from every starting state there exists a control policy that drives the state to zero. Determining whether a linear time-invariant system is stabilizable admits a particularly simple characterization. Henceforth, we denote by $\rho({A})$ the spectral radius of a matrix ${A}$, that is,
\[
    \rho({A}) = \max \{ |\lambda_1|,\ldots,|\lambda_d| \},
\]
the maximum modulus of all (possibly complex) eigenvalues. The stabilizability of a linear dynamical system can then be characterized as follows.

\begin{theorem}
\label{thm:stabilizability_condition}
A time-invariant LDS is stabilizable if and only if there exists a matrix \(K\) such that
\[
    \rho({A} + {B}{K}) < 1.
\]
\end{theorem}

We prove the sufficiency of this condition and leave the necessity as Exercise \ref{exer:necessary_K}. Both directions rely on the following characterization of the spectral radius.

\begin{fact}
$\rho({A}) < 1$ if and only if $\lim_{t \to \infty} {A}^t = 0$. Furthermore, if $\rho({A}) > 1$, then for any norm,
\[
    \lim_{t \to \infty} \|{A}^t\| = \infty.
\]
\end{fact}

The proof of this fact is left as Exercise \ref{exer:limit_specrtal}. Based on it, we can now prove the sufficiency part of Theorem \ref{thm:stabilizability_condition}.

\begin{proof}[Theorem \ref{thm:stabilizability_condition}, sufficiency]
Suppose that there exists a matrix ${K}$ such that \(\rho({A} + {B}{K}) < 1\). Choose the linear state-feedback controller \(\uv_t = {K}\x_t\). In the noiseless system, the closed-loop dynamics are
\[
    \x_{t+1} = ({A}+{B}{K})\x_t,
\]
and therefore
\[
    \x_t = ({A}+{B}{K})^t \x_0.
\]
By the fact above, \((A+BK)^t \to 0\) as \(t \to \infty\). Hence
\[
    \lim_{t\to\infty}\|\x_t\|
    =
    \lim_{t\to\infty}\|({A}+{B}{K})^t \x_0\|
    = 0.
\]
Thus the system is stabilizable.
\end{proof}

\subsection{Efficient Verification of Stabilizability}

Stabilizability admits a convenient convex characterization. We use the following
standard form of the discrete-time Lyapunov theorem: a matrix \(M\) satisfies
\(\rho(M)<1\) if and only if there exists a matrix \(P\succ 0\) such that
\[
    M P M^\top \prec P.
\]
By scaling \(P\), we may additionally impose \(P \succeq I\).

Applying this to the closed-loop matrix \(A+BK\) gives the following linear matrix
inequality.

\begin{theorem}
\label{thm:stabilizing}
There exists a matrix \(K\) such that \(\rho(A+BK)<1\) if and only if there exist
matrices \(P\in \mathcal{S}^{d_x\times d_x}\) and \(L\in \reals^{d_u\times d_x}\)
such that
\[
    \begin{bmatrix}
        P & ({A}{P}+{B}{L})^\top \\
        {A}{P}+{B}{L} & P
    \end{bmatrix} \succ 0,
    \qquad
    P \succeq I.
\]
Moreover, any feasible pair \((P,L)\) yields a stabilizing controller
\(K=L P^{-1}\).
\end{theorem}

\begin{proof}
Suppose first that \(\rho(A+BK)<1\). By the discrete-time Lyapunov theorem, there
exists \(P\succ 0\) such that
\[
    (A+BK) P (A+BK)^\top \prec P.
\]
After scaling, we may assume \(P\succeq I\). By the Schur complement, this is
equivalent to
\[
    \begin{bmatrix}
        P & ((A+BK)P)^\top \\
        (A+BK)P & P
    \end{bmatrix} \succ 0.
\]
Setting \(L=KP\) gives the displayed condition.

Conversely, suppose \(P\) and \(L\) satisfy the displayed condition, and define
\(K=L P^{-1}\). Another application of the Schur complement yields
\[
    (A+BK) P (A+BK)^\top \prec P.
\]
Hence \(\rho(A+BK)<1\) by the discrete-time Lyapunov theorem.
\end{proof}

Thus stabilizability can be checked efficiently via semidefinite feasibility, and
any feasible solution directly produces a stabilizing controller.

\section{Controllability of Linear Dynamical Systems}
\label{sec:controllability}

Recall that a system is controllable if and only if, in the absence of noise and for every target state, there is a sequence of controls that drives the system state to that target. Linear dynamical systems exhibit a particularly elegant characterization of controllability. First, we define an object of central importance for linear dynamical systems: the Kalman controllability matrix.

\begin{definition}
The \emph{Kalman controllability matrix} of a linear dynamical system is given by
\[
    {K}_r({A}, {B}) = [{B}, {A}{B}, {A^2}{B}, \ldots, {A^{r-1}}{B}] \in \reals^{d_x \times r d_u}.
\]
\end{definition}

The controllability of a linear dynamical system can now be characterized using this matrix. Recall that the rank of a matrix is the maximal number of linearly independent columns (or rows).

\begin{theorem}
A time-invariant linear dynamical system is controllable if and only if
\[
    \rank {K}_{d_x}({A}, {B}) = d_x.
\]
\end{theorem}

\begin{proof}
Consider the evolution of the noiseless linear dynamical system starting from the initial state \(\x_0\). The state at time \(T\) is given by
\begin{align*}
    \x_T
    &= {A}\x_{T - 1} + {B}\uv_{T - 1}\\
    &= {A}^2\x_{T - 2} + {A}{B}\uv_{T - 2} + {B}\uv_{T - 1}\\
    &= \ldots = \sum_{t=1}^T {A}^{t-1}{B} \uv_{T-t} + {A}^T \x_0.
\end{align*}
Thus,
\[
    \x_T - {A}^T \x_0
    =
    [{B}, {A}{B}, {A^2}{B}, \ldots, {A^{T - 1}}{B}]
    \begin{bmatrix}
        \uv_{T - 1}\\
        \vdots\\
        \uv_0
    \end{bmatrix}.
\]
Therefore, for every initial state \(\x_0\) and target state \(\x_T\), there exists a control sequence \(\uv_0,\ldots,\uv_{T-1}\) that satisfies this equation if and only if the column span of \({K}_T(A,B)\) is all of \(\reals^{d_x}\), i.e., if and only if \({K}_T(A,B)\) has rank \(d_x\).

Denote by \(r_t = \rank({K}_t)\) the rank of the Kalman matrix after \(t\) steps. Since adding columns cannot decrease rank, the sequence \((r_t)_{t\ge 1}\) is nondecreasing. We claim that once \(r_t = r_{t + 1}\), the rank can never increase again.

Indeed, if \(r_t = r_{t+1}\), then
\[
    A^t B \in \myspan [B, AB, \ldots, A^{t-1}B].
\]
Multiplying by \(A\) gives
\[
    A^{t+1}B \in \myspan [AB, A^2B, \ldots, A^tB]
    \subseteq
    \myspan [B, AB, \ldots, A^tB].
\]
Hence adding the next block \(A^{t+1}B\) does not increase the rank, so \(r_{t+2}=r_{t+1}\). By induction, once the rank stops increasing, it never increases again.

Since the rank can increase at most \(d_x\) times, it follows that \(r_t=r_{d_x}\) for all \(t\ge d_x\). Therefore, the system is controllable if and only if
\[
    \rank K_{d_x}(A,B) = d_x.
\]
\end{proof}

\section{Quantitative Definitions}

In the rest of this text our goal is to give efficient algorithms with finite-time
performance bounds for online control. These bounds naturally depend on how
difficult it is to stabilize and control a given system. Since the main focus of
the sequel is on linear time-invariant systems, we now introduce quantitative
notions of stabilizability and controllability for that setting.

\subsection{Stabilizability of Time-Invariant Systems}

Consider a linear time-invariant system given by the pair $({A},{B})$. We have already remarked that this system is classically stabilizable if and only if there exists a matrix \(K\) such that \(\rho(A+BK)<1\).

A stronger quantitative property can be obtained as follows, and in fact it is implied by BIBO stabilizability as defined in Definition \ref{def:gamma-stabilizing}. The proof of this fact is left as Exercise \ref{exer:bibo-implies-exp-convergence}.

\begin{lemma}
\label{lem:ltistable}
Consider a linear controller ${K}$ for the LTI system $({A},{B})$. If there exist $\kappa>0$ and $\delta\in (0,1]$ such that
\[
    \|(A+BK)^t\| \leq \kappa (1-\delta)^t
    \qquad \text{for all } t\ge 0,
\]
and moreover \(\kappa/\delta\le \gamma\), then \(K\) is a $\gamma$-stabilizing controller for \((A,B)\).
\end{lemma}

\begin{proof}
Let ${\tilde{K} = A+BK}$. Under the linear controller \(u_t=Kx_t\), the dynamics are
\[
    \x_{t+1} = \tilde{K}\x_t + \w_t.
\]
Unrolling the recursion gives
\[
    \x_{t+1}
    =
    \tilde{K}^{t+1}\x_0
    + \sum_{s=0}^{t}\tilde{K}^{t-s}\w_s.
\]
Define an augmented disturbance sequence by \(v_{-1}=\x_0\) and \(v_s=\w_s\) for \(s\ge 0\). Then
\[
    \x_{t+1}
    =
    \sum_{s=-1}^{t}\tilde{K}^{\,t-s} v_s.
\]
Therefore, whenever \(\|\x_0\|\le 1\) and \(\|\w_s\|\le 1\) for all \(s\),
\begin{align*}
\|\x_{t+1}\|
&\le \sum_{s=-1}^{t}\left\|\tilde{K}^{\,t-s}\right\| \|v_s\| \\
&\le \kappa \sum_{i=0}^{t+1}(1-\delta)^i \\
&\le \frac{\kappa}{\delta}
\le \gamma.
\end{align*}
Thus \(K\) is \(\gamma\)-stabilizing.
\end{proof}

A linear time-invariant system is thus $\gamma$-stabilizable if we can find a matrix \(K\) with the above properties.

\subsection{Controllability of Time-Invariant Systems}

Consider the following one-dimensional dynamical system for some small \(\varepsilon>0\):
\[
    x_{t+1} = x_t + \varepsilon u_t.
\]
Clearly, such a system is nominally controllable, as is indeed any one-dimensional system with nonzero \(B\). However, reaching a state \(x=1\) from an initial state \(x_0=0\) in a constant number of time steps requires a large control input \(u_t\) of magnitude at least \(1/\varepsilon\). Thus, such a system may be said to be barely controllable. This difficulty is captured by the definition below, which requires not only that \(K_k(A,B)\) has full row rank, but also that the corresponding Gram matrix be well conditioned.

\begin{definition}
\label{def:strongcon}
A time-invariant linear dynamical system \((A,B)\) is \((k,\kappa)\)-strongly controllable if and only if
\[
    \rank {K}_k({A}, {B}) = d_x,
    \qquad
    \left\| ({K}_k(A,B){K}_k(A,B)^\top)^{-1}\right\| \leq \kappa.
\]
\end{definition}

In the above definition, in addition to the full-row-rank constraint, the spectral norm of the inverse of the Gram matrix of \(K_k(A,B)\) is bounded by \(\kappa\). For example, in the simple one-dimensional system above with \(k=1\), the matrix \(K_1(A,B)\) is just \(\varepsilon\), and thus \(\kappa = 1/\varepsilon^2\), which indeed captures the complexity of controlling the system.

\ifarxiv
\newpage
\fi
\section{Bibliographic Remarks}

The idea of reducing a nonlinear dynamical system to a time-varying linear dynamical system has appeared independently in several fields. In the context of control, using linearization to characterize local stability properties of nonlinear systems was pioneered by Lyapunov \citep{lyapunov1992general}, and is called the Lyapunov direct method; see, e.g., \citet{slotine1991applied}.

The asymptotic notions of stability, stabilizability, and controllability discussed in this chapter constitute a core part of linear control theory, and are found in most texts on the subject (e.g. \citet{stengel1994optimal, hespanha2018linear}).

The efficient verification of stabilizability via a semidefinite program, together with the recovery of a stabilizing controller from a solution to that program, appears in \citet{aaac}. On the other hand, it is known that even imposing norm constraints on a plausibly stable controller makes this problem computationally hard \citep{blondel1997np,blondel2000survey,blondel1999three}. Somewhat distinct but related problems of stabilizability with even a 1-dimensional linear constraint remain open to this day.

The notion of \(\gamma\)-stabilizability has a precursor in the notion of strong stability introduced in \citet{cohen2018online}. An extension of this strong stability concept to time-varying linear dynamics was established in \citet{cohen2019learning}. The definition presented in this chapter has weaker requirements than those in the cited articles, allowing us to work with and guarantee regret later with respect to a larger class of comparator policies.

\ifarxiv
\newpage
\fi
\begin{exercises}

\begin{exercise}$^*$
\label{exer:necessary_K}
Prove that the existence of a matrix \(K\) such that \(\rho(A+BK)<1\) is a necessary condition for a linear dynamical system to be stabilizable. Hint: this exercise requires the optimal solution to LQR in the next chapter.
\end{exercise}

\begin{exercise}
\label{exer:limit_specrtal}
Prove that
\[
    \rho(A) < 1
    \quad \text{if and only if} \quad
    \lim_{t \to \infty} A^t = 0.
\]
Furthermore, if \(\rho(A) > 1\), then for any norm,
\[
    \lim_{t \to \infty} \|A^t\| = \infty.
\]
Finally, prove that
\[
    \rho(A) = \lim_{t\to \infty} \|A^t\|^{1/t}.
\]
\end{exercise}

\begin{exercise}
\label{exer:radius}
Prove that the spectral radius of a product of matrices is not necessarily bounded by the product of the spectral radii of the individual matrices. That is, show that in general,
\[
    \rho\!\left(\prod_{i=1}^n A_i\right) \nleq \prod_{i=1}^n \rho(A_i).
\]
\end{exercise}

\begin{exercise}
Assume \(\rho(A)<1\). Prove that the stochastic linear dynamical system with Gaussian perturbations
\[
    \x_{t+1} = A \x_t + \w_t,
    \qquad
    \w_t \sim \mathcal{N}(0,I),
\]
constitutes a reversible Markov chain if and only if \(A\) is symmetric, i.e. \(A=A^\top\).

{\it Hint:} One way to do this exercise is to explicitly calculate the stationary distribution of the chain and then verify detailed balance directly.
\end{exercise}

\begin{exercise}
\label{exer:bibo-implies-exp-convergence}
This exercise explores the consequences of \(\gamma\)-BIBO stability (Definition \ref{def:gamma-stabilizing}) for linear systems. Although we have already seen sufficient conditions to ensure BIBO stability, we ask for the necessity of such conditions.

\begin{subexercise}
	For a time-invariant linear system \((A,B)\) equipped with a linear policy \(K\), prove that BIBO stability implies \(\rho(A+BK)<1\). In fact, comment on how large \(\rho(A+BK)\) can be.
\end{subexercise}

\begin{subexercise}
	For a time-invariant BIBO-stable linear system, prove that there exist \(\kappa>0\) and \(\delta\in (0,1]\) such that, for all \(t\),
    \[
        \|(A+BK)^t\|\leq \kappa(1-\delta)^t.
    \]
    In particular, this proves that the conditions listed in Lemma~\ref{lem:ltistable} are necessary, in addition to being sufficient.
\end{subexercise}

\end{exercise}

\begin{exercise}
Prove that the strict condition \(\|A\|<1\) on the operator norm of \(A\) is sufficient, but not necessary, for stabilizability.
\end{exercise}

\end{exercises}

\chapter{Optimal Control of Linear Dynamical Systems}  \label{chap:optimalcontrol}
\chaptermark{Optimal Control}

In the chapters thus far, we have studied more basic notions than \emph{optimal} control, namely structural properties of dynamical systems such as stabilizability and controllability. Even for these more basic problems, we have seen that they are computationally intractable in general.

This motivated us to look into simpler dynamics, namely linear dynamical systems, for which we can at least answer basic questions about stabilizability and controllability. In this chapter, we consider how to control linear dynamical systems optimally.

The answer is positive in a strong sense.
An archetypal problem in optimal control theory is the control of linear dynamical systems with quadratic cost functions. This setting is called the \emph{Linear Quadratic Regulator}, which we define and study in this chapter.
The resulting efficient and practical algorithm is one of the most widely used subroutines in control.

\section{The Linear Quadratic Regulator}

The linear quadratic regulator (LQR) problem is that of optimal control of a linear dynamical system with known and fixed quadratic costs. For simplicity, we mostly consider the time-invariant version, where the dynamics and the costs do not vary with time.

\begin{definition}
The finite-horizon (time-invariant) Linear Quadratic Regulator (LQR) problem consists of a linear dynamical system evolving according to
\[
    \x_{t+1} = A \x_t + B \uv_t + \w_t,
    \qquad t=1,\dots,T,
\]
where the perturbations \(\w_t \in \reals^{d_x}\) are i.i.d. with
\[
    \mathbb{E}[\w_t]=0,
    \qquad
    \mathbb{E}[\w_t \w_t^\top]=\Sigma.
\]
Given positive definite matrices \(Q \in \reals^{d_x \times d_x}\) and \(R \in \reals^{d_u \times d_u}\), the goal is to choose policies \(\pi_1,\dots,\pi_T\), where \(\uv_t=\pi_t(\x_t)\), that minimize the expected quadratic cost
\[
    \min_{\pi_1,\dots,\pi_T}
    \mathbb{E}\!\left[
        \sum_{t=1}^{T}
        \x_t^{\top} Q \x_t + \uv_t^{\top} R \uv_t
    \right].
\]
\end{definition}

To remind the reader, the notation we use is as follows.
\begin{enumerate}
    \item \(\x_t \in \reals^{d_x}\) is the state.
    \item \(\uv_t \in \reals^{d_u}\) is the control input.
    \item \(\w_t \in \reals^{d_x}\) is the perturbation sequence, assumed i.i.d. with zero mean and covariance \(\Sigma\). A Gaussian disturbance is a common special case, but Gaussianity is not essential for the derivation below.
    \item \(A \in \reals^{d_x \times d_x}\) and \(B \in \reals^{d_x \times d_u}\) are the system matrices.
    \item \(Q \in \reals^{d_x \times d_x}\) and \(R \in \reals^{d_u \times d_u}\) are positive definite matrices. Thus, the stage costs are strictly convex in the control.
\end{enumerate}

This framework is very useful, since many physical problems, such as the centrifugal governor and the ventilator control problem, can be approximated by a linear dynamical system.
In addition, quadratic convex cost functions are natural in many physical applications.

The LQR is a special case of a Markov decision process, which we described in Chapter~\ref{chap:RL}, since the transitions are stochastic. However, we cannot naively apply value iteration, policy iteration, or linear programming since there are infinitely many states and actions.

We proceed to describe a particularly elegant solution to the LQR problem that makes use of its special structure.

\section{Optimal Solution of the LQR}

We know from Chapter~\ref{chap:RL} that the LQR is a special case of a finite-horizon Markov decision process and hence admits an optimal policy. However, the optimal policy is in general a mapping from states to actions, both of which are infinite Euclidean spaces in our case.

It turns out that the optimal policy for an LQR is a \textbf{linear policy}, that is, a linear mapping from \(\reals^{d_x}\) to \(\reals^{d_u}\). This remarkable fact, together with a corresponding quadratic form for the value function, is given in the following theorem.
We state the theorem for a finite horizon and a linear time-invariant system. Generalizing this to a time-varying linear system is left as Exercise~\ref{exer:gen_LDS_solution}.

\begin{theorem}\label{thm:lds_opt_solution}
For the finite-horizon LQR problem, there exist symmetric matrices \(S_t \succeq 0\), gains \(K_t \in \reals^{d_u \times d_x}\), and scalars \(c_t \in \reals\) such that, for every stage \(t\),
\[
    \bv_t^*(\x)=\x^{\top} S_t \x + c_t,
    \qquad
    \uv_t^*(\x)=K_t \x .
\]
\end{theorem}

\begin{proof}
For \(t=1,\dots,T\), let \(\bv_t^*(\x)\) denote the optimal cost-to-go from stage \(t\) onward when \(\x_t=\x\), and set \(\bv_{T+1}^*(\x)\equiv 0\).
We prove the claim by backward induction on \(t\).

\noindent\textbf{\textit{Base Case:}}
For \(t=T\), we have
\[
    \bv_T^*(\x)
    =
    \min_{\uv}
    \left\{
        \x^\top Q \x + \uv^\top R \uv
    \right\}.
\]
Since \(R\succ 0\), the unique minimum occurs at \(\uv=0\). Hence
\[
    S_T = Q,
    \qquad
    K_T = 0,
    \qquad
    c_T = 0.
\]

\noindent\textbf{\textit{Inductive Step:}}
Assume for some \(t>1\) that
\[
    \bv_t^*(\x)=\x^\top S_t \x + c_t
\]
for a symmetric matrix \(S_t \succeq 0\).
Then
\begin{align*}
    \bv_{t-1}^*(\x)
    &=
    \min_{\uv}
    \left\{
        \x^\top Q \x + \uv^\top R \uv
        + \mathbb{E}_{\w}\!\left[\bv_t^*(A\x+B\uv+\w)\right]
    \right\} \\
    &=
    \min_{\uv}
    \left\{
        \x^\top Q \x + \uv^\top R \uv
        + \mathbb{E}_{\w}\!\left[(A\x+B\uv+\w)^\top S_t (A\x+B\uv+\w)\right]
    \right\}
    + c_t.
\end{align*}
Using \(\mathbb{E}[\w]=0\) and \(\mathbb{E}[\w\w^\top]=\Sigma\), we obtain
\[
    \mathbb{E}_{\w}\!\left[(A\x+B\uv+\w)^\top S_t (A\x+B\uv+\w)\right]
    =
    (A\x+B\uv)^\top S_t (A\x+B\uv)
    + \trace(S_t \Sigma).
\]
Therefore,
\[
    \bv_{t-1}^*(\x)
    =
    \min_{\uv}
    \left\{
        \x^\top Q \x + \uv^\top R \uv
        + (A\x+B\uv)^\top S_t (A\x+B\uv)
    \right\}
    + c_t + \trace(S_t \Sigma).
\]

Since \(R\succ 0\) and \(S_t \succeq 0\), the expression inside the braces is a strictly convex quadratic function of \(\uv\). Its gradient is
\[
    2(R+B^\top S_t B)\uv + 2B^\top S_t A \x.
\]
Setting the gradient to zero yields
\[
    (R+B^\top S_t B)\uv = - B^\top S_t A \x.
\]
Since \(R\succ 0\), the matrix \(R+B^\top S_t B\) is positive definite and therefore invertible. Hence the optimal control is
\[
    \uv_{t-1}^*(\x) = K_{t-1}\x,
    \qquad
    K_{t-1}=-(R+B^\top S_t B)^{-1}B^\top S_t A.
\]

Substituting this back into the value function gives
\begin{align*}
    \bv_{t-1}^*(\x)
    &=
    \x^\top Q \x
    + \x^\top K_{t-1}^\top R K_{t-1}\x \\
    &\qquad\qquad
    + \x^\top (A+BK_{t-1})^\top S_t (A+BK_{t-1})\x
    + c_t + \trace(S_t \Sigma) \\
    &=
    \x^\top S_{t-1}\x + c_{t-1},
\end{align*}
where
\[
    S_{t-1}
    =
    Q
    + K_{t-1}^\top R K_{t-1}
    + (A+BK_{t-1})^\top S_t (A+BK_{t-1}),
\]
and
\[
    c_{t-1}=c_t+\trace(S_t \Sigma).
\]
Since each term in the definition of \(S_{t-1}\) is positive semidefinite, we have \(S_{t-1}\succeq 0\). This completes the induction.
\end{proof}

\section{Infinite Horizon LQR}

Although real control tasks always have finite duration, in practice the horizon length may be very large or unknown. This motivates the infinite-horizon formulation, which leads to an especially elegant stationary controller.

In this section we consider the deterministic infinite-horizon total-cost problem
\begin{align*}
    \min_{\pi} \quad & \sum_{t=1}^{\infty} \x_t^{\top} Q \x_t + \uv_t^{\top} R \uv_t \\
    \text{s.t.} \quad & \x_{t+1}=A \x_t+B \uv_t,
\end{align*}
where \(\uv_t=\pi(\x_t)\). We assume throughout that \(Q\succ 0\), \(R\succ 0\), and that the pair \((A,B)\) is stabilizable.

Persistent additive noise is more naturally treated with an average-cost formulation rather than a total-cost one; see Exercise~\ref{exer:infinite-horizon-lqr-with-noise}.

Under the above assumptions, the optimal policy is stationary and linear, and the value function has quadratic form.
Writing the Bellman optimality equation for the total cost, we obtain
\[
    \bv^{*}(\x)
    =
    \min_{\uv}
    \left\{
        \x^\top Q \x + \uv^\top R \uv + \bv^{*}(A \x + B \uv)
    \right\}.
\]

Assuming that the value function is quadratic, say \(\bv^*(\x)=\x^\top S \x\), we get
\[
    \x^\top S \x
    =
    \min_{\uv}
    \left\{
        \x^\top Q \x
        + \uv^\top R \uv
        + (A \x + B \uv)^\top S (A \x + B \uv)
    \right\}.
\]

Since the objective is strictly convex in \(\uv\), the optimum is characterized by a vanishing gradient:
\[
    2R\uv + 2B^\top S B \uv + 2B^\top S A \x = 0.
\]
Hence the optimal policy is linear and given by
\[
    \uv^{*}(\x)=K\x,
    \qquad
    K = -(R+B^\top S B)^{-1} B^\top S A.
\]

Plugging this back into the Bellman equation yields
\[
    \x^\top S \x
    =
    \x^\top Q \x
    + \x^\top K^\top R K \x
    + \x^\top (A+BK)^\top S (A+BK) \x.
\]
Since this equality holds for all \(\x\), we obtain
\[
    S
    =
    Q + K^\top R K + (A+BK)^\top S (A+BK).
\]

Simplifying this expression gives
\[
    S
    =
    Q + A^\top S A
    - A^\top S B \left(R+B^\top S B\right)^{-1} B^\top S A.
\]
This equation is called the \textbf{Discrete-Time Algebraic Riccati Equation} (DARE) and is one of the most important equations in control theory. Its stabilizing solution determines the optimal infinite-horizon controller.

\subsection{Solving the DARE}

In the one-dimensional case, although the right-hand side of the equation is a rational function, we can multiply both sides by \(R+B^\top S B\) to form a quadratic equation in \(S\). Solving the DARE therefore corresponds to finding a nonnegative root of a quadratic polynomial. We only consider the nonnegative root because the value function is nonnegative by our choice of quadratic convex costs.

In higher dimensions, a convenient convex formulation is obtained from the associated Riccati inequality. By the Schur complement, the matrix inequality
\[
    \begin{bmatrix}
        Q + A^\top S A - S & A^\top S B \\
        B^\top S A & R + B^\top S B
    \end{bmatrix}
    \succeq 0
\]
is equivalent to
\[
    S
    \preceq
    Q + A^\top S A
    - A^\top S B \left(R+B^\top S B\right)^{-1} B^\top S A.
\]
Under the standard stabilizability assumptions, the maximal positive semidefinite solution of this Riccati inequality is precisely the stabilizing solution of the DARE. Therefore, one can recover \(S\) via the semidefinite program
\[
    \max_{S \in \mathcal{S}^{d_x\times d_x}} \langle S, I\rangle
\]
subject to
\[
    \begin{bmatrix}
        Q + A^\top S A - S & A^\top S B \\
        B^\top S A & R + B^\top S B
    \end{bmatrix}
    \succeq 0,
    \qquad
    S \succeq 0.
\]

A popular alternative is the \textbf{Riccati iteration}: starting from some \(S_0 \succeq 0\), define
\[
    S_{t+1}
    =
    Q + A^\top S_t A
    - A^\top S_t B \left(R+B^\top S_t B\right)^{-1} B^\top S_t A.
\]
Under the same assumptions, this sequence converges to the stabilizing solution of the DARE.

\section{Robust Control}

The solution to the LQR problem is a hallmark of optimal control theory and one of the most widely used techniques in control. However, the assumption that the disturbances are stochastic and i.i.d. is a strong one, and numerous attempts have been made over the years to generalize the noise model.

One of the most important and prominent generalizations is robust control, also known as \(H_\infty\)-control. The goal of \(H_\infty\)-control is to come up with a more robust solution by comparing the controller against an adversary that chooses the disturbance sequence.

A standard finite-horizon formulation fixes a robustness parameter \(\gamma>0\) and considers
\[
\begin{array}{cl}
\min_{\pi_1,\dots,\pi_T} \max_{\psi_1,\dots,\psi_T}
&
\displaystyle
\sum_{t=1}^{T}
\left(
    \x_t^\top Q \x_t
    + \uv_t^\top R \uv_t
    - \gamma^2 \w_t^\top \w_t
\right)
\\[1em]
\text{s.t.}
&
\x_{t+1}=A_t \x_t + B_t \uv_t + \w_t, \\
&
\uv_t=\pi_t(\x_t),
\qquad
\w_t=\psi_t(\x_t).
\end{array}
\]
The parameter \(\gamma\) controls the tradeoff between nominal performance and robustness.

Notice that this problem does not fall into the category of a Markov decision process, since the disturbances are no longer stochastic. Instead, the appropriate techniques come from the theory of dynamic games. The above problem is a two-player zero-sum game: the controller tries to minimize the cost, while the adversary tries to maximize it, both subject to the same dynamical constraint.

Using techniques from dynamic games, one can show the following. References are provided in the bibliographic section.

\begin{theorem}\label{thm:robust}
Under the standard solvability conditions for finite-horizon \(H_\infty\) control, the above dynamic game admits a saddle point in linear state-feedback strategies. In particular, there exist matrices \(K_t\) and \(L_t\) such that
\[
    \uv_t = K_t \x_t,
    \qquad
    \w_t = L_t \x_t.
\]
\end{theorem}

The proof is somewhat more involved than the LQR derivation and is omitted from this text.

\ifarxiv
\newpage
\fi
\section{Bibliographic Remarks}
Optimal control is considered to be the hallmark of control theory, and numerous excellent introductory texts cover it in detail, e.g.~\citet{stengel1994optimal, hespanha2018linear}.

The LQR problem and its solution were introduced in the seminal work of \ifarxiv Kalman \fi \citet{kalman1960new}, who also introduced the eponymous filtering procedure for linear estimation.

While the finite-horizon derivation above follows the spirit of dynamic programming for Markov decision processes, the infinite-horizon theory is subtler because of the continuous nature of the state space and the role of stabilizability. A discussion of sufficient and necessary conditions for the existence of stabilizing solutions to Riccati equations can be found in \citet{bertsekas2007dynamic, hassibi1996linear}.

The development of \(H_\infty\) control was driven in part by the observation, made in the seminal paper of \ifarxiv Doyle \fi \citet{doyle1978guaranteed}, that optimal controllers need not be robust to misspecification of system parameters. This suggests that robustness specifications constitute yet another axis of performance for control algorithms and may at times be at odds with the objective of optimal control. For in-depth treatment of \(H_\infty\) control, see \citet{kemin}. The text \citet{bacsar2008h} presents a succinct derivation of robust controllers entirely in the state space, as the solution to a dynamic game between the controller and an adversary; see also \citet{TuBlog} for a recent exposition.

Although developed in the linear setting, adaptations of the LQR controller are also widely used to control nonlinear systems \citep{todorov2005generalized, levine2013guided}.

\ifarxiv
\newpage
\fi
\begin{exercises}

\begin{exercise}\label{exer:gen_LDS_solution}
Prove Theorem~\ref{thm:lds_opt_solution} for a time-varying linear dynamical system.
\end{exercise}

\begin{exercise}\label{exer:infinite-horizon-lqr-with-noise}
Formulate the infinite-horizon LQR with i.i.d.\ zero-mean Gaussian noise as an average-cost control problem, and derive the corresponding Bellman equation and optimal linear policy.
\end{exercise}

\begin{exercise}
Extend the proof of Theorem~\ref{thm:lds_opt_solution} to the case \(Q \succeq 0\) and \(R \succ 0\).
\end{exercise}

\begin{exercise}
Prove Theorem~\ref{thm:robust} by formulating the robust control problem above as a finite-horizon dynamic two-player zero-sum game.
\end{exercise}

\begin{exercise}
Prove that the function \(f(\x) = \x^\top M \x\) is convex if and only if the symmetric part \(\frac{1}{2}(M+M^\top)\) is positive semidefinite.
\end{exercise}

\end{exercises}

\part{Basics of Online Control}
\chapter{Regret in Control} \label{chap:policy-classes}
In the first part of this book and in chapter \ref{chap:optimalcontrol} we studied linear dynamical systems with quadratic costs, namely the LQR problem. We saw that in this case the optimal control takes a very special form: it is a linear function of the state. 
The Markovian property of the Markov Decision Process framework for reinforcement learning suggests that the optimal policy is always a function of the current state. It is characterized by the Bellman optimality equation, and can be computed in complexity proportional to the number of states and actions using, for example, the value iteration algorithm, which we analyzed in a previous chapter.  

However, this changes when we deviate from the MDP framework, which may be warranted for a variety of reasons:
\begin{enumerate}
    \item The state is not fully observable.
    \item The reward functions change online. 
    \item The transition mapping is not fixed and changes online. 
    \item The transitions or costs are subject to adversarial noise. 
\end{enumerate}

In such scenarios involving significant deviations from the MDP framework, classical notions of optimality may be ill-defined. Or if they do exist, the optimal policy may no longer be a mapping from immediate state to action. Rather, the learner may be required to use other signals available to her when choosing an action that is most beneficial in the long run. 
To this we add a significant consideration: the computation of optimal {\bf state-based policies may be computationally intractable}. 

This motivates the current chapter, which considers regret analysis as a computationally tractable alternative to optimal or robust control. 

We start by considering regret in the decision-making framework of online convex optimization. We then consider several classes of control policies and the relationships between them in the context of regret analysis.

\section{Online Convex Optimization} \label{chap:appendix-oco}

In this section, we give a brief introduction to the theory of Online Convex Optimization. The reader is referred to the bibliographic section for a comprehensive survey. 

In the decision-making framework of Online Convex Optimization, a player iteratively chooses a point in a convex set $\K \subseteq \reals^n$, and suffers a loss according to an adversarially chosen loss function $f_t: \K \mapsto \reals$. 

Let $\mA$ be an algorithm for OCO, which maps a certain game history to a decision in the decision set: 
$$ \x_t^\mA = \mA(f_1,...,f_{t-1}) \in \K . $$
We define the regret of $\mA$ after $T$ iterations as: 
\begin{equation} \label{eqn:regret-defn}
\regret_T(\mA) = \sup_{\{f_1,...,f_T\} \subseteq \F}  \left\{   \sum_{t=1}^T f_t(\x_t^\mA) -\min_{\x \in \K} \sum_{t=1}^T f_t(\x) \right\} .
\end{equation}

If the algorithm is clear from the context, we henceforth omit the superscript and denote the algorithm's decision at time $t$ simply as $\x_t$. 
Intuitively, an algorithm performs well if its
regret is sublinear as a function of $T$ (i.e.
$\regret_T(\mA) = o(T)$), since this implies that, on
average, the algorithm performs as well as the best fixed strategy
in hindsight.
We then describe a basic algorithm for online convex optimization.

\subsection{Online Gradient Descent} \label{section:ogd}

Perhaps the simplest algorithm that applies to the most general setting of online convex optimization is Online Gradient Descent (OGD).
The pseudocode is given in Algorithm \ref{alg:ogd}.

\begin{algorithm}[ht]
		\caption{\label{alg:ogd} Online Gradient Descent }
		\begin{algorithmic}[1]
			\STATE Input: convex set $\K$, $T$, $\x_1 \in \mathcal{K}$, step sizes $\{ \eta_t \}$
			\FOR {$t=1$ to $T$}
			\STATE Play $\x_t$ and observe cost $f_t(\x_t)$. 
			\STATE Update and project:
			\begin{align*}
			& \y_{t+1} = \bx_{t}-\eta_{t} \nabla f_{t}(\bx_{t}) \\
			& \bx_{t+1} = \argmin_{\x \in \K} \|\x - \y_{t+1}\|^2 = \prod_\K[\y_{t+1}]
			\end{align*}
			\ENDFOR
		\end{algorithmic}
\end{algorithm}

In each iteration, the algorithm takes a step from the previous point in the direction of the gradient of the previous cost.  This step may result in a point outside the underlying convex set. In such cases,
the algorithm projects the point back to the convex set, i.e.
finds its closest point in the convex set.
Despite the fact that the next cost function may be completely
different from the costs observed thus far, the regret
attained by the algorithm is sublinear. This is formalized in the following
theorem. Here we define the constants $D$ and $G$ as follows: 
\begin{itemize}
    \item $D$ = Euclidean diameter of the decision set, i.e. $$\max_{\x,\y \in \K} \|\x-\y\| . $$
    
    \item $G$ = an upper bound on the Lipschitz constant of the cost functions $f_t$, or equivalently, the subgradient norms of these functions. Mathematically, 
    $$ \forall t \ , \ \|\nabla f_t(\x_t)\| \leq G .$$
\end{itemize}

\begin{theorem}\label{thm:ogd}
Online gradient descent with step sizes $\{\eta_t = \frac{D}{G \sqrt{t}} , \ t \in [T] \}$ guarantees the following for all $T \geq 1$:
$$ \regret_T = \sum_{t=1}^{T} f_t(\bx_t) -\min_{\bx^\star \in \K}\sum_{t=1}^{T}
f_t(\bx^\star)\ \leq  \frac{3}{2} {G D}\sqrt{T} . $$
\end{theorem}

\begin{proof}
Let $\bx^\star \in \argmin_{\bx \in \K} \sum_{t=1}^T f_t(\bx)$.
Define $\nabla_t \equaltri \nabla f_{t}(\bx_{t})$. By convexity
\begin{eqnarray}  \label{eqn:gradient_inequality}
f_t(\bx_t) - f_t(\bx^\star) \leq   \nabla_t^\top (\bx_t - \bx^\star)
\end{eqnarray}
We first  upper-bound $\nabla_t^\top (\bx_t-\bx^\star)$ using the update
rule for $\bx_{t+1}$ and the Pythagorean Theorem, we have the projections only decrease the distance to the set $\K$:
\begin{equation} \label{eqn:ogdtriangle}
\| \by_{t+1}-\bx^\star \|^2\ =\  \left\|\proj_\K (\bx_t - \eta_t
\nabla_{t}) -\bx^\star\right\|^2 \leq  \left\|\bx_t - \eta_t \nabla_t-\bx^\star\right\|^2 .
\end{equation}

Hence,
\begin{eqnarray} \label{eqn:ogd_eq2}
\|\by_{t+1}-\bx^\star\|^2\ &\leq&\ \|\bx_t- \bx^\star\|^2 + \eta_t^2
\|\nabla_t\|^2 -2 \eta_t \nabla_t^\top (\bx_t -\bx^\star)\nonumber\\
2 \nabla_t^\top (\bx_t-\bx^\star)\ &\leq&\ \frac{ \|\bx_t-
\bx^\star\|^2-\|\bx_{t+1}-\bx^\star\|^2}{\eta_t} + \eta_t G^2 .
\end{eqnarray}
Summing \eqref{eqn:gradient_inequality} and \eqref{eqn:ogd_eq2}  from $t= 1$ to $T$, and setting $\eta_t =
\frac{D}{G \sqrt{t}}$ (with $\frac{1}{\eta_0} \equaltri 0$):
\begin{align*}
& 2 \left( \sum_{t=1}^T f_t(\bx_t)-f_t(\bx^\star) \right ) \leq 2\sum_{t=1}^T \nabla_t^\top (\xv - \x^\star) \\
&\leq  \sum_{t=1}^T \frac{ \|\bx_t-
	\bx^\star\|^2-\|\bx_{t+1}-\bx^\star\|^2}{\eta_t} + G^2 \sum_{t=1}^T \eta_t    \\
&\leq  \sum_{t=1}^T \|\bx_t - \bx^\star\|^2 \left(
\frac{1}{\eta_{t}} -
\frac{1}{\eta_{t-1}} \right) + G^2 \sum_{t=1}^T \eta_t & \frac{1}{\eta_0} \equaltri 0, \\
& &  \|\xv[T+1] - \xv[]^* \|^2 \geq 0 \\
&\leq D^2 \sum_{t=1}^T \left(
\frac{1}{\eta_{t}} -
\frac{1}{\eta_{t-1}} \right) + G^2 \sum_{t=1}^T \eta_t \\
& \leq  D^2  \frac{1}{\eta_{T}}  + G^2 \sum_{t=1}^T \eta_t  & \mbox{ telescoping series } \\
& \leq 3 DG \sqrt{T}.
\end{align*}
The last inequality follows since $\eta_t = \frac{D}{G \sqrt{t}}$ and $\sum_{t=1}^T \frac{1}{\sqrt{t}} \leq 2 \sqrt{T}$.
\end{proof}

The Online Gradient Descent algorithm is straightforward to implement, and updates
take linear time given the gradient. However, in general, the
projection step may take significantly longer via convex optimization.

\section{Regret for Control}

The most important consideration to explore beyond classical models of control, such as optimal and robust control, is computational. Even in linear control with convex loss functions, the optimal policy can be complex to describe, see the exercises section. 

A prominent approach to deal with computational difficulty is to consider a less stringent criterion for optimality. Instead of the optimal policy, we can consider all policies in a certain {\it policy class}, and try to compete with the optimal policy within the class in terms of regret.  

As alluded to earlier, the most natural classes of policies are computationally intractable. We thus resort to learning new classes that are computationally favorable. But to relate the performance of our methods, we require a language of comparing different policy classes. 
It is crucial to be able to relate the representation power of different policy classes. 
This motivates our main comparison metric as follows.

\begin{definition} \label{defn:regret-of-policies}
We say that a class of policies $\Pi_1$ $\eps$-approximates class $\Pi_2$ if the following is satisfied: for every 
$\pi_2 \in \Pi_2$, there exists $\pi_1 \in \Pi_1$ such that  for all sequences of $T$ disturbances and cost functions, it holds that
$$ \sum_{t=1}^T \left| c_t(\x_t^{\pi_1},\uv_t^{\pi_1}) - c_t(\x_t^{\pi_2},\uv_t^{\pi_2})  \right| \leq T \eps .$$
\end{definition}

The significance of this definition is that if a simple policy class approximates another more complicated class, then it suffices to consider only the first.  The usefulness of this definition comes from its implications for regret minimization, according to definition \ref{defn:regret}, as follows. 
\begin{lemma} \label{lem:regret_implication}
Let policy class $\Pi_1$ $\eps$-approximate policy class $\Pi_2$. Then for any control algorithm $\mA$ we have 
$$ \regret_T(\mA,\Pi_2) \leq  \regret_T(\mA,\Pi_1) + \eps T . $$
\end{lemma}
\begin{proof}
Recall that for policy class $\Pi$ and algorithm $\mathcal{A}$, 
$$\regret_T(\mathcal{A}, \Pi) = \max_{\mathbf{w}_{1:T}: \|\mathbf{w}_t\| \leq 1} \left(\sum_{t=1}^T c_t(\mathbf{x}_t, \mathbf{u}_t) - \min_{\pi \in \Pi} \sum_{t=1}^T c_t(\mathbf{x}_t^\pi, \mathbf{u}_t^\pi)\right)$$

It follows from definition \ref{defn:regret-of-policies} that 
\begin{align*}
\min_{\pi_2 \in \Pi_2}\sum_{t=1}^{T} c_t(\mathbf{x}_t^{\pi_2} , \mathbf{u}_t^{\pi_2}) & = \sum_{t=1}^{T} c_t(\mathbf{x}_t^{\pi_2^*} , \mathbf{u}_t^{\pi_2^*}) & \pi_2^* = \argmin_{\pi \in \Pi_2} \sum_t c_t (\x^\pi_t,\uv_t^\pi) \\ 
& \ge \sum_{t=1}^{T} c_t(\mathbf{x}_t^{\pi_1}, \mathbf{u}_t^{\pi_1}) - \epsilon T & \mbox{Definition \ref{defn:regret-of-policies}}   \\
& \geq  \min_{\pi_1 \in \Pi_1} \sum_{t=1}^T c_t(\mathbf{x}_t^{\pi_1}, \mathbf{u}_t^{\pi_1}) - \epsilon T
\end{align*}

Thus, for any fixed disturbance sequence $\mathbf{w}_{1:T}$,
\begin{align*}
& \sum_{t=1}^T c_t(\mathbf{x}_t, \mathbf{u}_t) - \min_{\pi_2 \in \Pi_2} \sum_{t=1}^T c_t(\mathbf{x}_t^{\pi_2}, \mathbf{u}_t^{\pi_2}) \\
& \leq \sum_{t=1}^T c_t(\mathbf{x}_t, \mathbf{u}_t) - \left(\min_{\pi_1 \in \Pi_1} \sum_{t=1}^T c_t(\mathbf{x}_t^{\pi_1}, \mathbf{u}_t^{\pi_1}) - \epsilon T\right) \\
& = \left(\sum_{t=1}^T c_t(\mathbf{x}_t, \mathbf{u}_t) - \min_{\pi_1 \in \Pi_1} \sum_{t=1}^T c_t(\mathbf{x}_t^{\pi_1}, \mathbf{u}_t^{\pi_1})\right) + \epsilon T
\end{align*}

Since this holds for any fixed disturbance sequence $\mathbf{w}_{1:T}$, it also holds for the maximum over all disturbance sequences $\mathbf{w}_{1:T}$ with $||\mathbf{w}_t|| \le 1$.
\end{proof}

We now proceed to define and study approximation relationships between different policy classes.

\section{Expressivity of Control Policy Classes}

In this section, we formally relate the power of different complexity classes for linear time-invariant linear dynamical systems. The definitions and results can be extended to time-varying linear dynamical systems, and recall from Chapter \ref{chap:lds} that time-varying linear dynamical systems were proposed by Lyapunov as a general methodology to study nonlinear dynamics. 

For the rest of this section we assume full observation of the state and let 
\begin{align*}
\x_{t+1} &  = A \x_{t} + B \uv_t + \bw_t . 
\end{align*}

\begin{figure}[h] 
    \centering
        \includegraphics[width=0.9\textwidth]{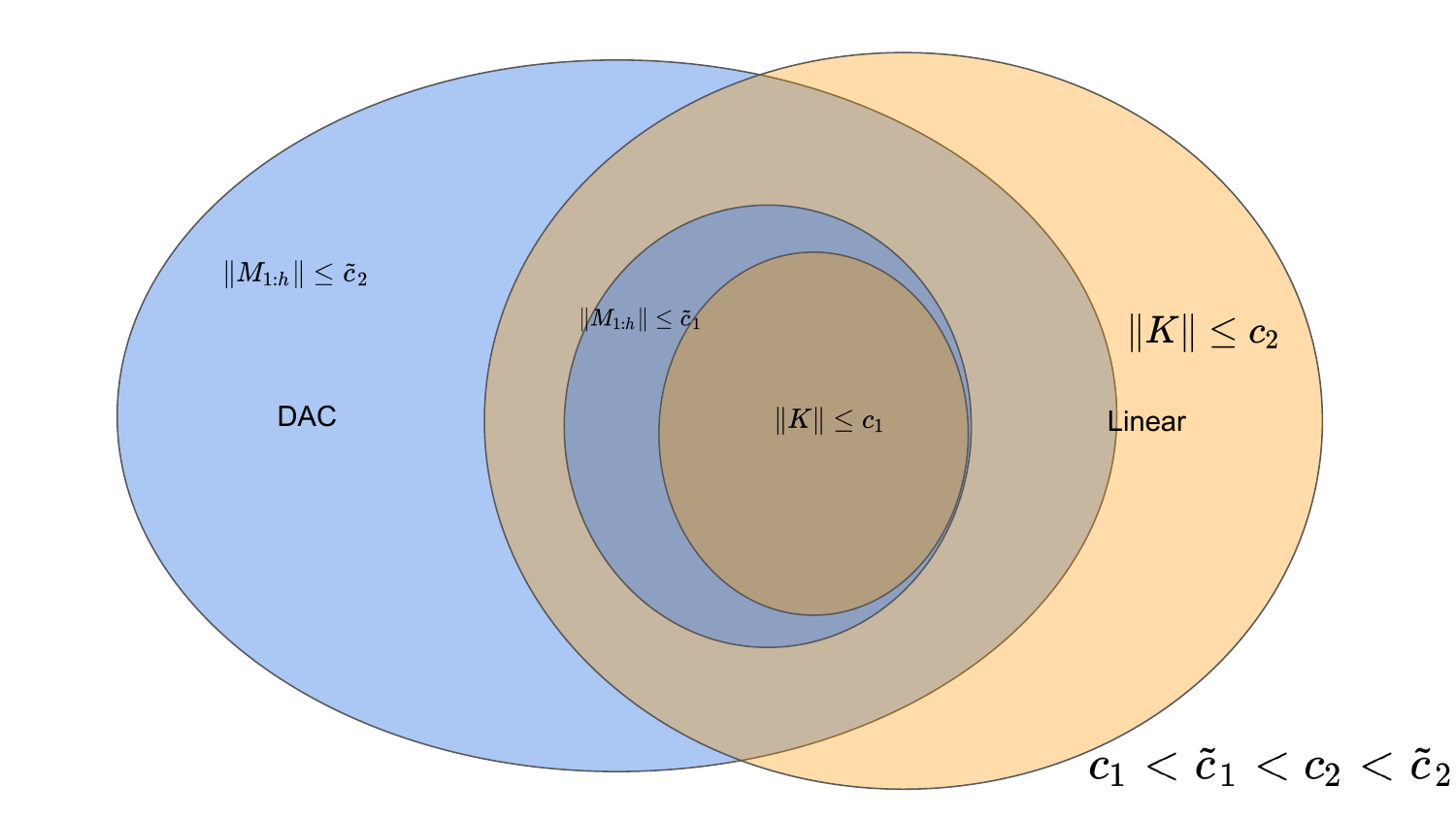} 
        \caption{Schematic relationship between linear Disturbance-Action Control policies and linear state-action policies of increasing diameter for linear time invariant systems.}\label{fig:policies1} 
\end{figure}

A sequence of stabilizing linear controllers for such a system is denoted by $\{K_t\}$. 
Schematically, the relationships we describe below are given in Figure \ref{fig:policies1} for fully observed systems. In later chapters of this text, we study policies for partially observed systems.

\subsection{Linear Policies}

The most basic and well-studied policy classes for control of linear dynamical systems are linear policies, i.e. those that choose a control signal as a (fixed) linear function of the state. The most important subcases of linear policies are those that are $\gamma$-stabilizing, meaning that the signal they create does not cause an unbounded state. This was formalized in Definition \ref{def:gamma-stabilizing}. Formally, 
\begin{definition}[Linear Policies] 
A linear controller $K \in \reals^{d_u \times d_x}$ is a linear operator that maps the state to the control. We denote by $\Pi_{\gamma}^{L}$ the set of all linear controllers that are $\gamma$-stabilizing according to definition \ref{def:gamma-stabilizing} and are bounded in Frobenius norm by $ \| K \| \leq \gamma $. 
\end{definition}

\subsection{Disturbance Action Controllers}

We next define the most important set of policies in online nonstochastic control, known as disturbance action controllers, or DAC. The main advantages of DAC, as opposed to linear controllers, are that the cost of control is convex in their parameterization and thus allows for provably efficient optimization.  Moreover, this set of policies is stronger than linear policies in the sense that it $\eps$-approximates them, as we show in the following. 
\begin{definition}[Disturbance Action Controller]\label{def:dac}
A (linear) disturbance action policy $\pi_{K_{1:T}, M_{1:h}}$ is parameterized by the matrices $M_{0:h} = [M_0,{M}_1, \ldots, {M}_{h}]$ and stabilizing controllers $\{K_t\}$. It outputs control $\mathbf{u}_t^\pi$ at time $t$ according to the rule
\begin{equation*}
\mathbf{u}_t^\pi = {K}_t \mathbf{x}_t + \sum_{i=1}^{h} {M}_{i} \mathbf{w}_{t-i} +M_0.
\end{equation*}
Without loss of generality, we take $M_0 = 0$, as we can add a fictitious unit coordinate to the perturbations $\{ \w_t \} $, and modify the system accordingly. This will allow for a constant shift in the control. 

Denote by $\Pi^{D}_{h,\gamma}$ the set of all disturbance action policies with $K_{1:T}$ being $\gamma$-stabilizing linear controllers for the system and 
$$ \sum_{i=1}^h \| M_i\| \leq \gamma .$$
\end{definition}

The class of DACs always produces a stabilizing control signal whenever $K_{1:T}$ are sequentially stabilizing for the entire system $\{A_t,B_t\}$. This important fact can be seen in the following lemma. 

\begin{lemma} \label{lem:dac-is-stable}
Any DAC policy in $\Pi_{h,\gamma}^{D}$ is $2\gamma^2$-stabilizing.
\end{lemma}
\begin{proof}
For convenience, we define a fictitious noise term $\tilde{\w}_t = B\sum_{i=1}^h M_i \w_{t-i}  $. First, due to the bound on $M_{1:h}$ and $\w_t$, we have that 
$$ \| \tilde{\w}_t\|  = \left\|  \sum_{i=1}^h M_i \w_{t-i}\right\| \leq  \sum_{i=1}^h \|M_i\| \| \w_{t-i}\| \leq \gamma , $$
for $\|M\|$ being the spectral norm of $M$. 

Next, since $K_t$'s are stabilizing, we can bound the historically distant terms as follows. For simplicity, assume that $K_t = K$ are all equal. 
Observe that the state at time $t$, following any DAC policy $\pi$, is given by 
\begin{align*}
\x_{t+1} & = A \x_t + B \uv_t + \w_t \\
& = A \x_t + B \left( K \x_t + \sum_{i=1}^h M_i \w_{t-i} \right) + \w_t \\
& = (A +  B K ) \x_t + B \sum_{i=1}^h M_i \w_{t-i} + \w_t  \\
& = \sum_{j=0}^{t} \left(   (A + B K )^{j}  \left(B \sum_{i=0}^h M_i \w_{t-i-j} + \w_{t-j}\right)\right) \\
& = \sum_{j=0}^{t-1} \left(  (A +  B K)^{j}  (B \tilde{\w}_{t-j} +\w_{t-j})\right) .
\end{align*}
We can thus view the states $\x_t$ as generated by a $\gamma$-stabilizing policy over noises that are of magnitude $2\gamma$, and hence bounded by $2\gamma^2$.
\ignore{
Thus, we have 
\begin{eqnarray*}
\| \x_{t+1}\| & = \| \sum_{j=0}^{t-1} \left( \left[ \prod_{\tau = t-j}^t (\bA_\tau + \bK_\tau \bB_\tau) \right] \sum_{i=0}^H \bM_i \w_{t-i-j} \right)  \| \\
& \leq \sum_{j=0}^t (1-\delta)^j \gamma \leq \frac{\gamma}{1-\delta}
\end{eqnarray*} 
}
\end{proof}

The above lemma is already a good indication about DAC: they generate controls that at least do not cause a blow-up. But how expressive are these policies in terms of reaching the optimal solution? 

One of the main appeals of the DAC class is the following approximation lemma which shows that for time-invariant systems DAC are in fact a stronger class than linear controllers. 
Notice that the $K_t$ terms for the DAC are only assumed to be stabilizing and not necessarily optimal for the actual sequence. 

\begin{lemma}\label{lemma:approx-dac-lin}
Consider a time-invariant linear dynamical system $\{A,B\}$. 
The class $\Pi^{D}_{h,\gamma}$ $\eps$-approximates the class $\Pi^{L}_{\gamma'}$ for
$h = \Omega( \gamma' \log (\frac{\gamma'}{\epsilon}))$ and $\gamma= \gamma'^2 $.
\end{lemma}
\begin{proof}
Disturbance-action policies use a $\gamma$-stabilizing linear policy on which they superimpose linear functions of disturbances -- fix such a policy, and let $K_0$ be this linear stabilizing component. The state sequence thus produced is identical to that obtained by running a DAC with $\mathbf{0}$ as the stabilizing controller on the system $(A+BK_0, B)$. Similarly, the state sequence produced by the execution of a linear policy $K$ on $(A,B)$ is identical to that of the linear policy $K-K_0$ on the system $(A+BK_0, B)$.

\begin{lemma}\label{lem:movesys}
	Let $\x_t^\pi(A,B), \uv_t^\pi(A,B)$ be the state-action sequence produced by a policy $\pi$ when executed on a linear dynamical system $(A,B)$. For any $\varepsilon>0$, if two policies $\pi_1, \pi_2$ satisfy for all $t$ that
	$$ \|\x_t^{\pi_1}(A+BK_0,B)-\x_t^{\pi_2}(A+BK_0,B)\|\leq \varepsilon,$$ $$\|\uv_t^{\pi_1}(A+BK_0,B)-\uv_t^{\pi_2}(A+BK_0,B)\|\leq \varepsilon$$
    when executed on some dynamical system $(A+BK_0, B)$, then when the corresponding policies $\pi'_1$ and $\pi'_2$, that is, $\pi_1$ and $\pi_2$ with $K_0$ superimposed, are executed on the dynamical system $(A,B)$ the iterates guarantee
	\begin{align*}
		 \max \{ &\|\x_t^{\pi'_1}(A,B)-\x_t^{\pi'_2}(A,B)\|,\\
		&\|\uv_t^{\pi'_1}(A,B)-\uv_t^{\pi'_2}(A,B)\| \} \leq (1+\|K_0\|)\varepsilon.
	\end{align*}
\end{lemma} 

With the aid of the above lemma, it will suffice to guarantee that the state-action sequence produced by the linear policy $K$ and an appropriately chosen DAC with the stabilizing part set to $\mathbf{0}$ on an intrinsically $\gamma$-stable system $(A,B)$ are similar enough. To prove this, we map any linear policy $K$ to a disturbance-action policy that approximates it arbitrarily well for some history parameter $h$. Specifically, for a given system $A,B$ and linear policy $K$, let $\pi_K$ be the policy given by
$$ \forall i \in [0,h] \ , \ M_i = K(A+BK)^i . $$

Now consider any trajectory that results from applying the linear policy for the given system. 
Let $\x_{1:T},\uv_{1:T}$ be the states and controls of the linear policy $K$ for a certain trajectory, as affected by a given noise signal. Let $\tilde{\x}_{1:T},\tilde{\uv}_{1:T}$ be the set of states and controls that would have been generated with the same noise sequence, but with the policy $\pi_K$. We first observe that control sequences produced by $K$ and $\pi_K$ are exponentially close.

\begin{align*}
\uv_{t+1} = K  \x_{t+1} & = K(A \x_t + B \uv_t + \w_t) \\
& = K (A + B K) \x_t + K \w_t \\
& =  \sum_{i=0}^t K \left( A + B K \right)^i  \w_{t-i}
\end{align*}

Now, we define $Z_t = \sum_{i=h+1}^t K \left( A + B K \right)^i  \w_{t-i} $, and continue as shown.

\begin{align*}
\uv_{t+1} & =  \sum_{i=0}^h K \left(A + B K \right)^i  \w_{t-i} + Z_t \\
& =  \sum_{i=0}^h M_i  \w_{t-i} + Z_t \\
& =  \tilde{\uv}_{t+1}  + Z_t 
\end{align*}

The magnitude of this residual term $Z_t$ can be bounded, due to the existence of $\frac{\kappa}{\delta} \leq \gamma'$ from Exercise \ref{exer:bibo-implies-exp-convergence}, as 
\begin{align*}
|Z_t| & \leq \left \| \sum_{i=h+1}^{t} K   \left( A + B K \right)^i  \w_{t-i} \right\| \\
& \leq   \sum_{i=h+1}^{t} \kappa (1-\delta)^{i+1} \| K \| \|  \w_{t-i} \| \\
& \le \kappa^2 \sum_{i=h+1}^{\infty} (1-\delta)^{i}  \leq \kappa^2 \int_{i=h}^\infty e^{- \delta i} di\\
& = \kappa^2 \frac{1}{\delta} e^{-\delta h} \le \eps.
\end{align*}
We thus conclude that $\|\uv_{t+1} - \tilde{\uv}_{t+1}\| \leq {\eps}$. 

Similarly, for the corresponding state sequences, first note that 
$$ \x_{t+1} = \sum_{i=0}^t A^i \left(\w_{t-i} + B \uv_{t-i}\right). $$
Therefore, since $\x_t$ and $\tilde{\x}_t$ share the same perturbation sequences as a part of their description, their difference can be bounded as a geometrically weighted sum of the difference in control inputs $\uv_t$ and $\tilde{\uv}_t$ that constitute them. Concretely, we have

\begin{align*}
	\|\x_t - \tilde{\x}_t\| &=  	\left\|\sum_{i=1}^t A^{i-1}B (\uv_{t-i} - \tilde{\uv}_{t-i})\right\|\\
	&\leq 	\left\|\sum_{i=0}^t A^iB\right\| \max_{i\leq t}\|\uv_i - \tilde{\uv}_i\|
	 \leq \frac{\kappa}{\delta}\varepsilon,
\end{align*}
by using the $\gamma$-stability of $(A,B)$.
The approximation in terms of states and controls implies $\eps$-approximation of policy class for any Lipschitz cost sequence, concluding the proof. 

\end{proof}

\ifarxiv
\newpage
\fi
\section{Bibliographic Remarks}

Regret minimization, online learning, and online convex optimization have a rich history and have been studied for decades in the machine learning and game theory literature. The reader is referred to the comprehensive texts of \citet{cesa2006prediction,hazan2016introduction} for a detailed treatment.   

Linear state-feedback policies are a mainstay of control theory in both applications, due to their simplicity, and in theory, due to their optimality in deterministic and stochastic settings, as demonstrated in \citet{kalman1960new}. The methods for computing such policies are often based on dynamic programming. Often motivated by the need to impose constraints beyond optimality, such as robustness, various other policy parameterizations were introduced. An early example is the Youla parameterization, sometimes called Q-parameterization \citep{youla1976modern}. Later, the introduction of system-level synthesis (SLS) \citep{doyle2017system, anderson2019system}, in a similar vein, proved useful for distributed control, especially given its emphasis on state-space representations. For a comparison of these and additional ones such as input-output parameterizations \citep{furieri2019input}, see \citet{zheng2020equivalence,zhang2023relationship}.

The Disturbance Action Control parameterization, introduced in \citet{agarwal2019online}, is similar to system-level synthesis. Like SLS, it is a convex parameterization that is amenable to the tools of convex optimization. However, unlike SLS, it is a more compact representation that has logarithmically many parameters in its natural representation in terms of the desired approximation guarantee against linear policies. This makes it better suited for machine learning.  

The study of regret in the context of control has roots in the work of \citet{abbasi2011regret,abbasi2014tracking}. These works, as well as many that followed \citep{dean2018regret,mania2019certainty,cohen2018online,agarwal2019logarithmic,pmlr-v119-simchowitz20a}, study regret in a stochastic setting, where the perturbations are stochastic.  In such settings, optimal control is well defined. The full power of regret analysis for control, involving both adversarial perturbations, adversarial loss functions, and computational considerations, was initiated in \citet{agarwal2019online}. 
The notion of regret as applied to policy classes was implicit in the work of \citet{even2004experts}. The regret metric when applied to policies is different from the counterfactual notion of policy regret defined by \citet{arora2012online}.

\ifarxiv
\newpage
\fi
\begin{exercises}

\begin{exercise}
Let us say that the total number of time steps $T$ is known in advance. Consider an implementation of online gradient descent where the step size is chosen as $\eta_t = D/G\sqrt{T}$, and therefore is uniform over time. Prove that the resultant regret is bounded by $GD\sqrt{T}$, which improves upon Theorem~\ref{thm:ogd} by a constant factor. In fact, for ease of exposition, we will use this uniform step size scheme going forward.
\end{exercise}

\begin{exercise}\label{exer:rockafellar}
In this exercise, we prove that the optimal solution to a linear control problem can be complex to describe.
\begin{enumerate}
    \item Consider a linear dynamical system in one dimension and loss functions of the form $f_t(\x) = \max\{0, \x - c\}$, for $c_t \in \reals$. Show that the optimal policy in hindsight can have as many domains as loss functions. 
    \item 
    Consider a similar setting in higher dimensions and give a bound on the number of domains as a function of the number of loss functions and the dimension. 

\end{enumerate}
\end{exercise}

\begin{exercise}
Consider the setting of time-varying linear dynamical systems, i.e. 
$$ \x_{t+1} = A_t \x_t + B_t \uv_t + \w_t. $$
Prove that any DAC according to definition \ref{def:dac} is stable, i.e., prove the analogue of Lemma \ref{lem:dac-is-stable} for time-varying systems. 
\end{exercise}



\begin{exercise}
	Prove Lemma~\ref{lem:movesys}.
\end{exercise}


\end{exercises}

\chapter{Online Nonstochastic Control}
\label{chap:gpc}

In this chapter, we explore recent advances in machine learning that have significantly reshaped the control problem. These developments mark a major departure from classical control theory and, in many cases, have led to methods that not only compete with, but often outperform traditional approaches. The key characteristics of these modern techniques are as follows:

\begin{enumerate}
    \item \textbf{Online Learning Over Precomputed Policies}  
    Unlike classical control, which relies on precomputing an optimal policy, online control adopts {adaptive learning techniques} to update policies dynamically. This aligns with the broader framework of {adaptive control}, where policies evolve in response to observed behavior of the system.

    \item \textbf{Regret Minimization as a Performance Metric}  
    While adaptive control focuses on learning a stabilizing policy, online control introduces a {rigorous performance metric}: regret, defined as the difference in cost between the learned policy and the best policy in hindsight. Unlike classical approaches, {online nonstochastic control algorithms provide finite-time regret guarantees}, ensuring competitive performance even under adversarial conditions.

    \item \textbf{General Loss Functions and Adversarial Perturbations}  
    Classical control typically assumes {quadratic cost functions} with known structures. In contrast, online nonstochastic control allows for {general convex loss functions}, which may be selected {adversarially}. This greater flexibility enables controllers to adapt to more diverse and complex real-world scenarios.

    \item \textbf{Improper Learning via Convex Relaxation}  
    A fundamental breakthrough in online nonstochastic control is the use of {improper learning}, where the algorithm competes against a given policy class while learning from a broader class of policies. Using {convex relaxation techniques}, these methods bypass computational hardness barriers that traditionally limit optimal control, and are accompanied by {efficient algorithms with strong theoretical guarantees}.
\end{enumerate}

These innovations redefine how we approach control problems, shifting the focus from {static policy optimization} to {adaptive, regret-minimizing strategies} that operate effectively in dynamic and adversarial environments.

\ignore{
In this chapter we consider the main new directions that have been taken in the machine learning literature for the control problem. 

These new developments are a major departure from classical control, and in recent years have produced methods that compete favorably and many times surpass classical methods. A few characteristics of the new approaches are listed here: 
\begin{enumerate}
    \item Instead of pre-computing an optimal policy, online control seeks to learn a policy using online learning techniques. This fits the framework known as ``adaptive control". 

    \item
    In contrast to adaptive control, online control optimizes a rigorous performance metric of regret vs. the best policy in hindsight. Online and non-stochastic control algorithms have provable finite-time regret guarantees. 
    
    \item
    In online nonstochastic control, the loss functions can be general Lipschitz convex losses that are chosen adversarially, as opposed to known quadratic losses in classical control. 
    
    \item
    Online and non-stochastic control deploy improper learning by convex relaxation: this means they compete with a certain policy class using a different policy class. This difference in learned policies allows us to sidestep the computational hardness results we have previously explored, and obtain rigorous guarantees for efficient algorithms. 
\end{enumerate}
}

\section{From Optimal and Robust to Online Control}

Classical control theory has been mainly shaped by two paradigms: \textbf{optimal control} and \textbf{robust control}. These approaches have been instrumental in various engineering applications, but they rely on specific assumptions about disturbances and system dynamics.

\begin{itemize}
    \item \textbf{Optimal control} assumes that disturbances follow a known stochastic model, often Gaussian. The objective is to design a control policy that minimizes the expected cumulative cost. Notable methods such as the {Linear Quadratic Regulator (LQR)} and {stochastic dynamic programming} fall into this category.
    \item \textbf{Robust control}, in contrast, takes a worst-case perspective. It ensures stability and performance even under the most adverse conditions, assuming perturbations belong to a predefined uncertainty set. A key example is {H-infinity control}.
\end{itemize}

Although these frameworks have been successful, real-world control tasks often involve {uncertainties that do not fit neatly into either a probabilistic model or a worst-case adversarial setting}. Consider the challenge of controlling a drone in an unpredictable environment:

\begin{itemize}
    \item A stochastic model of wind disturbances may be unrealistic since actual wind patterns can exhibit {long-term correlations or abrupt shifts}.
    \item A worst-case robust approach may result in {excessively conservative} control, leading to inefficient and overly cautious maneuvers.
\end{itemize}

These limitations motivate the need for a more adaptive approach, one that can {learn and adjust dynamically} based on observed disturbances rather than relying on rigid assumptions. This leads us to \textbf{online nonstochastic control}, a fundamentally different paradigm where controllers adapt in real-time without assuming a fixed model for disturbances.

\subsection{A Shift to Online Nonstochastic Control}

Rather than relying on predefined stochastic or worst-case assumptions, {online nonstochastic control} introduces a new approach based on {sequential decision-making and regret minimization}. The core principles are the following.

\begin{enumerate}
    \item \textbf{No Assumption on the Noise Model} -- Disturbances (e.g., wind forces, external perturbations) are treated as an {arbitrary sequence}, potentially chosen adversarially. The controller must adapt dynamically rather than relying on a fixed distribution.
    
    \item \textbf{Learning and Adaptation Over Time} -- Instead of precomputing an optimal policy, online control algorithms iteratively refine their strategies using techniques from {online convex optimization} and {sequential decision-making}.
    
    \item \textbf{Regret Minimization as the Performance Metric} -- Unlike optimal control, which minimizes expected cost, or robust control, which prepares for the worst case, online control seeks to minimize {regret}. Regret quantifies how much worse the controller performs compared to the best policy in hindsight.
\end{enumerate}

To build intuition, consider the problem of controlling an autonomous vehicle in an unknown city. If the vehicle had complete foresight of future traffic conditions, it could calculate the optimal route in advance. However, without such knowledge, the vehicle must continuously adjust its route based on real-time observations. {Regret measures how far the vehicle's performance deviates from the best possible path that could have been chosen in hindsight}. 

This shift in perspective enables online controllers to handle a wide range of real-world disturbances, achieving {adaptability, robustness, and computational efficiency} without requiring strong assumptions about uncertainty.

In the following sections, we formally define the online nonstochastic control problem and introduce key algorithms that provide provable performance guarantees.

\section{The Online Nonstochastic Control Problem}  

The shift from classical control to online nonstochastic control requires a fundamental rethinking of how we handle uncertainty. Unlike optimal control, which assumes disturbances follow a known probabilistic model, or robust control, which plans for the worst-case, online nonstochastic control operates in a more general and challenging setting.  

In this framework, both disturbances and cost functions are chosen by an adversary, meaning that they may follow no statistically reliable pattern and could even be designed to hinder the controller's decisions at every step. As a result, the goal is no longer to compute a fixed optimal control policy, but rather to learn and adapt dynamically in response to an evolving sequence of perturbations and costs.  

A fundamental difficulty in control is the computational tractability of finding the best policy. Classical control methods often rely on dynamic programming, which, in general, suffers from the curse of dimensionality. Even in simple settings, computing an optimal policy can be intractable due to the complexity of solving constrained optimization problems on trajectories. As pointed out by Rockafellar, finding an optimal control policy in arbitrary convex cost settings can be computationally prohibitive, particularly when state and control constraints are involved.  

To address this, online nonstochastic control shifts the focus from computing an optimal policy to competing with a benchmark class of policies using regret minimization. Instead of optimizing a predefined objective function, the controller seeks to minimize its total loss relative to the best policy that could have been chosen in hindsight. This approach bypasses the need to solve an intractable dynamic optimization problem in advance while still achieving performance guarantees over time.  

The fundamental challenge in online nonstochastic control can be summarized as follows:  
\begin{itemize}
    \item The controller does not have prior knowledge of the cost functions or disturbances.  
    \item At each time step, the controller must select a control input based only on past observations.  
    \item The performance of the controller is evaluated against a benchmark: the best fixed policy that could have been chosen in hindsight.  
\end{itemize}  

This formulation naturally connects to online learning and online convex optimization, where decisions must be made sequentially in uncertain environments. We now provide a formal definition of the problem.

\begin{definition}
Consider a dynamical system evolving according to the equation: $ \x_{t+1} = f_t(\x_t, \uv_t, \w_t),$ 
where:  
\begin{itemize}
    \item \( \x_t \in \mathbb{R}^{d_\x} \) is the system state at time \( t \),  
    \item \( \uv_t \in \mathbb{R}^{d_\uv} \) is the control input at time \( t \),  
    \item \( \w_t \in \mathbb{R}^{d_\w} \) represents an external perturbation, which is chosen adversarially.  
\end{itemize}  

At each time step, the controller $\mA$ selects \( u_t \) based on past observations, then incurs a cost defined by an arbitrary convex function $c_t : \mathbb{R}^{d_\x} \times \mathbb{R}^{d_\uv} \to \mathbb{R}$.

The objective is to design a control strategy that minimizes regret, defined as the difference between the total cost incurred by the controller and the best fixed policy in hindsight:
\begin{align*}
\regret_T(\mA,\Pi) = \max_{\mathbf{w}_{1:T}: \norm{\mathbf{w}_t} \le 1} &\left(\sum_{t=1}^{T} c_t (\mathbf{x}_t, \mathbf{u}_t) - \min_{\pi \in \Pi} \sum_{t=1}^{T} c_t (\x_t^\pi,  \uv_t^\pi)) \right) ,
\end{align*}

where \( \Pi \) is a benchmark class of policies, and \( (x_t^\pi, u_t^\pi) \) represent the state and control sequence obtained under the best policy \( \pi \) in hindsight.

\end{definition}

Henceforth, if $T,\Pi$ and $\mA$ are clear from the context, we refer to $\regret$ without quantifiers. 
The goal of an online control algorithm is to achieve sublinear regret,
\begin{equation}
    \regret_T = o(T),
\end{equation}

which ensures that, as \( T \to \infty \), the controller performs competitively with the best fixed policy.

\subsection{Improper Learning and Convex Relaxation}  

A key technique that enables online nonstochastic control algorithms to remain computationally efficient is improper learning through convex relaxation. Instead of directly competing with a predefined class of policies, these algorithms leverage a different policy class that allows for efficient optimization while still achieving strong performance guarantees.  

This approach is crucial because enforcing exact constraints on the policy space can lead to computational hardness results, as previously discussed. By relaxing the policy representation, for example, using convex approximations or surrogate loss functions, controllers can sidestep these intractabilities while still competing effectively with the best policy in hindsight.  

In the next section, we introduce algorithmic solutions that utilize this principle to achieve low-regret online control under adversarial conditions.

\section{The Gradient Perturbation Controller}

After motivating the framework, defining regret, and exploring the relationships between different policy classes, we are ready to prove the main new algorithm in non-stochastic control. 

The algorithm is described in figure \eqref{algo:GPC}, for time-varying linear dynamical systems. It is a specialization of the template introduced in Algorithm \ref{algo:GPC-generic} from Section \ref{sec:intro-gpc}. The main idea is to learn the parameterization of a DAC policy using known online convex optimization algorithms, namely online gradient descent. We henceforth prove that this parameterization is convex, allowing us to prove the main regret bound. 

The notation we use to describe the algorithm is as follows. We denote by $\x_t({M}_{1:h})$ the hypothetical state reached by playing a DAC policy ${M}_{1:h}$ from the beginning of time.  Note that this may not be equal to the true state $\x_t$ of the dynamical system at time $t$. The counterfactual $\uv_t({M}_{1:h})$ is defined similarly. 

\begin{algorithm}[H]
\begin{algorithmic}[1] 
\STATE {Input:}  $h$, $\eta$, initialization $M_{1:h}^1 \in \K$.
\STATE Observe the linear system $(A_t,B_t)$, compute a stabilizing linear controller ${K}_t$ as per Theorem \ref{thm:stabilizing}.
\FOR{$t$ = $1 \ldots T$}
        \STATE  $\mbox{Use control }\mathbf{u}_t = {K}_t \mathbf{x}_t + \sum_{i=1}^{h} {M}_i^t \mathbf{w}_{t-i} .$
        \STATE  Observe state $\mathbf{x}_{t+1}$; compute noise $\mathbf{w}_t = \mathbf{x}_{t+1} - {A}_t \mathbf{x}_t - {B}_t \mathbf{u}_t$.
        \STATE Record an instantaneous cost $c_t(\x_t, \uv_t)$.
        \STATE  Construct loss $\ell_t({M}_{1:h}) = c_t(\mathbf{x}_t({M}_{1:h}), \mathbf{u}_t({M}_{1:h})).$\\
       Update ${M}_{1:h}^{t+1} \leftarrow \prod_\K \left[ {M}_{1:h}^t -\eta \nabla \ell_t({M}_{1:h}^t) \right].$
\ENDFOR
 \caption{Gradient Perturbation Controller(GPC)}
 \label{algo:GPC}
\end{algorithmic}
\end{algorithm}

The GPC algorithm comes with a very strong performance guaranteee: it guarantees vanishing regret vs. the best Disturbance Action Control policy belonging to our reference class in hindsight. According to the relationship between DAC and other policies, this implies vanishing regret versus linear controllers as well. The formal statement is given as follows. 

\begin{theorem} \label{thm:gpc-regret}
Assuming that for all $t \in [T]$
\begin{itemize}
    \item[a]
    The costs $c_t$ are convex, bounded, and $L$-Lipschitz.
    
    \item[b]
     The matrices $\{{A}_t , {B}_t\}$ are $\gamma$-stabilizable and have a bounded $\ell_2$ norm. 
\end{itemize}
Then the GPC algorithm~(\ref{algo:GPC}) ensures that 
\begin{align*}
    \max_{\mathbf{w}_{1:T}: \norm{\mathbf{w}_t} \le 1} \left(\sum_{t=1}^{T} c_t (\mathbf{x}_t, \mathbf{u}_t) - \min_{\pi \in \Pi^{D}} \sum_{t=1}^{T} c_t (\x_t^\pi, \uv_t^\pi) \right) \le \mathcal{O}(  L G D \gamma^2 h \sqrt{T} ),
\end{align*}
where the $O$ notation hides constants that only depend on the norm of $A_t,B_t$. Furthermore, the time complexity of each loop of the algorithm is linear in the number of system parameters and logarithmic in T.
\end{theorem}

\textbf{Note:} we henceforth assume that the linear dynamical systems $(A_t,B_t)$ are a linear time-invariant system $(A,B)$ which is $\gamma$-stable, and there is no need for a stabilizing controller $K_t = 0$. The extension to time-varying stabilizable systems is left as exercise \ref{exercise:time-varying-gpc}.

\begin{proof}
According to Theorem \ref{thm:ogd}, the online gradient descent algorithm for convex loss functions ensures that the regret is bounded by $2 G D \sqrt{T}$.  However, to directly apply the OGD algorithm to our setting, we need to ensure that the following conditions hold true:
\begin{itemize}
    \item The loss function should be a convex function in the variables ${M}_{1:h}$.

    \item  We must ensure that optimizing the loss function $\ell_t$ as a function of parameters $M_{1:h}$ is similar to optimizing the cost $c_t$ as a function of the state and control.  
\end{itemize}

These facts are formalize in the following two lemmas, which we state here and prove at the end of this section. 

\begin{lemma}\label{lemma:convex}
The loss functions $\ell_t$ are convex in the variables ${M}_{1:h}$.
\end{lemma}

\begin{lemma} \label{lem:gpc-memory-lemma}
Let $G$ denote a upper bound on the gradients of the $L$-Lipschitz losses $c_t$, and $D$ denote the diameter of the matrices $M_{1:h}$. Then there exists a constant $C$ that depends only on $\|A\|,\|B|\|$, such that for all $t$,    
$$ |l_t(M^t_{1:h}) - c_t(\x_t, \uv_t)| \leq \frac{C L \gamma^2 h D}{\sqrt{T}} . $$
\end{lemma}

The regret bound of GPC can be derived from these lemmas as follows.  Note that $G$ is an upper bound on the gradients of $\ell_t$ as a function of the parameters $M_{1:h}$, while $L$ is the Lipschitz constant of $c_t$ as a function of the state and control.  
\begin{align*}
    &\sum_{t=1}^{T} c_t (\mathbf{x}_t, \mathbf{u}_t) - \min_{\pi \in \Pi^{D}} \sum_{t=1}^{T} c_t (\x_t^\pi,  \uv_t^\pi) \\
    &\leq \sum_{t=1}^T l_t(M_{1:h}^t) - \min_{\pi \in \Pi^{DAC}} l_t(M_{1:h}^\pi) + T\times \frac{L \gamma^2 C h D}{\sqrt{T}} & \mbox{Lemma \ref{lem:gpc-memory-lemma}} \\
    & \leq 2 GD \sqrt{T} + {\sqrt{T} L \gamma^2 C h D}  & \mbox{Theorem \ref{thm:ogd}}\\
    & \leq O(  G D L \gamma^2 h \sqrt{T} ).
\end{align*}
\end{proof}

\subsection{Convexity of the Losses}

To conclude the analysis, we first show that the loss functions are convex with respect to the variables ${M}_{1:h}$. This follows since the states and the controls are linear transformations of the variables.
\begin{proof}[Proof of Lemma \ref{lemma:convex}]
For simplicity, assume that the initial state $\mathbf{x_0}$ is $\mathbf{0}$.
    Using $\x_t$ to denote $\x_t(M_{1:h})$ and $\uv_t$ to denote $\uv_t(M_{1:h})$, the loss function $\ell_t$ is given by
    \begin{equation*}
        \ell_t({M}_{1:h}) = c_t(\mathbf{x}_t, \uv_t).
    \end{equation*}
    Since the cost $c_t$ is a convex function with respect to its arguments, we simply need to show that $\mathbf{x}_t$ and $\mathbf{u}_t$ depend linearly on ${M}_{1:h}$.
    The state is given by
    \begin{align*}
        \mathbf{x}_{t+1} = {A} \mathbf{x}_{t} + {B} \mathbf{u}_t + \mathbf{w}_t &= {A} \mathbf{x}_{t} + {B}\left(  \sum_{i=1}^{h} {M}_i \mathbf{w}_{t-i}\right) + \mathbf{w}_t  
    \end{align*}
By induction, we can further simplify
    \begin{align}\label{eq:countx}
        \mathbf{x}_{t+1} &= \sum_{i=0}^{t}  {A} ^i \left(B \sum_{j=1}^{h} {M}_j \mathbf{w}_{t-i-j} + \mathbf{w}_{t-i}\right) , 
    \end{align}
    which is a linear function of the variables.

    Similarly, the control $\uv_t$ is given by
    \begin{align*}
        \mathbf{u}_t =  \sum_{i=1}^{h} {M}_{i} \mathbf{w}_{t-i} .
    \end{align*}
    Thus, we have shown that $\mathbf{x}_t$ and $\mathbf{u}_t$ are linear transformations of ${M}_{1:h}$. A composition of convex and linear functions is convex (see Exercise~\ref{exer:linconvex}), which concludes our lemma. 
\end{proof}

\subsection{Loss Functions with Memory}


The actual loss $c_t$ at time $t$ is not calculated on $\x_t(M_{1:h})$, but rather on the true state $\x_t$, which in turn depends on different parameters $M_{1:h}^i$ for various historical times $i < t$. However, $c_t(\x_t, \uv_t)$ is well approximated by $l_t(M^t_{1:h})$, as shown next.

\begin{proof}[Proof of Lemma \ref{lem:gpc-memory-lemma}]

We first note that by the choice of step size by the Online Gradient Descent algorithm \ref{alg:ogd}, as per Theorem \ref{thm:ogd}, we have that  $\eta = \frac{D}{G \sqrt{T}}$. Thus,  
$$\|M_{1:h}^t - M_{1:h}^{t-i}\|\leq \sum_{s=t-i+1}^{t} \|M_{1:h}^s - M_{1:h}^{s-1}\| \leq i \eta G = \frac{i D}{ \sqrt{T}}.$$ 

We next use this fact to establish that $\x_t$ and $\x_t(M^t_{1:h})$ are close. Similar to the expansion of $\x_t(M_{1:h})$ in Equation~\ref{eq:countx},  $\x_t$ can be written as 
\begin{align*}
        \mathbf{x}_{t+1} &= \sum_{i=0}^{t} A^i \left(B \sum_{j=1}^{h} {M}^{t-i}_j \mathbf{w}_{t-i-j} + \mathbf{w}_{t-i}\right). 
    \end{align*}
    Recall that the stability of the time-invariant linear dynamical system $\{A,B\}$ implies that for any $i$, $\|A^i\|\leq \kappa (1-\delta)^{i}$, with $\frac{\kappa}{\delta} \leq \gamma$. Therefore, there exists a constant $C\geq 0$ such that
\begin{align*}\label{eq:realx}
        \|\mathbf{x}_{t} - \mathbf{x}_t(M_{1:h}^t)\| & =  \left\| \sum_{i=0}^{t}  A^i B \sum_{j=1}^{h} {M}^{t-i}_j \w_{t-i-j}  - \sum_{i=0}^{t}  A^i B \sum_{j=1}^{h} M^t_j \mathbf{w}_{t-i-j} \right\| 
        \\
         & \leq  \sum_{i=0}^{t} \left\| A^i \right\| \|B\| \sum_{j=1}^{h} \|{M}^{t-i}_j - M^t_j\| \|\mathbf{w}_{t-i-j}\| 
        \\
        &\leq  C \cdot \kappa \sum_{i=0}^t  (1-\delta)^{i} \cdot  i \frac{h D}{\sqrt{T}}   
        \\ 
        & = \frac{C h  \kappa D }{\sqrt{T}} \sum_{i=1}^\infty i (1-\delta)^{i-1}  \\
        & \leq  \frac{C \gamma^2  h D}{\sqrt{T}} ,
    \end{align*}
where on the second last line we use $\sum_{i=1}^\infty i (1-\delta)^i = \frac{1}{\delta^2}$.
    
    The control at time $t$ depends only on the parameters at the current iteration and not on historical parameters, hence $\uv_t =  \uv_t(M_{1:h}^t)$. By definition, $l_t(M_{1:h}^t) = c_t(\x_t(M_{1:h}^t), \uv_t(M_{1:h}^t))$. Thus, we have 
\begin{eqnarray*}
l_t(M_{1:h}^t) - c_t(\x_t,\uv_t) & = c_t(\x_t(M_{1:h}^t), \uv_t(M_{1:h}^t)) - c_t(\x_t, \uv_t) \\
& \leq L \cdot \| \x_t(M_{1:h}^t) - \x_t\| \leq \frac{L \gamma^2 C h D}{ \sqrt{T}} .
\end{eqnarray*}
    
\end{proof}


\ifarxiv
\newpage
\fi
\section{Bibliographic Remarks}
Online nonstochastic control is an extension of the framework of online convex optimization \citep{hazan2016introduction}, as it is applied to policies, and thus to loss functions with memory \citep{anava2013online}. 

The bulk of the material in this chapter is from \citet{agarwal2019online}. The work of \citet{agarwal2019online} defined the general online nonstochastic setting in which the disturbances and the cost functions are adversarially chosen, and where the cost functions are arbitrary convex functions. 

The follow-up work by \cite{agarwal2019logarithmic} achieves a logarithmic pseudo-regret for strongly convex, adversarially selected losses, and well-conditioned stochastic noise. Logarithmic regret algorithms for the full setting of online nonstochastic control with strongly convex costs were obtained in \citet{simchowitz2020making,foster2020logarithmic,cassel2020logarithmic}. 

Online nonstochastic control with known dynamics was further extended in several directions. The task of control with bandit feedback for the costs was considered in \citet{gradu2020non}.  
More refined regret metrics applicable to changing dynamics were studied in \citet{gradu2020adaptive,minasyan2021online,zhao2022non,singh2022nonstochastic}. More extensions suitable for changing environments were studied in \citet{baby2022optimal}. Online nonstochastic control of population dynamics was studied in \cite{golowich2024online}.

\citet{li2021online} studies algorithms that balance regret minimization while adhering to constraints. An application of nonstochastic control in iterative planning in the presence of perturbations is described in \citet{agarwal2021regret}. Various open-source software packages, such as \citep{gradu2020adaptive}, provide benchmarked implementations of many such algorithms.




An alternate performance metric for online control, which is also nonstochastic, is the competitive ratio. This objective tracks the cost of the online controller and compares it to the best policy in hindsight. Control methods that minimize the competitive ratio were initiated in \citet{goel2019online}.  
This approach was extended in a series of papers \citep{goel2019beyond, shi2020online,goel2021competitive}. Thus far, the success of this approach was limited to fully observed linear dynamical systems and further to quadratic costs when achieving the optimal competitive ratio. \citet{goel2022best} show that regret minimizers in the non-stochastic setting automatically achieve a near-optimal competitive ratio.


A precursor to online nonstochastic control is a line of work, beginning with \citet{abbasi2014tracking}, focused on the control of stochastic systems under sequentially revealed adversarial costs functions. See \citet{cassel2020bandit, cassel2022rate} for further developments.

\ifarxiv
\newpage
\fi
\begin{exercises}

\begin{exercise}\label{exer:linconvex}
Let $l:\reals^d\to \reals$ be a convex function. Prove that for any matrix $A\in \reals^{d\times n}$, $l(Ax)$ is convex for $x\in \reals^n$.
\end{exercise}

\begin{exercise}
\begin{enumerate}
    \item 
Let $l : \mathbb{R}^d \rightarrow \mathbb{R}$ be a $\beta$-smooth function and let $A \in \mathbb{R}^{d \times n}$ be a matrix. Show that $l(A)$ is necessarily a smooth function for $x \in \mathbb{R}^n$.

\item 
Let $l : \mathbb{R}^d \rightarrow \mathbb{R}$ be a $\mu$-strongly convex function and let $A \in \mathbb{R}^{d \times n}$ be a matrix. Show that $l(A)$ is not necessarily strongly-convex for $x \in \mathbb{R}^n$.
\end{enumerate}
\end{exercise}

\begin{exercise}
Prove that the loss functions $\ell_t$ are convex in $M_{1:h}$, as per Lemma \ref{lemma:convex}, even if the initial state is non-zero. 
\end{exercise}

\begin{exercise}
Consider control of a general, non-linear dynamical system. Give a version of the GPC algorithm \ref{algo:GPC} that can be applied to such a setting (without proving any performance guarantee).  
\end{exercise}

\begin{exercise} \label{exercise:time-varying-gpc}
In this exercise, we prove and extend Theorem \ref{thm:gpc-regret} for linear time-varying systems that are stabilizable, rather than stable. 
\begin{enumerate}
    \item Write down and prove the theorem for time-varying systems $\{A_t,B_t\}$ that are all $\gamma$-stable.

    \item Consider a sequence of $\gamma$-stabilizable time-varying systems $\{A_t,B_t\}$. Show how to reduce this case to the previous by considering a sequence of linear time-varying $\gamma$-stable systems $\tilde{A}_t,B_t$, where $\tilde{A}_t = A_t + B_t K_t$. 
\end{enumerate}
\end{exercise}

\end{exercises}

\chapter{Online Nonstochastic System Identification} \label{chap:unknown-system}
\chaptermark{System Identification in ONC}

In the previous chapters, we focused on control algorithms that assume knowledge of the system dynamics. However, in many real-world applications, system parameters are unknown or change over time. This necessitates an approach in which the controller must simultaneously learn the system model while making control decisions. This problem, known as system identification, is fundamental to adaptive control, reinforcement learning, and modern data-driven methods for control.  

\section{The Role of System Identification in Online Control}  

System identification plays a crucial role in control design, as it allows controllers to estimate unknown system parameters in real time. Unlike classical identification methods that rely on extensive offline data collection, an online approach must identify the system while actively controlling it. This presents several unique challenges:  

\begin{itemize}
    \item Feedback loops and nonstationarity: Unlike supervised learning settings, where data is collected passively, control decisions influence future observations, making the estimation process highly dynamic.  
    \item Adversarial or arbitrary disturbances: The disturbances affecting the system may not follow a probabilistic model but can be chosen adversarially, leading to challenges in maintaining robustness.  
    \item Balancing exploration and performance: The controller must explore sufficiently to estimate system parameters accurately while ensuring stable performance, a trade-off that directly impacts regret minimization.  
\end{itemize}  

\section{A Nonstochastic Approach to System Identification}  

Traditional system identification techniques often rely on stochastic assumptions, where noise is assumed to be drawn from a known distribution, and system parameters are estimated using statistical inference techniques. However, in nonstochastic system identification, no assumptions are made about the noise distribution or system perturbations. Instead, identification strategies are designed to be robust to worst-case disturbances and operate under adversarial uncertainty.  

This shift in perspective has several implications:  
\begin{itemize}
    \item Performance guarantees without probabilistic assumptions: Unlike stochastic estimation methods, which provide guarantees under specific noise models, nonstochastic approaches aim for regret bounds that hold in arbitrary environments.  
    \item Regret as a measure of learning efficiency: The effectiveness of an identification strategy is measured not in terms of statistical efficiency but in its ability to achieve low regret, ensuring that the system is identified quickly enough to enable near-optimal control.  
    \item Computational tractability in identification: Classical system identification methods often involve solving non-convex optimization problems, which makes them computationally demanding. Nonstochastic approaches frequently leverage convex relaxation techniques to sidestep intractability.  
\end{itemize}

The problem of online system identification is closely related to online convex optimization and sequential decision making. Many regret minimization techniques developed for online learning can be adapted to system identification settings. However, a key distinction is that, in control settings, the learned system model directly affects future decisions, creating a coupling between estimation and control that does not typically exist in standard online learning formulations.  

This connection between system identification and online learning leads to several fundamental questions:  
\begin{itemize}
    \item How should a controller optimally trade off exploration (learning system parameters) and exploitation (minimizing control cost)?  
    \item What are the best algorithmic strategies for estimating system parameters while controlling in adversarial settings?  
    \item What regret guarantees can be achieved in online nonstochastic system identification, and how do they compare to traditional learning-based control approaches?  
\end{itemize}

\section{Nonstochastic System Identification}

The problem we study in this section is online nonstochastic system identification of a linear time-invariant dynamical system. 

As a first step, we consider the question of whether the system matrices $A,B$ can be recovered from observations over the state, without regard to the actual cost of control. 

The method below to achieve this learns the system matrices by using a random control. Specifically, at each step the control $\mathbf{u}_t$ will be randomly selected from a Rademacher distribution, according to Algorithm \ref{alg:sys-id-mom}.

\begin{algorithm}
	\caption{System identification via method of moments.}
	\label{alg:sys-id-mom}
	\begin{algorithmic}[]
		\STATE Input: controllability index $k$, oversampling parameter $T_0$.
		\FOR{$t = 0, \ldots, T_0(k+1)$}
		\STATE Execute the control $\uv_t  = \eta_t$ with $\eta_t \sim_{i.i.d.} \{\pm 1\}^d$.
		\STATE Record the observed state $\x_t$. 
		\ENDFOR
		\STATE Set $\widehat{A^jB} = \frac{1}{T_0}\sum_{t=0}^{T_0-1} \x_{t(k+1)+j+1} \eta_{t(k+1)}^\top $, for all $j \in \{0,\dots k\} $.
		\STATE Define $\widehat{C_0}=(\widehat{B},\dots \widehat{A^{k-1}B})$, $\widehat{C_1}=(\widehat{AB},\dots \widehat{A^kB})$, and return $\widehat{A},\widehat{B}$ as
		\[ \widehat{B} = \widehat{B},\quad \widehat{A} = \argmin_{A} \|\widehat{C_1} - A\widehat{C_0}\|_F .  \]
	\end{algorithmic}
\end{algorithm}

The main guarantee from this simple system identification procedure is given as follows. 
\begin{theorem}\label{thm:sys-id-mom}
Assuming that
\begin{itemize}
	\item[a] the system $(A,B)$ is $\gamma$-stable,
	\item[b] the system is $(k,\kappa)$-strongly controllable,
	\item[c] the perturbations $\w_t$ do not depend on $\uv_t$,
\end{itemize}
and for $T_0 = \Omega\left(\frac{d^2k^2\kappa^2\gamma^6}{\varepsilon^2}\log \frac{1}{\Delta}\right) $, the following holds.  
Algorithm~\ref{alg:sys-id-mom} outputs a pair $(\widehat{A},\widehat{B})$ that satisfies, with probability $1-\Delta$, 
$$ \|\widehat{A}-A\| \leq \eps , \|\widehat{B}-B\| \leq \epsilon . $$
\end{theorem}

The proof of this theorem hinges upon the key observation that  
\begin{equation*}
    \E  [ \x_{t+k}   \uv_t^{\top} ] = \E \left[\sum_{i=0: k} A^{i}\left(B \uv_{t+k-i}+\w_{t}\right) \uv_{t}^{\top}\right]=A^{k} B ,
\end{equation*}
and the crucial fact that the controls are i.i.d. Thus, we can create an unbiased estimator for this matrix, and using many samples, we obtain an increasingly accurate estimate. 

\begin{proof}
First, we establish that $A^jB$ is close to $\widehat{A^jB}$ for all $j\in [k]$. 
\begin{lemma}
	As long as $T_0\geq \widetilde{\Omega}\left(\frac{d^2\gamma^2}{\varepsilon^2}\log\frac{1}{\Delta}\right)$, with probability $1-\Delta$, we have for all $j\in [k]$ that 
	$$ \|\widehat{A^j B}- A^jB\| \leq \varepsilon.$$
\end{lemma}
\begin{proof}
First note that for any $t$, since the system is $\gamma$ stable and the controls have norm $\sqrt{d}$, the state $\x_t$ is bounded as 
$$ \|\x_t\| = \left\|\sum_{i=1}^t A^{i-1}\left(B\eta_{t-i}+\w_{t-i}\right) \right\| \leq 2\sqrt{d}\gamma.$$
Therefore, defining $N_{j,t}=\x_{t(k+1)+j+1} \eta_{t(k+1)}^\top$ for any $j<k$ and $t<T_0$, we have that $\|N_{j,t}\| \leq 2d\gamma$. Note that $\{N_{j,t}\}_{t\leq T_0}$ is not a sequence of independent random variables, since the same instance of $\eta_t$ might occur in multiple of these terms. To remedy this, we claim that the sequence $\tilde{N}_{j,t} = N_{j,t} - A^j B$ forms a martingale difference sequence with respect to the filtration $\Sigma_t$, defined as $\Sigma_t = \{\eta_{i}:i< (k+1)(t+1)\}$. Since $N_{j,t}$ only involves the first $t(k+1)+k$ random control inputs, ${N}_{j,t}$ is $\Sigma_t$-measurable. Now, for any $t$, since $\eta_t$ is zero-mean and chosen independently of $\{w_j\}, \{\eta_j\}_{j\neq t}$ at each time step, we have
\begin{align*}
	\mathbb{E}\left[N_{j,t} | \Sigma_{t-1}\right] 
	&= \mathbb{E}\left[\left. \sum_{i>1} A^{i-1}\left(\w_{t(k+1)+j+1-i}+B\eta_{t(k+1)+j+1-i}\right) \eta_{t(k+1)}^\top \right\vert \Sigma_{t-1}\right] \\
	&= \mathbb{E}\left[\left. \sum_{i>1} A^{i-1}B\eta_{t(k+1)+j+1-i} \eta_{t(k+1)}^\top \mathbb{I}_{i=j+1}\right\vert \Sigma_{t-1}\right] = A^jB,
\end{align*} 
where the last equality follows from the fact that $\mathbb{E}[\eta_t\eta_t^\top]= I$.

\begin{theorem}[Matrix Azuma]
Consider a sequence of random symmetric matrices $\{X_k\}\in \reals^{d\times d}$ adapted to a filtration sequence $\{\Sigma_k\}$ such that $$\mathbb{E}[X_k|\Sigma_{k-1}]=0 \quad \text{ and } \quad X_k^2 \preceq \sigma^2 I \text{ holds almost surely},$$ then, for all $\varepsilon>0$, we have $$ P\left(\lambda_{\max}\left(\frac{1}{K}\sum_{k=1}^K X_k\right) \geq \varepsilon \right ) \leq d e^{-{K\varepsilon^2}/{8\sigma^2}}.$$
\end{theorem}

Applying the Matrix Azuma inequality (see the bibliography for a reference) on the symmetric dilation, i.e.
\begin{align*}
	\begin{bmatrix}
		0 & \tilde{N}_{j,t}^\top  \\
		\tilde{N}_{j,t} & 0
	\end{bmatrix},
\end{align*}
we have for all $j\in [k]$ with probability $1-\Delta$ that 
$$ \|\widehat{A^jB} - A^j B\| \leq {8d\gamma}\sqrt{\frac{1}{T_0}\log\frac{n+d}{\Delta}}.$$
A union bound over $j\in [k]$ concludes the claim.
\end{proof}

At this point, define $C_0 = (B, AB, \dots, A^{k-1}B)$ and $C_1 = (AB, A^2B,\dots A^kB)$ to observe that with probability $1-\Delta$, we have that
$$ \|\widehat{C_0}-C_0\|, \|\widehat{C_1}-C_1\| \leq \sqrt{k}\varepsilon.$$
By this display, we already have $\|B-\widehat{B}\|\leq \varepsilon$. We will make use of this observation to establish the proximity of $\widehat{A}$ and $A$, using the following lemma, the proof of which is left as Exercise~\ref{exer:linpert}.
\begin{lemma}\label{lem:linpert}
Let $X,X+\Delta X$ be the solutions to the linear system of equations $XA=B$ and $(X+\Delta X)(A+\Delta A) = B+\Delta B$. Then, as long as $\|\Delta A\|< \sigma_{\min} (A)$, we have 
$$ \|\Delta X\| \leq \frac{\|\Delta B\|+ \|\Delta A\|\|X\|}{\sigma_{\min}(A)-\|\Delta A\|}$$ 
\end{lemma}
Note that $\widehat{A}$ does not necessarily satisfy $\widehat{A}\widehat{C_0} = \widehat{C_1}$. We now use the above lemma on the normal equations, that is, on $AC_0C_0^\top = C_1C_0^\top$ and $\widehat{A}\widehat{C_0} \widehat{C_0}^\top = \widehat{C_1}\widehat{C_0}^\top$. First, observe that for any $\varepsilon<1$
\begin{align*}
	\left\|C_0C_0^\top  - \widehat{C_0}\widehat{C_0}^\top \right\| &= \left\|C_0C_0^\top  - C_0\widehat{C_0}^\top + C_0\widehat{C_0}^\top- \widehat{C_0}\widehat{C_0}^\top \right\| \\
	&\leq \|C_0\| \|C_0-\widehat{C_0}\| + \|\widehat{C_0}\| \|C_0-\widehat{C_0}\| \\
	&\leq 2\|C_0\|\sqrt{k}\varepsilon + k\varepsilon^2
	\leq 3\gamma k\varepsilon ,
\end{align*}
where the last line uses that $\|C_0\|\leq \gamma$ following the stability of $(A,B)$. A similar display implies the same upper bound on $\|C_1C_1^\top-\widehat{C_1}\widehat{C_1}^\top\|$. Therefore, using $\|A\|\leq \gamma$ and the perturbation result for linear systems, we have
$$ \|A-\widehat{A}\| \leq  \frac{12\gamma^2 k}{\sigma_{\min}(C_0C_0^\top)} \varepsilon, $$
as long as $6k\gamma\varepsilon < \sigma_{\min} (C_0C_0^\top)$. Finally, note that $(k, \kappa)$-strong controllability implies $\sigma_{\min}(C_0C_0^\top) \geq 1/\kappa$.
\end{proof}

\section{From Indentification to Nonstochastic Control}
Having recovered approximate estimates $(\widehat{A},\widehat{B})$ of the system matrices, a natural question arises: how well does the Gradient Perturbation Algorithm \ref{algo:GPC} perform on the original system $(A,B)$ when provided with only approximate estimates of the same? We briefly state a result to this extent, whose proof we leave to the explorations of an interested reader.

\begin{lemma}\label{lem:idtocon}
	Given estimates of the system matrices $(\widehat{A}, \widehat{B})$ such that
	$$ \|\widehat{A}-A\|, \|\widehat{B}-B\|\leq \varepsilon,$$
	the GPC algorithm when run on the original system $(A,B)$ guarantees a regret bound of
	$$\max_{\mathbf{w}_{1:T}: \norm{\mathbf{w}_t} \le 1} \left(\sum_{t=1}^{T} c_t (\mathbf{x}_t, \mathbf{u}_t) - \min_{\pi \in \Pi^{DAC}} \sum_{t=1}^{T} c_t (\hat{\mathbf{x}}_t,  \pi(\hat{\mathbf{x}}_t)) \right) \le \mathcal{O}(\sqrt{T}+\varepsilon T).
	$$
\end{lemma}

This result translates to a regret bound of $\tilde{O}(T^{2/3})$ for nonstochastic control of unknown, yet stable, linear systems, by first running the nonstochastic system identification procedure, and subsequently using the estimates thus obtained to run GPC. Details of how to derive this regret bound are left as Exercise~\ref{exer:23bound}.


\section{Summary and Key Takeaways}  

This chapter explored online nonstochastic system identification, where a controller must simultaneously learn an unknown system model while making control decisions in an adversarial setting. Unlike classical stochastic system identification, which assumes structured noise and statistical convergence, the nonstochastic approach makes no such assumptions and instead relies on regret minimization techniques to ensure efficient learning.

The key insights from this chapter are as follows. 

\begin{itemize}  
    \item \textbf{The Exploration-Exploitation Tradeoff}  
    A fundamental challenge in online system identification is balancing exploration, which enables learning the system dynamics, and exploitation, which minimizes control cost. Too much exploration leads to high immediate costs, while too much exploitation can result in poor long-term learning and suboptimal control performance. The algorithms in this chapter address this by incorporating controlled perturbations in the control inputs, ensuring sufficient exploration while keeping regret low. However, the trade-off results in a regret of $O(T^{2/3})$, rather than $O(\sqrt{T})$ we had in the last chapter. 

    More sophisticated techniques for balancing exploration and exploitation can result in better regret bounds, some of which are surveyed in the bibliographic section. 

    \item \textbf{Regret Minimization as a Learning Metric}  
    Unlike traditional identification methods that aim at asymptotic convergence, online nonstochastic identification is evaluated in terms of regret, measuring the performance relative to the best fixed model in hindsight.

\end{itemize}  

\subsection{Looking Ahead}  

This part of the book considered the online nonstochastic control problem, the regret metric, and online learning techniques applied to control.  We have thus far considered the case where the state is observable to the controller, an assumption which does not always hold. 

In the next parts of the book, we consider the setting of partial observability, where the system state is unknown, but an observed signal is available. We start with the problem of learning and move onto online control.

\ifarxiv
\newpage
\fi
\section{Bibliographic Remarks}
Controlling linear dynamical systems given unknown or uncertain system parameters has long been studied in adaptive control; see the text of \citet{sastry1990adaptive} for example. However, the results from adaptive control typically concern themselves with stability and, even when discussing implications for optimality, present asymptotic bounds. 

In contrast, much of the work at the intersection of machine learning and control is focussed on precise non-asymptotic bounds on sample complexity or regret. \citet{fiechter1997pac} which studies the discounted LQR problem under {\bf stochastic noise} is the earliest example that provides a precise sample complexity bound. Similarly, the early work of \citet{abbasi2011regret} uses an optimism-based approach to achieve optimal regret in terms of dependence on horizon. This regret bound was later improved for sparse systems in \citet{ibrahimi2012efficient}. \citet{dean2018regret,dean2020sample} established regret and sample complexity results for the general stochastic LQR formulation with a polynomial dependence on natural system parameters. \citet{fazel2018global} studies the efficacy of first-order optimization-based approaches in synthesizing controllers for the LQR problem. The work of \citet{mania2019certainty} establishes that simple explore-then-commit strategies achieve the optimal dependence of regret on horizon, with \citet{pmlr-v119-simchowitz20a} proving that such approaches are optimal in terms of their dependence on dimension of the system, as well. \citet{cohen2019learning} also gave a first optimal-in-horizon polynomial regret result simultaneously with the former work, albeit using an optimism-based approach. The result of \citet{plevrakis2020geometric} gives a rate-optimal regret bound for unknown systems with convex costs. 

In terms of {\bf nonstochastic} control, the method-of-moments system identification procedure and the corresponding analysis presented here appeared in \citet{hazan2019nonstochastic}, which also gave the first end-to-end regret result for nonstochastic control starting with unknown system matrices. Even earlier, \citet{simchowitz2019learning} presented a regression procedure for identifying Markov operators under nonstochastic noise. Both of these approaches have roots in the classical Ho-Kalman system identification procedure \citep{ho1966effective}, for which the first non-asymptotic sample complexity, although for stochastic noise, was furnished in \citet{oymak2019non}. For strongly convex costs,  logarithmic regret was established in \citet{simchowitz2020making}. Nonstochastic control of unknown, unstable systems without any access to prior information, also called black-box control, was addressed in \citet{chen2021black}.

The Matrix Azuma inequality used in this chapter appears in \citet{tropp2012user}.

\ifarxiv
\newpage
\fi
\begin{exercises}
\begin{exercise}\label{exer:23bound}
Fix a value of $T$. Consider running the nonstochastic system identification procedure for $T_0 =\tilde{\Theta}(T^{2/3})$ steps to obtain system matrix estimates $\widehat{A}, \widehat{B}$, and subsequently using these estimates as inputs to the GPC algorithm when executed over the system $(A,B)$. Given Lemma~\ref{lem:idtocon}, conclude that the resultant control procedure produces a regret of at most $\tilde{O}(T^{2/3})$ without any prior knowledge of system matrices, as long as the underlying system is stable.
\end{exercise}

\begin{exercise}
Design an analogous control procedure as that for Exercise~\ref{exer:23bound} with an identical regret upper bound (up to constants), when the horizon $T$ is not known in advance.
\end{exercise}

\begin{exercise}\label{exer:linpert}
Prove Lemma~\ref{lem:linpert}.
\end{exercise}
\end{exercises}

\part{Learning and Filtering}
\chapter{Learning in Unknown Linear Dynamical Systems}  \label{chap:LDS-learning}

In previous chapters, we focused on control in dynamical systems, often assuming that the system state was fully observable and that system dynamics were either known or could be estimated through system identification. However, in many real-world scenarios, these assumptions do not hold. Instead, an agent may only have access to partial observations of the system and must learn to make accurate predictions without directly influencing the system's evolution.

This chapter focuses on learning in a dynamical system, which means predicting future system outputs based on past observations while the underlying system parameters remain unknown. Unlike control, where an agent actively influences system trajectories to minimize a given cost, learning in this setting is a passive task; the goal is to accurately forecast future outputs without modifying the system's behavior.

The ability to predict future states and outputs of an unknown system is fundamental to many domains, including time series forecasting, financial modeling, sensor networks, autonomous systems, and most recently large language models. Moreover, learning a predictive model is often a key stepping stone toward solving more complex problems, such as control under partial observability or reinforcement learning in dynamical systems. In this chapter, we study online learning algorithms for prediction in unknown linear dynamical systems, with an emphasis on regret minimization.

\section{Learning in Dynamical Systems}

The term learning in dynamical systems can refer to a variety of tasks, including state estimation, system identification, and policy learning. In this chapter, we focus on the specific task of prediction, where the objective is to estimate future outputs of a dynamical system based on past observations. Unlike classical system identification, which seeks to recover the system parameters explicitly, the learning approach taken here aims to optimize the predictions directly.

\begin{figure}[h]
    \centering
    \includegraphics[width=.7\linewidth]{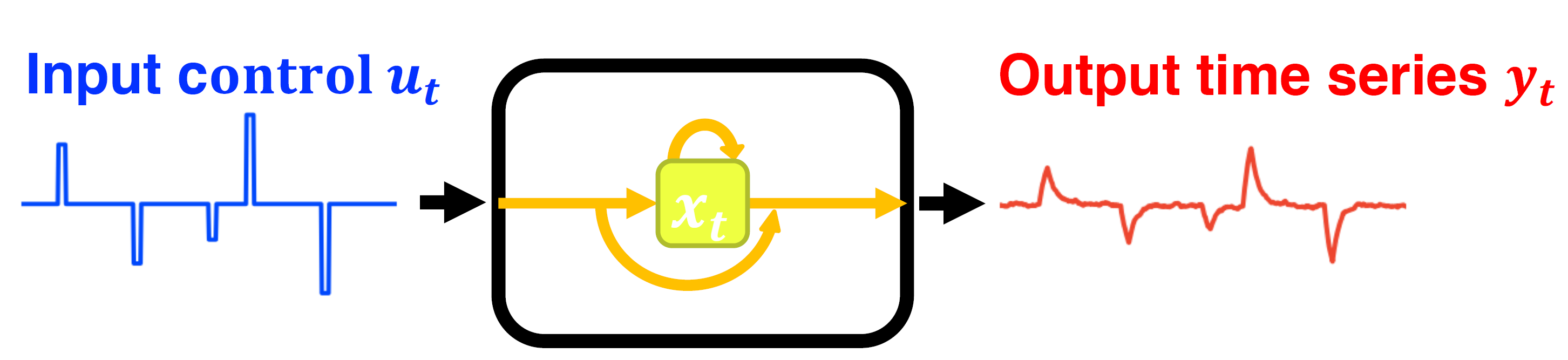}
    \caption{In learning we passively observe the input sequence and attempt to predict output.}
    \label{fig:lds}
\end{figure}

To formally define the problem, consider a sequence of observable inputs fed into a dynamical system, as illustrated in Figure \ref{fig:lds}. The learner observes these inputs and aims to predict the corresponding outputs while minimizing a given loss function.

For the remainder of this chapter, we consider a partially observable linear time-invariant dynamical system described by the following equations. 
\begin{align*}
    \x_{t+1} &= A \x_t + B \uv_t + \w_t,\\
    \y_t &= C \x_t + \vv_t .
\end{align*}

Here, \( \x_t \) represents the hidden system state, \( \uv_t \) is the observed control input, and \( \y_t \) is the measured system output. The terms \( \w_t \) and \( \vv_t \) account for the process noise and the observation noise, respectively. A crucial challenge in this setting is that the system matrices \( A, B, C \) are unknown to the learner.

The learner's objective is to construct an efficient prediction rule that estimates \( \y_t \) based on past observations, despite not knowing the underlying parameters of the system.

\section{Online Learning of Dynamical Systems}

In the problem of online learning in a linear dynamical system, sequentially for $t = 1, \ldots, T$, the online learning algorithm:
\begin{itemize}
    \item observes input $\uv_{t-1} \in \reals^{d_u}$,
    \item predicts output $\hat{\y}_t \in \reals^{d_y}$,
    \item observes true system output $\y_t$,
    \item suffers loss according to a pre-specified loss function, $\ell_t(\hat{\y}_t,\y_t)$, e.g.
    $$ \ell_t(\hat{\y}_t,\y_t) = \| \hat{\y}_t - \y_t\|^2 . $$
\end{itemize}

The goal of the online learner is to minimize total loss over all game iterations. If the observations ${\y}_t$ are arbitrary, this is an ill-defined goal, and rather than minimizing loss, we consider a comparative analogue.

More precisely, we consider a family of prediction rules from available information to prediction, and consider the metric of {\it regret} with respect to the best of these rules in hindsight. Formally, let $\Pi$ be a family of mappings from $\{\uv_\tau,\y_\tau\}_{\tau=1}^{t-1}$ to a prediction $\hat{\y}_t$, i.e.
$$ \pi \in \Pi \ , \ \pi(\uv_{1:t-1},\y_{1:t-1}) = \hat{\y}_t^\pi \in \reals^{d_y}. $$
Let $\mA$ be a prediction algorithm that iteratively produces predictions $\hat{\y}_t$. Then its regret with respect to $\Pi$ is defined as
$$ \regret_T(\mA,\Pi) =  \sum_{t=1}^T \ell_t(\hat{\y}_t,\y_t) - \min_{\pi \in \Pi} \sum_{t=1}^T \ell_t(\hat{\y}^\pi_t,\y_t) . $$

We proceed to define common prediction classes in the context of linear dynamical systems.

\section{Classes of Prediction Rules}

We now explore what constitutes a reasonable prediction rule for the output of a linear dynamical system.
However, before considering the prediction rules themselves, we discuss how to compare classes of such rules. This is analogous to the definitions and discussion in Chapter \ref{chap:policy-classes} where we discuss comparison between policy classes for control. However, the definitions and the corresponding conclusions for making predictions are simpler.

\begin{definition} \label{defn:eps-approximation-of-predictors}
We say that a class of predictors $\Pi_1$ $\eps$-approximates class $\Pi_2$ over horizon $T$ if the following holds. For every sequence of inputs and outputs $\{\uv_{1:T},\y_{1:T}\}$ and every $\pi_2 \in \Pi_2$, there exists $\pi_1 \in \Pi_1$ such that
$$ \sum_{t=1}^T \| \hat{\y}_t^{\pi_1} - \hat{\y}_t^{\pi_2}  \| \leq T \eps ,$$
where
$$ \hat{\y}_t^{\pi_1} = \pi_1( \y_{1:t-1} , \uv_{1:t-1} ) \ ,\ \hat{\y}_t^{\pi_2}= \pi_2( \y_{1:t-1} , \uv_{1:t-1} ) . $$
\end{definition}

The significance of this definition is that if a simple predictor class approximates another more complicated class, then it suffices to consider only the first. The implication of $\eps$-approximation on the regret is given below.

\begin{lemma} \label{lem:approximation_of_open_loop_predictors}
Assume that each loss function $\ell_t(\hat{\y},\y_t)$ is $G$-Lipschitz in its first argument on the relevant prediction domain. If predictor class $\Pi_1$ $\eps$-approximates class $\Pi_2$, then for any prediction algorithm $\mA$ we have
$$ \regret_T(\mA,\Pi_2) \leq  \regret_T(\mA,\Pi_1) + G \eps T . $$
\end{lemma}

The proof of this lemma is left as an exercise to the reader at the end of this chapter. We now proceed to consider specific prediction rules.

\subsection{Linear Predictors}

The simplest prediction rule is a linear function of past observations and past inputs.

\begin{definition}[Linear Predictors]  \label{defn:linear-predictors}
A linear predictor parameterized by matrices
$$ M_{1,i} \in \reals^{d_y \times d_y} \ \text{ for } \ i=1,\ldots,h, \qquad
   M_{2,j} \in \reals^{d_y \times d_u} \ \text{ for } \ j=1,\ldots,k, $$
maps past outputs and inputs to a prediction according to
$$ \hat{\y}_{t} = \sum_{i=1}^h M_{1,i} \y_{t-i} + \sum_{j=1}^k M_{2,j} \uv_{t-j} . $$
We denote by $\Pi_{h, k,\kappa}^{L}$ the set of all such predictors satisfying
$$ \sum_{i=1}^h \| M_{1,i} \| + \sum_{j=1}^k \|M_{2,j}\|  \leq \kappa . $$
\end{definition}

This class is computationally attractive because it is parameterized by a convex set of matrices. At first sight, however, it may appear too restrictive to model the predictions generated by a linear dynamical system with hidden state. The next subsection explains why this concern is misleading.

\subsection{Linear Dynamical Predictors}

An intuitive class of prediction rules that makes use of the special structure of linear dynamical systems is that of open-loop linear dynamical predictors.

The motivation for this class is the case that the system transformations are known and the system is fully observed, namely $C = I$, the identity matrix. In this situation, the optimal prediction can be calculated by unrolling the system evolution equation $\x_{t+1} = A \x_t + B \uv_t$. It can be seen that this prediction rule is optimal for any zero-mean noise or perturbation model. However, this optimality is only true if the state is fully observable. In the next chapter, we consider optimal predictors for zero-mean i.i.d.\ noise in the more general partially observed case.

For a partially observed linear dynamical system without noise we can write
\begin{align*}
    \y_{t}^\pi = \pi_{\{A,B,C,\x_0\}}(t) &= C \x_t \\
                         &= C A^{t}\x_0 + \sum_{i=1}^{t} C A^{i-1}B \uv_{t-i} .
\end{align*}

Thus, the operator $\pi_{\{A,B,C,\x_0\}}$ takes a history of inputs and gives the next output of the system accurately. We denote by $\Pi^{\star}$ the set of all such predictors that in the case of zero noise give a perfectly accurate prediction.

The class $\Pi^{\star}$ is a very natural comparator class for the noiseless open-loop prediction problem. It is important to note, however, that in the general partially observed noisy setting the optimal predictor need not belong to this class, since it may depend on past outputs in a nontrivial way. We return to this more general question in the next chapter. For the present chapter, $\Pi^{\star}$ serves as a clean structural benchmark whose induced predictors we can approximate or represent by convex classes.

\section{Convex Relaxations of Linear Dynamical Predictors}

In the previous sections we have defined what it means to learn in a dynamical system, and described two natural classes of predictors. The class of linear dynamical predictors is most natural, as it exactly captures the noiseless evolution of a linear dynamical system. However, it is nonconvex in its parametrization. 

In this section we describe convex relaxation of this class that approximate it extremely well and allow for efficient learning.

\subsection{Convex Relaxation of Stable Systems}

The first relaxation we discuss is that which assumes additional structure from the dynamics, namely stability. 
For a noiseless system, define the truncated open-loop predictor
$$
\hat{\y}_{t} = \pi_{\{A,B,C,h\}}(t) = \sum_{i=1}^h CA^{i-1}B \uv_{t-i},
$$
where by convention $\uv_s = 0$ for all $s \leq 0$. Denote by $\Pi_h^\star$ the class of all such truncated predictors.

When the system is stable, this truncation yields a simple approximation to the
class of open-loop linear dynamical predictors. To state this in the language of
Definition~\ref{defn:eps-approximation-of-predictors}, we restrict attention to a
stable subclass with uniformly bounded parameters and zero initial state.

\begin{lemma}\label{lem:lr_approx_prediction}
Fix $M \ge 1$, and define
\[
\Pi^\star_{\mathrm{stab}}(M)
\defeq
\left\{
\pi_{\{A,B,C\}} \in \Pi^\star :
\|A^t\| \le M e^{-t/M}\ \text{for all } t\ge 0,
\quad
\|B\|+\|C\| \le M
\right\}.
\]
Assume that $\x_0=0$ and that
\[
\max_{t\in[T]} \|\uv_t\| \le M.
\]
Then $\Pi_h^\star$ $\eps$-approximates $\Pi^\star_{\mathrm{stab}}(M)$ over horizon
$T$ provided
\[
h \ge M \log\!\left(\frac{2M^5}{\eps}\right).
\]
\end{lemma}

\begin{proof}
Fix any predictor $\pi_{\{A,B,C\}} \in \Pi^\star_{\mathrm{stab}}(M)$, and let
$\pi_{\{A,B,C,h\}} \in \Pi_h^\star$ be its truncated predictor. For each $t\in[T]$,
\begin{align*}
\hat{\y}_t^{\pi_{\{A,B,C\}}}
-
\hat{\y}_t^{\pi_{\{A,B,C,h\}}}
&=
\sum_{i=h+1}^{t} CA^{i-1}B \uv_{t-i}.
\end{align*}
Hence
\begin{align*}
\left\|
\hat{\y}_t^{\pi_{\{A,B,C\}}}
-
\hat{\y}_t^{\pi_{\{A,B,C,h\}}}
\right\|
&\le
\sum_{i=h+1}^{t}\|C\|\,\|A^{i-1}\|\,\|B\|\,\|\uv_{t-i}\| \\
&\le
M^4 \sum_{i=h+1}^{t} e^{-(i-1)/M}.
\end{align*}
Since $M\ge 1$, we have $1-e^{-1/M}\ge \frac{1}{2M}$, and therefore
\[
\sum_{i=h+1}^{t} e^{-(i-1)/M}
\le
\sum_{i=h+1}^{\infty} e^{-(i-1)/M}
\le
\frac{e^{-h/M}}{1-e^{-1/M}}
\le
2M e^{-h/M}.
\]
Thus
\[
\left\|
\hat{\y}_t^{\pi_{\{A,B,C\}}}
-
\hat{\y}_t^{\pi_{\{A,B,C,h\}}}
\right\|
\le
2M^5 e^{-h/M}.
\]
Summing over $t=1,\dots,T$ yields
\[
\sum_{t=1}^T
\left\|
\hat{\y}_t^{\pi_{\{A,B,C\}}}
-
\hat{\y}_t^{\pi_{\{A,B,C,h\}}}
\right\|
\le
2TM^5 e^{-h/M}.
\]
Choosing
\[
h \ge M \log\!\left(\frac{2M^5}{\eps}\right)
\]
makes the right-hand side at most $T\eps$, proving the claim.
\end{proof}

\subsection{Exact Representation via Cayley Hamilton}

In contrast to the previous approximation result, which is dimension-free but
requires stability, the Cayley--Hamilton argument below gives an exact
finite-memory representation for arbitrary noiseless finite-dimensional systems,
at the price of dependence on the hidden-state dimension $d_x$.

For a given matrix $A$ that describes the state evolution of a linear dynamical system with hidden-state dimension $d_x$, let
$$ p_A(\lambda) = \lambda^{d_x} + a_1 \lambda^{d_x-1} + \cdots + a_{d_x}  $$
be the characteristic polynomial of $A$, and define $a_0 = 1$. 
The following theorem is the key structural observation of this chapter.

\begin{theorem} \label{thm:cayley-hamilton-linear-predictors}
Consider the noiseless linear dynamical system
\begin{align*}
    \x_{t+1} &= A \x_t + B \uv_t, \\
    \y_t &= C \x_t ,
\end{align*}
with characteristic polynomial $ p_A(\lambda)$. Then there exist matrices
$$ M_{1,i} \in \reals^{d_y \times d_y}, \qquad M_{2,j} \in \reals^{d_y \times d_u}, \qquad i,j=1,\ldots,d_x, $$
such that for every $t > d_x$,
$$
\y_t = \sum_{i=1}^{d_x} M_{1,i}\y_{t-i} + \sum_{j=1}^{d_x} M_{2,j}\uv_{t-j}.
$$
In particular, one may take
$$ M_{1,i} = -a_i I_{d_y}, \qquad M_{2,j} = C\left(\sum_{i=0}^{j-1} a_i A^{j-1-i}\right)B. $$
\end{theorem}

\begin{proof}
Write $d=d_x$. By unrolling the state dynamics for $d$ steps, we obtain
\begin{equation}\label{eqn:ch-proof}
\x_t = A^d \x_{t-d} + \sum_{j=1}^{d} A^{j-1} B \uv_{t-j}.
\end{equation}
Similarly, for each $i=1,\ldots,d$,
\[
\x_{t-i} = A^{d-i}\x_{t-d} + \sum_{j=i+1}^{d} A^{j-i-1} B \uv_{t-j},
\]
so
\[
A^{d-i}\x_{t-d}
=
\x_{t-i} - \sum_{j=i+1}^{d} A^{j-i-1} B \uv_{t-j}.
\]

By the Cayley--Hamilton theorem, $A^d + a_1 A^{d-1} + \cdots + a_d I = 0$, hence $ A^d = -\sum_{i=1}^{d} a_i A^{d-i}$.
Substituting this into \eqref{eqn:ch-proof}, and then using the previous identity, gives
\begin{align*}
\x_t
&= -\sum_{i=1}^{d} a_i A^{d-i}\x_{t-d} + \sum_{j=1}^{d} A^{j-1} B \uv_{t-j} \\
&= -\sum_{i=1}^{d} a_i \x_{t-i}
   + \sum_{i=1}^{d} a_i \sum_{j=i+1}^{d} A^{j-i-1} B \uv_{t-j}
   + \sum_{j=1}^{d} A^{j-1} B \uv_{t-j} \\
&= -\sum_{i=1}^{d} a_i \x_{t-i}
   + \sum_{j=1}^{d}
     \left(
       A^{j-1} + \sum_{i=1}^{j-1} a_i A^{j-1-i}
     \right) B \uv_{t-j} \\
&= -\sum_{i=1}^{d} a_i \x_{t-i}
   + \sum_{j=1}^{d}
     \left(
       \sum_{i=0}^{j-1} a_i A^{j-1-i}
     \right) B \uv_{t-j}.
\end{align*}
Multiplying both sides by $C$ yields
\[
\y_t
=
-\sum_{i=1}^{d} a_i \y_{t-i}
+
\sum_{j=1}^{d} C\left(\sum_{i=0}^{j-1} a_i A^{j-1-i}\right) B \uv_{t-j},
\]
which is exactly the claimed recurrence.
\end{proof}

The theorem shows that every noiseless $d_x$-dimensional linear dynamical system induces an exact linear predictor of memory at most $d_x$. Thus, if an upper bound on the hidden-state dimension is known, then it suffices to compete with a class of linear predictors of that memory length.

\begin{corollary}\label{cor:Pi-star-contained-in-linear}
Let $\pi_{\{A,B,C,\x_0\}} \in \Pi^\star$ be any noiseless linear dynamical predictor
with hidden-state dimension at most $d_x$, and let $\{\y_t\}_{t=1}^T$ be the output
sequence it generates under an arbitrary input sequence $\{\uv_t\}_{t=1}^T$. Then
there exists $\kappa < \infty$ and a predictor $\pi \in \Pi^{L}_{d_x,d_x,\kappa}$
such that
\[
\hat{\y}_t^{\pi}
=
\hat{\y}_t^{\pi_{\{A,B,C,\x_0\}}}
=
\y_t
\qquad
\text{for all } t > d_x .
\]
Consequently, along every realizable trajectory of a noiseless
$d_x$-dimensional linear dynamical system, the class
$\Pi^{L}_{d_x,d_x,\kappa}$ exactly represents the induced linear dynamical predictor
after a warm-up of $d_x$ rounds.
\end{corollary}

\begin{proof}
Theorem~\ref{thm:cayley-hamilton-linear-predictors} explicitly constructs matrices
$M_{1,1:d_x}$ and $M_{2,1:d_x}$ that realize the predictor as a linear predictor of
memory $d_x$ for all times $t>d_x$. Choosing $\kappa$ larger than the corresponding
sum of matrix norms yields the claim.
\end{proof}

This exact finite-memory representation is one of the main reasons linear predictors are useful. It says that, in the noiseless setting, learning a linear predictor is not merely a heuristic surrogate for learning an unknown linear dynamical system: after an initial transient of length at most $d_x$, it is expressive enough to represent the predictor exactly.

In the noisy setting, the same argument gives the same linear recurrence plus an additive disturbance term that depends on recent process and observation noises. Thus, linear predictors remain a natural benchmark even in the stochastic case, though they are no longer exact.

\section{Efficient Learning of Linear Predictors}

The previous section gives several natural classes of predictors as well as some of the relationships between them. In particular, the class of linear dynamical predictors is natural but nonconvex in the matrices $A,B,C$.
The Cayley--Hamilton theorem circumvents this challenge: every noiseless $d_x$-dimensional LDS induces a predictor in the convex class $\Pi^{L}_{d_x,d_x,\kappa}$ for an appropriate $\kappa$.

The following algorithm learns the class of linear predictors in the sense that it attains a sublinear regret bound for any sequence of convex loss functions.

\begin{algorithm}[h]
		\caption{\label{alg:prediction1} Learning $\Pi^{L}$ by Online Gradient Descent}
		\begin{algorithmic}[1]
			\STATE Input: initial predictor matrices $M^1 = (M^1_{1,1:h},M^1_{2,1:k}) \in \K$, convex constraint set $\K$
			\FOR {$t=1$ to $T$}
			\STATE Predict
            $$ \hat{\y}_t(M^t) = \sum_{i=1}^h M^t_{1,i} \y_{t-i} + \sum_{j=1}^k M^t_{2,j} \uv_{t-j}, $$
            and observe true $\y_t$.
			\STATE Suffer loss $\ell_t(\hat{\y}_t(M^t),\y_t)$ and update
			\begin{align*}
			& M^{t+\frac{1}{2}} = M^{t}- \eta_{t} \nabla \ell_{t}(M^{t}) \\
			& M^{t+1} = \proj_\K(M^{t+\frac{1}{2}})
			\end{align*}
			\ENDFOR
		\end{algorithmic}
\end{algorithm}

Here $\ell_t(M)$ is shorthand for the loss incurred by predictor $M$ at time $t$, namely $\ell_t(\hat{\y}_t(M),\y_t)$.

This algorithm is an instance of the Online Gradient Descent algorithm, and Theorem \ref{thm:ogd} directly gives the following bound for it. We denote by $D$ the diameter of the set $\K$ and by $G$ an upper bound on the norm of the gradients of the loss functions $\ell_t(M)$ over $\K$.
\begin{corollary}
\label{cor:learn-lds-1}
For the choice of step sizes $\eta_t = \frac{D}{G \sqrt{t}}$, and $\mA$ being Algorithm \ref{alg:prediction1}, we have
$$ \regret_T ( \mA , \Pi^L_{h,k,\kappa} ) \leq 2 GD \sqrt{T}. $$
\end{corollary}

Combining Corollary \ref{cor:Pi-star-contained-in-linear} with Corollary \ref{cor:learn-lds-1}, we conclude that if the observations are generated by an unknown noiseless linear dynamical system of hidden-state dimension at most $d_x$, then learning $\Pi^L_{d_x,d_x,\kappa}$ suffices to obtain sublinear regret with respect to the best linear dynamical predictor in hindsight, up to the lower-order cost of the first $d_x$ initialization rounds.

\section{Summary}

In this chapter, we explored the problem of learning in unknown linear dynamical systems, where the goal is to predict future system outputs based on past observations while the underlying system parameters remain unknown. Unlike traditional control settings where an agent actively influences system behavior, the learning problem here is passive, focusing on accurate forecasting without intervention.

We began by formulating the problem in terms of partially observable LDS, highlighting the challenges posed by unknown system matrices and noisy observations. We then introduced several natural classes of predictors and explained how to compare them through regret minimization.

The key technical observation of the chapter is that every noiseless finite-dimensional linear dynamical system induces an exact finite-memory linear predictor. This follows from the Cayley--Hamilton theorem, and it explains why simple linear predictors are far more expressive than they might initially appear. Rather than identifying the unknown system matrices explicitly, we can compete directly with the induced input-output predictor.

Finally, we showed that linear predictors can be learned efficiently by online convex optimization, using Online Gradient Descent to obtain sublinear regret.

\ifarxiv
\newpage
\fi
\section{Bibliographic Remarks}
The task of predicting future observations in an online environment is often called ``filtering''. Resting at the crossroads between statistical estimation and dynamical systems, the theory of filtering had its early beginnings in the works of Norbert Wiener (see, e.g., \citet{wiener1949extrapolation}). 

Learning linear predictors with finite history is a classical consequence of the Cayley--Hamilton theorem and realization theory; see, e.g., \cite{kailath1980linear,chen1999linear}.

For a comprehensive treatment of online learning in games and online convex optimization, see \citet{cesa2006prediction,hazan2016introduction}.

In the statistical noise setting, a natural approach for learning is to perform system identification and then use the identified system for prediction. This approach was taken by \citet{simchowitz2018learning,pmlr-v119-simchowitz20a,pmlr-v97-sarkar19a}. The work of \citet{ghai2020no} extends identification-based techniques to adversarial noise and marginally stable systems.

The class $\Pi^\star$, which is the class of optimal open-loop predictors for systems with full observation, was studied in \citet{hazan2017learning}. That paper showed how to learn partially observable and even marginally stable dynamical systems whose spectral radius approaches one, as long as their transition matrix is symmetric.

Under probabilistic assumptions, the work of \citet{hardt2018gradient} shows that the matrices $A,B,C,D$ of an unknown and partially observed time-invariant linear dynamical system can be learned using first-order optimization methods. Learning without recovery was studied in \cite{ghai2020no} even for marginally stable systems. More recently, tensor methods were used in \cite{bakshi2023new}, and in \cite{bakshi2023tensor} to learn a mixture of linear dynamical systems.

The idea that a finite-dimensional state-space model induces a finite-order input-output recurrence is classical in linear systems theory. In the context of this chapter, the Cayley--Hamilton theorem provides a particularly clean explanation for why linear predictors can exactly represent noiseless finite-dimensional LDS.

\ifarxiv
\newpage
\fi
\begin{exercises}

\begin{exercise}
Prove that Lemma \ref{lem:approximation_of_open_loop_predictors} holds.
\end{exercise}

\begin{exercise}
Assume the system is fully observed, so that $C=I$, and that the process noise
has zero conditional mean given the past. Show that the open-loop predictor
obtained by unrolling the dynamics minimizes the expected squared prediction
error among all predictors measurable with respect to the past inputs.
\end{exercise}

\begin{exercise}\label{exercise:h-is-zero-for-olop}
Show that every truncated open-loop predictor of the form
$$ \hat{\y}_t = \sum_{i=1}^h H_i \uv_{t-i} $$
is a special case of a linear predictor in the class $\Pi^{L}_{0,h,\kappa}$ for an appropriate choice of $\kappa$.
\end{exercise}

\begin{exercise}
Assume that $\|\uv_t\| \le U$, $\|\y_t\| \le Y$ for all $t$, and that the
parameter space defining $\Pi^L_{h,k,\kappa}$ is endowed with a norm under
which its diameter is at most $D$. Derive an explicit upper bound on
$\regret_T(\mA,\Pi^L_{h,k,\kappa})$ for Online Gradient Descent in terms of
$h,k,\kappa,U,Y,D$, and $T$.
\end{exercise}

\begin{exercise}
Assume the square loss, and suppose moreover that predictions, outputs, and the
parameter domain are uniformly bounded. Give an algorithm based on Online
Newton Step for learning $\Pi^L_{h,k,\kappa}$, and derive the corresponding
regret bound.
\end{exercise}

\end{exercises}

\chapter{Kalman Filtering}
In the previous chapter we defined what it means to learn in dynamical systems, and established relationship between prediction classes. Most notably, we showed that despite nonconvexity of the class of linear dynamical predictors, they can be learned exactly using the convex relaxation of linear predictors. 

In this chapter we show another approach to learn the class of linear dynamical predictors. This approach is more limited in terms of the generality where it applies, it is based on generative data and statistical assumptions about the noise. However, it allows for more than just prediction: Kalman filtering allows state estimation which is very useful by its own right in various applications.

\section{Observable Systems}

Observability asks whether the hidden state of a partially observed dynamical system can be recovered from its observations. This is the basic structural condition that makes state estimation meaningful: if two different hidden states can produce the same observation history, then no estimator---Kalman filter or otherwise---can determine the state uniquely.

This notion is dual to controllability, but conceptually different. Controllability asks whether one can drive the state by choosing controls; observability asks whether one can infer the state from what is observed. For linear systems, observability admits a clean rank characterization through the Kalman observability matrix.

\begin{definition}
Consider the noiseless partially observed linear system
\[
\x_{t+1}=A\x_t+B\uv_t,
\qquad
\y_t=C\x_t,
\]
with known input sequence $\{\uv_t\}_{t\ge 0}$. We say that the pair $(A,C)$ is
observable if for every two initial states $\x_0,\x_0'$, equality of the
resulting observation sequences under the same inputs implies $\x_0=\x_0'$.
Equivalently, the observation history uniquely determines the initial state.
\end{definition}

A simple example of an unobservable system is the following variant of the double integrator from Section \ref{section:double-integrator}. Consider the double integrator dynamical system with $\Delta =1$ and partial observability where the observation matrix is given by
$$
A = \begin{bmatrix} 1 & 1 \\ 0 & 1 \end{bmatrix},
\qquad
C = \begin{bmatrix} 0 & 1 \end{bmatrix}.
$$
This system allows only the velocity of the object to be observed, while the position remains hidden. It can be seen that this setting is unobservable, as the initial position cannot be determined on the basis of the velocities and forces applied to the system alone. Indeed, this is a special case of the Galilean relativity principle of Newtonian mechanics, which states that the laws of physics are the same in all inertial reference frames.

For linear systems, observability can be verified using a rank condition on the Kalman observability matrix, given by
$$
\hat{K}(A,C) \defeq \begin{bmatrix} C \\ CA \\ \dots \\ CA^{d_\x-1} \end{bmatrix}.
$$
where $A$ and $C$ are the system matrices that define the state transition and observation model, respectively. Notice that in our example, the Kalman observability matrix is given by
$$
\hat{K}(A,C) = \begin{bmatrix} 0 & 1 \\ 0 & 1 \end{bmatrix},
$$
which is degenerate. This is not by chance: the system is observable if and only if the observability matrix has full rank. We ask the reader to provide a proof of the following result in Exercise~\ref{exer:obsrank}.

\begin{theorem}\label{thm:obsrank}
A partially-observed dynamical system with system matrices $(A,C)$ is observable if and only if $\rank (\hat{K}(A,C))=d_{\x}$.
\end{theorem}

\section{The Kalman Filter}

Chapter 9 focused on open-loop predictors, which map past observations and controls to a forecast of the next output. Such predictors do not maintain or update an explicit estimate of the hidden state.  The Kalman filter is the canonical closed-loop predictor for linear systems with zero-mean process and observation noise. It recursively combines the system dynamics with the latest measurement, producing the best linear least-squares estimate of the hidden state when the system matrices and noise covariances are known.

Formally, let $\w_t$ and $\vv_t$ denote the process and observation noise, assumed to be zero-mean and i.i.d., with covariance matrices
$$
\Sigma_x \defeq \mathbb{E}[\w_t\w_t^\top],
\qquad
\Sigma_y \defeq \mathbb{E}[\vv_t\vv_t^\top].
$$

Although our ultimate goal is to predict observations, it is convenient to first estimate the hidden state. Once we have a least-squares estimate $\hat{\x}_t$ of $\x_t$ given the past, the corresponding least-squares prediction of the observation is $\hat{\y}_t = C\hat{\x}_t$; see Exercise~\ref{exer:xtoy}.

The following theorem is the essence of the Kalman filter. It constructs a recursive state estimator that is optimal among linear estimators based on past observations and controls.

\begin{theorem}\label{thm:kalman}
	Let the control sequence $\uv_t$ be an arbitrary affine function of $\y_{1:t}$.
	There exists an efficiently computable sequence of linear state estimates $\hat{\x}_{t+1}$ defined recursively as
	$$ \hat{\x}_{t+1} = (A-L_t C)\hat{\x}_t + B\uv_t + L_t \y_t, $$
	such that
	$$
	\mathbb{E}\|\x_t-\hat{\x}_t\|^2
	\le
	\min_{L}
	\mathbb{E}\|\x_t-L[\y_{1:t-1},\uv_{1:t-1}]\|^2,
	$$
	where the minimum ranges over all linear maps from $(\y_{1:t-1},\uv_{1:t-1})$ to $\mathbb{R}^{d_\x}$.
\end{theorem}

To prove this theorem, we require some basic tools from linear regression which we state next.

\subsection{Tools from Linear Regression}

We begin with an observation about least-squares predictors, which we ask the reader to prove in Exercise~\ref{exer:decomp-lin}.
\begin{lemma}\label{lem:decomp-lin}
	Let $X, Y, Z$ be matrix-valued random variables. Let
	$$ A^* = \argmin_{A}\mathbb{E}\|X-AY\|^2, \text{ and } B^* = \argmin_{B}\mathbb{E}\|X-BZ\|^2.$$
	If $\mathbb{E}\left[YZ^\top \right]=0$, then
	$$ \mathbb{E} \|X-A^*Y-B^*Z\|^2 \leq \min_{A, B} \mathbb{E} \|X-AY-BZ\|^2.$$
\end{lemma}

In the remaining proof, we use this observation to construct an optimal least-squares estimator $\hat{\x}_{t+1}$ for $\x_{t+1}$ from the available information. In particular, by the induction hypothesis,
$$
\mathbb{E} \|\x_t - \hat{\x}_t\|^2
\leq
\min_{L} \mathbb{E} \|\x_t - L\left[\y_{1:t-1}, \uv_{1:t-1}\right]\|^2,
$$
where the minimum ranges over all linear maps from $(\y_{1:t-1},\uv_{1:t-1})$ to $\mathbb{R}^{d_\x}$. But is $\hat{\x}_t$ any good for predicting $\x_{t+1}$? To answer this, we note the following structural characterization of least-squares (Exercise~\ref{exer:ls}).
\begin{lemma}\label{lem:ls}
	Let $X, Y$ be possibly correlated random variables. Then
	$$ A^*\defeq \argmin_A \mathbb{E} \|X-AY\|^2 = \mathbb{E}\left[XY^\top \right] \left(\mathbb{E}\left[YY^\top\right]\right)^{-1}. $$
	Furthermore, if $\hat{X}= A^* Y$, then $\mathbb{E} (X-\hat{X})Y^\top=0$.
\end{lemma}

\subsection{Analysis}
We are now ready to prove the main property of the Kalman filter below. Throughout this section, assume that $\Sigma_y \succ 0$. 
\begin{proof}[Proof of Theorem \ref{thm:kalman}]
	Since the estimator is defined recursively, a natural strategy is to proceed by induction. Let us say $\hat{\x}_t$ is such a predictor. Furthermore, let
	$$ \Sigma_t \defeq \mathbb{E} \left[(\x_t - \hat{\x}_t) (\x_t - \hat{\x}_t)^\top\right]. $$

	Since for any $s<t$,
	$$
	\mathbb{E} \left[(\x_{t+1} - B\uv_t)\y_s^\top \right]
	=
	\mathbb{E} \left[(A\x_t + \w_t)\y_s^\top \right]
	=
	A\mathbb{E} \left[\x_t\y_s^\top \right],
	$$
	using the first part of Lemma~\ref{lem:ls}, we have that
	$$
	\mathbb{E} \|\x_{t+1} - A\hat{\x}_t- B\uv_t\|^2
	\leq
	\min_{L}
	\mathbb{E} \|\x_{t+1} - B\uv_t- L[\y_{1:t-1}, \uv_{1:t-1}]\|^2,
	$$
	where the minimum ranges over all linear maps from $(\y_{1:t-1},\uv_{1:t-1})$ to $\mathbb{R}^{d_\x}$.

	But, when predicting $\x_{t+1}$, we have more information. In particular, we also have $\y_t$. Consider the random variable $\y_t - C\hat{\x}_t$. Can we use this covariate alone to predict $\x_{t+1}$? Note that
	\begin{align*}
		\mathbb{E}\left[(\y_t-C\hat{\x}_t)(\y_t-C\hat{\x}_t)^\top \right]
		&= \Sigma_y + C\mathbb{E} \left[(\x_t-\hat{\x}_t)(\x_t-\hat{\x}_t)^\top \right] C^\top  \\
		&= \Sigma_y + C\Sigma_t C^\top,
	\end{align*}
	\begin{align*}
		\mathbb{E}\left[(\x_{t+1}-B\uv_t)(\y_t-C\hat{\x}_t)^\top\right]
		&= \mathbb{E}\left[(A\x_t + \w_t ) (C(\x_t-\hat{\x}_t) + \vv_t)^\top\right]  \\
		&= A \mathbb{E}\left[(\x_t-\hat{\x}_t)(\x_t-\hat{\x}_t)^\top\right]C^\top + A \underbrace{\mathbb{E}\left[\hat{\x}_t(\x_t-\hat{\x}_t)^\top\right]}_{=0}C^\top\\
		&= A \Sigma_t C^\top,
	\end{align*}
	where the equality indicated in braces follows from the second part of Lemma~\ref{lem:ls}, by noting that $\hat{\x}_t$ is a linear function of $\y_{1:t-1}$.

	In the preceding paragraphs, we have constructed two least-squares predictors for ${\x}_{t+1}$. Observe that for any $s<t$,
	$$
	\mathbb{E}(\y_t-C\hat{\x}_t)\y_s^\top
	=
	C\mathbb{E}(\x_t -\hat{\x}_t)\y_s^\top
	=
	0,
	$$
	where the last equality once again follows from Lemma~\ref{lem:ls}. This orthogonality allows us to invoke Lemma~\ref{lem:decomp-lin} to arrive at
	$$
	\hat{\x}_{t+1} = A\hat{\x}_t + B\uv_t + \underbrace{A \Sigma_t C^\top \left(\Sigma_y + C\Sigma_tC^\top\right)^{-1}}_{\defeq L_t} (\y_t-C\hat{\x}_t),
	$$
	where the last component follows from Lemma~\ref{lem:ls} and the correlations we just computed.

	To establish that such predictors can be computed efficiently, note that
	$$
	\x_{t+1}-\hat{\x}_{t+1}
	=
	(A-L_t C)(\x_t - \hat{\x}_t) + \w_t - L_t \vv_t.
	$$
	Finally, we give a recursive relation on $\Sigma_t$ as follows:
	\begin{align*}
		\Sigma_{t+1}
		&\defeq \mathbb{E} \left[ (\x_{t+1} - \hat{\x}_{t+1})(\x_{t+1} - \hat{\x}_{t+1})^\top \right]  \\
		&= (A-L_tC) \Sigma_t (A-L_tC)^\top + \Sigma_x + L_t \Sigma_y L_t^\top\\
		&= A \Sigma_t A^\top + \Sigma_x + L_t (C\Sigma_t C^\top + \Sigma_y) L_t^\top - A\Sigma_t C^\top L_t^\top - L_t C\Sigma_t A^\top\\
		&= A\Sigma_t A^\top - A \Sigma_t C^\top (C\Sigma_t C^\top +\Sigma_y)^{-1} C \Sigma_t A^\top + \Sigma_x.
	\end{align*}
\end{proof}

\subsection{Infinite Horizon Kalman Filtering}

In the infinite-horizon setting, when the covariance recursion converges, the Kalman filter admits a steady-state description with a fixed gain matrix $L$. The resulting estimator is itself a linear dynamical system driven by the control inputs and the incoming observations:
\begin{align*}
	\hat{\x}_{t+1} &= (A-LC)\hat{\x}_t + B\uv_t + L\y_t, \\
	\hat{\y}_t &= C\hat{\x}_t,
\end{align*}
where $L=A\Sigma C^\top (C\Sigma C^\top + \Sigma_y)^{-1}$ and $\Sigma$ satisfies the Riccati equation
$$
\Sigma
=
A\Sigma A^\top
-
A\Sigma C^\top (C\Sigma C^\top +\Sigma_y)^{-1} C \Sigma A^\top
+
\Sigma_x.
$$

The existence of a stabilizing solution to this Riccati equation is a classical question. Under standard assumptions---typically detectability of $(A,C)$ together with an appropriate stabilizability condition for the state-noise model---the Riccati equation admits a stabilizing solution. In that case the steady-state Kalman gain $L$ is well defined and the closed-loop matrix
$A-LC$ is stable. We do not pursue the sharp conditions here.

\subsection{Linear Predictors Approximate the Kalman Filter}

At first sight, the steady-state Kalman filter appears to lie outside the class of predictors studied in Chapter 9 because it maintains an internal state estimate and uses feedback from the newest observation. The next lemma gives a relationship between the class of predictors. Once the closed-loop matrix $A-LC$ is stable, the Kalman recursion can be unrolled and truncated, yielding a finite-memory linear predictor with exponentially small approximation error. This is the closed-loop analogue of the truncation phenomenon from Chapter 9.

\begin{lemma}\label{lem:kalmancontain}
Fix $M \ge 1$. Consider the family of steady-state Kalman filters
\[
\hat{\x}_{t+1}=F\hat{\x}_t + B\uv_t + L\y_t,
\qquad
\hat{\y}_t = C\hat{\x}_t,
\]
with $\hat{\x}_1=0$, where $F=A-LC$ satisfies
\[
\|F^s\| \le M e^{-s/M}
\qquad\text{for all } s\ge 0,
\]
and $\|B\|+\|L\|+\|C\| \le M$, as wel as $\max_{t\in[T]} \|\uv_t\| \le M, \qquad \max_{t\in[T]} \|\y_t\| \le M$. 
Then the class $\Pi^L_{h,h,\kappa}$ $\eps$-approximates this family over horizon
$T$ provided
\[
h \ge M \log\!\left(\frac{4M^5}{\eps}\right),
\qquad
\kappa \ge 4M^4.
\]
\end{lemma}

\section{Bayes-Optimality under Gaussian Noise}

The previous section established the Kalman recursion as the optimal linear closed-loop estimator. Under Gaussian noise, this linear optimality strengthens to full Bayes optimality.\footnote{This is also called the Bayes-optimal predictor. See bibliographic materials for more details and literature.} In this section we assume that $\w_t$ and $\vv_t$ are i.i.d. Gaussian and show that the Kalman filter computes the conditional expectation of the hidden state, and hence the mean-square optimal predictor of the next observation.

Concretely, assume that $\w_t \sim \mathcal{N}(0,\Sigma_x)$ and $\vv_t \sim \mathcal{N}(0,\Sigma_y)$ are i.i.d. Gaussian and that $\x_0=0$. Given realizations $\y_1,\y_2,\dots,\y_t$ under a control sequence $\uv_1,\uv_2,\dots,\uv_t$, we would like to compute
\begin{align*}
	\hat{\y}_{t+1} = \mathbb{E}[\y_{t+1}\mid \y_{1:t},\uv_{1:t}] .
\end{align*}
Since $\hat{\y}_{t+1}=C\hat{\x}_{t+1}$, it suffices to recursively compute
$$
\hat{\x}_{t+1} = \mathbb{E}[\x_{t+1}\mid \y_{1:t},\uv_{1:t}] .
$$
\begin{theorem}
The conditional expectation of the state $\hat{\x}_{t+1}$ satisfies
	\begin{align*}
		\hat{\x}_{t+1} = (A-L_t C)\hat{\x}_t + B\uv_t + L_t \y_t,
	\end{align*}
for a sequence of matrices $L_t\in \mathbb{R}^{d_{\x}\times d_y}$ that can be efficiently computed given $(A,B,C)$, and is independent of the realizations $\uv_{1:T}, \y_{1:T}$.
\end{theorem}
\begin{proof}
From the Gaussian nature of perturbations, we observe that
\begin{align*}
	\begin{bmatrix}
		\x_t | (\y_{1:t-1}, \uv_{1:t-1}) \\
		\y_t | (\y_{1:t-1}, \uv_{1:t-1})
	\end{bmatrix}
	\sim
	\mathcal{N}\left(
	\begin{bmatrix}
		\hat{\x}_t \\
		C\hat{\x}_t
	\end{bmatrix},
	\begin{bmatrix}
		\Sigma_t & \Sigma_t C^\top \\
		C\Sigma_t & C\Sigma_t C^\top + \Sigma_{y}
	\end{bmatrix}\right).
\end{align*}
for some $\Sigma_t\in \mathbb{R}^{d_{\x}\times d_{\x}}$, for which we provide a recursive formula towards the end of the proof. We make note of the following lemma for Gaussian distributions, whose proof we ask the reader to complete in Exercise~\ref{exer:condgauss}.
\begin{lemma}\label{lem:condgauss}
	If a random variable pair $(\x, \y)$ are distributed as
	$$ \begin{bmatrix}
		\x \\ \y
	\end{bmatrix} \sim \mathcal{N} \left(\begin{bmatrix}
		\mu_x \\ \mu_y
	\end{bmatrix}, \begin{bmatrix}
		\Sigma_{11} & \Sigma_{12} \\
		\Sigma_{21} & \Sigma_{22}
	\end{bmatrix}\right), $$
	then $\x|\y \sim \mathcal{N}(\mu_x + \Sigma_{12}\Sigma_{22}^{-1} (\y-\mu_y), \Sigma_{11} - \Sigma_{12}\Sigma_{22}^{-1} \Sigma_{21}).$
\end{lemma}
Therefore, we have that
\begin{align*}
	\x_t | (\y_{1:t}, \uv_{1:t-1})
	&\sim
	\mathcal{N}\Big( \hat{\x}_t + \Sigma_t C^\top (C\Sigma_tC^\top +\Sigma_y)^{-1} (\y_t - C\hat{\x}_t), \Sigma'_t\Big), \\
	\Sigma'_t &= \Sigma_t - \Sigma_t C^\top (C\Sigma_tC^\top +\Sigma_y)^{-1} C \Sigma_t.
\end{align*}
Now, using this, we arrive at
\begin{align*}
	\hat{\x}_{t+1}
	&= A \mathbb{E}[\x_t | \y_{1:t}, \uv_{1:t-1}] + B\uv_t\\
	&= (A -  \underbrace{A\Sigma_t C^\top(C\Sigma_tC^\top +\Sigma_y)^{-1}}_{\defeq L_t}C) \hat{\x}_t + B\uv_t + \underbrace{A\Sigma_t C^\top (C\Sigma_tC^\top +\Sigma_y)^{-1}}_{\defeq L_t} \y_t.
\end{align*}
It remains to obtain a recursive relation on $\Sigma_t$. To this extent, from the state evolution of a linear dynamical system, we have that
$$
\Sigma_{t+1}
=
A\Sigma'_t A^\top + \Sigma_{x}
=
A\Sigma_t A^\top - A \Sigma_t C^\top (C\Sigma_tC^\top +\Sigma_y)^{-1} C \Sigma_t A^\top+ \Sigma_{x}.
$$
\end{proof}

\section{Conclusion}

This chapter extends the story of Chapter 9 from open-loop prediction to closed-loop state estimation. The main conceptual point is that feedback changes the benchmark but not the convex relaxation: the Kalman filter is the natural optimal closed-loop predictor, yet stable Kalman filters are still approximable by finite-memory linear predictors.

We first introduced observability as the structural condition that makes state estimation possible. We then derived the Kalman recursion and its steady-state form, and showed that under Gaussian noise it is Bayes-optimal. Finally, we connected this classical estimator back to the learning perspective of the previous chapter by showing that finite-memory linear predictors can approximate stable Kalman filters. Consequently, the online learning machinery developed in Chapter 9 continues to apply in partially observed stochastic systems.

The next chapter moves from Kalman filtering to spectral filtering. There the emphasis shifts again from explicit state estimation to improper learning of predictors, but the central theme remains the same: exploit linear structure while avoiding direct nonconvex identification of the underlying dynamical system.

\ifarxiv
\newpage
\fi
\section{Bibliographic Remarks}

In 1960, Rudolph Kalman gave a state-space solution to the filtering problem that was highly amenable to further extensions \citep{kalman1960new}. See \citet{anderson1991kalman} for a historical perspective on its development and aftermath.

For unknown systems, in the stochastic case, one may estimate the system matrices, as we have discussed before, before applying optimal filtering. Improper learning approaches to this problem were studied in \citet{kozdoba2019line}.

State estimation is a fundamental problem in control and signal processing, with numerous approaches beyond the classical Kalman filter. Alternative methods include:

\begin{itemize}
    \item {\bf Nonlinear Filtering Extensions.}
The Kalman filter assumes that the dynamics is linear and the noise is Gaussian, which limits its effectiveness in nonlinear settings. The Extended Kalman Filter (see, for example, \citet{urrea2021kalman}) addresses this issue by linearizing the dynamics at each time step via a first-order Taylor series expansion. However, linearization can introduce significant errors in highly nonlinear systems. Notable extensions using more sophisticated sampling include the Unscented Kalman Filter \citep{julier2004unscented}, leading to better estimation accuracy in strongly nonlinear systems.

\item {\bf Particle Filtering (Sequential Monte Carlo Methods).}
For highly nonlinear and non-Gaussian settings, particle filters \citep{gordon1993novel} provide a flexible alternative. These methods approximate the posterior distribution using a set of weighted samples (particles) and employ importance sampling and resampling techniques.

\item {\bf Adversarial \& Robust Filtering.}
Robust filtering methods seek to provide guarantees under adversarial noise. The $H_\infty$ filter \citep{bacsar2008h} minimizes the worst-case estimation error rather than the mean-squared error, making it suitable for applications with unknown disturbances.
\end{itemize}

In the next chapter, we will study an alternative using a technique which circumvents the need for state estimation altogether and has dimension-free guarantees for certain cases of sequential prediction.

\ifarxiv
\newpage
\fi
\begin{exercises}

\begin{exercise}\label{exer:obsrank}
In this exercise we prove Theorem~\ref{thm:obsrank}.

\begin{enumerate}
    \item Suppose $\rank(\hat K(A,C)) < d_{\x}$. Show that there exists a
    nonzero vector $v \in \ker(\hat K(A,C))$. Conclude that the two initial
    states $\x_0$ and $\x_0+v$ generate the same observation sequence under the
    same input sequence, and hence the system is not observable.

    \item Suppose $\rank(\hat K(A,C)) = d_{\x}$. Show that the map
    \[
    \x_0 \mapsto
    \begin{bmatrix}
    C\x_0\\
    CA\x_0\\
    \vdots\\
    CA^{d_{\x}-1}\x_0
    \end{bmatrix}
    \]
    is injective. Conclude that, given the input sequence, the first $d_{\x}$
    observations uniquely determine the initial state.
\end{enumerate}
\end{exercise}

\begin{exercise}\label{exer:ls}
Prove Lemma~\ref{lem:ls}, which states the following. Let $X, Y$ be possibly correlated random variables. Then
		$$ A^*\defeq \argmin_A \mathbb{E} \|X-AY\|^2 = \mathbb{E}\left[XY^\top \right] \left(\mathbb{E}\left[YY^\top\right]\right)^{-1}. $$
		Furthermore, if $\hat{X}= A^* Y$, then $\mathbb{E} (X-\hat{X})Y^\top=0$.
\end{exercise}

\begin{exercise}\label{exer:decomp-lin}
Using Lemma~\ref{lem:ls}, prove Lemma~\ref{lem:decomp-lin}, which states the following.
		Let $X, Y, Z$ be matrix-valued random variables. Let
		$$ A^* = \argmin_{A}\mathbb{E}\|X-AY\|^2, \text{ and } B^* = \argmin_{B}\mathbb{E}\|X-BZ\|^2.$$
		If $\mathbb{E}\left[YZ^\top \right]=0$, then
		$$ \mathbb{E} \|X-A^*Y-B^*Z\|^2 \leq \min_{A, B} \mathbb{E} \|X-AY-BZ\|^2.$$
\end{exercise}

\begin{exercise}\label{exer:kalmancontain}
Prove Lemma~\ref{lem:kalmancontain} by unrolling the steady-state Kalman
recursion and truncating the resulting linear predictor. Conclude that, under
the assumptions of the lemma, the family of stable steady-state Kalman filters
can be learned improperly using Algorithm~\ref{alg:prediction1}.
\end{exercise}

\begin{exercise}\label{exer:condgauss}
Prove Lemma~\ref{lem:condgauss}, which is stated below. If a random variable pair $(\x, \y)$ are distributed as
	$$ \begin{bmatrix}
		\x \\ \y
	\end{bmatrix} \sim \mathcal{N} \left(\begin{bmatrix}
		\mu_x \\ \mu_y
	\end{bmatrix}, \begin{bmatrix}
		\Sigma_{11} & \Sigma_{12} \\
		\Sigma_{21} & \Sigma_{22}
	\end{bmatrix}\right), $$
	then the conditional distribution of $\x$ given $\y$ can be written as $\x|\y \sim \mathcal{N}(\mu_x + \Sigma_{12}\Sigma_{22}^{-1} (\y-\mu_y), \Sigma_{11} - \Sigma_{12}\Sigma_{22}^{-1} \Sigma_{21}).$
\end{exercise}

\begin{exercise}\label{exer:xtoy}
Using Lemma~\ref{lem:ls}, prove that the optimal least-squares estimator for the observation at time $t$ given past observations $y_{1:t-1}$ and control inputs $u_{1:t-1}$ is given by $\hat{\y}_t = C\hat{\x}_t$, where $\hat{\x}_t$ is as defined in Theorem~\ref{thm:kalman}.
\end{exercise}

\end{exercises}

\chapter{Spectral Filtering} \label{chap:sf}

In the previous chapters, we explored two major approaches for state estimation and prediction in unknown linear dynamical systems:
\begin{enumerate}
    \item State estimation via Kalman filtering, which recursively estimates both the latent system state and the projection matrices that map the system state to observations. Kalman filtering provides an optimal solution under Gaussian noise, and updates the state estimates over time. However, it requires an accurate model of system dynamics, including transition and observation matrices, which may be difficult to estimate, especially in adversarial or high-noise environments.

\item 
Online learning of linear predictors, which offers a relaxation of the Kalman filtering approach by directly learning predictive models without explicit system identification. Instead of estimating the full state and transition matrices, this approach constructs a forecasting model using past inputs and outputs. Although this avoids the complexity of full system identification, it introduces a major limitation: the number of parameters required depends on the hidden dimension of the system, or the stability constant of the system, which can be large and unknown, making learning impractical for high-dimensional or poorly conditioned systems.
\end{enumerate}

In this chapter, we introduce Spectral Filtering, an alternative approach that bypasses these limitations. Rather than relying on state estimation or direct system identification, spectral filtering constructs predictors on a transformed basis, which does not depend on the stability properties of the system. This method leverages spectral decompositions to efficiently extract the relevant dynamical structure, enabling more robust and scalable learning in both stable and unstable systems.

To illustrate the power of spectral filtering, we first present a one-dimensional case, which provides an intuitive understanding of how spectral methods can be applied to system identification and forecasting. We then extend the discussion to higher-dimensional settings, demonstrating how spectral predictors enable a structured and computationally efficient representation of system dynamics.

Although this chapter primarily focuses on symmetric system matrices for clarity, the principles of spectral filtering extend to asymmetric and more general linear dynamical system settings. These extensions require additional mathematical tools, which we reference in the bibliographic section.

\section{Spectral filtering in One Dimension}

In this section, we give a simple introduction to the technique of spectral filtering, restricting ourselves to a single dimension. 
The restriction to a single dimension simplifies the exposition and is yet sufficient to illustrate the main ideas. In the next section, we consider the high-dimensional case. 

The one-dimensional systems which we consider next are scalar systems, with $d_u=d_y=d_x=1$. For simplicity, we assume that the initial state is zero $\x_0=0$.
Recall the set of optimal predictors, the class $\Pi_\star$, which for scalar systems can be written and simplified to, 
\begin{align} \label{eqn:mu_alpha}
\y_t   & =  \sum_{i=1}^{t-1} C A^{i-1} B \uv_{t-i}  =  \beta   \sum_{i=1}^{t-1}  \alpha^{i-1} \uv_{t-i} = \beta \cdot \tilde{\uv}_t^\top \mu_\alpha,
\end{align}
where $\beta = C  B \in \reals$ is a scalar and we used the vector notation for 
$$\mu_\alpha =  \begin{bmatrix}  1 & \alpha & \alpha^2 & \ldots & \alpha^{T-1}  \end{bmatrix}^\top \in \reals^T,  $$ 
and we define $\tilde{\uv}_{t}$ to be the padded vector $\uv_{t-1:1}$:
$$ \tilde{\uv}_t = \begin{bmatrix} \uv_{t-1} &  \ldots & \uv_1 & 0 & \ldots & 0  \end{bmatrix}^\top \in \reals^T  . $$

The main idea of spectral filtering is to rewrite the vectors $\mu_\alpha$ in a more {efficient} basis, as opposed to the standard basis. The notion of efficiency comes from the definition of $\eps$-approximation that we have studied in previous chapters. 
To define the basis, we consider the following matrix:
$$Z_T = \int_0^1\mu_\alpha  \mu_\alpha^\top d\alpha \in \reals^{T \times T}  .$$

The matrix $Z_T$ is symmetric and is independent of $\alpha$. 
Let $\phi_1, ...,\phi_h, ... \in \reals^T$ be the eigenvectors of $Z_T$ ordered by the magnitude of their corresponding eigenvalues. Our basis for representing $\mu_\alpha$ is spanned by the eigenvectors of $Z_T$, and we proceed to study their properties. 

\begin{figure}[ht]
    \centering
    \includegraphics[scale = 0.5]{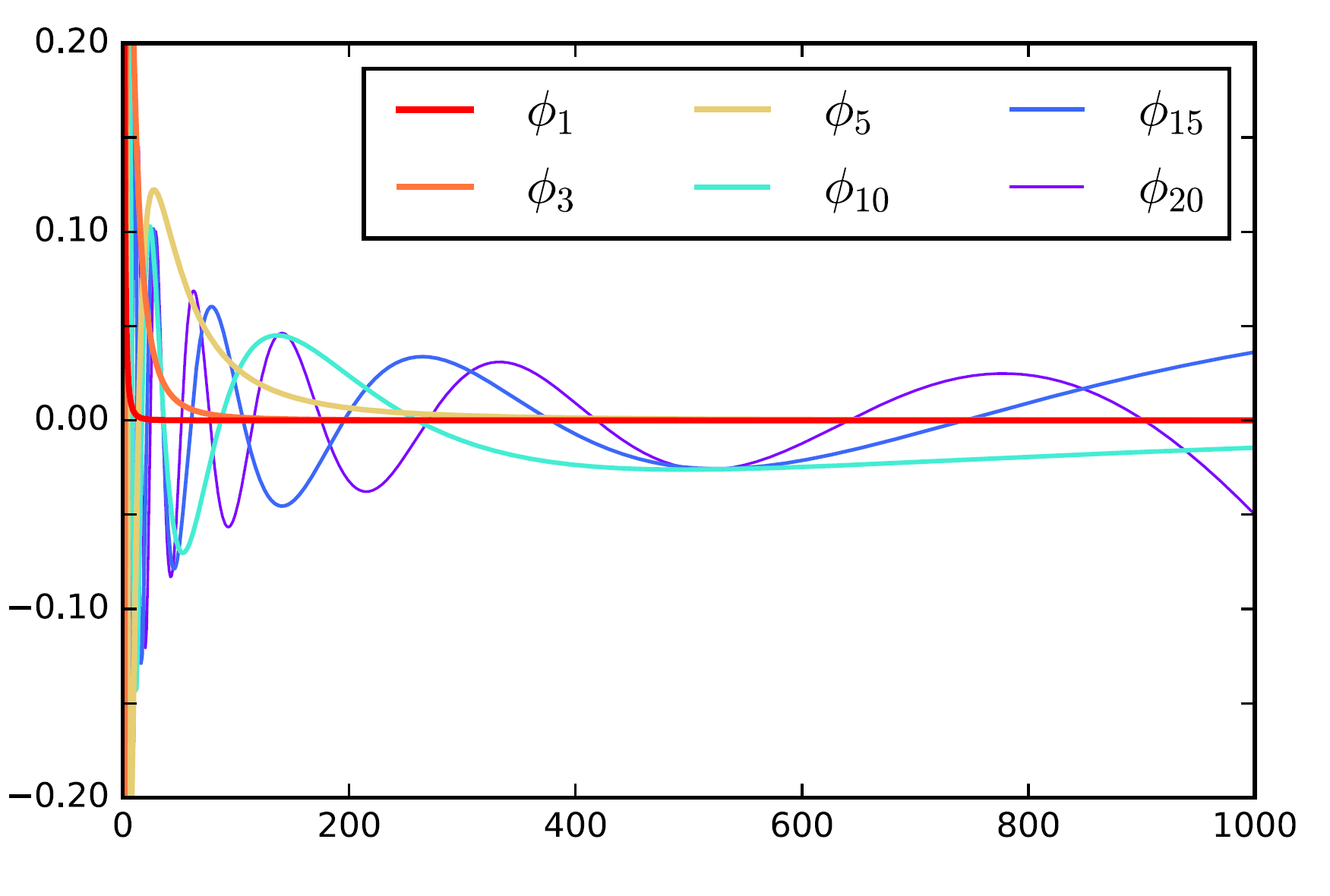}
    \caption{The filters obtained by the eigenvectors of $Z_T$.}
    \label{fig:filterbank}
\end{figure}

\subsection{The Magic of Hankel Matrices}

The matrix $Z_T$ is called the Hilbert matrix, and it is a special case of a {Hankel Matrix}, which is a square Hermitian matrix with each ascending skew-diagonal from left to right having the same value. To spell out $Z_T$ more precisely, 
\begin{align*}
    Z_T &= \int_0^1\begin{bmatrix}
            1 & \alpha & \alpha^2 & & \\
            \alpha & \alpha^2 & \alpha^3 & & \\
            \alpha^2 & \alpha^3 & \alpha^4 & & \\
            & & & ... & \\
            & & & & \alpha^{2T-2}
            \end{bmatrix} d\alpha
      = \begin{bmatrix}
            1 & \frac{1}{2}& \frac{1}{3} & & \\
            \frac{1}{2} & \frac{1}{3} & \frac{1}{4} & & \\
            \frac{1}{3} & \frac{1}{4} & \frac{1}{5} & & \\
            & & & ... & \\
            & & & & \frac{1}{2T-1}
            \end{bmatrix} .
\end{align*}

The properties of Hankel matrices have been studied extensively, and the following theorem captures one of their crucial properties. 
\begin{theorem}
\label{thm:BT}
Let $\sigma_k$ be the $k$-th largest singular value of the Hilbert matrix $Z_T \in \reals^{T \times T}$, then for $k,T > 1$: 
$$\sigma_k(Z_T) \leq 4 \pi^2 \cdot e^{-\frac{k}{\log T}}$$
\end{theorem}
The proof of this theorem is beyond our scope, and references to details are given in the bibliographic material at the end of this chapter. 

\begin{figure}[h]
    \centering
    \includegraphics[width=.5\linewidth]{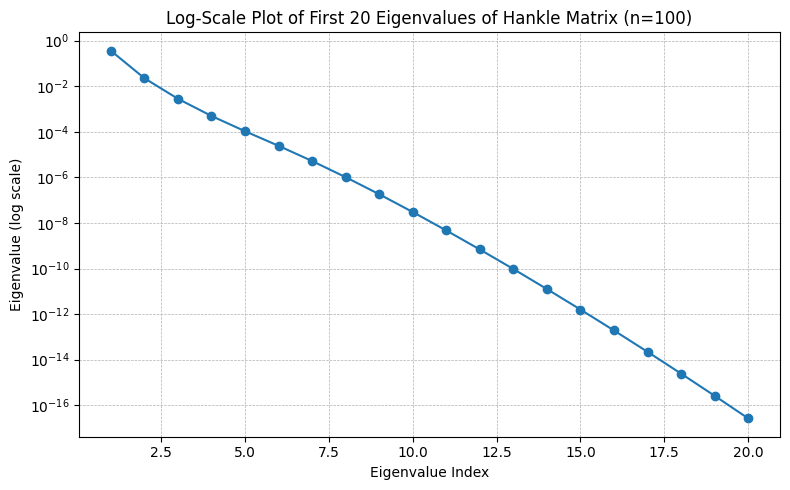}
    \caption{Eigenvalues of Hankel matrices decrease geometrically. These are the eigenvalues of the matrix $Z_T$, where $T=100$, plotted on a logarithmic scale.}
    \label{fig:hankeleigen}
\end{figure}

This theorem shows that the eigenvalues of a Hankel matrix decay very rapidly. An additional illustration of this fact is given in Figure \ref{fig:hankeleigen}, which plots the eigenvalues of the matrix $Z_T$ for $T = 100$ on a logarithmic scale. 

\section{Spectral Predictors}

The basis of eigenvectors of a Hankel matrix gives rise to an interesting class of predictors. This class of mappings does not directly apply to the history of inputs and/or observations. Instead, we first apply a linear transformation over these vectors, and then apply a linear map. 
More formally, 
\begin{definition} \label{def:phi_alpha_operator}
Define the class of spectral linear predictors of order $h$ as 
$$ \Pi_{h} = \left\{ \pi_{M} \ \ \left| \ \ \pi_{M} (\uv_{1:t},\y_{1:t})  =  \sum_{i=1}^h M_i \big( \tilde{\uv}_{t}^\top \phi_i \big) \right. \right\} . $$  
Here $\tilde{\uv}_{t}^\top \phi_i$ evaluates the inner product of the $i$'th eigenvector of $Z_T$ with the padded history vector $\tilde{\uv}_{t}$, where $ \tilde{\uv}_{t} = \begin{bmatrix} \uv_{t-1} &  \ldots & \uv_1 & 0 & \ldots & 0  \end{bmatrix}^\top \in \reals^T$.
\end{definition}

Notice that as $h \to T$, this class is equivalent to the class of all linear predictors on the inputs $\Pi^{LIN}$. The reader may wonder at this point: Since a linear transformation composed with another linear transform remains linear, what is the benefit here?  

The answer is that linear transformations are equivalent when we have full-rank transforms. In our case, however, we consider low-rank approximations to the transform, and the approximation properties of different subspaces can differ drastically! This is the case in our setting: the spectral basis arising from the matrix $Z_T$ gives excellent approximations to the class of (standard) linear predictors, as we formally prove next. 
The following Theorem is analogous to Lemma \ref{lem:lr_approx_prediction}, with the significant difference that the number of parameters $h$ does not depend on the stabilizability parameter of the system $\gamma$ at all! 
\begin{theorem} \label{thm:spectral_approx_prediction}
Suppose that $\x_0 = 0$ and $|u_t|\le 1$ for all $t$.
Then the class $\Pi_h$ for 
$h = O\!\left(\log^2 \frac{T}{\eps}\right)$ 
$\eps$-approximates the class $\Pi_\star$.
\end{theorem}

The proof of this lemma relies on the following crucial property of the spectral basis.

\begin{lemma} \label{lem:spectral-technical}
Let $\{\phi_j\}$ be the orthonormal eigenvectors of $Z_T$, with corresponding eigenvalues $\sigma_j$. 
Then for all $j \in [T]$, $j,T > 1$, and $\alpha \in [0,1]$, we have
\begin{align*}
|\phi_j^\top  \mu_\alpha|
\;\le\;
(4T^2 \sigma_j)^{1/4}.
\end{align*}
In particular, using Theorem~\ref{thm:BT},
\begin{align*}
|\phi_j^\top  \mu_\alpha|
=
O\!\left(
T^{1/2} e^{-\frac{j}{4\log T}}
\right).
\end{align*}
\end{lemma}

Using this property, we can prove Theorem \ref{thm:spectral_approx_prediction} by shifting to a representation on the spectral basis as follows.

\begin{proof}[Proof of Theorem \ref{thm:spectral_approx_prediction}]

Consider any sequence and any predictor from $\Pi_\star$, say $\pi_1=\pi_{A,B,C,0}$. 
Following equation \eqref{eqn:mu_alpha}, we can write
\begin{align*}
\y_t^{\pi_1}
&= \beta \cdot \tilde{\uv}_{t}^\top \mu_\alpha  \\
&= \beta \cdot \tilde{\uv}_{t}^\top 
\left( \sum_{i=1}^T \phi_i \phi_i^\top \right)
\mu_\alpha 
\qquad
\mbox{(basis property } \sum_{i=1}^T \phi_i \phi_i^\top = I ) \\
&= \sum_{i=1}^T M_i^\star \big( \tilde{\uv}_{t}^\top \phi_i \big),
\qquad
M_i^\star = \beta (\phi_i^\top  \mu_\alpha) .
\end{align*}

By Lemma \ref{lem:spectral-technical},
\begin{align}
|M_i^\star|
&=
|\beta|\, |\phi_i^\top  \mu_\alpha|
\le
|\beta|\,(4T^2 \sigma_i)^{1/4}
\notag\\
&=
O\!\left(
|\beta|\,T^{1/2} e^{-\frac{i}{4\log T}}
\right).
\label{eqn:shalom111-fixed}
\end{align}

Consider the predictor $\pi_2 \in \Pi_h$ that keeps only the first $h$ coefficients.
Then we have
\begin{eqnarray*}
| \y_t^{\pi_1} - \y_t^{\pi_2} |
&=
\left|
\sum_{i=1}^T M_i^\star \big( \tilde{\uv}_{t}^\top \phi_i \big)
-
\sum_{i=1}^h M_i^\star \big( \tilde{\uv}_{t}^\top \phi_i \big)
\right| \\
&=
\left|
\sum_{i=h+1}^T M_i^\star \big( \tilde{\uv}_{t}^\top \phi_i \big)
\right|  \\
&\le
\sum_{i=h+1}^T |M_i^\star|\, \big| \tilde{\uv}_{t}^\top \phi_i \big|  \\
&\le
\sum_{i=h+1}^T |M_i^\star| \, \|\phi_i\|_2 \|\tilde{\uv}_{t}\|_2 &
 \mbox{(Cauchy--Schwarz)}  \\
&\le \sqrt{T}\,
\sum_{i=h+1}^T |M_i^\star| & \|\phi_i\|_2=1,\; |u_t|\le 1  \\
&\le \sqrt{T}\,|\beta|\,T^{1/2}
\sum_{i=h+1}^\infty
e^{-\frac{i}{4\log T}} &  \eqref{eqn:shalom111-fixed}
\end{eqnarray*}

The geometric tail can be strictly bounded via an integral:
\begin{align*}
\sum_{i=h+1}^\infty
e^{-\frac{i}{4\log T}}
\le 
\int_h^\infty e^{-\frac{x}{4\log T}} dx
= 4\log T \cdot e^{-\frac{h}{4\log T}}.
\end{align*}

Thus, the overall error is bounded by:
\begin{align*}
| \y_t^{\pi_1} - \y_t^{\pi_2} | \le |\beta|\,T \big( 4\log T \big) e^{-\frac{h}{4\log T}}.
\end{align*}

Choosing
\begin{align*}
h
\ge
4\log T
\left(
\log T + \log(4\log T) + \log\frac{|\beta|}{\eps}
\right)
\end{align*}
ensures the per-round error is at most $\eps$, completing the proof.
\end{proof}

\begin{proof}[Proof of Lemma \ref{lem:spectral-technical}]

Consider the scalar function
\begin{align*}
g(\alpha) = (\mu_{\alpha}^\top \phi_j)^2
\end{align*}
over the interval $[0,1]$.

First,
\begin{align*}
\int_{0}^1 g(\alpha)\, d\alpha
&=
\int_{0}^1
(\phi_j^\top\mu_\alpha)^2
d\alpha \\
&=
\phi_j^{\intercal}
\left(
\int_{0}^1
\mu_{\alpha}\mu_{\alpha}^{\intercal}
d\alpha
\right)
\phi_j \\
&=
\phi_j^{\intercal}Z_T \phi_j
=
\sigma_j.
\end{align*}

Next, we bound the Lipschitz constant of $g$.

We have
\begin{align*}
\|\mu_\alpha\|_2^2
=
\sum_{t=0}^{T-1} \alpha^{2t}
\le T,
\end{align*}
and
\begin{align*}
\left\|
\frac{\partial}{\partial\alpha}
\mu_\alpha
\right\|_2^2
=
\sum_{t=0}^{T-1} (t \alpha^{t-1})^2
\le T^3.
\end{align*}

Thus
\begin{align*}
|g'(\alpha)|
&=
2 |\mu_{\alpha}^\top \phi_j|
\left|
\frac{\partial}{\partial \alpha}
(\mu_{\alpha}^\top \phi_j)
\right| \\
&\le
2 \|\mu_\alpha\|_2 \|\phi_j\|_2
\left\|
\frac{\partial}{\partial\alpha}
\mu_\alpha
\right\|_2 \\
&\le
2 \sqrt{T}\sqrt{T^3}
=
2T^2.
\end{align*}

Hence $g$ is $L$-Lipschitz with $L\le 2T^2$.

Let $g_{\max} = \max_{\alpha\in[0,1]} g(\alpha)$.
Among all nonnegative $L$-Lipschitz functions attaining value $g_{\max}$,
the minimal area is achieved by a triangle. Because the half-width of this triangle's base is at most $g_{\max}/L \le T / (2T^2) = 1/(2T) \le 1/2$, the triangle cannot be truncated by both boundaries of the interval $[0,1]$ simultaneously. Thus, the area strictly within $[0,1]$ is at least half the total area of the unbounded triangle, yielding
\begin{align*}
\int_0^1 g(\alpha)\, d\alpha
\ge
\frac{g_{\max}^2}{2L}.
\end{align*}

Therefore,
\begin{align*}
g_{\max}^2
\le
2L\sigma_j
\le
4T^2\sigma_j.
\end{align*}

Taking fourth roots,
\begin{align*}
\max_{\alpha}
|\phi_j^\top \mu_\alpha|
\le
(4T^2\sigma_j)^{1/4}.
\end{align*}

Using Theorem~\ref{thm:BT},
\begin{align*}
\sigma_j
\le
4\pi^2 e^{-\frac{j}{\log T}},
\end{align*}
which yields
\begin{align*}
|\phi_j^\top \mu_\alpha|
=
O\!\left(
T^{1/2} e^{-\frac{j}{4\log T}}
\right).
\end{align*}

\end{proof}

\subsection{Generalization to High Dimension} 

Consider a linear dynamical system with matrices $A,B,C$. The expression $\tilde{\uv}_t$ is now a block matrix of dimensions $T \times d_u$, and thus $\tilde{\uv}_{t}^\top \mu_\alpha$ is a vector of dimension $d_u$. 

We further assume that $A$ is symmetric. This implies that $A$ can be diagonalized over the set of reals as $A = U^{\top}DU$. Decomposing over each dimension $ j \in [d_x]$ separately, and denoting $D_{jj} = \alpha_j$, we can write \eqref{eqn:mu_alpha}:
\begin{align*}
    & \y_{t}  =  \sum_{i=1}^{t-1} C U^\top D^{i-1}  U B \uv_{t-i}  \\
    & =\sum_{i=1}^{t-1} C U^\top  \left( \sum_{j=1}^{d_x} \alpha^{i-1}_j   e_j e_j^\top \right) U B \uv_{t-i}  \\
    & = \sum_{j=1}^{d_x}  c_jb_j^\top  \sum_{i=1}^{t-1}  \alpha^{i-1}_j   \uv_{t-i} & c_j,b_j \leftarrow j^{th}\mbox{ col of }CU^\top, UB\\
    & = \sum_{j=1}^{d_x}  M_j \big( \tilde{\uv}_{t}^\top \mu_{\alpha_j} \big) & \text{where } M_j := c_j b_j^\top \in \reals^{d_y \times d_u}.
\end{align*}
Thus, we can apply the same approximation techniques from the previous section, and derive an analogous conclusion to Theorem \ref{thm:spectral_approx_prediction} for dimension higher than one. Crucially, the number of parameters required does not depend on the hidden dimension of the original system. The details are left as Exercise \ref{exer:high-dim-spectral-filtering}. 

\section{Online Spectral Filtering} 

The prediction class $\Pi_{h}$ can be learned by online gradient descent in the online convex optimization framework, spelled out in Algorithm~\ref{alg:spectral}. 
The regret guarantees of the resulting algorithm, as given in Theorem~\ref{thm:ogd}, directly imply the following. 

\begin{theorem} 
\label{thm:learn-lds-2}
Let $D$ be the diameter of the parameter set $\K$, and $G$ an upper bound on the norm of the gradients for the loss functions $f_t$ as in Algorithm~\ref{alg:spectral}.  
For choice of step sizes $\eta_t = \frac{D}{G \sqrt{t}} $, we have that 
\[
\sum_{t=1}^T \|\y_t - \hat{\y}_t\|^2 
-
\min_{\pi \in \Pi_{h}} 
\sum_{t=1}^T \|\y_t - \y_t^\pi \|^2 
\leq 
2 GD \sqrt{T}. 
\]
\end{theorem}

\begin{algorithm}[h]
\caption{\label{alg:spectral} Online Spectral Filtering with OGD}
\begin{algorithmic}[1]
    \STATE Input: $M^1_{1:h}$, convex constraints set $\K \subseteq \reals^{h \times d_y \times d_u}$, horizon $T$
    \STATE Compute $h$ top eigenvectors of $Z_T$, denoted $\phi_1,\dots,\phi_h$. 
    \FOR {$t=1$ to $T$}
        \STATE Predict 
        $\hat{\y}_t = \sum_{j=1}^h M^t_{j} \big( \tilde{\uv}_{t}^\top \phi_j \big)$
        and observe true $\y_t$. 
        \STATE Define loss $f_t(M) = \|\y_t - \hat{\y}_t  \|^2$. 
        \STATE Update and project:
        \begin{align*}
            M^{t+1} &= M^{t}- \eta_{t} \nabla f_{t}(M^{t}) \\
            M^{t+1} &= \proj_\K(M^{t+1})
        \end{align*}
    \ENDFOR
\end{algorithmic}
\end{algorithm}

However, the gradients of the loss can depend polynomially on $T$, since the feature vectors can have norms that grow with the horizon. 
This renders the regret guarantee of OGD uninformative in our setting. 

To address this issue, we turn instead to the Vovk-Azoury-Warmuth (VAW) forecaster from Appendix~\ref{chap:ons-square}. 
In the spectral filtering setting, the feature vector is known before the prediction is made, so the VAW protocol is a natural fit. 
For simplicity of exposition, we present the scalar setting ($d_y=d_u=1$). 
The resulting regret depends multiplicatively only on the label bound $Y$, while the dependence on the feature magnitudes appears only through a logarithmic determinant term.

\begin{algorithm}[h!]
\caption{Spectral Filtering with VAW}
\label{alg:ons_spectral}
\begin{algorithmic}[1]
    \STATE \textbf{Input:} Filter size $h$, $\lambda>0$, horizon $T$.
    \STATE \textbf{Precompute:} Top $h$ eigenvectors of $Z_T$, stored in $\Phi_h \in \reals^{T \times h}$.
    \STATE \textbf{Initialize:} $A_0 = \lambda I_h$, $v_0 = 0 \in \reals^h$.
    \FOR{$t = 1,2,\dots,T$}
        \STATE $\tilde{\uv}_t = [u_{t-1}, \dots, u_1, 0, \dots, 0]^\top$.
        \STATE $h_t = \Phi_h^\top \tilde{\uv}_t$.
        \STATE Curvature update: $A_t = A_{t-1} + h_t h_t^\top$.
        \STATE Predict $\hat{y}_t = h_t^\top A_t^{-1} v_{t-1}$ and observe $y_t$.
        \STATE Update target vector: $v_t = v_{t-1} + y_t h_t$.
    \ENDFOR
\end{algorithmic}
\end{algorithm}

\begin{corollary}[VAW regret for spectral filtering]
\label{cor:ons_spectral_regret}
Assume $|y_t|\le Y$ for all $t$, and let $\Theta \subseteq \reals^h$ satisfy
$\|\theta\|_2 \le B$ for all $\theta \in \Theta$.
Then
\[
\sum_{t=1}^T (y_t-\hat y_t)^2
-
\min_{\theta\in\Theta}
\sum_{t=1}^T (y_t-\theta^\top h_t)^2
\;\le\;
\lambda B^2
+
Y^2 \log \det\!\left(
I_h + \frac{1}{\lambda}\sum_{t=1}^T h_t h_t^\top
\right).
\]
If in addition $|u_s|\le 1$ for all $s$, and setting $\lambda = 1/B^2$, gives
\[
\sum_{t=1}^T (y_t-\hat y_t)^2
-
\min_{\theta\in\Theta}
\sum_{t=1}^T (y_t-\theta^\top h_t)^2
\;\le\;
1 + h Y^2 \log\!\left(1+B^2T^2\right).
\]
\end{corollary}

Indeed, applying the VAW regret bound from Appendix~\ref{chap:ons-square} with $a_t = h_t$ and $b_t = y_t$ yields
\[
\sum_{t=1}^T (y_t-\hat y_t)^2
-
\sum_{t=1}^T (y_t-\theta^\top h_t)^2
\;\le\;
\lambda \|\theta\|_2^2
+
Y^2 \log \det\!\left(
I_h + \frac{1}{\lambda}\sum_{t=1}^T h_t h_t^\top
\right)
\]
for every $\theta \in \reals^h$.
Taking $\theta$ to be the best predictor in $\Theta$ gives the corollary.

Since $\Phi_h$ has orthonormal columns,
$\|\Phi_h^\top\|_{2}=1$.
If $|u_s|\le 1$, then $\|\tilde{\uv}_t\|_2\le \sqrt{T}$ and hence
$\|h_t\|_2\le \sqrt{T}$.
Therefore $\|h_t h_t^\top\|_2 \le T$, so
\[
\left\|
\sum_{t=1}^T h_t h_t^\top
\right\|_2
\le T^2,
\]
which implies
\[
\log \det\!\left(
I_h + \frac{1}{\lambda}\sum_{t=1}^T h_t h_t^\top
\right)
\le
h \log\!\left(1+\frac{T^2}{\lambda}\right).
\]

Even if the feature magnitudes grow polynomially with $T$, their contribution remains only logarithmic. 
Thus VAW avoids the large multiplicative constants that arise in first-order methods, and the leading multiplicative dependence is only on the label magnitude $Y$.

\section{Conclusion}

In this chapter, we introduced Spectral Filtering as a novel approach to state estimation and prediction in linear dynamical systems, offering an alternative to classical methods like Kalman filtering and online learning of linear predictors. Using spectral decompositions, we bypass the need for explicit system identification and provide a framework that is robust to the stability properties of the system.

We began by outlining the core challenges of traditional state estimation approaches. Kalman filtering, while optimal under Gaussian noise assumptions, requires knowledge of system matrices, which may be difficult to estimate in high-noise or adversarial settings. Online learning of linear predictors relaxed these requirements but introduced a dependence on system stability constants, which can be large and unknown. Spectral Filtering circumvents these limitations by constructing predictors in a transformed basis, allowing for efficient representation without requiring explicit system matrices.

We demonstrated the method first in the one-dimensional case, providing an intuition for how spectral predictors extract relevant dynamical structure. This was then extended to higher-dimensional settings, where spectral filtering enables a structured and computationally efficient approach to forecasting. Crucially, we showed that this approach does not rely on system stability constants, making it particularly appealing for learning in both stable and unstable dynamical systems.

From a methodological standpoint, we connected Spectral Filtering to convex optimization techniques, showing that it can be implemented via online learning in a spectral basis. This formulation allows for efficient sequential learning and prediction with provable performance guarantees.

Looking ahead, Spectral Filtering opens the door for several exciting directions. Extensions to asymmetric system matrices, as referenced in the bibliographic section, provide a pathway toward broader applicability. Moreover, integrating spectral methods with control strategies may lead to novel approaches for nonstochastic control with partial observations, which we explore in the next part of this text.

\ifarxiv
\newpage
\fi
\section{Bibliographic Remarks}

Deviating from classical recovery and into the world of improper prediction,
the work of \citet{hazan2017learning} devised an improper learning technique
using the eigenvectors of the Hankel moment matrix $Z$.
The methods we surveyed in previous chapters for learning linear dynamical systems
had statistical and computational complexity that depended on the hidden dimension.
In contrast, the spectral filtering technique studied in this chapter has regret bounds
and computational complexity guarantees that do not depend on the hidden dimension.

The spectral decay properties of the Hankel (Hilbert) matrices used in this chapter
can be derived from the displacement-structure bounds of \citet{beckermann2017singular},
and are closely connected to the classical approximation theory of Hankel operators;
see, e.g., \citet{Peller2003} for a comprehensive treatment.

Connections to realization and model-reduction theory trace back to the Ho--Kalman
construction \citet{ho1966effective} and to balanced truncation and related approximation
methods \citet{Antoulas2005}.

Spectral filtering can also be interpreted through the lens of operator approximation:
it approximates the input--output map via projection onto leading modes of a compact
Hankel operator. This viewpoint is closely related to Koopman/operator-theoretic
approaches and data-driven spectral methods such as Dynamic Mode Decomposition
\citet{tu2014dynamic}.

The first extension of spectral filtering beyond symmetric settings to general (asymmetric)
linear dynamical systems is due to \citet{hazan2018spectral}. More recent dimension-free
advances for learning asymmetric linear dynamical systems were developed in
\citet{marsden2025dimension}.

The technique of spectral filtering has found applications in control and sequence prediction
since its introduction. In \citet{arora2018towards}, the authors use a spectral filtering
representation to cast control of unknown linear dynamical systems as a convex program.
Closed-loop control using spectral filtering techniques was recently advanced in
\citet{brahmbhatt2025newapproachcontrollinglinear}.

A recent line of work makes use of spectral bases to design neural architectures for sequence
prediction. This was first suggested in the context of spectral state space models
\citet{agarwal2023spectral}, and was extended to hybrid models such as FlashSTU
\citet{liu2024flash}. Later extensions explored efficient implementation \citet{agarwal2024futurefill}
and the concept of length generalization \citet{marsden2024provable}.
Structured state-space models such as S4 \citep{Gu2022S4} also exploit diagonalizations
of linear recurrences, though derived from continuous-time parameterizations rather than
Hankel moment structure.

At a conceptual level, spectral filtering replaces explicit state estimation by operator
approximation: instead of reconstructing latent states, it approximates the input--output
map in a near-optimal spectral basis. This perspective unifies ideas from realization theory,
compact operator approximation, and modern sequence modeling.

\ifarxiv
\newpage
\fi
\begin{exercises}

\begin{exercise}
Generalize Theorem \ref{thm:spectral_approx_prediction} to all real numbers in the range $[-1,1]$, instead of $[0,1]$. 
\end{exercise}

\begin{exercise}\label{exer:high-dim-spectral-filtering}
Generalize Theorem \ref{thm:spectral_approx_prediction} to any dimension $d_x > 1$. 
\end{exercise}

\begin{exercise}
The fact that $Z_T$ is positive semidefinite
is important to ensure that the eigenvalues of $Z_T$ decay.
Give an example of a family of $n \times n$ non-positive semidefinite
Hankel matrices that have eigenvalues with constant absolute value.
\end{exercise}

\begin{exercise}
The trace of a matrix is the sum of its eigenvalues, which measures the ``total energy'' of the system.
\begin{enumerate}
    \item Using the integral definition $Z_T = \int_0^1 \mu_\alpha \mu_\alpha^\top d\alpha$, show that the $(i, j)$-th entry of $Z_T$ (where indices $i, j$ run from $0$ to $T-1$) is given by:
    \[
    (Z_T)_{i,j} = \frac{1}{i+j+1}.
    \]
    \item Compute the trace of $Z_T$ as a summation.
    \item Show that $\text{Tr}(Z_T) \approx \frac{1}{2} \ln T$.
\end{enumerate}
\textit{Note: This contrasts with the operator norm $\|Z_T\|_2$, which is bounded by $\pi$. The fact that the trace grows logarithmically while the norm is bounded implies that the vast majority of eigenvalues must be extremely small, reinforcing the motivation for spectral filtering.}
\end{exercise}

\begin{exercise}
Consider the simplest non-trivial case where the horizon is $T=2$.
\begin{enumerate}
    \item Compute the matrix $Z_2$ explicitly using the formula derived in the previous exercise.
    \item Calculate the two eigenvalues of $Z_2$.
    \item Calculate the ratio $\lambda_1 / \lambda_2$. What does this ratio tell you about how much information is captured by the first spectral feature compared to the second?
\end{enumerate}
\end{exercise}

\begin{exercise}
In spectral filtering, we learn a parameter vector $\theta \in \reals^h$ such that our prediction is $\hat{y}_t = \theta^\top (\Phi_h^\top \tilde{\uv}_t)$, where $\Phi_h \in \reals^{T \times h}$ contains the top $h$ eigenvectors of $Z_T$.
Standard linear predictors (FIR filters) use a weight vector $\mathbf{w} \in \reals^T$ such that $\hat{y}_t = \mathbf{w}^\top \tilde{\uv}_t$.
\begin{enumerate}
    \item Find the expression for the equivalent FIR weights $\mathbf{w}_{\text{spec}}$ implied by the learned spectral parameters $\theta$.
    \item If $h=T$ (we use all eigenvectors), can we represent \textit{any} arbitrary FIR filter $\mathbf{w}$ using spectral parameters $\theta$? Why or why not?
\end{enumerate}
\end{exercise}

\end{exercises}
\chapter{Luenberger Observers} \label{chap:luenberger}

In Chapter 11 we developed spectral filtering for learning linear dynamical systems (LDS) without dependence on the hidden state dimension.
However, those guarantees relied implicitly on favorable spectral structure.
In particular, systems with symmetric or real spectrum are significantly easier to learn than fully asymmetric systems with complex eigenvalues near the unit circle.

In this chapter we address the learning of asymmetric linear dynamical systems.
We begin by showing that without spectral restrictions, learning can be information-theoretically hard.
We then introduce the Luenberger observer as a mechanism for spectrally reshaping the observer error dynamics.
This leads to a regret bound governed by a new quantity: the Luenberger complexity $Q_\star$.

\section{Lower Bounds for Learning Asymmetric LDS}

We first show that without spectral restrictions, learning an LDS may require a number of samples proportional to the hidden dimension.

\begin{theorem}
\label{thm:lds_lower_bound}
For any online learning algorithm $\mathcal{A}$, there exists a linear dynamical system with hidden state dimension $d_h$ such that the expected cumulative squared loss of $\mathcal{A}$ over $T \ge d_h$ rounds is at least $\Omega(d_h)$.
\end{theorem}

\begin{proof}
We construct an autonomous LDS for which prediction is equivalent to predicting independent coin flips.

Consider the system:
\begin{align*}
x_{t+1} &= A x_t, \\
y_t &= C x_t,
\end{align*}
where $A$ is the $d_h \times d_h$ cyclic permutation matrix
\[
A =
\begin{pmatrix}
0 & 1 & 0 & \cdots & 0 \\
0 & 0 & 1 & \cdots & 0 \\
\vdots & \vdots & \vdots & \ddots & \vdots \\
0 & 0 & 0 & \cdots & 1 \\
1 & 0 & 0 & \cdots & 0
\end{pmatrix}
\]
and $C = (1,0,\dots,0)$.

Let the initial state $x_0 \in \{-1,1\}^{d_h}$ be drawn uniformly at random, with independent Rademacher entries.

Then $x_t = A^t x_0$, and for $t = 0,\dots,d_h-1$,
\[
y_t = (x_0)_{t+1}.
\]

Thus each observation reveals a fresh independent coordinate of $x_0$.

Fix any realization of the learner's internal randomness.
Conditional on this realization, the prediction at time $t$ has the form
\[
\hat y_t = f_t(y_0,\dots,y_{t-1}),
\]
which depends only on $(x_0)_1,\dots,(x_0)_t$.

But $y_t = (x_0)_{t+1}$ is independent of these values, hence
\[
\mathbb{E}[y_t \mid y_0,\dots,y_{t-1}] = 0.
\]

Therefore,
\begin{align*}
\mathbb{E}[(y_t - \hat y_t)^2]
&= \mathbb{E}[y_t^2 - 2 y_t \hat y_t + \hat y_t^2] \\
&= 1 + \mathbb{E}[\hat y_t^2] \\
&\ge 1.
\end{align*}

Summing over $t = 0,\dots,d_h-1$ yields expected cumulative loss at least $d_h$.
\end{proof}

This lower bound shows that without restricting spectral structure,
learning asymmetric systems may require a number of samples proportional to the hidden dimension.

We now show that this hardness can be overcome when the system admits an observer representation that reshapes the error dynamics.

\section{Pole Placement}

In this section we prove a fundamental and elegant result from classical control theory: the eigenstructure of the observer error dynamics can be modified completely under the observability assumption.

The key structural condition that allows this radical shift in eigenstructure is \emph{observability}, as defined earlier in the book.
Recall that $(A,C)$ is observable if the observability matrix
\[
\mathcal{O}_{d_h}
=
\begin{pmatrix}
C \\
CA \\
\vdots \\
CA^{d_h-1}
\end{pmatrix}
\]
has full column rank.
Equivalently, the only state $x$ satisfying $C A^k x = 0$ for all $k \ge 0$ is $x=0$.
Observability ensures that the hidden state can be completely reconstructed from the outputs.

The following result shows that this same condition allows for the arbitrary assignment of the spectrum of the observer error dynamics.

\begin{theorem}[Observer Pole Placement]
\label{thm:observer_pole_placement}
Let $(A,C)$ be observable.
Then for any self-conjugate set of $d_h$ complex numbers $\Lambda$,
there exists an observer gain matrix $L \in \mathbb{R}^{d_h \times d_y}$ such that the spectrum (the set of eigenvalues) of $A - LC$ is exactly $\Lambda$.
\end{theorem}

\begin{proof}
For clarity we provide a constructive, self-contained proof for the single-output case ($d_y = 1$), where $C$ is a $1 \times d_h$ row vector and $L$ is a $d_h \times 1$ column vector. 

Let the characteristic polynomial of the unobserved dynamics $A$ be
\[
p(z) = \det(zI - A) = z^{d_h} + a_{d_h-1}z^{d_h-1} + \dots + a_1 z + a_0.
\]
Because the pair $(A, C)$ is observable, there exists an invertible change-of-basis matrix $T$ such that in the new coordinates $x = T \bar{x}$, the system matrices $\bar{A} = T^{-1} A T$ and $\bar{C} = C T$ are in \emph{Observable Canonical Form}:
\[
\bar{A} =
\begin{pmatrix}
0 & 0 & \dots & 0 & -a_0 \\
1 & 0 & \dots & 0 & -a_1 \\
0 & 1 & \dots & 0 & -a_2 \\
\vdots & \vdots & \ddots & \vdots & \vdots \\
0 & 0 & \dots & 1 & -a_{d_h-1}
\end{pmatrix},
\quad
\bar{C} = \begin{pmatrix} 0 & 0 & \dots & 0 & 1 \end{pmatrix}.
\]
Let the desired target spectrum $\Lambda$ be defined by the roots of a monic polynomial:
\[
q(z) = z^{d_h} + \tilde{a}_{d_h-1}z^{d_h-1} + \dots + \tilde{a}_1 z + \tilde{a}_0.
\]
We wish to find an observer gain in the transformed coordinates, $\bar{L} = (l_0, l_1, \dots, l_{d_h-1})^\top$, such that $\bar{A} - \bar{L}\bar{C}$ has characteristic polynomial $q(z)$.

Observe the structure of the outer product $\bar{L}\bar{C}$: it is a $d_h \times d_h$ matrix of all zeros, except for its last column which equals exactly $\bar{L}$. Therefore, subtracting it from $\bar{A}$ simply modifies the last column:
\[
\bar{A} - \bar{L}\bar{C} =
\begin{pmatrix}
0 & 0 & \dots & 0 & -(a_0 + l_0) \\
1 & 0 & \dots & 0 & -(a_1 + l_1) \\
0 & 1 & \dots & 0 & -(a_2 + l_2) \\
\vdots & \vdots & \ddots & \vdots & \vdots \\
0 & 0 & \dots & 1 & -(a_{d_h-1} + l_{d_h-1})
\end{pmatrix}.
\]
Because this resulting matrix is also in canonical form, its characteristic polynomial can be read directly from its last column:
\[
\det(zI - (\bar{A} - \bar{L}\bar{C})) = z^{d_h} + (a_{d_h-1} + l_{d_h-1})z^{d_h-1} + \dots + (a_0 + l_0).
\]
To assign the spectrum $\Lambda$, we simply match the coefficients to $q(z)$ by choosing $l_i = \tilde{a}_i - a_i$ for all $i = 0, \dots, d_h-1$.

Finally, we transform the gain back to the original coordinates by setting $L = T \bar{L}$. The closed-loop matrix is $A - LC = T(\bar{A} - \bar{L}\bar{C})T^{-1}$, which is similar to $\bar{A} - \bar{L}\bar{C}$ and therefore possesses the exact assigned spectrum $\Lambda$.
\end{proof}

\paragraph{Constructing the observable canonical form.}
In the single-output case ($d_y=1$), the observability matrix
\[
\mathcal{O}_{d_h}(A,C)
=
\begin{pmatrix}
C\\
CA\\
\vdots\\
CA^{d_h-1}
\end{pmatrix}
\]
is a $d_h\times d_h$ matrix. Since $(A,C)$ is observable, it has full rank and is therefore invertible.
Likewise, the observability matrix of the observable companion pair $(A_{\mathrm{obs}},C_{\mathrm{obs}})$ is invertible.
Because under the coordinate change $x = T\bar x$ we have
\[
\mathcal O_{d_h}(\bar A,\bar C)=\mathcal O_{d_h}(A,C)\,T,
\]
we may compute the similarity transform from
\[
\mathcal O_{d_h}(A,C)\,T
=
\mathcal O_{d_h}(A_{\mathrm{obs}},C_{\mathrm{obs}}),
\qquad\text{so}\qquad
T
=
\mathcal O_{d_h}(A,C)^{-1}\mathcal O_{d_h}(A_{\mathrm{obs}},C_{\mathrm{obs}}).
\]

\paragraph{Output coordinates and dynamical interpretation.}
Still in the single-output case, there is a second coordinate change,
closely related to observability, that makes the output dynamics
transparent. Define
\[
z_t := \mathcal O_{d_h}(A,C)x_t
=
\begin{pmatrix}
y_t\\
y_{t+1}\\
\vdots\\
y_{t+d_h-1}
\end{pmatrix}.
\]
Because $(A,C)$ is observable, $\mathcal O_{d_h}(A,C)$ is invertible, so
this is a valid state coordinate. In these coordinates the dynamics become
\[
z_{t+1}
=
\begin{pmatrix}
0 & 1 & 0 & \cdots & 0\\
0 & 0 & 1 & \cdots & 0\\
\vdots & & \ddots & \ddots & \vdots\\
0 & 0 & \cdots & 0 & 1\\
-a_0 & -a_1 & \cdots & -a_{d_h-2} & -a_{d_h-1}
\end{pmatrix}
z_t,
\qquad
y_t = e_1^\top z_t.
\]
The last row is exactly the Cayley--Hamilton relation, so unrolling the
recursion yields
\[
y_{t+d_h}+a_{d_h-1}y_{t+d_h-1}+\cdots+a_1 y_{t+1}+a_0 y_t = 0.
\]
Thus an observable $d_h$-dimensional autonomous linear system may be
viewed as a degree-$d_h$ autoregressive model: in suitable coordinates,
the state is encoded by a block of $d_h$ consecutive outputs, and the
dynamics are a shift together with a linear recurrence. From this
perspective, pole placement simply changes the coefficients of that
recurrence, and hence its roots.

We proceed to show how pole placement can be used to create the Luenberger observer system: a powerful technique in learning of dynamical systems.

\section{The Luenberger Observer}

Consider the linear dynamical system
\begin{equation*}
x_{t+1} = A x_t + B u_t,
\qquad
y_t = C x_t.
\end{equation*}

\begin{definition}[Luenberger Observer]
Given a gain matrix $L \in \mathbb{R}^{d_h \times d_y}$, the Luenberger observer is the dynamical system
\begin{equation*}
\tilde x_{t+1}
=
A \tilde x_t + B u_t + L (y_t - \tilde y_t),
\qquad
\tilde y_t = C \tilde x_t.
\end{equation*}
\end{definition}

Define the estimation error
\[
e_t := x_t - \tilde x_t.
\]
Subtracting the observer dynamics from the true dynamics yields
\begin{align*}
e_{t+1}
&= x_{t+1} - \tilde x_{t+1} \\
&= (A x_t + B u_t) - (A \tilde x_t + B u_t + L(y_t - \tilde y_t)) \\
&= (A - LC)e_t.
\end{align*}

Thus the observer error evolves autonomously under the matrix $A - LC$, entirely independent of the control inputs $u_t$.
By Theorem~\ref{thm:observer_pole_placement}, if $(A,C)$ is observable then $L$ can be chosen so that the spectrum of $A-LC$ matches any desired target spectrum. We define the resulting closed-loop matrix as $\tilde{A} := A - LC$.

\subsection{Luenberger Complexity}

Although pole placement guarantees that we may assign the eigenvalues of $\tilde{A}$
arbitrarily, the transient behavior of the observer depends not only on the
spectrum, but also on the conditioning of the eigenbasis.

We henceforth suppose that the chosen closed-loop matrix $\tilde A = A - LC$ is diagonalizable.
For any diagonalization $\tilde A = H D H^{-1}$, the factor $\|H\|\,\|H^{-1}\|$ measures the conditioning of the eigenbasis.

\begin{definition}[Luenberger Complexity]
For a strictly stable, diagonalizable matrix $\tilde A$, let $\gamma = 1 - \rho(\tilde A) > 0$ be its stability margin.
Define
\[
\kappa_{\mathrm{diag}}(\tilde A)
=
\min\{\|H\|\,\|H^{-1}\| : \tilde A = H D H^{-1}, \ D \text{ diagonal}\},
\]
and its Luenberger complexity by
\[
Q(\tilde A)
:= \frac{1}{\gamma}
\max(1, \log {\kappa_{\mathrm{diag}}(\tilde A) }).
\]
\end{definition}

Thus $Q(\tilde A)$ measures the transient amplification and time scale
required before contraction dominates. By definition, $Q(\tilde{A}) \ge 1/\gamma$.

\begin{theorem}[Error Contraction and Burn-In]
\label{thm:luenberger_contraction}
Let $(A,C)$ be observable, choose $L$, and let $\tilde A = A-LC$.
Assume $\tilde A$ is strictly stable and diagonalizable with stability margin $\gamma = 1 - \rho(\tilde A) > 0$.
Then
\[
\|e_t\|
\le
\kappa_{\mathrm{diag}}(\tilde A)\,(1-\gamma)^t\,\|e_0\|.
\]
Consequently, for any $\varepsilon>0$, to guarantee $\|e_t\|\le\varepsilon$ it suffices that
\[
t \ge 
Q(\tilde{A})   \log \frac{1+ \|e_0\|}{\varepsilon} .
\]
\end{theorem}

\begin{proof}
Let $\tilde A = H D H^{-1}$ be a diagonalization attaining the minimum in the definition of $\kappa_{\mathrm{diag}}(\tilde A)$.
Since $e_{t+1}=\tilde A e_t$, we have $e_t=\tilde A^t e_0$.
Using the diagonalization $\tilde A = H D H^{-1}$,
\[
\tilde A^t = H D^t H^{-1}.
\]
Thus
\[
\|\tilde A^t\|
\le
\|H\|\,\|H^{-1}\|\,\|D^t\|
\le
\kappa_{\mathrm{diag}}(\tilde A)\,(1-\gamma)^t.
\]
Multiplying by $\|e_0\|$ yields the first inequality.
The burn-in bound follows by solving
\[
\kappa_{\mathrm{diag}}(\tilde A)(1-\gamma)^t\|e_0\|\le\varepsilon,
\]
and using the standard inequality $\log(1-\gamma) \le -\gamma$.
\end{proof}

Pole placement ensures existence of a stable observer.
However, the effective convergence speed depends on both the spectral gap $\gamma$
and the conditioning of the eigenbasis (which is constrained because we can only assign the eigenvalues, not the exact matrix $\tilde A$).
The Luenberger complexity $Q(\tilde A)$ captures precisely the scale
that governs this burn-in time.

\subsection{One Dimensional Example}

Consider the simple one dimensional (and thus fully observable) linear dynamical system
\[
x_{t+1} = 2 x_t,
\qquad
y_t = x_t.
\]
Define
\[
\tilde x_{t+1}
=
2 \tilde x_t + L (x_t - \tilde x_t).
\]
Then
\[
e_{t+1} = (2 - L)e_t.
\]
Choosing $L=2.5$ yields
\[
e_{t+1} = -0.5 e_t,
\]
so the error contracts exponentially even though the system diverges.

\section{Reduction to Stable Learning}

We now exploit the reshaping principle as a general reduction for online learning.
Section~12.3 showed that if $(A,C)$ is observable, we may choose an observer gain $L$
so that the error dynamics evolve under a strictly stable closed-loop matrix $\tilde A = A - LC$. The effective transient of this observer is governed by the Luenberger complexity $Q(\tilde A)$.

Crucially, this transforms the problem of predicting an unstable, partially-observed system with infinite memory into the problem of predicting the outputs of a strictly stable, well-behaved system.

\subsection{A Learnable Spectral Region}

To guarantee that the observer forgets its initial estimation error $e_0$, we must ensure that the closed-loop matrix $\tilde{A}$ is strictly stable. Fix a desired stability margin $\gamma \in (0,1)$ and define the target spectral region simply as the strictly stable complex disk:
\[
\Sigma_{\gamma} := \mathbb{D}_{1-\gamma} = \{z\in\mathbb{C} : |z| \le 1-\gamma\}.
\]

We enforce the margin $\gamma$ entirely to guarantee rapid error dissipation (burn-in). Marginally stable modes (e.g., $|\lambda| = 1$) are strictly excluded. Because the Luenberger observer is a closed-loop system, an eigenvalue of $1$ would cause the unforced initial estimation error $e_t = \tilde{A}^t e_0$ to persist indefinitely as a constant offset. Because online learning algorithms build predictive features solely from forced inputs and outputs, they cannot accurately predict this unforced hidden offset, resulting in linear regret.

All eigenvalues of the original system $A$ --- in particular unstable modes or complex eigenvalues near the unit circle --- constitute the difficult modes; we aim to move them strictly into the stable region $\Sigma_{\gamma}$ via the observer.

\subsection{The Luenberger Program}

Pole placement guarantees feasibility under observability, but not necessarily good conditioning.
Even if there exists an $L$ such that the spectrum satisfies $\sigma(A-LC)\subseteq \Sigma_{\gamma}$, the resulting observer may have a poorly conditioned eigenbasis, leading to a large Luenberger complexity.

We therefore define an optimization program to find the best-conditioned observer among all feasible pole placements for an arbitrary stable target set $\Sigma$.

\begin{definition}[Luenberger Program]
Fix an arbitrary strictly stable spectral constraint set $\Sigma \subseteq \mathbb{D}_{1-\epsilon}$ for some $\epsilon > 0$. Define the optimal Luenberger complexity as:
\[
Q_\star(A,C;\Sigma)
:=
\min_{L}
\;
Q(A-LC)
\quad
\text{s.t.}
\quad
\sigma(A-LC)\subseteq \Sigma,
\]
with the convention $Q_\star(A,C;\Sigma)=+\infty$ if infeasible.
For our specific stable disk $\Sigma=\Sigma_{\gamma}$, we write $Q_\star^{\gamma}$.  We call $Q_\star$ the {\bf Optimal Luenberger Complexity}.
\end{definition}

\subsection{Regret Reduction}

Assume $Q_\star(A,C;\Sigma_{\gamma})<\infty$
and let $L^\star$ attain the minimum in the Luenberger program.
Define the optimal observer matrix
\[
\tilde A := A - L^\star C,
\qquad
Q_\star := Q(\tilde A).
\]

By recursively expanding the observer state, we can express the true system output $y_t$ entirely as a function of the past observations $y_{t-i}$ and controls $u_{t-i}$, plus an unforced error term:
\[
y_t
=
\underbrace{
C\tilde{A}^t \tilde x_0
+
\sum_{i=1}^t C \tilde{A}^{i-1} (B u_{t-i} + L^\star y_{t-i})
}_{=:\bar{y}_t}
+
C \tilde{A}^t e_0.
\]
The sequence $\bar{y}_t$ represents a strictly stable mapping from past data to the current output. Because $\tilde{A}$ is strictly stable, the sequence of coefficient matrices $C\tilde{A}^{i-1}B$ and $C\tilde{A}^{i-1}L^\star$ decay to zero exponentially fast.

This provides a powerful, general reduction for online learning in dynamical systems: {we have reduced the problem to learning a stable transfer function}.

Because the coefficients decay exponentially fast, the mapping can be effectively approximated using a finite memory length $m = O(Q_\star \log T)$. We may now employ \emph{any} suitable online sequence prediction algorithm to learn this mapping. For example:
\begin{itemize}
    \item \textbf{Linear Regression (FIR Filters):} We can parameterize the predictor as a standard finite impulse response filter of length $m$ and learn the coefficients using Online Gradient Descent.
    \item \textbf{Spectral Filtering \& ONS:} We can project the observations onto the highly efficient Hankel basis (which requires only $O(\log^2 T)$ features for real modes) and use the Online Newton Step to achieve logarithmic regret.
\end{itemize}

Regardless of the chosen learning algorithm, the total regret against the optimal observer is bounded by the algorithm's inherent regret for stable sequence prediction, plus an initial, bounded penalty from the Luenberger burn-in period.

The true observation $y_t$ differs from the idealized stable prediction $\bar{y}_t$ only by the unforced observer error $C\tilde{A}^t e_0$. By Theorem~\ref{thm:luenberger_contraction}, the total squared error introduced by the observer's transient burn-in over all time is bounded by a constant independent of $T$:
\begin{align*}
E_{\text{burn-in}} := \sum_{t=1}^T \|y_t - \bar{y}_t\|^2 
&\le \|C\|^2 \kappa_{\mathrm{diag}}(\tilde A)^2 \|e_0\|^2 \sum_{t=1}^\infty (1-\gamma)^{2t} \\
&\le \frac{\|C\|^2 \kappa_{\mathrm{diag}}(\tilde A)^2 \|e_0\|^2}{2\gamma} . 
\end{align*}
Notice that this burn-in constant is governed exactly by $\kappa_{\mathrm{diag}}(\tilde A)$ and $1/\gamma$, which are the quantities minimized by the Luenberger complexity $Q_\star$.

Let $\mathcal{F}_{\gamma}$ denote any class of stable linear predictors with margin $\gamma$ that contains the stable surrogate sequence $\bar{y}_t$ defined above.

\begin{theorem}[Regret Reduction]
Let $\mathcal{A}$ be an online learning algorithm that guarantees $\mathrm{Regret}_{\mathcal{A}}(T)$ against the class $\mathcal{F}_{\gamma}$. If we apply $\mathcal{A}$ to the sequence of observations from the partially observed system $(A,C)$, then
\[
\sum_{t=1}^T \|y_t - \hat y_t\|^2
\le
\inf_{f \in \mathcal{F}_{\gamma}}
\sum_{t=1}^T \|y_t - f_t\|^2
+
\mathrm{Regret}_{\mathcal{A}}(T)
\le
\mathrm{Regret}_{\mathcal{A}}(T) + E_{\mathrm{burn-in}}(L^\star).
\]
\end{theorem}
This reduction completely separates the \emph{control} problem (using observability to reshape the error dynamics and minimize $Q_\star$) from the \emph{learning} problem (running an algorithm on a stable system). The regret against the true underlying (potentially unstable and asymmetric) system is therefore governed by the regret of learning a stable system, plus a constant burn-in penalty.

\section{Summary}

This chapter shows that partial observation fundamentally changes the learning problem for linear dynamical systems.
Without additional structure, one cannot hope for dimension-free guarantees in general:
the lower bound demonstrates that a learner may be forced to uncover the hidden state one coordinate at a time, incurring loss of order $d_h$ before the system becomes predictable.
Thus the hidden dimension is a genuine barrier if we work directly with the original dynamics.

The Luenberger observer provides a way around this obstacle.
When $(A,C)$ is observable, we may introduce an output-injection matrix $L$ and study the estimation error dynamics
\[
e_{t+1} = (A-LC)e_t.
\]
Pole placement shows that the spectrum of this closed-loop matrix can be moved into any prescribed stable region.
This does not by itself solve the learning problem, because fast asymptotic decay is not enough: one must also control transient growth.
That is the role of the Luenberger complexity $Q(\tilde A)$, and of the optimized quantity $Q_\star$, which measure how quickly the observer forgets its initial error and how badly the non-normal geometry can amplify that error before contraction dominates.

Once a well-conditioned observer is chosen, the original partially observed system is converted into a stable input-output prediction problem, up to a burn-in term coming from the decaying initial estimation error.
The chapter therefore separates into two clean pieces.
Control theory supplies the observer that reshapes the error dynamics into a stable form, and online learning supplies an algorithm for the resulting stable sequence-prediction task.
In this way, the effective difficulty of learning is governed not directly by the hidden dimension $d_h$, but by the best attainable observer geometry encoded by $Q_\star$.

\ifarxiv
\newpage
\fi
\section{Bibliographic Remarks}

The fundamental hardness of learning linear dynamical systems without spectral assumptions, as discussed in the beginning of this chapter, contrasts with the symmetric or stable cases. While \citet{hardt2018gradient} showed that gradient descent efficiently learns stable systems, the lower bounds for general asymmetric systems typically rely on constructions where the system dynamics obscure the initial state for long periods, similar to the hardness results in \citet{simchowitz2018learning}. The information theoretic lower bound described here is due to \cite{hazan2026spectralfilteringlearningquantum}.

The theoretical foundation for reconstructing internal states from outputs lies in the work of \citet{luenberger1964observing}, who introduced the concept of the state observer. The specific architecture discussed in this chapter is widely known as the Luenberger Observer. \citet{luenberger1971introduction} provides a comprehensive introduction to this theory. The powerful result that observability allows for arbitrary assignment of the closed-loop spectrum (Theorem 12.2) is known as the Pole Placement Theorem, originally established by \citet{wonham1967pole}.

While classical control theory focuses on asymptotic stability, this chapter emphasizes the transient behavior of the error dynamics, which is crucial for finite-time learning guarantees. The study of transient growth in stable but non-normal matrices (where eigenvectors are ill-conditioned) is famously treated in the text on pseudospectra by \citet{trefethen2005spectra}. The concept of \emph{Luenberger Complexity} ($Q_\star$) serves as a quantitative measure of this transient behavior specifically optimized for the learning setting. It was introduced in \cite{dogariu2025universal}.

The reduction framework presented in this chapter, which utilizes the Luenberger observer to transform an unstable or asymmetric learning problem into a stable prediction task, is based on the work of \citet{dogariu2025universal}. This work showed that dimension-free regret bounds are achievable for general linear systems provided one can find a well-conditioned observer, shifting the complexity burden from the state dimension $d_h$ to the spectral conditioning of the dynamics.

\ifarxiv
\newpage
\fi
\begin{exercises}

\begin{exercise}
\label{ex:peaking}
This exercise illustrates the "peaking phenomenon," where placing eigenvalues far into the stable region can cause the error norm to grow transiently before decaying (i.e., high Luenberger complexity).
Consider the two-dimensional system with $A = \begin{bmatrix} 1 & 1 \\ 0 & 1 \end{bmatrix}$ and $C = \begin{bmatrix} 1 & 0 \end{bmatrix}$.
\begin{enumerate}
    \item Verify that the system is observable.
    \item Find the observer gain vector $L = [l_1, l_2]^\top$ such that the closed-loop matrix $\tilde{A} = A - LC$ has both eigenvalues equal to some $\lambda \in [0, 1)$.
    \item Show that as we demand faster convergence (i.e., as $\lambda \to 0$), the norm of the gain $\|L\|$ tends to infinity.
    \item Calculate the matrix $\tilde{A}$ explicitly for $\lambda=0$. Compute $\|\tilde{A} e_0\|$ for an initial error $e_0 = [0, 1]^\top$. What happens to the error magnitude in the first step?
\end{enumerate}
\end{exercise}

\begin{exercise}
\label{ex:deadbeat}
A \emph{deadbeat observer} is one where the closed-loop poles are all placed at the origin ($0$).
\begin{enumerate}
    \item Let $\tilde{A} = A - LC$. Prove that if all eigenvalues of $\tilde{A}$ are $0$, then $\tilde{A}$ is nilpotent, meaning $\tilde{A}^{d_h} = 0$.
    \item Conclude that for a deadbeat observer, the estimation error $e_t$ becomes exactly zero for all $t \ge d_h$, regardless of the initial error.
    \item Why might we not always choose a deadbeat observer in practice, particularly in the presence of observation noise $\mathbf{v}_t$? (Hint: Consider the magnitude of $L$ derived in Exercise \ref{ex:peaking}).
\end{enumerate}
\end{exercise}

\begin{exercise}
\label{ex:cyclic_observability}
In Theorem 12.1, we used a cyclic permutation matrix to prove a lower bound on learning. Let $A$ be the $d_h \times d_h$ cyclic permutation matrix where $A_{i, i+1} = 1$ for $1 \le i < d_h$, $A_{d_h, 1} = 1$, and $0$ elsewhere. Let $C = [1, 0, \dots, 0]$.
\begin{enumerate}
    \item Write down the observability matrix $\mathcal{O}_{d_h}$ for this pair $(A, C)$.
    \item Prove that this system is fully observable.
    \item Since it is observable, Theorem 12.2 implies we can stabilize it. Find a gain vector $L$ such that $A - LC$ is nilpotent.
\end{enumerate}
\end{exercise}

\begin{exercise}
Prove that the set of stable matrices is not convex. specifically, give two matrices $A_1, A_2 \in \mathbb{R}^{2 \times 2}$ such that $\rho(A_1) < 1$ and $\rho(A_2) < 1$, but $\rho(\frac{1}{2}A_1 + \frac{1}{2}A_2) > 1$.
\textit{Note: This geometric fact implies that the Luenberger Program (Definition 12.6), which optimizes over stable matrices, involves a non-convex constraint set.}
\end{exercise}

\end{exercises}
\chapter{Learning in Nonlinear Dynamical Systems} \label{chap:koopman}

In the previous chapters, we developed a comprehensive theory for learning
and predicting in linear dynamical systems. We moved from symmetric systems
in Chapter \ref{chap:sf} to general, asymmetric, and potentially unstable
linear systems in Chapter \ref{chap:luenberger} via the Luenberger observer
framework.

However, the physical world is rarely strictly linear. In this chapter, we
address the fundamental problem of learning \emph{nonlinear} dynamical
systems. Formally, we consider a system of the form
\begin{align}
    \x_{t+1} &= f(\x_t) \label{eq:nonlinear_dyn} \\
    \y_t &= h(\x_t), \notag
\end{align}
where $f: \mathcal{X} \to \mathcal{X}$ is the nonlinear transition function
and $h: \mathcal{X} \to \reals^{d_y}$ is the observation function. The state
$\x_t$ is hidden, and the functions $f$ and $h$ are unknown. For clarity, we
treat the autonomous case in this chapter; the controlled extension
$\x_{t+1} = f(\x_t,\uv_t)$ is significantly more complex.

Historically, learning such systems involved \emph{system identification}:
attempting to approximate the function $f$ using neural networks or kernels.
In this chapter, we take a different approach rooted in \emph{improper
learning}. Rather than trying to recover the nonlinearity explicitly, we
compete with a linear surrogate that predicts well.

More precisely, we will show two things. First, every bounded nonlinear
system with non-expansive dynamics admits a finite-dimensional approximate
linear representation over any fixed horizon via discretization. Second, if
such a lifted linear approximation admits a stable, well-conditioned
observer, then the learning guarantees from Chapter \ref{chap:luenberger}
allow us to compete with it. Thus, nonlinear prediction can be reduced to
linear prediction, provided the corresponding lifted system has moderate
observer complexity.

\paragraph{A passive prediction viewpoint.}
A useful way to think about the learning problem in this chapter is as
\emph{passive sequential prediction}. Here the learner does not act on the
system and does not choose any control input. Instead, it only observes a
sequence generated by hidden nonlinear dynamics and attempts to predict the
next observation.

\paragraph{Example 1: Lorenz dynamics.}
A canonical example is the Lorenz system, given in continuous time by
\begin{align*}
    \dot x^{(1)}(\tau) &= \sigma\bigl(x^{(2)}(\tau)-x^{(1)}(\tau)\bigr),\\
    \dot x^{(2)}(\tau) &= x^{(1)}(\tau)\bigl(\rho-x^{(3)}(\tau)\bigr)-x^{(2)}(\tau),\\
    \dot x^{(3)}(\tau) &= x^{(1)}(\tau)x^{(2)}(\tau)-\beta x^{(3)}(\tau),
\end{align*}
where $\sigma,\rho,\beta > 0$ are fixed parameters. Writing
\[
\x(\tau) =
\begin{bmatrix}
x^{(1)}(\tau)\\
x^{(2)}(\tau)\\
x^{(3)}(\tau)
\end{bmatrix}
\in \reals^3,
\]
this system may be viewed abstractly as
\[
\dot \x(\tau) = F(\x(\tau)).
\]
Fix a sampling interval $\Delta > 0$, and define the discrete-time state
\[
\x_t = \x(t\Delta).
\]
Then the sampled trajectory evolves according to an autonomous nonlinear
dynamical system
\[
\x_{t+1} = f(\x_t),
\]
where $f$ is the time-$\Delta$ flow map of the differential equation. If we
observe only a scalar signal
\[
\y_t = h(\x_t),
\]
for example one coordinate of the state, then the learning problem is to
predict the next observation $\y_t$ from the past observations
$\y_1,\ldots,\y_{t-1}$. Thus a nonlinear physical system gives rise
naturally to the passive prediction problem studied in this chapter.

\paragraph{Example 2: large language models.}
A large language model can be viewed in the same way. Let $\y_t \in V$ be
the token emitted at time $t$, where $V$ is a finite vocabulary. The model
maintains a hidden state $\x_t$ that summarizes its internal activations
together with the current context, cache, and recent tokens. The evolution
of this hidden state is governed by the nonlinear dynamics of the underlying
deep neural network, and the emitted token is a function of the current
state. Abstractly, this may be written as
\[
\x_{t+1} = f(\x_t),
\qquad
\y_t = h(\x_t),
\]
where $f$ is induced by one forward pass of the model together with the
decoding rule. In the deterministic case, for example under greedy decoding,
this fits exactly into the autonomous framework studied in this chapter.

If the decoding rule is randomized, for example when the next token is
sampled from a truncated distribution such as top-$k$ or nucleus sampling,
then the dynamics become stochastic rather than deterministic. In that case
one obtains a stochastic analogue of the same framework; see
Exercise~\ref{ex:llm-stochastic-dynamics}.

In the sequel we return to the Euclidean observation model
$\y_t \in \reals^{d_y}$. The language-model example is included as
motivation; one may embed discrete tokens in this framework, for example,
via one-hot or probability-vector representations.

\section{Global Linear Approximation via Discretization}
\label{sec:nonlinear-discretization}

The core mathematical insight enabling linear learning of nonlinear dynamics
is that nonlinear dynamics in a finite-dimensional space can be lifted to
linear dynamics in a higher, potentially infinite-dimensional space. This is
often associated with the \emph{Koopman operator}, which we discuss in
Section~\ref{sec:koopman-perspective}.

We begin with a concrete finite-horizon construction based on
discretization.

\subsection{The $\epsilon$-Net Construction}

We make the following standard assumptions about the regularity of the
system:
\begin{enumerate}
    \item \textbf{Boundedness:} The state space $\mathcal{X}$ is contained in
    a ball of radius $R$.

    \item \textbf{Lipschitz continuity:} The dynamics $f$ and observation map
    $h$ are $1$-Lipschitz. That is,
    \[
    \|f(x) - f(z)\| \le \|x-z\|,
    \qquad
    \|h(x) - h(z)\| \le \|x-z\|
    \]
    for all $x,z \in \mathcal{X}$.
\end{enumerate}

Let $\mathcal{S}$ be an $\epsilon$-net of the state space $\mathcal{X}$.
This is a finite set of points
\[
\mathcal{S} = \{s^{(1)}, \dots, s^{(N)}\}
\]
such that for every $x \in \mathcal{X}$, there exists some
$s \in \mathcal{S}$ with $\|x-s\| \le \epsilon$. Its cardinality satisfies
\[
N = O\!\left((R/\epsilon)^{d_x}\right).
\]

We define a lifted state space of dimension $N$, where the lifted state
$\z_t \in \reals^N$ is a one-hot vector encoding the current approximate
grid location. Let $\pi(x)$ denote a nearest grid point to $x$ in
$\mathcal{S}$, with ties broken arbitrarily:
\[
\pi(x) \in \argmin_{s \in \mathcal{S}} \|x-s\|.
\]

We now define a linear transition matrix $A' \in \reals^{N \times N}$ and
observation matrix $C' \in \reals^{d_y \times N}$ as follows:
\begin{enumerate}
    \item \textbf{Dynamics ($A'$):} If the system is at grid point $s^{(j)}$,
    the next state is approximated by the grid point nearest to $f(s^{(j)})$.
    Thus
    \[
    A'_{ij} = 1
    \quad\text{if}\quad
    s^{(i)} = \pi(f(s^{(j)})),
    \qquad
    A'_{ij} = 0
    \quad\text{otherwise.}
    \]

    \item \textbf{Observations ($C'$):} The $j$-th column of $C'$ is the
    observation vector at the grid point $s^{(j)}$:
    \[
    C'_{:,j} = h(s^{(j)}).
    \]
\end{enumerate}

This defines a linear system
\[
\z_{t+1} = A' \z_t,
\qquad
\y'_t = C' \z_t.
\]
If $s^{(j)} = \pi(\x_0)$ and we initialize $\z_0 = \mathbf{e}_j$, then
$\z_t$ remains a one-hot vector for all $t$.

\subsection{Approximation Guarantee}

We now prove that this high-dimensional linear system remains close to the
true nonlinear trajectory over a finite horizon.

\begin{theorem}[Finite-Horizon Linear Approximation via Discretization]
\label{thm:nonlinear_approx}
Let $\y_0, \dots, \y_{T-1}$ be generated by the nonlinear system
\eqref{eq:nonlinear_dyn} under the boundedness and $1$-Lipschitz assumptions
above. Let $A', C'$ be the matrices constructed from an $\epsilon$-net
$\mathcal{S}$, and let $\y'_t$ be the output of the lifted linear system
initialized at the one-hot vector corresponding to $\pi(\x_0)$.

Then for every $t = 0,1,\dots,T-1$,
\[
\|\y_t - \y'_t\| \le (t+1)\epsilon.
\]
Consequently,
\[
\sum_{t=0}^{T-1} \|\y_t - \y'_t\|^2
\le
\epsilon^2 \sum_{t=0}^{T-1} (t+1)^2
=
\epsilon^2 \frac{T(T+1)(2T+1)}{6}
\le
T^3 \epsilon^2.
\]
\end{theorem}

\begin{proof}
Let $\x_0,\dots,\x_T$ be the true trajectory of the nonlinear system. Let
$s_t$ be the grid point occupied by the lifted linear system at time $t$.
Then
\[
s_0 = \pi(\x_0),
\qquad
s_{t+1} = \pi(f(s_t)).
\]

Define the tracking error
\[
\xi_t := \|s_t - \x_t\|.
\]
We analyze its evolution:
\begin{align*}
\xi_{t+1}
&= \|s_{t+1} - \x_{t+1}\| \\
&= \|\pi(f(s_t)) - f(\x_t)\|.
\end{align*}
By the triangle inequality,
\[
\xi_{t+1}
\le
\|\pi(f(s_t)) - f(s_t)\|
+
\|f(s_t) - f(\x_t)\|.
\]
The first term is at most $\epsilon$ by the definition of the $\epsilon$-net
and nearest-neighbor projection. The second term is at most
$\|s_t-\x_t\| = \xi_t$ by the $1$-Lipschitz property of $f$. Hence
\[
\xi_{t+1} \le \xi_t + \epsilon.
\]
Since
\[
\xi_0 = \|\pi(\x_0)-\x_0\| \le \epsilon,
\]
an induction gives
\[
\xi_t \le (t+1)\epsilon
\qquad
\text{for all } t \ge 0.
\]

Now $\y_t = h(\x_t)$ and $\y'_t = h(s_t)$, so by the $1$-Lipschitz property
of $h$,
\[
\|\y_t - \y'_t\|
=
\|h(\x_t) - h(s_t)\|
\le
\|\x_t - s_t\|
=
\xi_t
\le
(t+1)\epsilon.
\]
This proves the pointwise bound. Summing the squared bound over
$t=0,\dots,T-1$ yields
\[
\sum_{t=0}^{T-1} \|\y_t - \y'_t\|^2
\le
\epsilon^2 \sum_{t=0}^{T-1}(t+1)^2
=
\epsilon^2 \frac{T(T+1)(2T+1)}{6}
\le
T^3 \epsilon^2.
\]
\end{proof}

Thus, by choosing $\epsilon \asymp T^{-3/2}$, the cumulative squared
approximation error is $O(1)$ over the horizon $T$. The price is that the
lifted dimension
\[
N = O\!\left((R/\epsilon)^{d_x}\right)
\]
may be exponentially large in the original state dimension $d_x$.

\section{The Koopman Operator}
\label{sec:koopman-perspective}

The discretization argument above gives a finite-dimensional linear
approximation over a fixed horizon. The Koopman perspective explains why
linear structure is present more fundamentally.

The \textbf{Koopman operator} $\mathcal{K}$ is an infinite-dimensional
linear operator acting on scalar observable functions
$g : \mathcal{X} \to \reals$ by
\[
(\mathcal{K} g)(x) = g(f(x)).
\]
Even if $f$ is nonlinear, the operator $\mathcal{K}$ is linear.

Thus every nonlinear dynamical system admits an exact linear representation
at the level of observables, but in general this representation lives in an
infinite-dimensional function space. The discretization in
Section~\ref{sec:nonlinear-discretization} should be viewed as a concrete
finite-state approximation to this principle: by tracking which cell of an
$\epsilon$-net the system occupies, we obtain a finite-dimensional lifted
linear system.

For the purposes of prediction, our perspective is pragmatic. We do not need
to recover the full infinite-dimensional Koopman operator. We only need to
know that a finite-dimensional linear approximation exists and that this
approximation can be learned competitively by the online algorithms
developed earlier.

\section{Improper Learning of Nonlinear Dynamics}

We can now combine the existence result of
Theorem~\ref{thm:nonlinear_approx} with the learning algorithms from
Chapter~\ref{chap:luenberger}.

The key point is existential: the lifted linear system constructed above is
used only as a surrogate comparator in the analysis. The learner never
constructs the grid $\mathcal{S}$, the matrices $(A',C')$, or an observer
gain $L$ explicitly.

The lifted linear system can have enormous hidden dimension
\[
N = O\!\left((R/\epsilon)^{d_x}\right),
\]
yet the regret bounds from Chapter~\ref{chap:luenberger} do not depend
directly on that hidden dimension. This makes improper learning possible:
we never manipulate the lifted state $\z_t$ explicitly, but instead compete
with the predictions of the lifted linear surrogate on the observed sequence
$\y_t$.

Dimension-independence alone, however, is not enough. To apply the
Luenberger reduction from Chapter~\ref{chap:luenberger}, the lifted system
must admit a stable, well-conditioned observer. This leads to the following
complexity measure.

\begin{definition}[Nonlinear Luenberger Complexity]
\label{def:nonlinear_luenberger_program}
Fix $\epsilon > 0$ and a stability margin $\gamma \in (0,1)$. Define
\[
Q^\star_\gamma(f,h;\epsilon)
:=
\inf_{\substack{\mathcal{S}\ \text{an }\epsilon\text{-net of }\mathcal X \\
                \pi\ \text{a nearest-neighbor map}}}
\;
\inf_{\substack{L \in \reals^{N \times d_y} \\
                A' - LC' \text{ diagonalizable} \\
                \sigma(A' - LC') \subseteq D_{1-\gamma}}}
Q(A' - LC'),
\]
where $(A',C')$ are the lifted matrices induced by $(\mathcal{S},\pi)$ as in
Section~\ref{sec:nonlinear-discretization}. If no feasible choice exists, we
set
\[
Q^\star_\gamma(f,h;\epsilon) = +\infty.
\]
\end{definition}

This quantity asks whether some $\epsilon$-resolution linearization of the
nonlinear system can be stabilized by a Luenberger observer with moderate
transient behavior. Feasibility is not automatic: unlike the setting of
Chapter~\ref{chap:luenberger}, the discretized pair $(A',C')$ need not be
observable, and the nonlinear Luenberger program may therefore have value
$+\infty$.

Because the setting is autonomous, the relevant comparator class now
consists of stable linear predictors based only on past outputs.

We now combine the discretization error with the linear regret bound.

\begin{theorem}[Improper Learning Guarantee]
\label{thm:nonlinear_improper_learning}
Fix $\epsilon > 0$ and $\gamma \in (0,1)$. Let $(A',C')$ be any lifted
linear system obtained from an $\epsilon$-net construction in
Section~\ref{sec:nonlinear-discretization}, and suppose there exists a gain
matrix $L \in \reals^{N \times d_y}$ such that $A' - LC'$ is diagonalizable
and
\[
\sigma(A' - LC') \subseteq D_{1-\gamma}.
\]
Let $\bar{\y}_t$ denote the stable surrogate sequence obtained by applying
the Luenberger reduction of Chapter~\ref{chap:luenberger} to the lifted
linear system $(A',C')$, and let
$E_{\mathrm{burn\mbox{-}in}}(A',C',L)$ denote the corresponding burn-in
term.

If $\mathcal{A}$ is any online learning algorithm with regret
$\mathrm{Regret}_{\mathcal{A}}(T)$ against the class of stable linear
predictors with margin $\gamma$, then its predictions $\hat{\y}_t$ on the
nonlinear sequence satisfy
\[
\sum_{t=0}^{T-1} \|\y_t - \hat{\y}_t\|^2
\le
\mathrm{Regret}_{\mathcal{A}}(T)
+
2T^3 \epsilon^2
+
2E_{\mathrm{burn\mbox{-}in}}(A',C',L).
\]
\end{theorem}

\begin{proof}
Let $\y'_t$ denote the output of the lifted linear system $(A',C')$
constructed from the $\epsilon$-net. By Theorem~\ref{thm:nonlinear_approx},
\[
\sum_{t=0}^{T-1} \|\y_t - \y'_t\|^2 \le T^3 \epsilon^2.
\]

Next, applying the reduction from Chapter~\ref{chap:luenberger} to the
lifted linear system $(A',C')$ yields a stable surrogate sequence
$\bar{\y}_t$ such that
\[
\sum_{t=0}^{T-1} \|\y'_t - \bar{\y}_t\|^2
\le
E_{\mathrm{burn\mbox{-}in}}(A',C',L).
\]

Since $\bar{\y}_t$ belongs to the comparator class of stable linear
predictors with margin $\gamma$, the regret guarantee of $\mathcal{A}$ gives
\[
\sum_{t=0}^{T-1} \|\y_t - \hat{\y}_t\|^2
\le
\sum_{t=0}^{T-1} \|\y_t - \bar{\y}_t\|^2
+
\mathrm{Regret}_{\mathcal{A}}(T).
\]

Finally, using
\[
\|a+b\|^2 \le 2\|a\|^2 + 2\|b\|^2,
\]
with
\[
a = \y_t - \y'_t,
\qquad
b = \y'_t - \bar{\y}_t,
\]
we obtain
\begin{align*}
\sum_{t=0}^{T-1} \|\y_t - \bar{\y}_t\|^2
&\le
2\sum_{t=0}^{T-1} \|\y_t - \y'_t\|^2
+
2\sum_{t=0}^{T-1} \|\y'_t - \bar{\y}_t\|^2 \\
&\le
2T^3 \epsilon^2
+
2E_{\mathrm{burn\mbox{-}in}}(A',C',L).
\end{align*}
Substituting this into the regret bound proves the claim.
\end{proof}

In particular, choosing $\epsilon \asymp T^{-3/2}$ makes the discretization
term $O(1)$. Therefore, instantiating $\mathcal{A}$ with any stable-predictor
learner from Chapter~\ref{chap:luenberger} yields sublinear regret whenever
$Q^\star_\gamma(f,h;\epsilon)$ and the burn-in term are moderate. The lifted
dimension $N$ does not appear explicitly in the guarantee; instead, the
difficulty is pushed into the observer complexity and burn-in term.

\ifarxiv
\newpage
\fi
\section{Bibliographic Remarks}

The operator-theoretic viewpoint on nonlinear dynamics goes back to the work of
\citet{koopman1931hamiltonian}, who observed that nonlinear state evolution induces a
linear evolution on observables. Modern Koopman theory, especially as it relates to model
reduction, spectral analysis, and control, was further developed in
\citet{mezic2005spectral,mezic2013analysis}. Useful modern overviews include the review of
\citet{otto2021koopman}, the survey of \citet{brunton2022modern}, and the volume
\citet{mauroy2020koopman}.

The finite-state discretization used in Section~\ref{sec:nonlinear-discretization} is also closely
related to the older literature on transfer operators and set-oriented numerics. In particular,
our $\epsilon$-net construction is more naturally viewed as an Ulam-type finite approximation
than as a standard EDMD construction: one partitions the state space, represents the dynamics
by a finite matrix on those cells, and then studies the resulting linear surrogate. See
\citet{ulam1960collection}, \citet{froyland1999entropy}, and the overview
\citet{froyland2001extracting} for this line of work.

On the data-driven side, \citet{schmid2010dynamic} popularized Dynamic Mode Decomposition
(DMD) in fluid mechanics, and \citet{tu2014dynamic} clarified its operator-theoretic
interpretation. Extended Dynamic Mode Decomposition (EDMD), which replaces the identity
observable by a user-chosen dictionary, was introduced by \citet{williams2015data}. Related
developments include generator-based methods such as gEDMD \citep{klus2020generator} and
time-delay liftings \citep{kamb2020time}. These methods aim to approximate Koopman
eigenfunctions or invariant subspaces directly from data.

A separate strand of the literature studies when nonlinear systems admit exact or nearly exact
finite-dimensional Koopman representations. Finite-dimensional invariant subspaces for control
were explored by \citet{brunton2016koopman}, while learned Koopman coordinates and invariant
subspaces were developed by \citet{takeishi2017learning}. This distinction is conceptually
important for the present chapter: exact finite-dimensional Koopman closures are special,
whereas the discretization argument here gives a universal but approximate lifting.

For systems with inputs and control, operator-theoretic liftings lead naturally to linear or
bilinear predictor models. Particularly relevant references are \citet{proctor2018generalizing}
on Koopman operators with inputs and control, and \citet{korda2018linear} on linear predictors
for nonlinear controlled systems and model predictive control. These works are natural points of
comparison for the autonomous presentation adopted in this chapter.

Finally, the traditional alternative to the improper-learning viewpoint taken here is nonlinear
system identification, which attempts to estimate the nonlinear transition map or latent
state-space model directly from data. Good entry points to the classical literature are the perspective
article of \citet{ljung2010perspectives} and the state-space identification treatment of
\citet{schon2011system}. More recently, there has been significant progress in providing non-asymptotic, 
finite-sample guarantees for learning nonlinear dynamical systems directly from a single trajectory. 
For instance, \citet{foster2020learning} establish optimal sample complexity bounds for learning 
generalized nonlinear dynamics without relying on strict mixing assumptions. Similarly, 
\citet{sattar2022nonasymptotic} develop noise-sensitive, uniform convergence guarantees for learning 
nonlinear systems via gradient descent. The specific improper-learning reduction developed in this chapter,
which combines a lifted linear approximation with the Luenberger framework from
Chapter~\ref{chap:luenberger}, is based on \citet{dogariu2025universal}.

\ifarxiv
\newpage
\fi
\begin{exercises}

\begin{exercise}
Consider the proof of Theorem~\ref{thm:nonlinear_approx}. How does the dimension $N$ of the approximating linear system scale with the dimension of the true state $d_x$? Why does this curse of dimensionality not appear explicitly in the regret bounds derived from Chapter 12?
\end{exercise}

\begin{exercise}
Theorem~\ref{thm:nonlinear_approx} assumed the dynamics $f$ are $1$-Lipschitz. Suppose instead that $f$ is $L$-Lipschitz with $L > 1$.
\begin{enumerate}
    \item Derive the recurrence satisfied by the tracking error $\xi_t$.
    \item Show that $\xi_t$ can grow exponentially with $t$.
    \item What does this imply about the grid scale $\epsilon$ required to maintain low approximation error over a horizon $T$?
\end{enumerate}
\end{exercise}

\begin{exercise}
Assume $x_0 > 0$, and consider the nonlinear system
\[
x_{t+1} = x_t^2,
\qquad
y_t = x_t.
\]
\begin{enumerate}
    \item Is this system linear in the state variable $x$?
    \item Show that the observable
    \[
    g(x) = \log x
    \]
    evolves linearly along the dynamics, in the sense that
    \[
    g(x_{t+1}) = 2 g(x_t).
    \]
    \item Explain why this gives a one-dimensional Koopman-style lifting of the dynamics on the positive half-line.
\end{enumerate}
\end{exercise}
\begin{exercise}
\label{ex:llm-stochastic-dynamics}
Consider the large-language-model example in the introduction to this
chapter. Assume that the hidden state $\x_t$ contains the current internal
activations together with the prefix, cache, and other information needed to
generate the next token.

\begin{enumerate}
    \item Formulate deterministic decoding as an autonomous nonlinear
    dynamical system of the form
    \[
    \x_{t+1} = f(\x_t),
    \qquad
    \y_t = h(\x_t),
    \]
    where $\y_t \in V$ is the emitted token.

    \item Now consider randomized decoding, where the next token is sampled
    from a distribution determined by the current hidden state. A natural
    formulation is
    \[
    \y_t \sim p(\cdot \mid \x_t),
    \qquad
    \x_{t+1} = f(\x_t,\y_t).
    \]
    Explain why this gives a stochastic nonlinear dynamical system, and
    rewrite it equivalently either in noise-driven form
    \[
    \x_{t+1} = F(\x_t,\xi_t),
    \qquad
    \y_t = H(\x_t,\xi_t),
    \]
    or in Markov-kernel form.

    \item Suppose that, after an appropriate discretization or lifting, the
    relevant hidden-state space becomes finite. Explain why the evolution of
    the resulting distribution over lifted states is linear and governed by a
    stochastic matrix, rather than by a zero-one transition matrix as in the
    deterministic setting.
\end{enumerate}
\end{exercise}

\end{exercises}

\part{Online Control with Partial Observation}
\chapter{Policy Classes for Partially Observed Systems} \label{sec:partially-observed-policies}
\chaptermark{Policies for PO-LDS}

\ignore{
In this chapter we consider linear time varying systems with partial observation, i.e. the state is not available to the controller. Instead, a (known or unknown) linear projection of the state is revealed. In full generality, recall from definition \ref{def:dynamical-system-general} partially observed dynamical system is given by the equations
$$ \x_{t+1} = f_t(\x_t, \uv_t, \w_t), \quad \y_t = g_t(\x_t). $$
As per our notation thus far, here  $\x_t \in \reals^{d_x} $ is the state of the system, $\y_t \in \reals^{d_y} $ is the observation,  $\uv_t \in \reals^{d_u}$ is the control input, and $\w_t \in \reals^{d_w}$ is the perturbation. 
The function $g_t$ is a mapping from the state to the observed state, and the function $f_t$ is the transition function, given current state $\x_t$ and control input $\uv_t$. The subscript indicates the relevant quantities at time step $t.$

Many of the control policies we consider in this chapter are general and can be applied to any dynamical system of the above formulation. However, following the paradigm of chapter \ref{chap:policy-classes}, we relate the representation power of policy classes only for partially observed linear dynamical systems. These are given by the equations 
}

In previous chapters, we explored control and learning techniques in fully observed linear dynamical systems, where the system state is directly accessible to the controller. However, in many real-world scenarios, such as robotics, economics, and signal processing, the state is only partially observed, requiring the controller to infer system dynamics from noisy or incomplete observations.

This chapter focuses on partially observed linear dynamical systems, where the state is hidden, and only a linear projection of it is observed. For the remainder of this part of the text, we restrict ourselves to time-invariant systems and refer to the bibliographic section for extensions.  The system follows the general form:
\begin{align*}
\x_{t+1} &  = A \x_{t} + B \uv_t + \bw_t  \\
\y_{t} & = C \x_{t}  .
\end{align*}
Here, $\x_t$ represents the hidden system state, $\uv_t$ the control input, $\y_t$ the observed output, and $\w_t$ the disturbance. Often, partially observed LDSs are defined with an additional noise term that corrupts the observations. Exercise \ref{exer:generalpolds} asks the reader to prove that even in such circumstances we may assume the previously stated form for partially observed systems without loss of any generality.

Unlike in the fully observed setting, where controllers can react based on the full state, partial observability introduces fundamental limitations on policy design.

In the partial observation setting, some of the control policy classes from Chapter \ref{chap:policy-classes} are not well defined. We then define policy classes and describe the relationship between them. 

Not all control policies designed for fully observed systems extend naturally to partially observed settings. In this chapter, we systematically explore different policy classes, analyzing their expressivity, computational tractability, and suitability for partially observed LDS.

\begin{itemize}
\item
Linear Observation-Action Controllers (or simply Linear Controllers): These controllers directly map observations to actions, but are highly inexpressive, as they ignore past observations and fail to capture hidden state dependencies.

\item 
Generalized Linear Observation-Action Controllers (or generalized linear controllers): These extend linear controllers by incorporating the history of observations, significantly improving expressiveness.

\item 
Linear Dynamic Controllers (LDCs): These policies model internal system dynamics, making them more powerful than simple observation-based policies.

\item 
Disturbance Response Controllers (DRCs): Introduced as an alternative to state-based policies, DRCs act directly on a counterfactual signal, the ``Nature's y's'', which captures the response of the system that result by uniformly setting the control input to zero. We show that DRCs can approximate LDCs, indicating that they are a highly expressive and yet convex policy class.
\end{itemize}

We conclude by analyzing the relationships between these policy classes, showing how they fit into the broader framework of nonstochastic control with partial observation. This chapter lays the groundwork for online learning and regret minimization in partially observed LDS, which we will explore in the next chapter.

\begin{figure}[h] 
    \centering
     \includegraphics[width=0.5\textwidth]{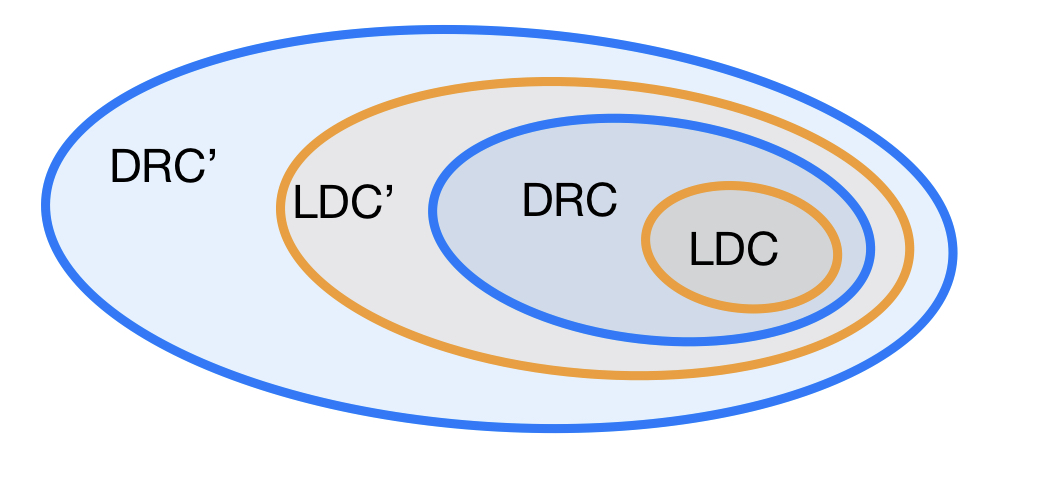} 
               \caption{Schematic relationship between linear Disturbance-Response Control policies and Linear Dynamical Control policies of increasing dimensions (or number of parameters) for linear time invariant systems.}\label{fig:policies2}
\end{figure}


\section{Linear Observation Action Controllers}

Linear Observation-action controllers are the most natural generalization of linear controllers to the partially observed case, and are defined as follows. 

\begin{definition}
A linear observation-action controller $K \in \reals^{d_u \times d_y}$ is a linear operator mapping observation to control as $\uv_t^\pi = K \y_t$. We denote by $\Pi_{\gamma}^{L}$ the set of all linear controllers bounded in the Fronbenius norm by $ \| K \| \leq \gamma $, which are also $\gamma$-stabilizing for given LDS. 
\end{definition}

Although this class of policies is natural to consider, it turns out to be very inexpressive. Even for optimal control with stochastic noise, the optimal policy cannot be approximated by a linear observation-action policy (see Exercise~\ref{exer:oacbad}).  Therefore, we now define more expressive and useful policies. However, linear controllers will be useful in simplifying proofs later on.

\subsection{Generalized Linear Observation Action Controllers}

We can generalize linear observation-action controllers to apply to a history of past observations. This additional generality, of taking a history of observations, greatly increases the expressive power of this policy class. 
The formal definition is given below. 

\begin{definition}
Generalized linear observation-action controllers parameterized by matrices $K_{0:h}$ choose a control input as
	 $$\uv_t^\pi = \sum_{i=0}^h K_i \y_{t-i}.$$ 
By $\Pi^{G}_{h,\gamma}$, we denote the class of generalized linear observation-action controllers that satisfy $\sum_{i=0}^h\|K_i\|\leq \gamma$ and are also $\gamma$-stabilizing.
\end{definition}

Generalized observation-action policies can be seen as a special case of observation-action controllers using the technique of dimension lifting as follows. 
Consider any partially-observed system $(A, B, C)$ and perturbation sequence $\w_t$. 
Consider  the {\em lifted} system $(\tilde{A}, \tilde{B}, \tilde{C}, \tilde{D})$ given by
$$ \z_{t+1} = \tilde{A} \z_t + \tilde{B}  \uv_t +  \tilde{D}\w_t, \quad \tilde{\y}_t = C\z_t,$$
where 
\begin{eqnarray*}
\tilde{A} = 
\begin{array}{|c|c|c|c|c|}
\hline
A & 0 &  \ldots & 0 & 0 \\ \hline
I & 0 &   \ldots & 0 & 0 \\ \hline
0 & I &   \ldots & 0 & 0\\ \hline
\vdots &  \vdots &\cdots & \vdots & \vdots \\ \hline
0 & 0 &  \cdots  &  I & 0 \\
\hline
\end{array} 
\ \ , \ \
\tilde{B} = 
\begin{array}{|c|}
\hline
B_t \\ \hline
0 \\ \hline
0 \\ \hline
\vdots \\ \hline
0 \\ \hline
\end{array} 
\ \ ,  \\ \ \ 
\tilde{C} = \begin{array}{|c|c|c|c|}
\hline
	C & C & \dots & C \\
\hline
\end{array} \ \ , \ \
\tilde{D} = 
\begin{array}{|c|}
\hline
I \\ \hline
0 \\ \hline
0 \\ \hline
\vdots \\ \hline
0 \\ \hline
\end{array}.
\end{eqnarray*}

The following lemma relates the original and lifted system (for a proof, see exercise \ref{exer:augpolds}).

\begin{lemma}\label{lem:augpolds}
For any sequence of control inputs $\uv_t$, it holds that $\tilde{\y}_t = [\y_t, \y_{t-1}, \dots \y_{t-h}]$. 
\end{lemma}

The above lemma shows that generalized linear observation-action controllers are a special case of simple observation-action controllers such that $\uv_t^\pi = [K_0 \dots K_h]\tilde{\y}_t$ for the lifted system. 

\section{Linear Dynamic Controllers}

The most well-known class of controllers for partially observable linear dynamical systems is that of linear dynamical controllers due to its connection to Kalman filtering.  
\begin{definition}[Linear Dynamic Controllers for Partially Observed Systems] A linear dynamic controller $\pi$ has parameters  $\left(A_{\pi}, {B}_{\pi}, C_{\pi}\right)$, and chooses a control at time $t$ as 
$$
\uv_{t}^\pi = C_{\pi} \s_{t} \ ,  \ \s_{t}= A_{\pi} \s_{t-1}+B_{\pi} \y_{t}.
$$
\end{definition}
We denote by $\Pi^{LDC}_{\gamma}$ the class of all LDC that are $\gamma$-stabilizing as per Definition \ref{def:gamma-stabilizing} and exercise \ref{exer:bibo-implies-exp-convergence}, with parameters $\frac{\kappa}{\delta} \leq \gamma$ that satisfy
$$ \forall i\ , \ \| A_\pi^i \| \leq \kappa (1-\delta)^i \ , \ (\|B_\pi\|+\|C_\pi\|) \leq \kappa . $$

Note that LDCs are $\varepsilon$-approximated by generalized observation-action controllers. The proof is left as Exercise \ref{exer:generalglc}.

\begin{lemma}\label{lemma:approx-ldc-gloc-po}
Consider a partially-observed time-invariant linear dynamical system $\{A,B,C\}$.
The class $\Pi^{G}_{h',\gamma'}$ $\eps$-approximates the class $\Pi^{LDC}_{\gamma}$ for $h'=O\left(\gamma \log \frac{\gamma}{\varepsilon}\right), \gamma'=2\gamma^3$.
\end{lemma}

\section{Disturbance Response Controllers}

The most important policy class in the context of nonstochastic control is that of Disturbance Response Controllers, or DRC.  These policies act on a signal called {\bf Nature's y's}, denoted $\ynat_t$, which is the would-be observation at time $t$ assuming that all inputs to the system were zero from the beginning of time. 


For ease of exposition, we only consider linear time invariant linear dynamical systems that are intrinsically stable. This allows us to ignore the stabilizing controller component of disturbance response control and greatly simplifies the presentation of the main ideas. The extension to time-varying systems is explored in the exercises. 

\begin{definition}[Disturbance Response Controller]
A disturbance response policy for stable systems $\pi_{M_{1:h}}$ is parameterized by matrices $M_{0:h} = [{M}_0, \ldots, {M}_{h}]$. It outputs control $\mathbf{u}_t^\pi$ at time $t$ according to the rule
\begin{equation*}
\mathbf{u}_t^\pi =  \sum_{i=0}^{h} {M}_{i} \ynat_{t-i},
\end{equation*}
where $\ynat_t$ is given by the system response to the zero control at time $t$, i.e. 
\begin{align*}
\xnat_{t+1} &  = A \xnat_{t} + \bw_t 
 = \sum_{i=0}^t A^i \bw_{t-i}   \ , \ \ynat_{t}  = C \xnat_{t} . 
\end{align*}
Denote by $\Pi^{R}_{h,\gamma}$ the set of all disturbance response policies that satisfy\footnote{Recall that for this section, we restrict our discussion to stable systems. Otherwise, we would also ask the policy class to be $\gamma$-stabilizing, which can, for example, be achieved by superimposing a fixed $\gamma$-stabilizing linear dynamical controller.} 
$$ \sum_{i=0}^h \| M_i\| \leq \gamma .$$

\end{definition}

If the perturbations are all zero, seemingly DRC policies produce zero control. Therefore, we could add a constant offset to the control that can be incorporated as in Definition~\ref{def:dac}.

The main advantage of this policy class, as opposed to previously considered policies in optimal control, is that they yield convex parametrization. Thus, this policy class allows for provably efficient optimization and algorithms that we explore in the next chapter.

We conclude with the following important lemma on an expression that is important to compute nature's y's. The proof of this lemma follows directly from the definition and is left as Exercise \ref{exer:nature-y-representation}.

\begin{lemma} \label{lem:nature-y-representation}
The sequence $\{ \ynat_t\}$ can be iteratively computed using only the sequence of observations as
$$ \ynat_t = \y_t - C \sum_{i=1}^t   A^i B \uv_{t-i}  $$
Alternatively, the above computation can be carried out recursively as stated below starting with $\z_0=\mathbf{0}$.
\begin{align*}
	\z_{t+1} &= A\z_t + B\uv_t,\\
	\ynat_t &= \y_t - C \z_t.
\end{align*}
\end{lemma}

We proceed to show that the class of DRC approximates the class of LDC.

\subsection{Expressivity of Disturbance Response Controllers}
\label{sec:DRC-LDC-power}

A DRC always produces controls that ensure that the system remains state-bounded. This can be shown analogously to Lemma \ref{lem:dac-is-stable} (see Exercise \ref{exer:drcstab}).

Although DRCs are contained within the policy class of DACs, they are still very expressive. Note that a DAC policy can not be executed in a partially observed system from observations alone; even the knowledge of the entire observation sequence $\yv$ may not uniquely determine $\w_t$.

We have already seen that LDCs are approximately captured by generalized linear controllers in Lemma \ref{lemma:approx-ldc-gloc-po}. It can be shown that the class of DRCs with appropriate parameters $\eps$-approximates the class of generalized linear policies for linear time-invariant systems. This observation in conjunction with Lemma~\ref{lemma:approx-ldc-gloc-po} implies that DRCs also approximate LDCs.

To demonstrate that DRCs approximate generalized linear controllers, we instead prove DRCs approximate linear controllers, a weaker class of policies. However, this is sufficient to conclude the stronger claim. To see this, first recall that generalized linear controllers acting on the last $h$ states can be written as linear controllers on a lifted system with $h$ states, as noted in Lemma \ref{lem:augpolds}. Together with the next lemma, this implies that a DRC on the lifted system can approximate generalized linear controllers on the original system. However, this is not quite what we promised. To wrap up the argument, one can verify that a DRC with history $h'$ on the lifted system can be written as a DRC with history $h'+h$ on the original system. 

\begin{lemma}\label{lemma:approx-drc-ldc}
Consider a $\gamma$-stable time-invariant linear dynamical system $(A,B,C)$.
The class $\Pi^{R}_{h',\gamma'}$ $\eps$-approximates the class $\Pi^{L}_{\gamma}$ for the parameters $h' =  O\left(\gamma\log\frac{\gamma}{\varepsilon}\right)$, $\gamma'=O(\gamma^6)$.
\end{lemma}

Recall that for fully observed LDS, the state-action pairs produced by execution of a linear policy can be written as linear functions of the perturbations. This observation allowed us to prove that DACs capture linear policies for fully observed systems. We prove an analogue for partially observed systems: The observation-action pairs produced by the execution of linear policies can be written as linear functions of nature's $y$'s. Thereafter, we argue that parameterizing the controllers to be linear in nature's $y$'s  allows us to encompass linear observation-action policies.

\begin{proof}
We introduce an augmented partially-observed linear system that helps us write the observation-action pairs produced by a linear policy as a linear function of nature's $y$'s. In fact, we will shortly see that, for any sequence of control inputs, the augmented linear system produces the same sequence of obeservations as the original system. Thus, the new system has the same nature's $y$'s as the original system. 

For any partially-observed system $(A, B, C)$ and perturbation sequence $\w_t$ producing observations $\y_t$, construct a {\em lifted} system $(\tilde{A}, \tilde{B}, \tilde{C})$ given by 

$$ \begin{bmatrix}
	\xnat_{t+1} \\ \Delta_{t+1}
\end{bmatrix} = \begin{bmatrix}
	A & 0\\
	0 & A
\end{bmatrix}\begin{bmatrix}
	\xnat_{t} \\ \Delta_{t}
\end{bmatrix} + \begin{bmatrix}0 \\ B
\end{bmatrix} \uv_t +  \begin{bmatrix}
	I \\ 0
\end{bmatrix}\w_t, \quad \tilde{\y}_t = \begin{bmatrix}
	C & C
\end{bmatrix}\begin{bmatrix}
	\xnat_{t} \\ \Delta_{t}
\end{bmatrix}.$$
Fix any sequence of control inputs $\uv_t$. Let $\x_t, \y_t$ be the states and observations produced by the original partially-observed system. The equivalents for the augmented system are $[{\xnat_{t}}^\top, \Delta_t^\top]^\top, \tilde{\y}_t$, as we have described above. Now we can verify for all time $t$ that $\tilde{\y}_t = \y_t$ and $\x_t=\xnat_t +\Delta_t$.

Since this new system has the same input-output behavior as the original system, we might equivalently investigate how the execution of a linear controller $\pi$ affects the augmented system. Substituting $\uv_t^\pi = K\y^\pi_t = KC(\xnat_t +\Delta^\pi_t)$, the state-action sequence produced by a linear policy $\uv_t^\pi = K\y_t^\pi$ can be written as
$$ \begin{bmatrix}
	\xnat_{t+1} \\ \Delta^\pi_{t+1}
\end{bmatrix} = \begin{bmatrix}
	A & 0\\
	BKC & A+BKC
\end{bmatrix}\begin{bmatrix}
	\xnat_{t} \\ \Delta^\pi_{t}
\end{bmatrix} +  \begin{bmatrix}
	I \\ 0
\end{bmatrix}\w_t, \quad \begin{bmatrix} \y^\pi_t \\ \uv^\pi_t \end{bmatrix} = \begin{bmatrix}
	C & C\\
	KC & KC
\end{bmatrix}\begin{bmatrix}
	\xnat_{t} \\ \Delta^\pi_{t}
\end{bmatrix},$$
or using that $\ynat_t = C\xnat_t$, equivalently as 
$$ \Delta^\pi_{t+1} = (A+BKC)\Delta^\pi_{t} + BK\ynat_t , \quad \y^\pi_t =  C\Delta^\pi_{t} + \ynat_t, \quad \uv^\pi_t = KC\Delta_t + K\ynat_t.$$
Now, define $M_{i} = KC(A+BKC)^{i-1} BK$ for all positive $i\in [h']$ and $M_0 = K$. These parameters specify our DRC. Let $\y_t, \uv_t$ be the observation-action sequence produced by this DRC. We have
\begin{align*}
	\uv^\pi_t &= K\ynat_t + KC \Delta^\pi_t \\ 
	&= K\ynat_t + KC BK \ynat_{t-1} + KC(A+BKC)\Delta^\pi_{t-1} \\ 
	&= K\ynat_t + \sum_{i>1} KC (A+BKC)^{i-1} BK \ynat_{t-i} \\
	&= \sum_{i=0}^{h'} M_i \ynat_{t-i} + \sum_{i>H'} KC (A+BKC)^{i-1} BK \ynat_{t-i} \\
	&= \uv_t + Z_t .
\end{align*}
Given $\gamma$-stabilizability of the linear controller $K$ and $\gamma$-intrinsic stability of the of the system $(A,B,C)$, we have that for some $\kappa, \delta$ such that $\frac{\kappa}{\delta}\leq \gamma$, 
\begin{align*}
	\|Z_t\| & \leq\sum_{i>h'} \|KC(A+BKC)^{i-1}BK\|\|\ynat_{t-i}\|\\
	 & \leq\sum_{i>h'}  \|KC(A+BKC)^{i-1}BK\|\left(\sum_{j>1}\left\| CA^{j-1}\w_{t-i-j}\right\|\right)\\
     & \leq\sum_{i>h'} \|K\|^2 \kappa (1-\delta)^{i-1}\sum_{j>1}\kappa (1-\delta)^j\\
	&\leq \frac{\kappa^4}{\delta} \sum_{i>h'} (1-\delta)^{i-1} \leq \frac{\kappa^4}{\delta^2} (1-{\delta})^{h'} \leq \frac{\kappa^4}{\delta^2}e^{-\delta h'}
	\end{align*}
Therefore, $\|\uv_t - \uv^\pi_t\|\leq \varepsilon$ by the appropriate choice of $h'$.
Once again using the {\em intrinsic} stability of the LTI $(A,B,C)$ , we have
\begin{align*}
	\|\y^\pi_t - \y_t\| &=  	\left\|\sum_{i=1}^t CA^{i-1}B (\uv^\pi_{t-i} - {\uv}_{t-i})\right\|\\
	&\leq 	\left\|\sum_{i=0}^t CA^iB\right\| \max_{i\leq t}\|\uv^\pi_i - {\uv}_i\|
	 \leq \frac{\kappa}{\delta}\varepsilon.
\end{align*}
The approximation in terms of observations and controls implies $\eps$-approximation of the policy class for any Lipschitz cost sequence, concluding the proof. 
\end{proof}

\section{Summary}

In this chapter, we extended our study of control to the partially observed setting, where the controller no longer has direct access to the system state but must instead rely on noisy or incomplete observations. This shift fundamentally changes the nature of the control problem, requiring the design of policies that can infer and act on the basis of limited information.

We introduced a hierarchy of policy classes, starting with Linear Observation-Action Controllers, which directly map observations to actions, and progressing to generalized linear policies and Linear Dynamic Controllers (LDCs), which incorporate history to improve decision-making. This increasing expressiveness highlighted a key limitation: memoryless policies are insufficient for optimal control, making structured state-dependent controllers necessary.

A major contribution of this chapter was the introduction of Disturbance Response Controllers (DRCs). Instead of explicitly estimating the state of the system, DRCs act directly on a counterfactual signal, the ``Nature's y's'', capturing what the system would have observed in the absence of control inputs. We established that DRCs are a convex relaxation of LDCs, making them computationally attractive while maintaining strong expressive power. This result provides a principled approach to designing policies that are efficient and robust in partially observed environments.

This chapter also set the stage for online nonstochastic control under partial observability, which we will explore in the next chapter. By establishing policy expressiveness results and identifying convex approximations for optimal control, we have built the necessary framework for developing learning-based algorithms that operate effectively in uncertain and adversarial settings.

\ifarxiv
\newpage
\fi
\section{Bibliographic Remarks}

The study of control in partially observed linear dynamical systems has a long history, with foundational contributions dating back to optimal estimation and control theories. This section provides an overview of key references related to the topics discussed in this chapter, including classical works, expressivity of policy classes, convex relaxations, and connections to online learning.

The problem of control under partial observation was first rigorously formulated in the context of the Linear Quadratic Gaussian (LQG) regulator, which combines Kalman filtering with optimal control \citep{anderson1979optimal}. The certainty equivalence principle, which states that optimal controllers can be designed assuming a perfect state estimate, was shown to have limitations in more general partially-observed settings \citep{wonham1968separation}.

This chapter introduced a hierarchy of policy classes, including linear observation-action controllers, generalized linear controllers, Linear Dynamic Controllers (LDCs), and Disturbance Response Controllers (DRCs). The concept of disturbance response controllers was inspired by parameterizations used in robust control, particularly the Youla-Kucera parameterization \citep{youla1976modern}.

Convex relaxations have been a powerful tool for simplifying control design in partially observed LDS. We have surveyed convex relaxations for learning in dynamical systems in the previous part of this book. Recently, spectral methods for control \citep{arora2018towards} have provided an alternative approach to convex relaxation, utilizing spectral filtering to bypass direct system identification.

The Disturbance Response Parametrization was introduced in \citet{simchowitz2020improper} as an analogue of DAC to control partially-observed systems. It can be viewed as a succinct state-space representation of the Youla parameterization \citep{youla1976modern}. For time-varying dynamics, the power of various policy classes was contrasted in \citet{minasyan2021online}.

\ifarxiv
\newpage
\fi
\begin{exercises}

\begin{exercise}\label{exer:nature-y-representation}
Prove Lemma \ref{lem:nature-y-representation}, stating that the sequence $\{ \ynat_t\}$ can be iteratively computed using only the sequence of observations as
$$ \ynat_t = \y_t - C \sum_{i=1}^t   A^i B \uv_{t-i}  $$
Alternatively, the above computation can be carried out recursively as stated below starting with $\z_0=\mathbf{0}$.
\begin{align*}
	\z_{t+1} &= A\z_t + B\uv_t,\\
	\ynat_t &= \y_t - C \z_t.
\end{align*}
\end{exercise}

\begin{exercise}\label{exer:generalpolds}
	Consider a seemingly more general definition of a partially observed linear dynamical system, given below:
	\begin{align*}
\x_{t+1} &  = A_t \x_{t} + B_t \uv_t + \bw_t,  \\
\y_{t} & = C_{t} \x_{t}  +\vv_t.
\end{align*}
Construct a linear dynamical system parameterized via matrices $(A'_t, B'_t, C'_t)$ and perturbations $\bw'_t$ with a hidden state $\z_t$ of the form
	\begin{align*}
\x_{t+1} &  = A'_t \z_{t} + B'_t \uv_t + \bw'_t,  \\
\y'_{t} & = C'_{t} \z_{t}  
\end{align*}
such that for any sequence of $\uv_t$, $\y_t=\y'_t$ holds uniformly over time. Additionally, verify that there exists a construction of $\z_t$ where $\bw'_t$ are i.i.d. as long as $(\bw_t, \vv_t)$ are i.i.d. across time steps. 

{\bf Hint:} Consider the choice $\bw'_t=D'_t\begin{bmatrix}\bw_t \\ \vv_t\end{bmatrix}$ for some $D'_t$. Also, consider using a hidden state $\z_t$ with a larger number of coordinates than $\x_t$.
\end{exercise}

\begin{exercise}\label{exer:augpolds}
\begin{enumerate}
    \item Prove Lemma~\ref{lem:augpolds}.
\item Extend the lemma and prove it for time-varying linear dynamical systems. 
\end{enumerate}
\end{exercise}

\begin{exercise}\label{exer:generalglc}
Prove Lemma~\ref{lemma:approx-ldc-gloc-po}.
\end{exercise}

\begin{exercise}\label{exer:drcstab}
For any $\gamma$-stable partially observed system, prove that any DRC policy in $\Pi_{H,\gamma}^{D}$ is $\gamma^2$-stabilizing.
\end{exercise}

\begin{exercise}\label{exer:oacbad}
Construct a partially observed linear dynamical system subject to Gaussian i.i.d. perturbations with quadratic costs such that simultaneously it is true that (a) there exists a genearlized linear policy that incurs zero (or at most constant) aggregate cost, (b) yet any linear observation-action policy incurs $\Omega(T)$ aggregate cost when run for $T$ time steps.

{\bf Hint:} Consider a 3-dimensional hidden state so that $u_t$ can be chosen as a function of $y_{t-1}$ to ensure zero cost at each time step. 
\end{exercise}

\end{exercises}
\chapter{Online Nonstochastic Control with Partial Observation}
\chaptermark{ONC with Partial Observation}

In the previous chapter, we introduced Disturbance Response Controllers (DRCs) as a structured class of policies for controlling partially observed linear dynamical systems (PO-LDS). Unlike classical observer-based controllers, which rely on explicit state estimation, DRCs act directly on Nature’s y’s, the system’s counterfactual response in the absence of control. Although DRCs offer a powerful alternative to state-based policies, we have not yet integrated them into an online learning algorithm, leaving open the question of how to adapt them dynamically in uncertain environments.

This chapter takes the next step by introducing an online learning algorithm for optimizing DRCs in real time. We introduce the Gradient Response Controller (GRC), which efficiently updates DRC policies using online convex optimization techniques. The GRC algorithm enables controllers to:
\begin{itemize}
\item 
Learn from past disturbances and actions, refining policy parameters without requiring full system identification.
\item 
Handle adversarial disturbances, extending the policy regret framework to the partially observed setting.
\item 
Optimize control decisions dynamically, adapting to system behavior as new observations arrive.
\end{itemize}
Unlike previous approaches that relied on fixed policy classes, GRC provides an adaptive control mechanism that bridges the gap between structured controllers and learning-based methods. A key focus of this chapter is to extend policy regret minimization from the fully observed case to partial observability, introducing new challenges and solutions in learning-based control.

To simplify the theoretical analysis, we assume system stability, allowing us to focus on the core learning problem without additional stabilization constraints. However, we discuss possible extensions to stabilizable systems in the Bibliographic Remarks section.

\section{The Gradient Response Controller}

Partial observation requires modification to the Gradient Perturbation Controller we have studied in the previous chapter. 
The Gradient Response Controller algorithm is described in figure \eqref{algo:GRC}, for linear time-invariant dynamical systems with partial observation. Similarly to GPC, the main idea is to learn the parameterization of a disturbance response control policy using known online convex optimization algorithms, such as online gradient descent. We henceforth prove that this parameterization is convex, allowing us to prove the main regret bound. 

In algorithm \eqref{algo:GRC} we denote $\y_t({M}_{1:h})$ as the observation arising from playing the DRC policy ${M}_{1:h}$ from the beginning of time, and the same for $\uv_t({M}_{1:h})$. Although it may be impossible to determine $\w_t$, especially if $C$ has fewer rows than columns, from the observations $\y_t$, $\y_t({M}_{1:h})$ can be determined from $\ynat_t$ alone by the application of Lemma~\ref{lem:nature-y-representation}, as
$$ \y_t({M}_{1:h}) = \ynat_t + C \sum_{i=1}^t   A^i B \uv_{t-i}({M}_{1:h}). $$

\begin{algorithm}[H]
\begin{algorithmic}[1] 
\STATE {Input:} system $\{ A, B , C \}$ , $h$, $\eta$, initialization $M_{0:h}^1 \in \K$.
\FOR{$t$ = $1 \ldots T$}
        \STATE  $\mbox{Use Control }\mathbf{u}_t =\sum_{i=0}^{h} {M}_i^t \ynat_{t-i} $
        \STATE  Observe $\mathbf{y}_{t+1}$, compute for $t+1$: 
        $$ \ynat_t = \y_t - C \sum_{i=0}^t     A^i B \uv_{t-i} . $$        \STATE  Construct loss $\ell_t({M}_{0:h}) = c_t(\y_t({M}_{0:h}), \mathbf{u}_t({M}_{0:h}))$\\
       \STATE Update $M_{0:h}^{t+1} \leftarrow \prod_\K \left[  {M}_{0:h}^t -\eta \nabla \ell_t({M}_{0:h}^t) \right]$
\ENDFOR
 \caption{Gradient Response Controller(GRC)}
 \label{algo:GRC}
\end{algorithmic}
\end{algorithm}

The GRC algorithm comes with a finite-time performance guarantee: it guarantees vanishing worst-case regret versus the best disturbance response control policy in hindsight. Following the relationship between disturbance response control and other policies studied in Section \ref{sec:DRC-LDC-power}, this implies vanishing regret versus linear dynamical controllers. The formal statement is given as follows. 

\begin{theorem} \label{thm:grc-main}
Assuming that 
\begin{itemize}
    \item[a]
    The costs $c_t$ are convex, bounded and have bounded gradients with respect to the arguments $\mathbf{y}_t$ and $\mathbf{u}_t$. 

    \item[b] The diameter of the constraint set for the parameters $M_{0:h}$ is bounded by $D$. 
    
    \item[c] The matrices $\{{A} , {B}, C \}$ have bounded $\ell_2$ norms.
     
    \item[d] The linear dynamical system $\{{A} , {B}, C \}$ is $\gamma$-stable.
\end{itemize}
Then the GRC algorithm~(\ref{algo:GRC}) ensures that 
\begin{align*}
    \max_{\mathbf{w}_{1:T}: \norm{\mathbf{w}_t}  \le 1} \left(\sum_{t=1}^{T} c_t (\mathbf{y}_t, \mathbf{u}_t) - \min_{\pi \in \Pi^{R}} \sum_{t=1}^{T} c_t (\y_t^\pi, \uv_t^\pi) \right) \le \mathcal{O}(  L G D \gamma^3 h \sqrt{T} ).
\end{align*}
Furthermore, the time complexity of each loop of the algorithm is polynomial in the number of system parameters and logarithmic in T.
\end{theorem}

The proof is largely analogous to the regret bound of the GPC presented in Theorem \ref{thm:gpc-regret}. 

\begin{proof}
According to Theorem \ref{thm:ogd}, the online gradient descent algorithm for convex loss functions ensures that the regret is bounded as a sublinear function of the number of iterations. However, to directly apply the OGD algorithm to our setting, we need to make sure that the following conditions hold true:
\begin{itemize}
    \item The loss function should be a convex function in the variables ${M}_{0:h}$.

    \item We must ensure that optimizing the loss function $\ell_t$ as a function of parameters $M_{0:h}$ is similar to optimizing the cost $c_t$ as a function of observation and control.
\end{itemize}

First, we show that the loss function is convex with respect to the variables ${M}_{0:h}$. This follows since the states and the controls are linear transformations of the variables.
\begin{lemma}\label{lemma:convex2}
The loss functions $\ell_t$ are convex in the variables ${M}_{0:h}$.
\end{lemma}

\begin{proof}
    Using $\y_t$ to denote $\y_t(M_{0:h})$ (and similarly for $\uv_t$), the loss function $\ell_t$ is given by
    \begin{equation*}
        \ell_t({M}_{0:h}) = c_t(\y_t, \uv_t).
    \end{equation*}
    Since the cost $c_t$ is a convex function with respect to its arguments, we simply need to show that $\mathbf{y}_t$ and $\mathbf{u}_t$ depend linearly on ${M}_{0:h}$.
    Using Lemma~\ref{lem:nature-y-representation}, 
    \begin{align*}
    	\mathbf{u}_t &= \sum_{i=0}^h M_i \ynat_{t-i},\\
        \mathbf{y}_{t} &= \ynat_t + C \sum_{i=1}^t A^i B \uv_{t-i} \\
        &= \ynat_t + C \sum_{i=1}^t A^i B \sum_{j=0}^h M_j \ynat_{t-i-j} 
    \end{align*}
    which is a linear function of the variables. Thus, we have shown that $\mathbf{y}_t$ and $\mathbf{u}_t$ are linear transformations in ${M}_{0:h}$ and, hence, the loss function is convex in ${M}_{0:h}$.
\end{proof}

Next, we use the following lemma. 
\begin{lemma} \label{lem:grc-lemma2}
For all $t$, $|l_t(M^t_{0:h}) - c_t(\y_t, \uv_t)| \leq \frac{L \gamma^3 C h D}{\sqrt{T}} $.
\end{lemma}

The regret bound of GRC can be derived from these lemmas as follows.  Note that $G$ is an upper bound on the gradients of $\ell_t$ as a function of the parameters $M_{0:h}$, while $L$ is the Lipschitz constant of $c_t$ as a function of the state and control.  
\begin{align*}
    &\sum_{t=1}^{T} c_t (\y_t, \mathbf{u}_t) - \min_{\pi \in \Pi^{R}} \sum_{t=1}^{T} c_t ({\y}^\pi_t,  \uv_t^\pi)) \\
    &\leq \sum_{t=1}^T l_t(M_{1:h}^t) - \min_{\pi \in \Pi^{R}} l_t(M_{1:h}^\pi) + T\times \frac{L \gamma^3 C h D}{\sqrt{T}} & \mbox{Lemma \ref{lem:grc-lemma2}} \\
    & \leq 2 GD \sqrt{T} + {\sqrt{T} L \gamma^3 C h D}  & \mbox{Theorem \ref{thm:ogd}}\\
    & \leq O(  G D L \gamma^3 h \sqrt{T} ).
\end{align*}
\end{proof}

\subsection{Finishing up the Regret Bound}
\begin{proof}[Proof of Lemma \ref{lem:grc-lemma2}]
We first note that by the choice of step size by the Online Gradient Descent algorithm \ref{alg:ogd}, as per Theorem \ref{thm:ogd}, we have $\eta = \frac{D}{G \sqrt{T}}$. Thus,  
$$\|M_{0:h}^t - M_{0:h}^{t-i}\|\leq \sum_{s=t-i+1}^{t} \|M_{0:h}^s - M_{0:h}^{s-1}\| \leq i \eta G = \frac{i D}{ \sqrt{T}}.$$

Now, we use the fact that the iterates of OGD $M_{1:h}^t$ move slowly to establish that $\y_t$ and $\y_t(M^t_{0:h})$ are close in value. Before that, by definition, we have $\uv_t(M^t_{0:h}) = \uv_t$ since the sequence $\ynat_t$ is determined independently of the choice of the policy that is being executed. Due to the $\gamma$-stability of the underlying system, for any $i > 0$, $\|C {A}^{i} B\|\leq \kappa (1-\delta)^{i}$ for some $\kappa>0, \gamma<1$. Using Lemma~\ref{lem:nature-y-representation}, we have for some $C>0$ that
\begin{align*}
\|\y_t(M^t_{0:h})-\y_t\| &= \left\|C \sum_{i=1}^t \left[  A_{i-1} B (\uv_{t-i}(M^t_{0:h}) - \uv_{t-i}) \right] \right\|\\
	&\leq \sum_{i=1}^t \left\|C   A^{i-1} B \right\| \|(\uv_{t-i}(M^t_{0:h}) - \uv_{t-i})\| \\
	&\leq  C h \kappa \sum_{i=1}^t (1-\delta)^{i} \frac{i}{\sqrt{T}} \max_j\|\ynat_j\| \\
	& \leq \frac{C h \kappa D }{\delta^2\sqrt{T}} \max_j\|\ynat_j\| \leq \frac{C h D \gamma^3}{\sqrt{T}} 	
\end{align*}
Here we used the fact that $\|\ynat_t\| \leq \gamma $, which is left as an exercise. 

The control depends only on the parameters of the current iteration, and we have $\uv_t =  \uv_t(M_{1:h}^t)$. By definition, $l_t(M_{0:h}^t) = c_t(\y_t(M_{0:h}^t), \uv_t(M_{0:h}^t))$. Thus, we have 
\begin{eqnarray*}
l_t(M_{0:h}^t) - c_t(\y_t,\uv_t) & = c_t(\y_t(M_{0:h}^t), \uv_t(M_{0:h}^t)) - c_t(\y_t, \uv_t) \\
& \leq L \cdot \| \y_t(M_{1:h}^t) - \y_t\| \leq \frac{L \gamma^3 C h D}{ \sqrt{T}} .
\end{eqnarray*}

\end{proof}

\section{Extension to Time Varying Systems}

The GRC algorithm can be extended to time-varying dynamics as we explain in this section. The linear dynamical system at time $t$ is denoted by the system matrices $A_t,B_t,C_t$, and we assume that they are known to the controller. We further assume that these dynamical systems are all $\gamma$-stable. 
The algorithm is formally spelled out in figure \ref{algo:GRC2}.

\begin{algorithm}[h]
\begin{algorithmic}[1] 
\STATE {Input:} sequence $\{ A_t, B_t , C_t \}$ , $h$, $\eta$, initialization $M_{0:h}^1 \in \K $.
\FOR{$t$ = $1 \ldots T$}
        \STATE  $\mbox{Use Control }\mathbf{u}_t =\sum_{i=0}^{h} {M}_i^t \ynat_{t-i} $
        \STATE  Observe $\mathbf{y}_{t+1}$, compute for $t+1$: 
        $$ \ynat_t = \y_t - C_t \sum_{i=0}^t   \left[ \prod_{j=0}^{i}  A_{t-j} \right] B_{t-i} \uv_{t-i} . $$        \STATE  Construct loss $\ell_t({M}_{0:h}) = c_t(\y_t({M}_{0:h}), \mathbf{u}_t({M}_{0:h}))$\\
       Update $M_{0:h}^{t+1} \leftarrow \prod_\K \left[   {M}_{0:h}^t -\eta \nabla \ell_t({M}_{0:h}^t) \right]$
\ENDFOR
 \caption{Gradient Response Controller(GRC)}
 \label{algo:GRC2}
\end{algorithmic}
\end{algorithm}

It is possible to prove a similar regret bound to the one we proved for linear time-invariant systems. An important step is to note that the observation $\y_t({M}_{1:h})$ can be determined from $\ynat_t$ alone by the application of Lemma~\ref{lem:nature-y-representation}, as
$$ \y_t({M}_{1:h}) = \ynat_t + C_t \sum_{i=1}^t   \left[ \prod_{j=1}^{i-1}  A_{t-j}  \right] B_{t-i} \uv_{t-i}({M}_{1:h}). $$

\section{Conclusion}

In this chapter, we extended the framework of online nonstochastic control to the partially observed setting, where the controller must make decisions without direct access to the system state. Unlike the fully observed case, where state-feedback policies can be applied directly, partial observability introduces fundamental challenges, requiring policies that infer system behavior from past observations and actions.

Building on the Disturbance Response Controller (DRC) policies introduced in the previous chapter, we developed the Gradient Response Controller (GRC), an online learning algorithm that optimizes DRC policies in real time.

The results in this chapter demonstrate that adaptive control under partial observability is possible without explicit state estimation, bridging the gap between classical observer-based controllers and modern learning-based methods. Using regret minimization techniques, we provide a robust and flexible approach that applies even in adversarial and uncertain environments.

\ifarxiv
\newpage
\fi
\section{Bibliographic Remarks}

Online nonstochastic control with partial observability was studied in \citet{simchowitz2020improper}, who defined the Gradient Response Controller as well as the policy class of Disturbance Response Controllers. 

The setting was further extended to time-varying linear dynamics in \citet{minasyan2021online}, where the expressive power of different classes of policy was also studied. 

The analogue of online nonstochastic control in the stochastic noise setting is called the Linear Quadraic Gaussian problem, and it is further limited by assuming quadratic noise. This classical problem in optimal control was proposed in Kalman's work \citep{kalman1960new}. Regret bounds for controlling unknown partially-observed systems were studied in \citet{anima1,anima3,mania2019certainty}. \citet{anima2}, which use DRC policies even in the context of stochastic control, obtain rate-optimal logarithmic bounds for the LQG problem.

In \citet{sun2023optimal}, the task of controlling a partially-observed system subject to semi-adversarial noise under bandit feedback was considered. The tight rates for quadratic control with bandit feedback, in the fully adversarial model, were obtained in \citet{suggala2024second}, and more general cost functions in the bandit setting were considered in \citet{sun2024tight}.

\ifarxiv
\newpage
\fi
\begin{exercises}

\begin{exercise}
In this exercise, we consider the computation of nature's y's for time-varying systems.  Let 
\begin{align*}
\xnat_{t+1} &  = A_t \xnat_{t} + \bw_t 
 = \sum_{i=0}^t \left[ \prod_{j=0}^{i-1}  A_{t-j}  \right] \bw_{t-i} = \Phi_t \w_{1:t}. \\
\ynat_{t+1} & = C_{t+1} \xnat_{t+1} = C_{t+1} \Phi_t \w_{1:t} .
\end{align*}
Here $\Phi_t$ is the linear operator such that $ \Phi_t(t-i) = \prod_{j=0}^{i-1}  A_{t-j}  $.
Prove that 
The sequence $\{ \ynat_t\}$ can be iteratively computed using only the sequence of observations as
$$ \ynat_t = \y_t - C_t \sum_{i=1}^t   \left[ \prod_{j=1}^{i-1}  A_{t-j}  \right] B_{t-i} \uv_{t-i}  $$
Alternatively, the above computation can be carried out recursively as stated below starting with $\z_0=\mathbf{0}$.
\begin{align*}
	\z_{t+1} &= A_t\z_t + B_t\uv_t,\\
	\ynat_t &= \y_t - C_t \z_t.
\end{align*}

\end{exercise}

\begin{exercise}\label{exer:DRCstable}
Consider a partially observed linear dynamical system that is $\gamma$-stable in the absence of control inputs. Prove that when a DRC policy $\pi\in \Pi^{R}_{H,\gamma}$ is executed on such a dynamical system, the resulting system is $\gamma^2$-stable.
\end{exercise}

\begin{exercise}\label{exer:opt-LQG}
Prove that the optimal controller that minimizes the aggregate cost for a partially observed linear dynamical system assuming that the perturbations are stochastic is a linear dynamical controller. 
\end{exercise}

\begin{exercise}
Prove that the regret of the GRC controller for time-varying linear dynamical systems is bounded by $O(\sqrt{T})$. \\
{\bf Hint:} Follow the proof of Theorem \ref{thm:grc-main} and modify as required. 
\end{exercise}

\end{exercises}

\appendix

\chapter{Second Order Online Learning with the Square Loss} \label{chap:ons-square}

This appendix presents second order methods for the special case of online convex optimization with the square loss.  The first algorithm is a specialization of the Online Newton Step (ONS) 
algorithm for the square loss. ONS operates strictly within the standard Online Convex Optimization (OCO) protocol, meaning no feature vector is revealed to the learner prior to making a decision. In addition, ONS gracefully handles arbitrary constrained convex sets for the learned parameters. 
The proof follows the same potential-based argument as in \cite[Theorem 4.5]{hazan2016introduction}, but exploits the exact quadratic structure of square loss.

The second method we describe is the classical VAW algorithm, which does not strictly fall within the framework of online convex optimization due to two limitations. First, it requires knowing the feature vector before prediction, and secondly, it does not allow for constraints. However, in many control scenarios studied in this text, these limitations are acceptable, and in return it offers superior regret bounds in some respects.

\section{Setting}

We consider the standard Online Convex Optimization (OCO) protocol. At each round $t=1,\dots,T$:

\begin{itemize}
\item The learner chooses a parameter $\x_t \in \K \subseteq \reals^d$, where $\K$ is a closed convex set with diameter $D$.
\item The environment reveals a feature vector $\a_t \in \reals^d$ such that $\|\a_t\| \leq R$, and a true label $b_t \in \reals$, defining the loss function $\ell_t(\x) = \frac12(\x^\top \a_t - b_t)^2$.
\item The learner suffers loss $\ell_t(\x_t)$.  Assume that the losses  suffered by the learner are bounded by a constant $C>0$:$$\max_{t=1,\dots,T} \ell_t(\x_t) = \max_t \frac{1}{2} (\x_t ^\top \a_t - b_t)^2 \le C .$$
\end{itemize}


Define regret against $\x^\star \in \K$:
\[
\regret_T(\x^\star)
=
\sum_{t=1}^T
\ell_t(\x_t)
-
\ell_t(\x^\star).
\]

\section{Algorithm: ONS for Square Loss}

\begin{algorithm}[h]
\caption{ONS for Square Loss}
\label{alg:ons-square}
\begin{algorithmic}[1]
\STATE Input: regularization $\lambda>0$
\STATE Initialize $A_0 = \lambda I_d$, $\x_1 \in \K$
\FOR{$t=1$ to $T$}
    \STATE Play $\x_t$
    \STATE Observe $\a_t, b_t$ and suffer loss $\ell_t(\x_t) = \frac12(\x_t^\top \a_t - b_t)^2$
    \STATE Compute gradient $\nabla_t = \nabla \ell_t(\x_t)
        = (\x_t^\top \a_t - b_t) \a_t$
    \STATE Update curvature matrix:
    \[
    A_t = A_{t-1} + \a_t \a_t^\top
    \]
    \STATE Update parameter:
    \[
    \x_{t+1}
    =
    \proj_{\K}^{A_t}
    \left(
    \x_t - A_t^{-1} \nabla_t
    \right)
    \]
\ENDFOR
\end{algorithmic}
\end{algorithm}

\section{Regret Bound}

\begin{theorem}[Logarithmic regret for square loss]
\label{thm:ons-square}
For Algorithm~\ref{alg:ons-square}, assuming losses are bounded by $C$ and $\K$ has diameter $D$, for all $\x^\star\in\K$,
\[
\regret_T(\x^\star)
\le
\frac{\lambda}{2} D^2
+
{C}
\log\det\!\left(
I_d + \frac{1}{\lambda}
\sum_{t=1}^T \a_t \a_t^\top
\right).
\]
In particular, setting $\lambda = 1/D^2$, and $d,C,T,R,D > 2$ we get
\[
\regret_T(\x^\star)
\le
\frac{\lambda}{2} D^2
+
{d C}
\log\!\left(1 + \frac{T R^2}{\lambda}\right) \leq 4 dC \log\!\left(T R D \right) .
\]
\end{theorem}

\begin{proof}
Fix $\x^\star \in \K$. Let $r_t = \x_t^\top \a_t - b_t$ denote the prediction residual.
For square loss we have the exact identity
\[
\ell_t(\x_t) - \ell_t(\x^\star)
=
\nabla_t^\top (\x_t - \x^\star)
-
\frac12 ((\x_t - \x^\star)^\top \a_t)^2.
\]
Indeed, expanding the difference in losses yields
\begin{align*}
\ell_t(\x_t) - \ell_t(\x^\star)
&=
\frac12\left(r_t^2 - (r_t-(\x_t - \x^\star)^\top \a_t)^2\right)\\
&=
r_t\,(\x_t - \x^\star)^\top \a_t
-
\frac12((\x_t - \x^\star)^\top \a_t)^2\\
&=
\nabla_t^\top (\x_t - \x^\star)
-
\frac12((\x_t - \x^\star)^\top \a_t)^2.
\end{align*}

We define the potential $\Phi_t = \|\x_t - \x^\star\|_{A_{t-1}}^2$.
By the generalized Pythagorean theorem for projection onto a convex set in the $A_t$-norm,
\[
\|\x_{t+1} - \x^\star\|_{A_t}^2
\le
\|\x_t - A_t^{-1} \nabla_t - \x^\star\|_{A_t}^2.
\]
Expanding the right-hand side gives
\[
\|\x_t - A_t^{-1} \nabla_t - \x^\star\|_{A_t}^2
=
\|\x_t - \x^\star\|_{A_t}^2
- 2 \nabla_t^\top (\x_t - \x^\star)
+ \nabla_t^\top A_t^{-1} \nabla_t.
\]
Rearranging this inequality to bound the linear term yields
\[
\nabla_t^\top (\x_t - \x^\star)
\le
\frac12
\left(
\|\x_t - \x^\star\|_{A_t}^2
-
\|\x_{t+1} - \x^\star\|_{A_t}^2
\right)
+
\frac12 \nabla_t^\top A_t^{-1} \nabla_t.
\]
Since the curvature matrix is updated via $A_t = A_{t-1} + \a_t \a_t^\top$, we can relate the norms as
\[
\|\x_t - \x^\star\|_{A_t}^2
=
\|\x_t - \x^\star\|_{A_{t-1}}^2
+
((\x_t - \x^\star)^\top \a_t)^2.
\]
Plugging this expansion into our previous inequality and moving the resulting quadratic term to the left-hand side gives
\[
\nabla_t^\top (\x_t - \x^\star) - \frac12 ((\x_t - \x^\star)^\top \a_t)^2
\le
\frac12
\left(
\|\x_t - \x^\star\|_{A_{t-1}}^2
-
\|\x_{t+1} - \x^\star\|_{A_t}^2
\right)
+
\frac12 \nabla_t^\top A_t^{-1} \nabla_t.
\]
By the exact identity established above, the left-hand side is exactly the instantaneous regret $\ell_t(\x_t) - \ell_t(\x^\star)$.
Summing over $t=1$ to $T$ telescopes the potential terms:
\[
\regret_T(\x^\star)
\le
\frac12 \|\x_1-\x^\star\|_{A_0}^2
+
\frac12 \sum_{t=1}^T \nabla_t^\top A_t^{-1} \nabla_t.
\]
Since $A_0=\lambda I_d$, we have $\frac12 \|\x_1-\x^\star\|_{A_0}^2 = \frac{\lambda}{2}\|\x^\star-\x_1\|_2^2 \le \frac{\lambda}{2} D^2$, where the last inequality uses the assumption that $\K$ has diameter $D$.
Using $\nabla_t = (\x_t^\top \a_t - b_t)  \a_t$, we can rewrite the gradient term as $\nabla_t^\top A_t^{-1} \nabla_t = 2 \ell_t(\x_t) (\a_t^\top A_t^{-1} \a_t)$.
By our assumption that the losses are bounded by $C$, we have $\ell_t(\x_t)  \le C $, and thus
\[
\sum_{t=1}^T \nabla_t^\top A_t^{-1} \nabla_t 
\le 
2 C \sum_{t=1}^T \a_t^\top A_t^{-1} \a_t.
\]
By the matrix determinant lemma, $A_t = A_{t-1} + \a_t \a_t^\top$, and the vector harmonic series lemma from \cite{hazan2016introduction} Theorem 4.5,  we have 
\[
\sum_t \a_t^\top A_t^{-1} \a_t = \sum_t A_t^{-1} \bullet (A_t - A_{t-1}) \le \log\frac{\det(A_T)}{\det(A_{0})}.
\]
Thus, the regret is bounded by
\[
\regret_T(\x^\star)
\le
\frac{\lambda}{2} D^2
+
{C}
\log\frac{\det(A_T)}{\det(A_0)}.
\]

To explicitly bound the log-determinant ratio, we bound the determinant of $A_T$ using its trace. Since $A_T = \lambda I_d + \sum_{t=1}^T \a_t \a_t^\top$, the trace is linearly bounded:$$\text{Tr}(A_T) = \text{Tr}(\lambda I_d) + \sum_{t=1}^T \|\a_t\|_2^2 \le \lambda d + T R^2.$$By the AM-GM inequality, the determinant of a positive definite matrix is bounded by the average of its eigenvalues raised to the dimension $d$, which gives:$$\det(A_T) \le \left( \frac{\text{Tr}(A_T)}{d} \right)^d \le \left( \frac{\lambda d + T R^2}{d} \right)^d \le (\lambda + T R^2)^d.$$Since $A_0 = \lambda I_d$, we have $\det(A_0) = \lambda^d$. Taking the logarithm of the ratio yields:$$\log\frac{\det(A_T)}{\det(A_0)} \le \log\frac{(\lambda + T R^2)^d}{\lambda^d} = d \log\left( 1 + \frac{T R^2}{\lambda} \right).$$Substituting this upper bound into our regret inequality yields the final stated bound.

\end{proof}

\section{The VAW Algorithm for Online Least Squares}
\label{sec:vaw}

We explicitly compare this approach to the well-known Vovk-Azoury-Warmuth (VAW) forecaster \cite{vovk2001competitive, azoury2001relative} to highlight two critical distinctions. First, ONS operates strictly within the standard Online Convex Optimization (OCO) protocol, meaning no feature vector is revealed to the learner prior to making a decision. Second, ONS gracefully handles arbitrary constrained convex sets $\K$, whereas VAW is natively an unconstrained algorithm. 

The projection step $\proj_{\K}^{A_t}$ in Algorithm~\ref{alg:ons-square} plays a crucial mathematical role in keeping the algorithm stable within the strict OCO protocol. To see why, consider the \emph{unprojected} version of the algorithm where $\K = \reals^d$. The parameter update simplifies to:
\[
\x_{t+1} = \x_t - A_t^{-1} \a_t (\x_t^\top \a_t - b_t).
\]
Multiplying both sides by $A_t = A_{t-1} + \a_t \a_t^\top$ gives:
\[
A_t \x_{t+1} = (A_{t-1} + \a_t \a_t^\top) \x_t - \a_t \a_t^\top \x_t + b_t \a_t = A_{t-1} \x_t + b_t \a_t.
\]
By induction (and assuming $A_0 \x_1 = 0$), we obtain:
\[
\x_{t+1} = A_t^{-1} \sum_{s=1}^t b_s \a_s.
\]

The vulnerability of unprojected ONS is that, because the decision $\x_t$ must be made \emph{before} seeing $\a_t$, the resulting prediction $\x_t^\top \a_t$ can be overly aggressive.

\subsection{The VAW Forecaster}

The \textbf{Vovk-Azoury-Warmuth (VAW)} forecaster achieves logarithmic regret \emph{without} requiring projections, but it must relax the strict OCO protocol to do so. In the VAW framework, the unlabelled feature $\a_t$ is revealed \emph{before} the learner predicts. 

\begin{algorithm}[h]
\caption{Vovk-Azoury-Warmuth (VAW) Forecaster}
\label{alg:vaw}
\begin{algorithmic}[1]
\STATE Input: regularization $\lambda>0$
\STATE Initialize $A_0 = \lambda I_d$, target vector $\bv_0 = \bzero \in \reals^d$
\FOR{$t=1$ to $T$}
    \STATE Observe feature vector $\a_t$
    \STATE Update curvature matrix: $A_t = A_{t-1} + \a_t \a_t^\top$
    \STATE Predict $\hat{b}_t = \a_t^\top A_t^{-1} \bv_{t-1}$
    \STATE Observe true label $b_t$ and suffer loss $\ell_t(\hat{b}_t) = \frac12(\hat{b}_t - b_t)^2$
    \STATE Update target vector: $\bv_t = \bv_{t-1} + b_t \a_t$
\ENDFOR
\end{algorithmic}
\end{algorithm}

Because $\a_t$ is known in advance, VAW exploits this by updating the precision matrix with the current feature before committing to a prediction. By the Sherman-Morrison formula, $A_t^{-1} \a_t = \frac{A_{t-1}^{-1} \a_t}{1 + \a_t^\top A_{t-1}^{-1} \a_t}$. Consequently, the VAW prediction can be written in terms of the unprojected ONS prediction (where $\x_t = A_{t-1}^{-1} \bv_{t-1}$):
\[
\hat b_t^{\text{VAW}} = \frac{\x_t^\top \a_t}{1 + \a_t^\top A_{t-1}^{-1} \a_t}.
\]
This demonstrates that VAW fundamentally \emph{shrinks} the standard unprojected Ridge Regression prediction by the factor $(1 + \a_t^\top A_{t-1}^{-1} \a_t)$. Whenever the model is highly uncertain in the direction of $\a_t$, the denominator grows and shrinks the prediction. 

This implicit regularization automatically bounds the loss relative to the true labels, entirely bypassing the need for worst-case domain or magnitude bounds. 

\begin{theorem}[Regret of the VAW Forecaster]
\label{thm:vaw}
For Algorithm~\ref{alg:vaw}, assume the true labels are bounded such that $|b_t| \le Y$ for all $t=1,\dots,T$. For any benchmark $\x^\star \in \reals^d$, the regret is bounded by:
\[
\sum_{t=1}^T \frac12(\hat{b}_t - b_t)^2 - \sum_{t=1}^T \frac12(\x^\star{}^\top \a_t - b_t)^2
\le
\frac{\lambda}{2} \|\x^\star\|_2^2
+
\frac{Y^2}{2} \log\det\!\left( I_d + \frac{1}{\lambda} \sum_{t=1}^T \a_t \a_t^\top \right).
\]
In particular, if $\|\a_t\|_2 \le R$, and setting $\lambda = 1/\|\x^\star\|^2$, with $d,Y,T,R > 2$ we get:
\[
\regret_T(\x^\star) \le \frac{\lambda}{2} \|\x^\star\|_2^2 + \frac{d Y^2}{2} \log\!\left(1 + \frac{T R^2}{\lambda}\right) \leq 4 d Y^2 \log (TR\|\x^*\|) .
\]
\end{theorem}

Comparing Theorem~\ref{thm:ons-square} to Theorem~\ref{thm:vaw} highlights a distinct advantage of VAW: the regret bound depends on $Y^2$, which is determined strictly by the maximum magnitude of the observed labels. In contrast, the ONS bound depends on $C$, which scales with both the diameter of the decision set $\K$ and the feature norm $R$. However, this elegant shrinkage in VAW relies on the domain being unconstrained ($\reals^d$) and requires violating the strict zero-lookahead OCO timeline, making projected ONS the necessary algorithm when dealing with strict constraints or true zero-lookahead settings.

\subsection{VAW with Known Additive Offsets}
\label{sec:vaw-offset}

We now consider a shifted prediction setting in which, at each round $t$, the learner observes both the feature vector $\a_t \in \reals^d$ and a known scalar offset $c_t \in \reals$ before making a prediction. The learner then predicts $\hat z_t \in \reals$, after which the environment reveals the true label $z_t \in \reals$, and the learner suffers square loss
\[
\ell_t(\hat z_t) = \frac12(\hat z_t - z_t)^2.
\]
We compare against benchmarks of the form $\x^\star{}^\top \a_t + c_t$, i.e. predictors that are allowed to use the same known offset:
\[
\regret_T^{\mathrm{off}}(\x^\star)
=
\sum_{t=1}^T \frac12(\hat z_t-z_t)^2
-
\sum_{t=1}^T \frac12(\x^\star{}^\top \a_t + c_t - z_t)^2.
\]

The key point is that, since $c_t$ is known prior to prediction, one can simply subtract it from the labels and run the usual VAW forecaster on the centered targets.

\begin{algorithm}[h]
\caption{VAW Forecaster with Known Additive Offsets}
\label{alg:vaw-offset}
\begin{algorithmic}[1]
\STATE Input: regularization $\lambda>0$
\STATE Initialize $A_0 = \lambda I_d$, target vector $\bv_0 = \bzero \in \reals^d$
\FOR{$t=1$ to $T$}
    \STATE Observe feature vector $\a_t$ and known offset $c_t$
    \STATE Update curvature matrix:
    \[
    A_t = A_{t-1} + \a_t \a_t^\top
    \]
    \STATE Predict the centered label:
    \[
    \hat y_t = \a_t^\top A_t^{-1}\bv_{t-1}
    \]
    \STATE Output the shifted prediction:
    \[
    \hat z_t = \hat y_t + c_t
    \]
    \STATE Observe true label $z_t$ and suffer loss
    \[
    \ell_t(\hat z_t) = \frac12(\hat z_t-z_t)^2
    \]
    \STATE Update target vector using the centered label:
    \[
    \bv_t = \bv_{t-1} + (z_t-c_t)\a_t
    \]
\ENDFOR
\end{algorithmic}
\end{algorithm}

\begin{theorem}[VAW regret with known additive offsets]
\label{thm:vaw-offset}
For Algorithm~\ref{alg:vaw-offset}, assume the centered labels are bounded:
\[
|z_t-c_t| \le Y
\qquad
\text{for all } t=1,\dots,T.
\]
Then for any benchmark $\x^\star \in \reals^d$,
\[
\sum_{t=1}^T \frac12(\hat z_t-z_t)^2
-
\sum_{t=1}^T \frac12(\x^\star{}^\top \a_t + c_t - z_t)^2
\le
\frac{\lambda}{2}\|\x^\star\|_2^2
+
\frac{Y^2}{2}
\log\det\!\left(
I_d + \frac{1}{\lambda}\sum_{t=1}^T \a_t \a_t^\top
\right).
\]
In particular, the regret bound depends only on the centered labels $z_t-c_t$, and has no dependence on the magnitude of the offsets $c_t$.

If furthermore $\|\a_t\|_2 \le R$ for all $t$, then
\[
\sum_{t=1}^T \frac12(\hat z_t-z_t)^2
-
\sum_{t=1}^T \frac12(\x^\star{}^\top \a_t + c_t - z_t)^2
\le
\frac{\lambda}{2}\|\x^\star\|_2^2
+
\frac{dY^2}{2}\log\!\left(1+\frac{TR^2}{\lambda}\right).
\]
\end{theorem}

\begin{proof}
Define the centered labels and centered predictions by
\[
y_t := z_t - c_t,
\qquad
\hat y_t := \hat z_t - c_t.
\]
By construction of Algorithm~\ref{alg:vaw-offset}, we have
\[
\hat y_t = \a_t^\top A_t^{-1}\bv_{t-1},
\qquad
\bv_t = \bv_{t-1} + y_t \a_t.
\]
Therefore, when written in terms of $(\a_t,y_t)$, Algorithm~\ref{alg:vaw-offset} is exactly the standard VAW forecaster of Algorithm~\ref{alg:vaw} applied to the centered labels $y_t$.

Since $|y_t| = |z_t-c_t| \le Y$, Theorem~\ref{thm:vaw} implies that for every $\x^\star \in \reals^d$,
\[
\sum_{t=1}^T \frac12(\hat y_t-y_t)^2
-
\sum_{t=1}^T \frac12(\x^\star{}^\top \a_t-y_t)^2
\le
\frac{\lambda}{2}\|\x^\star\|_2^2
+
\frac{Y^2}{2}
\log\det\!\left(
I_d + \frac{1}{\lambda}\sum_{t=1}^T \a_t \a_t^\top
\right).
\]

It remains to relate this centered regret to the original shifted problem. By definition of $y_t$ and $\hat y_t$,
\[
\hat z_t = \hat y_t + c_t,
\qquad
z_t = y_t + c_t.
\]
Hence the square loss is exactly translation invariant:
\[
\frac12(\hat z_t-z_t)^2
=
\frac12\big((\hat y_t+c_t)-(y_t+c_t)\big)^2
=
\frac12(\hat y_t-y_t)^2.
\]
Likewise, for any benchmark $\x^\star$,
\[
\frac12(\x^\star{}^\top \a_t + c_t - z_t)^2
=
\frac12\big(\x^\star{}^\top \a_t + c_t - (y_t+c_t)\big)^2
=
\frac12(\x^\star{}^\top \a_t-y_t)^2.
\]
Substituting these identities into the previous bound yields
\[
\sum_{t=1}^T \frac12(\hat z_t-z_t)^2
-
\sum_{t=1}^T \frac12(\x^\star{}^\top \a_t + c_t - z_t)^2
\le
\frac{\lambda}{2}\|\x^\star\|_2^2
+
\frac{Y^2}{2}
\log\det\!\left(
I_d + \frac{1}{\lambda}\sum_{t=1}^T \a_t \a_t^\top
\right),
\]
which proves the first claim.

The simplified bound under $\|\a_t\|_2 \le R$ follows exactly as in Theorem~\ref{thm:vaw}, by bounding the log-determinant term via
\[
\log\det\!\left(
I_d + \frac{1}{\lambda}\sum_{t=1}^T \a_t \a_t^\top
\right)
\le
d\log\!\left(1+\frac{TR^2}{\lambda}\right).
\]

Thus the regret bound is completely independent of the magnitude of the offsets $c_t$; only the centered labels $z_t-c_t$ matter.
\end{proof}

\backmatter
\bibliographystyle{alpha}

\addcontentsline{toc}{chapter}{Bibliography}
     \markboth{\sffamily\slshape Bibliography}
       {\sffamily\slshape Bibliography}
\bibliography{bookbib}

\printindex
\end{document}